\documentclass[acmlarge,screen,nonacm]{acmart} %
\usepackage{array}
\usepackage{colortbl} %
\usepackage{multicol}

\makeatletter\providecommand\@currentauthors{}\makeatother
\usepackage{subcaption}
\usepackage{tikz}
\ifdefined\pdfinclusioncopyfonts\pdfinclusioncopyfonts=1\fi
\usetikzlibrary{arrows.meta,decorations.pathreplacing,fit,positioning,calc}
\newcommand{\TODO}[1][]{\textcolor{red}{\textbf{TODO}\ifx\relax#1\relax\else: #1\fi}}

\newcommand{\ChanBorder}{9}

\newcommand{\ChanNumFew}{13}
\newcommand{\ChanNumMany}{16}
\newcommand{\ChanRhoWithinFew}{0.55}
\newcommand{\ChanRhoWithinMany}{0.58}
\newcommand{\ChanRhoBetween}{0.41}
\newcommand{\ChanRhoChannelsSubjects}{-0.48}

\newcommand{\ChanBestMany}{CNN-HAR}
\newcommand{\ChanRhoRandomForest}{-0.67}
\newcommand{\ChanPartialRhoRandomForest}{-0.58}
\newcommand{\ChanRhoLITEWAY}{-0.75}
\newcommand{\ChanPartialRhoLITEWAY}{-0.69}
\newcommand{\ChanRhoDeepConvLSTM}{0.60}
\newcommand{\ChanPartialRhoDeepConvLSTM}{0.52}

\newcommand{\ChanRankFewCNNHAR}{4.7}
\newcommand{\ChanRankManyCNNHAR}{5.1}
\newcommand{\ChanMcuFitsFewCNNHAR}{12}
\newcommand{\ChanMcuFitsManyCNNHAR}{5}
\newcommand{\ChanRankFewDANA}{5.5}
\newcommand{\ChanRankManyDANA}{9.0}

\newcommand{\ChanRankFewDeepConvLSTM}{10.6}
\newcommand{\ChanRankManyDeepConvLSTM}{13.5}

\newcommand{\ChanRankFewDeepConvLSTMAttn}{8.2}
\newcommand{\ChanRankManyDeepConvLSTMAttn}{12.2}

\newcommand{\ChanRankFewLITEWAY}{12.7}
\newcommand{\ChanRankManyLITEWAY}{5.8}

\newcommand{\ChanMcuFitsManyLITEWAY}{16}

\newcommand{\ChanRankFewMLPHAR}{13.2}
\newcommand{\ChanRankManyMLPHAR}{9.0}

\newcommand{\ChanRankFewMLPMixer}{8.3}
\newcommand{\ChanRankManyMLPMixer}{13.4}

\newcommand{\ChanRankFewRandomForest}{13.1}
\newcommand{\ChanRankManyRandomForest}{6.1}

\newcommand{\ChanRankFewSVM}{18.4}
\newcommand{\ChanRankManySVM}{16.3}

\newcommand{\ChanMcuFitsManyTinierHAR}{14}
\newcommand{\ChanRankFewTinyHAR}{5.2}
\newcommand{\ChanRankManyTinyHAR}{5.4}

\newcommand{\ChanMcuFitsManyTinyHAR}{10}
\newcommand{\ChanRankFewTripleCrossAttn}{4.7}
\newcommand{\ChanRankManyTripleCrossAttn}{5.5}

\newcommand{\ChanMcuNumFew}{13}
\newcommand{\ChanMcuNumMany}{16}

\newcommand{\ClassBorder}{12}

\newcommand{\ClassMin}{2}
\newcommand{\ClassMax}{33}

\newcommand{\ClassRhoWithinFew}{0.44}
\newcommand{\ClassRhoWithinMany}{0.51}
\newcommand{\ClassRhoBetween}{0.49}
\newcommand{\ClassRhoClassesChannels}{0.05}

\newcommand{\ClassRankFewDynamicWHAR}{8.2}
\newcommand{\ClassRankManyDynamicWHAR}{5.6}

\newcommand{\CompRNNWithFewChannels}{8.2}
\newcommand{\CompRNNWithoutFewChannels}{9.3}
\newcommand{\CompRNNWithManyChannels}{10.4}
\newcommand{\CompRNNWithoutManyChannels}{8.2}

\newcommand{\CompAttnWithFewChannels}{8.5}
\newcommand{\CompAttnWithoutFewChannels}{9.1}
\newcommand{\CompAttnWithManyChannels}{8.2}
\newcommand{\CompAttnWithoutManyChannels}{10.4}

\newcommand{\CompRNNNumWith}{8}
\newcommand{\CompRNNNumWithout}{8}

\newcommand{\PairNumBetween}{208}

\newcommand{\PairNumNegative}{9}

\newcommand{\PropRateBorder}{50}
\newcommand{\PropRateNumLow}{8}
\newcommand{\PropRateWithinHigh}{0.56}
\newcommand{\PropRateWithinLow}{0.36}
\newcommand{\PropRateBetween}{0.41}
\newcommand{\PropDeviceDifference}{0.04}
\newcommand{\PropNoEffectMax}{0.03}
\newcommand{\PropRateFallModelLow}{9.4}
\newcommand{\PropRateFallModelHigh}{15.3}
\newcommand{\PropRateGainCNNLow}{7.1}
\newcommand{\PropRateGainCNNHigh}{4.0}
\newcommand{\PropRateGainTinyLow}{7.4}
\newcommand{\PropRateGainTinyHigh}{4.5}

\newcommand{\GenGapMedian}{15.0}
\newcommand{\GenGapMax}{50.5}
\newcommand{\GenGapMaxDataset}{WISDM-19 (Phone)}
\newcommand{\GenGapMin}{3.9}
\newcommand{\GenGapMinDataset}{UCI-HAR}
\newcommand{\GenGapTopTwo}{46.2}
\newcommand{\GenGapTopTwoDataset}{RealLifeHAR}
\newcommand{\GenGapTopThree}{37.8}
\newcommand{\GenGapTopThreeDataset}{AICOS-HAR}

\newcommand{\GenGapRhoWindows}{0.98}
\newcommand{\GenGapModelStd}{6.6}
\newcommand{\GenGapSubjectsMin}{4}
\newcommand{\GenGapSubjectsMax}{94}
\newcommand{\GenGapMaxSubjects}{51}
\newcommand{\GenGapTopThreeSubjects}{94}
\newcommand{\GenGapRhoSubjects}{-0.07}
\newcommand{\GenGapRhoChannels}{0.20}
\newcommand{\GenGapRhoTestFOne}{-0.57}
\newcommand{\GenGapRhoChannelsPartial}{0.19}
\newcommand{\GenGapRhoChannelsValFOne}{0.63}
\newcommand{\GenGapRhoChannelsTestFOne}{0.32}

\newcommand{\HwRhoMacsMcuLat}{0.97}
\newcommand{\HwRhoParamsMcuLat}{0.88}
\newcommand{\HwRhoMacsPhoneLat}{0.93}
\newcommand{\HwRhoParamsPhoneLat}{0.69}
\newcommand{\HwRhoParamsMcuRam}{0.47}
\newcommand{\HwRhoParamsMcuFlash}{0.89}

\newcommand{\HwEnergyLatR}{1.000}

\newcommand{\HwPowerCv}{5}
\newcommand{\HwLatencySpan}{33,962}
\newcommand{\HwPhoneGmacsMin}{0.18}
\newcommand{\HwPhoneGmacsMinModel}{LITEWAY}
\newcommand{\HwPhoneGmacsMax}{10.8}
\newcommand{\HwPhoneGmacsMaxModel}{DeepConvShallowLSTM}
\newcommand{\HwPhoneGmacsRatio}{61}
\newcommand{\HwMcuMmacsRatio}{13}

\newcommand{\HwNoFitFlash}{181}
\newcommand{\HwNoFitRam}{39}

\newcommand{\InsYearRho}{-0.23}
\newcommand{\InsYearP}{0.39}
\newcommand{\InsBestModel}{CNN-HAR}
\newcommand{\InsBestModelYear}{2015}

\newcommand{\InsParamsRho}{0.38}
\newcommand{\InsParamsP}{0.14}

\newcommand{\InsArchSpread}{4.8}
\newcommand{\InsNumSubjectLarger}{23}
\newcommand{\InsNumSpreadDatasets}{29}

\newcommand{\InsWinNumPairs}{551}
\newcommand{\InsWinMedianDiff}{+0.5}
\newcommand{\InsWinShareBetter}{58}

\newcommand{\InsWinChanRho}{-0.45}
\newcommand{\InsWinRankTau}{0.94}

\newcommand{\InsEnergyHourNumSave}{4}
\newcommand{\InsEnergyHourNumCost}{5}
\newcommand{\InsEnergyHourNumModels}{15}

\newcommand{\DsHighMedian}{80}
\newcommand{\DsLowMedian}{50}

\newcommand{\DsNumDatasets}{29}

\newcommand{\DsNumModelsWithWin}{8}
\newcommand{\DsMeanEasiest}{SAD}
\newcommand{\DsMeanEasiestValue}{93.02}
\newcommand{\DsMeanHardest}{WISDM-19 (Phone)}
\newcommand{\DsMeanHardestValue}{23.91}
\newcommand{\DsMeanEasiestStd}{4.86}
\newcommand{\DsMeanHardestStd}{3.02}
\newcommand{\DsStdLargestOneMean}{59.63}

\newcommand{\DsStdLargestTwoMean}{77.82}

\newcommand{\DsNumMeanHigh}{7}
\newcommand{\DsNumMeanLow}{4}
\newcommand{\DsStdLargestOne}{UTD-MHAD}
\newcommand{\DsStdLargestOneValue}{15.89}

\newcommand{\DsStdLargestTwo}{REALDISP}
\newcommand{\DsStdLargestTwoValue}{14.43}

\newcommand{\DsStdMean}{7.89}
\newcommand{\DsStdStd}{2.92}
\newcommand{\DsWinNumNear}{26}

\newcommand{\RankNumDatasets}{29}

\newcommand{\RankFirstModel}{CNN-HAR}
\newcommand{\RankSecondModel}{Triple-Cross-Attn}
\newcommand{\RankThirdModel}{TinyHAR}
\newcommand{\ModelMeanRankCNNHAR}{4.90}

\newcommand{\ModelMeanFOneCNNHAR}{74.4}
\newcommand{\ModelMeanGapCNNHAR}{3.1}
\newcommand{\ModelStdGapCNNHAR}{2.8}

\newcommand{\ModelMaxGapCNNHAR}{10.3}
\newcommand{\ModelWinsCNNHAR}{5}

\newcommand{\ModelWithinCloseCNNHAR}{24}

\newcommand{\ModelMeanRankTripleCrossAttn}{5.14}

\newcommand{\ModelMeanFOneTripleCrossAttn}{74.0}

\newcommand{\ModelBeatsFirstTripleCrossAttn}{15}
\newcommand{\ModelMeanRankTinyHAR}{5.28}

\newcommand{\ModelMeanFOneTinyHAR}{73.6}

\newcommand{\ModelBeatsFirstTinyHAR}{12}

\newcommand{\ModelMeanRankRandomForest}{9.24}

\newcommand{\ModelMeanFOneRandomForest}{70.0}
\newcommand{\ModelMeanGapRandomForest}{7.6}
\newcommand{\ModelStdGapRandomForest}{7.4}

\newcommand{\ModelWinsRandomForest}{6}

\newcommand{\ModelTopThreeRandomForest}{9}
\newcommand{\ModelBeatsFirstRandomForest}{11}

\newcommand{\LeadThreshold}{2.5}
\newcommand{\LeadNum}{8}
\newcommand{\LeadModels}{CNN-HAR, Triple-Cross-Attn, TinyHAR, DynamicWHAR, TinierHAR, LITEWAY-Light, DANA, and Attend-and-Discriminate}

\newcommand{\LeadFOneSpread}{2.2}

\newcommand{\TopThreeFOneSpread}{0.8}
\newcommand{\GapNear}{2}
\newcommand{\GapClose}{5}

\newcommand{\McuRelRamParetoModelsThreeS}{Random Forest, LITEWAY-Light, TinierHAR, TinyHAR, Triple-Cross-Attn, and CNN-HAR}
\newcommand{\McuRelFlashParetoModelsThreeS}{LITEWAY-Light and CNN-HAR}

\newcommand{\McuFitsCNNHARThreeS}{17}

\newcommand{\McuFitsTinyHARThreeS}{22}
\newcommand{\McuFitsTinierHARThreeS}{25}

\newcommand{\McuFitsAttendAndDiscriminateThreeS}{3}

\newcommand{\McuFitsRandomForestThreeS}{25}

\newcommand{\McuTotalThreeS}{29}

\newcommand{\ParetoShareTopTwoModel}{LITEWAY-Light}
\newcommand{\ParetoShareTopTwo}{63}

\newcommand{\ParetoNumLatencyCases}{120}
\newcommand{\ParetoNumAllPlatforms}{68}
\newcommand{\ParetoOverlapEnergy}{0.99}
\newcommand{\ParetoOverlapRAM}{0.75}
\newcommand{\ParetoOverlapFlash}{0.63}
\newcommand{\ParetoShareRAMLITEWAY}{48}
\newcommand{\ParetoShareRAMRandomForest}{86}
\newcommand{\ParetoShareFlashLITEWAYLight}{90}
\newcommand{\ParetoShareFlashRandomForest}{38}
\newcommand{\ParetoShareLatencyLITEWAY}{16}
\newcommand{\ParetoShareLatencyRandomForest}{95}

\newcommand{\WatchNumModelsInBudget}{18}
\newcommand{\WatchMostAccurateInBudget}{CNN-HAR}

\newcommand{\McuMedianLatSCNNHAR}{2.6}

\newcommand{\McuMedianLatSTripleCrossAttn}{3.5}

\newcommand{\McuNumModelsInBudget}{9}
\newcommand{\McuMostAccurateInBudget}{TinyHAR}
\newcommand{\McuFitParetoModels}{LITEWAY-Light, TinierHAR, TinyHAR, and CNN-HAR}

\newcommand{\PhoneMemMaxMb}{216}
\newcommand{\PhoneMemMaxShare}{2}
\newcommand{\WatchMemMaxMb}{174}
\newcommand{\WatchMemMaxShare}{11}

\newcommand{\BudgetS}{1.5}

\newcommand{\RankNumClusters}{4}
\newcommand{\KendallW}{0.50}

\newcommand{\RepNumModels}{16}
\newcommand{\RepNumDatasets}{29}
\newcommand{\RepNumFamiliar}{15}
\newcommand{\RepNumNever}{14}
\newcommand{\RepTopDatasetOne}{Opportunity (Loc.)}
\newcommand{\RepTopDatasetOneCount}{13}
\newcommand{\RepTopDatasetTwo}{PAMAP2}
\newcommand{\RepTopDatasetTwoCount}{11}
\newcommand{\RepNumOnlyTop}{6}
\newcommand{\RepOnlyTopModels}{Attend-and-Discriminate, CNN-HAR, DeepConvLSTM, DeepConvLSTM-Attn, DynamicWHAR, and Guan-LSTM}
\newcommand{\RepNumBetter}{8}
\newcommand{\RepNumWorse}{8}
\newcommand{\RepManyThreshold}{4}
\newcommand{\RepManyNum}{6}
\newcommand{\RepManyNumBetter}{4}

\newcommand{\RepMedianAbsDiff}{1.7}
\newcommand{\RepManyBetterMaxFOne}{2.3}

\newcommand{\RepFewNum}{10}
\newcommand{\RepFewNumWorse}{6}
\newcommand{\RepWithoutTopNumModels}{10}
\newcommand{\RepWithoutTopNumBetter}{5}
\newcommand{\RepWithoutTopNumWorse}{5}

\newcommand{\RepSplitLeadNum}{8}

\newcommand{\RepSplitLeadInTopNeverText}{All 8}
\newcommand{\RepSplitTau}{0.80}
\newcommand{\RepRandomTauMedian}{0.77}

\newcommand{\RepSplitMoveThreshold}{3}
\newcommand{\RepSplitNumMovers}{4}

\newcommand{\SubsetNumDatasets}{29}
\newcommand{\SubsetNumDraws}{2000}
\newcommand{\SubsetLowPercent}{5}
\newcommand{\SubsetHighPercent}{95}
\newcommand{\SubsetFewK}{5}
\newcommand{\SubsetSomeK}{10}

\newcommand{\SubsetMeanTauOne}{0.21}
\newcommand{\SubsetMeanTauOneLow}{-0.16}

\newcommand{\SubsetMeanTauTwo}{0.35}
\newcommand{\SubsetMeanTauFour}{0.47}
\newcommand{\SubsetMeanTauFew}{0.52}
\newcommand{\SubsetMeanTauSome}{0.63}
\newcommand{\SubsetMeanTauMaxK}{14}
\newcommand{\SubsetMeanTauMax}{0.68}

\newcommand{\SubsetNumModels}{17}

\newcommand{\SubsetExcludedModels}{k-NN and SVM}

\newcommand{\SubsetMeanROne}{0.32}

\newcommand{\SubsetMeanRMax}{0.87}

\newcommand{\PhoneLatGrowth}{1.39}

\newcommand{\WatchLatGrowth}{1.37}

\newcommand{\McuLatGrowth}{1.55}

\newcommand{\WinMcuCombos}{551}
\newcommand{\WinMcuFitsTwoS}{297}
\newcommand{\WinMcuFitsThreeS}{278}

\newcommand{\WinFlashKbLITEWAYLightTwoS}{179}

\newcommand{\WinFlashKbLITEWAYLightThreeS}{182}

\newcommand{\WinBigFlashModels}{Attend-and-Discriminate, DeepConvLSTM-Attn, and DeepConvLSTM}

\newcommand{\WinBigFlashMinMbTwoS}{1.4}
\newcommand{\WinBigFlashMaxMbTwoS}{1.5}
\newcommand{\WinBigFitsRangeTwoS}{5 to 6}
\newcommand{\WinBigFlashMinMbThreeS}{1.7}
\newcommand{\WinBigFlashMaxMbThreeS}{1.9}
\newcommand{\WinBigFitsRangeThreeS}{3}

\newcommand{\WatchSlowdownMedian}{11}

\newcommand{\WatchNumModelsOverBudget}{7}

\newcommand{\McuPowerMinMw}{24.9}
\newcommand{\McuPowerMaxMw}{30.1}
\newcommand{\McuPowerMedianMw}{28.2}
\newcommand{\McuEnergyLatencySpearman}{0.999}
\newcommand{\McuLowestEnergyModel}{LITEWAY-Light}
\newcommand{\McuLowestEnergyMj}{8.3}

\newcommand{\EffParamsLITEWAYLight}{5.4k}
\newcommand{\EffMcuRamLITEWAYLight}{78.3}
\newcommand{\EffMcuLatLITEWAYLight}{311}

\newcommand{\EffParamsLITEWAY}{5.6k}
\newcommand{\EffMcuRamLITEWAY}{53.4}
\newcommand{\EffMcuLatLITEWAY}{648}

\newcommand{\LiteRamHigherShare}{84}

\newcommand{\LiteLatRatio}{2.1}
\newcommand{\RelMcuReference}{LITEWAY-Light}
\newcommand{\RelMcuMinFits}{10}

\newcommand{\RelMcuLatCNNHARThreeS}{25.9}

\newcommand{\RelMcuLatDANAThreeS}{61.3}

\newcommand{\RelMcuLatMLPMixerThreeS}{72.2}

\newcommand{\RelMcuLatRandomForestThreeS}{0.02}

\newcommand{\RelMcuFlashRandomForestThreeS}{1.89}

\newcommand{\RelMcuLatTinierHARThreeS}{2.33}

\newcommand{\RelMcuLatTinyHARThreeS}{8.62}

\newcommand{\FOneTabNumDatasetsTwoS}{29}
\newcommand{\FOneTabPartOneNumDatasetsTwoS}{15}
\newcommand{\FOneTabPartTwoNumDatasetsTwoS}{14}
\newcommand{\FOneTabNumDatasetsThreeS}{29}
\newcommand{\FOneTabPartOneNumDatasetsThreeS}{15}
\newcommand{\FOneTabPartTwoNumDatasetsThreeS}{14}

\newcommand{\NumModels}{19}
\newcommand{\NumDeepModels}{16}

\newcommand{\NumDatasets}{29}

\newcommand{\TrainingLosoFitsTotal}{25\,688}
\newcommand{\TrainingFitsTotal}{10\,564}

\newcommand{\HarenaURL}{\anon[\url{https://anonymous.4open.science/r/HARena-F9E1/}]{\url{https://github.com/teco-kit/HARena}}}
\newcommand{\WharModelsURL}{\anon[\url{https://anonymous.4open.science/r/whar-models/}]{\url{https://github.com/teco-kit/whar-models}}}
\newcommand{\WharDatasetsURL}{\anon[\url{https://anonymous.4open.science/r/whar-datasets-8141/}]{\url{https://github.com/teco-kit/whar-datasets}}}
\AtBeginDocument{%
  }

\AtBeginDocument{%
  }

\setcopyright{none}
\renewcommand\footnotetextcopyrightpermission[1]{}

\title{HARENA: A Living Benchmark for Sensor-Based Human Activity Recognition on Resource-Constrained Edge Devices}

\author{Maximilian Burzer}
\authornote{Both authors contributed equally to this research.}
\email{maximilian.burzer@kit.edu}
\orcid{0009-0000-9628-8667}
\affiliation{%
  \institution{Karlsruhe Institute of Technology}
  \city{Karlsruhe}
  \country{Germany}
}

\author{Tobias King}
\authornotemark[1]
\email{tobias.king@kit.edu}
\orcid{0009-0006-7289-3356}
\affiliation{%
  \institution{Karlsruhe Institute of Technology}
  \city{Karlsruhe}
  \country{Germany}
}

\author{Till Riedel}
\email{till.riedel@kit.edu}
\orcid{0000-0003-4547-1984}
\affiliation{%
  \institution{Karlsruhe Institute of Technology}
  \city{Karlsruhe}
  \country{Germany}
}

\author{Michael Beigl}
\email{michael.beigl@kit.edu}
\orcid{0000-0001-5009-2327}
\affiliation{%
  \institution{Karlsruhe Institute of Technology}
  \city{Karlsruhe}
  \country{Germany}
}

\author{Tobias Röddiger}
\email{tobias.roeddiger@ipai-foundation.ai}
\orcid{0000-0002-4718-9280}
\affiliation{%
  \institution{IPAI Foundation gGmbH} %
  \city{Heilbronn}
  \country{Germany}
}

\begin{document}

\begin{abstract}
Sensor-based human activity recognition runs on edge devices, trading performance against latency, memory, and energy.
Yet new architectures are usually compared with few, often outdated baselines on few datasets with study-specific data preparation and hardware, leaving open which perform best at what cost.
We present HARENA, a living benchmark with a public leaderboard, to which new architectures and datasets can be added under one protocol.
Its initial suite evaluates \NumDeepModels{} neural architectures and three classical baselines on \NumDatasets{} datasets, measuring performance on unseen subjects and deployment cost on a smartphone, smartwatch, and microcontroller.
\RankFirstModel{}, the suite's oldest deep architecture, leads overall, but \LeadNum{} architectures lie within \LeadThreshold{} points of its mean test macro-F1, while their cost differs by orders of magnitude.
On the microcontroller, Random Forest leads on latency and peak RAM, LITEWAY-Light on flash, and \McuMostAccurateInBudget{} is most accurate within the real-time budget.
Rankings shift with dataset and channel count, so general claims require many datasets, which a growing benchmark can provide.
We find no sign of benchmark overfitting, but performance drops by a median of \GenGapMedian{} points on unseen subjects, about three times the typical spread between architectures, making generalization and adaptation the most promising future directions.
\end{abstract}




\maketitle

\section{Introduction}
\label{sec:introduction}

Sensor-based Human Activity Recognition (HAR) infers human activities
from signals captured by smartphones, smartwatches, and other wearable
devices~\cite{bulling2014tutorial,lara2012survey,wang2019deep}.
Many applications, from fitness tracking to healthcare, require continuous
recognition, and streaming raw data to a server is often impractical, as
connectivity is not always available and wireless transmission consumes
considerable energy and exposes personal data
\cite{trotta2023optimizing,banbury2021mlperf}. Inference is therefore
performed on the device, within a tight computational, memory, and energy
budget. This setting favors compact task-specific models trained for a given
sensing configuration over large foundation models~\cite{bian2026foundation},
whose memory and compute demands are costly to meet continuously and whose
scale does not remove the need to ground inference in the specific user,
device, and sensor placement~\cite{kawsar2026calibration}.

HAR models have evolved from classifiers based on hand-crafted features to
end-to-end neural networks \cite{chen2021deep}, complemented by efficient
architectures for edge devices such as TinyHAR~\cite{zhou2022tinyhar},
TinierHAR \cite{bian2025tinierhar}, and LITEWAY~\cite{nshimyimana2026liteway}.
Broader evaluations have begun with the multi-dataset evaluation released with
TinyHAR, which subsequent studies have adopted. However, most studies still
compare a new design mainly against its immediate predecessors or
long-established baselines such as DeepConvLSTM~\cite{ordonez2016deep},
and each relies on its own selection of datasets, preparation choices, and
hardware. It therefore remains unclear which architectures perform best, under
which sensing conditions, and at what deployment cost.

A major obstacle is that results from different publications cannot be aggregated into a reliable comparison. Public HAR datasets require dataset-specific decisions about
parsing, resampling, windowing, and normalization that can substantially affect
recognition performance~\cite{alam2023open,nguyen2024sok,jordao2018human},
studies differ in subject separation, training procedures, and model selection,
and deployment costs are reported as complexity counts or measured on
study-specific hardware. Evaluations also rest on few recurring datasets: of the
\NumDeepModels{} architectures considered in this work, \RepTopDatasetOneCount{}
report results on Opportunity~\cite{roggen2010opportunity} and
\RepTopDatasetTwoCount{} on PAMAP2~\cite{reiss2012pamap2}. Since rankings may
change with the sensing configuration, such a selection does not establish a
generally better architecture, and it is unknown how many datasets a reliable
comparison needs. Repeated use of familiar datasets may also favor choices
tailored to them, an effect known as \emph{benchmark overfitting}
\cite{roelofs2019meta,dwork2015generalization}.

To address these challenges, we introduce \emph{HARENA}, a living benchmark for
sensor-based HAR on resource-constrained edge devices. Since a fixed comparison
becomes incomplete as new architectures and datasets appear, HARENA is maintained
with a public leaderboard and a contribution workflow, through which future
contributions are evaluated under the same protocol and compared directly with
established ones. Its initial suite comprises \NumDatasets{}
datasets, \NumDeepModels{} neural architectures, and three classical baselines,
prepared by a standardized pipeline and evaluated under a common protocol. We
investigate which architectures achieve the strongest recognition performance
across datasets (\textbf{RQ1}) and which offer favorable trade-offs between
recognition performance and deployment cost (\textbf{RQ2}). We further ask how
consistently their rankings hold across datasets, including those not reported
in their original publications, and which dataset properties explain the
differences (\textbf{RQ3}).

Our contributions are as follows:

\begin{enumerate}
    \item \textbf{HARENA.} A living, open benchmark with a public website\footnote{\HarenaURL{}}, leaderboard, and
    contribution workflow, through which new datasets and architectures are evaluated
    under identical, reproducible conditions.

    \item \textbf{Dataset and architecture suite.} An initial suite of \NumDatasets{}
    heterogeneous HAR datasets\footnote{\WharDatasetsURL{}}, \NumDeepModels{} task-specific neural architectures\footnote{\WharModelsURL{}} from
    different publications, and three classical baselines.

    \item \textbf{Standardized data preparation.} A configuration-driven
    pipeline from raw recordings to the samples of each fold, so that each fold
    can be reproduced exactly from a single configuration.

    \item \textbf{Large-scale evaluation.} An evaluation of recognition
    performance on unseen subjects, latency and memory on a smartphone, a
    smartwatch, and a microcontroller, and energy and flash on the
    microcontroller. It shows that a broad group of architectures performs
    similarly while their cost differs by orders of magnitude, that rankings
    require many datasets to stabilize, and that the drop on unseen subjects far
    exceeds the differences between architectures.

    \item \textbf{Practical guidance.} Pareto fronts for each platform and
    resource, and a shortlist of candidate architectures per device and
    channel group.
\end{enumerate}

\section{Background and Related Work}
\label{sec:background}

Assessing architectural progress requires understanding the evidence
behind reported advances. We outline the architectures and their
evaluations (\autoref{sec:related-architectures}), examine how far this
evidence extends across subjects (\autoref{sec:related-evaluation}) and
datasets (\autoref{sec:related-overfitting}), how comparable it is across
publications (\autoref{sec:related-data}), and how deployment cost is
measured (\autoref{sec:related-efficiency}), and then review existing
benchmarks (\autoref{sec:related-benchmarks}).

\subsection{Sensor-Based HAR Architectures}
\label{sec:related-architectures}

The architectures compared in this work span more than a decade of HAR
research. Early systems applied conventional learners such as k-nearest
neighbors, support vector machines, and random forests to handcrafted
features \cite{lara2012survey}. Convolutional networks replaced this
manual design with learned representations \cite{yang2015deep}, and
DeepConvLSTM added recurrent layers \cite{ordonez2016deep},
becoming a common baseline. Later architectures added recurrent
ensembles, shallower recurrent stages, and attention
\cite{guan2017ensembles,bock2021improving,murahari2018attention,mahmud2020human,tang2022triple,abedin2021attend},
adapted to multi-sensor inputs
\cite{malekzadeh2021dana,miao2022towards}, or replaced convolution
and recurrence with fully connected layers \cite{ojiako2023mlps}. We
focus on such task-specific architectures, trained separately for each
dataset, rather than on foundation models that rely on large-scale pretraining \cite{bian2026foundation}. Scale alone does not remove the
need to adapt models to a given user and device
\cite{kawsar2026calibration}, and compact models trained for a specific
sensing configuration meet this need at a cost compatible with
continuous on-device inference.

Edge deployment has made efficiency a design objective in its own right
\cite{trotta2023optimizing}. TinyHAR often matches or surpasses
DeepConvLSTM at several times smaller size \cite{zhou2022tinyhar},
MLP-HAR reduces complexity by at least an order of magnitude relative to
TinyHAR \cite{zhou2024mlp}, TinierHAR requires 2.7$\times$ fewer
parameters than TinyHAR at comparable average F1-scores
\cite{bian2025tinierhar}, and LITEWAY reduces model size by up to
9.5$\times$ through fully convolutional temporal modeling
\cite{nshimyimana2026liteway}. Each thus claims a better balance between
recognition performance and deployment cost than its predecessors.

The evidence for these claims has broadened from two datasets for
CNN-HAR and DeepConvLSTM \cite{yang2015deep,ordonez2016deep} to
six for TinyHAR and MLP-HAR \cite{zhou2022tinyhar,zhou2024mlp}, 14 for
TinierHAR \cite{bian2025tinierhar}, and 16 for LITEWAY
\cite{nshimyimana2026liteway}, but it concentrates on recurring datasets:
Opportunity is reported for \RepTopDatasetOneCount{} and PAMAP2 for
\RepTopDatasetTwoCount{} of the \NumDeepModels{} architectures, whereas
\RepNumNever{} of the \RepNumDatasets{} datasets in our suite are reported
by none of them (\autoref{fig:dataset-usage}). Each study also compares
against its immediate predecessors and reports efficiency in its own
terms, from parameter and operation counts to latency and energy on a
microcontroller \cite{bian2025tinierhar,zhou2024mlp,nshimyimana2026liteway}.
Each publication therefore provides controlled evidence for its own
architecture, but this evidence is bound to its subjects and datasets,
and its deployment costs are not measured under common conditions.

\subsection{Subject-Independent Evaluation and Generalization}
\label{sec:related-evaluation}

This evidence must first hold for new users. Because movement patterns, sensor placement, and execution style vary between individuals, architectures must be evaluated on subjects unseen during training. Earlier studies often split windows at random, so that the
same subject, and with overlapping windows even the same measurements,
appeared in both training and test data. Avoiding such overlap lowers
accuracy by ten percentage points on average
\cite{jordao2018human}, and the gains commonly attributed to
overlapping windows disappear under subject-independent evaluation
\cite{dehghani2019quantitative}.

A fixed split into training and test subjects removes scarce subjects from
training, since many public HAR datasets contain few participants, such as
Opportunity \cite{roggen2010opportunity} and PAMAP2 \cite{reiss2012pamap2}
with four and nine, and it yields an estimate that depends strongly on who
is held out. Leave-one-subject-out (LOSO) cross-validation instead tests each
subject once while training on all others. It has become a common protocol in
HAR \cite{alam2023open,jordao2018human} and is applied by
recent architecture studies for most of their datasets
\cite{zhou2022tinyhar,zhou2024mlp,bian2025tinierhar,nshimyimana2026liteway}.

Subject-independent evaluation exposes the loss on new users but does
not reduce it \cite{weiss2012impact,kawsar2026calibration}. Two lines of work address this
shift. Foundation models aim to learn more general representations through
large-scale pretraining with self-supervised objectives such as
contrastive learning \cite{bian2026foundation}, whereas calibration and
adaptation ground inference in the specific user, device, and placement
\cite{kawsar2026calibration}, with efficient on-device approaches being
explored \cite{burzer2026bayahar}.

\subsection{Evidence Across Datasets}
\label{sec:related-overfitting}

Results must also hold beyond the datasets on which they were obtained,
and comparing algorithms over multiple datasets is a methodological
question of its own \cite{demvsar2006statistical}. In computer vision, new
test sets lowered accuracy while largely preserving model rankings
\cite{recht2019imagenet}, consistent with the broader observation that
rankings replicate better than accuracy numbers
\cite{hardt2026emerging}. HAR evaluations are smaller and more
heterogeneous. Their datasets differ in sensors, channels, sampling rates,
and activities, and such differences affect recognition
\cite{nguyen2024sok}, so a ranking need not carry over from one sensing
configuration to another. To our knowledge, no study has
examined how many datasets are needed before a ranking of HAR
architectures stabilizes, or whether the number of channels or
activities explains changes in it.

Reusing evaluation data adds a second concern. LOSO is commonly used to
retain or discard components according to the same mean score that is later
reported \cite{varma2006bias}, and new architectures are designed in light of
results on the same public datasets. Such adaptive reuse can overfit holdout
data even when no model is trained on them \cite{dwork2015generalization},
and designs may exploit characteristics of the populations and recording
conditions in a benchmark \cite{torralba2011unbiased}, an effect known as
benchmark overfitting. Since training and model-selection procedures are
often reported in insufficient detail \cite{huang2023standardizing}, examining it
requires evaluating the published architecture unchanged on datasets that did
not inform its design. Concern about overfitting has been raised \cite{cai2025towards} for long-used datasets such as UCI-HAR \cite{anguita2013public}, but to our
knowledge, no study has measured it in HAR or examined whether reported
advantages of HAR architectures persist on such datasets.

\subsection{Data Preparation and Reproducibility}
\label{sec:related-data}

Such results cannot be assembled from the existing literature, as results from different publications are not directly comparable. Public
Public HAR datasets come from independent sensing studies and require extensivepreprocessing \cite{alam2023open}, whose decisions shape the task itself. Sampling rates affect
classification accuracy \cite{nguyen2024sok}, window length trades recognition
speed against accuracy \cite{banos2014window}, overlap can cause leakage
\cite{jordao2018human}, activity selection and transition handling
change the target labels \cite{geissler2024beyond}, and metrics
differ in how they weight skewed classes \cite{ward2011performance}.
Architecture studies differ in these decisions as well: PAMAP2 alone is evaluated with windows
of 1.28\,s \cite{zhou2024mlp}, 4\,s \cite{nshimyimana2026liteway}, and 5.12\,s
\cite{zhou2022tinyhar}. The datasets
also contain irregularities that are easy to miss and that each study must
resolve on its own. UCI-HAR is distributed only as fixed windows with 50\%
overlap and a predefined split, so concatenating
these windows repeats half of the measurements. In HARTH \cite{logacjov2021harth}, one of the 22 recordings is stored at 100\,Hz although the dataset is described as 50\,Hz. HHAR \cite{stisen2015hhar} and AICOS-HAR \cite{barandas2026human} combine devices with different sampling rates, and in AICOS-HAR
only 94 of 106 participants have a recording with all four sensors. Training
procedures add further variation \cite{huang2023standardizing}. Results bearing
the same dataset name can therefore reflect different procedures, and
differences between publications cannot be attributed to architecture alone.

For new datasets, Niemann et al.\ propose a framework, taxonomy, and
checklist for dataset creation
\cite{niemann2025towards}, but datasets already in use must be
standardized after publication. Several frameworks do so for parts of the
pipeline. B-HAR provides a
configurable baseline workflow on nine datasets \cite{turetta2024b}, DAGHAR
harmonizes six smartphone datasets for domain adaptation research
\cite{napoli2024domain}, and the initial WHAR Datasets Library introduced a
common format and configuration-driven processing for nine datasets
\cite{burzer2025whar}. None has yet been used to compare a broad set
of architectures across many datasets, for which we extend the WHAR Datasets
Library (\autoref{sec:method-library}).

\subsection{Deployment Efficiency}
\label{sec:related-efficiency}

Comparability also extends to the deployment cost at which this performance is achieved, yet efficiency claims in HAR rest on different quantities, which are not interchangeable. For continuous recognition on embedded devices, inference
latency determines whether a window is processed before the next one arrives,
and peak memory, dominated by intermediate activations, must fit into budgets
about two orders of magnitude smaller than those of conventional machine
learning systems. Model size determines the required storage, and energy
governs battery life but requires dedicated measurement hardware
\cite{banbury2021mlperf}. All of these depend on the device, runtime,
numerical precision, and input configuration, which in HAR varies between
datasets. Parameter and operation counts are common stand-ins but track these
costs only loosely: architectures with nearly the same number of operations can
differ substantially in latency on the same phone \cite{tan2019mnasnet}, and
predictors based on operation counts are far less accurate than those that
model the device \cite{zhang2021nn}. On microcontrollers, models are
typically quantized to integer arithmetic \cite{jacob2018quantization} and
executed by a constrained runtime \cite{david2021tensorflow}, so that operators that
remain in floating point and the way a runtime represents recurrent layers can
change memory and flash. Classical models can be especially cheap in this
setting: in an accelerometer-based study, Random Forests matched the accuracy
of a small CNN on a microcontroller while running several times faster
\cite{elsts2021microcontrollers}.

Because recognition performance and deployment cost cannot generally be
optimized at the same time, architectures must be compared by their trade-off.
MLPerf Tiny compares cost at a fixed quality target
\cite{banbury2021mlperf}. Without such a target, Pareto analysis identifies
the architectures for which no alternative achieves higher recognition
performance at equal or lower cost, and unlike scalar efficiency scores, it
requires no weighting. Any such comparison requires that performance and cost
refer to the same input and that costs are measured under identical conditions,
which the architecture studies meet at most within each publication.

\subsection{HAR Benchmarks and Leaderboards}
\label{sec:related-benchmarks}

\begin{table*}[b]
    \centering
    \caption{Representative multi-dataset work in the standard supervised
    HAR setting (\checkmark: yes, --: no or not reported).
    \emph{Classical baselines}: includes non-neural classifiers.
    \emph{Subject-indep.}: test subjects are excluded from training
    (for frameworks: such splits are supported).
    \emph{Standardized data prep.}: a reusable pipeline prepares raw recordings
    for training.
    \emph{Ranking consist.}: analyzes how rankings hold across datasets.
    \emph{Measured latency}, \emph{memory}, \emph{energy}: devices on which this cost is
    measured, without complexity counts (HARENA also measures flash on the MCU).
    \emph{Trade-off analysis}: joint analysis of recognition performance
    and cost across the compared models.
    \emph{Living}: designed to incorporate new models and datasets with
    maintained, publicly reported results.
    Benchmarks for other learning problems are discussed in
    \autoref{sec:related-benchmarks}.}
    \label{tab:related-work}
    \footnotesize
    \setlength{\tabcolsep}{1.5pt}
    \renewcommand{\arraystretch}{1.1}

    \begin{tabular*}{\linewidth}{@{\extracolsep{\fill}}l c c c c c c >{\centering\arraybackslash}p{0.07\linewidth} >{\centering\arraybackslash}p{0.07\linewidth} >{\centering\arraybackslash}p{0.07\linewidth} c c@{}}
        \toprule
        \scriptsize\textbf{Study}
        & \scriptsize\textbf{\shortstack{HAR\\datasets}}
        & \scriptsize\textbf{Models}
        & \scriptsize\textbf{\shortstack{Classical\\baselines}}
        & \scriptsize\textbf{\shortstack{Subject-\\indep.}}
        & \scriptsize\textbf{\shortstack{Standardized\\data prep.}}
        & \scriptsize\textbf{\shortstack{Ranking\\consist.}}
        & \scriptsize\textbf{\shortstack{Measured\\latency}}
        & \scriptsize\textbf{\shortstack{Measured\\memory}}
        & \scriptsize\textbf{\shortstack{Measured\\energy}}
        & \scriptsize\textbf{\shortstack{Trade-off\\analysis}}
        & \scriptsize\textbf{Living}
        \\
        \midrule
        \multicolumn{12}{@{}l}{\emph{Frameworks}} \\
        B-HAR~\cite{turetta2024b}
            & 9 & 10 & \checkmark & \checkmark & \checkmark & -- & -- & -- & -- & -- & -- \\
        WHAR Datasets~\cite{burzer2025whar}
            & 9 & 2 & -- & \checkmark & \checkmark & -- & -- & -- & -- & -- & -- \\
        \midrule
        \multicolumn{12}{@{}l}{\emph{Model benchmarks}} \\
        MONSTER~\cite{dempster2025monster}
            & 7 & 7 & \checkmark & \checkmark & -- & -- & -- & -- & -- & -- & -- \\
        Hossain et al.~\cite{hossain2025benchmarking}
            & 5 & 14 & \checkmark & -- & -- & -- & -- & -- & -- & -- & -- \\
        \midrule
        \multicolumn{12}{@{}l}{\emph{Architecture studies}} \\
        TinyHAR~\cite{zhou2022tinyhar}
            & 6 & 2 & -- & \checkmark & -- & -- & -- & -- & -- & -- & -- \\
        MLP-HAR~\cite{zhou2024mlp}
            & 6 & 11 & -- & \checkmark & -- & -- & Watch, MCU & -- & -- & -- & -- \\
        TinierHAR~\cite{bian2025tinierhar}
            & 14 & 3 & -- & \checkmark & -- & -- & -- & -- & -- & -- & -- \\
        LITEWAY~\cite{nshimyimana2026liteway}
            & 16 & 4 & -- & \checkmark & -- & -- & MCU & MCU & MCU & \checkmark & -- \\
        \midrule
        \textbf{HARENA}
            & \textbf{\NumDatasets} & \textbf{\NumModels} & \checkmark & \checkmark & \checkmark & \checkmark
            & Phone, watch, MCU & Phone, watch, MCU & MCU & \checkmark & \checkmark \\
        \bottomrule
    \end{tabular*}
\end{table*}

The preceding subsections define the requirements for a shared reference:
evaluation on datasets beyond those used during development, common data
preparation and training, and deployment costs measured under identical
conditions. Existing benchmarks meet these only in part. MONSTER includes seven
HAR datasets with participant-based folds but evaluates general-purpose
time-series classifiers \cite{dempster2025monster}, and Hossain et al.\ and
Abdel-Salam et al.\ compare classical and deep models on five and six HAR
datasets \cite{hossain2025benchmarking,abdel2021human}, with Hossain et al.\
reporting CNNs ahead on all five and Random Forest competitive only on smaller
datasets. Their model suites do not cover the task-specific architectures of
\autoref{sec:related-architectures}, and none measures deployment costs. HAR benchmarks complement this work by addressing different learning problems,
including domain adaptation and out-of-distribution generalization
\cite{napoli2024domain,lu2026harood}, continual learning
\cite{jha2021continual}, and self-supervised and foundation-model transfer
\cite{tanigaki2026harbench,cai2026benchhar}.

A shared reference must also remain current as new architectures and datasets
appear. Other fields address this with continuously maintained benchmarks:
TabArena is a living benchmark for tabular data that can revise its datasets and
evaluation as flaws surface or new methods appear \cite{erickson2026tabarena},
and the RobustBench leaderboard aims to reflect the state of the art in
adversarial robustness \cite{croce2020robustbench}. Shared evaluation in HAR has
a long tradition \cite{kawaguchi2011hasc,alvarez2013evaluation}, and
HARBench recently introduced a public leaderboard for foundation models
\cite{tanigaki2026harbench}. To our knowledge, no comparable, continuously
maintained reference yet exists for the compact, task-specific architectures
intended for deployment on resource-constrained edge devices.

\autoref{tab:related-work} summarizes representative work. HARENA combines the
components that prior work provides separately. To our knowledge, it is the
first benchmark to evaluate task-specific HAR architectures from different
publications under one protocol on this many datasets, together with measured
deployment costs on a smartphone, a smartwatch, and a microcontroller, with
results that can be extended as new architectures and datasets emerge.

\section{Benchmark Design}
\label{sec:methodology}

HARENA is designed around the requirements identified in \autoref{sec:background}.
The dataset suite (\autoref{sec:method-datasets}) is prepared by the WHAR Datasets Library (\autoref{sec:method-library}), the model suite (\autoref{sec:method-models}) is implemented through the WHAR Models Library, and fixed protocols specify how architectures are trained and evaluated on unseen subjects (\autoref{sec:method-protocol}) and how their deployment costs are measured (\autoref{sec:method-efficiency}).
All components are defined through documented configurations, so new datasets and architectures can be evaluated under identical conditions (\autoref{sec:platform}).

\subsection{Dataset Suite}
\label{sec:method-datasets}

\begin{table*}[t]
    \centering
    \caption{Datasets of the benchmark suite, sorted by
    publication year. Act.: activity classes used in the benchmark. ADL: activities of daily living,
    Loc: locomotion, Fit: fitness, Acc: accelerometer, Gyr: gyroscope, Mag: magnetometer,
    HR: heart rate, ECG: electrocardiogram, EEG: electroencephalogram, EMG: electromyogram,
    ALS: ambient light sensor, Baro: barometer, Resp: respiration.}
    \label{tab:dataset-suite}
    \footnotesize
    \setlength{\tabcolsep}{3pt}
    \begin{tabular*}{\linewidth}{@{\extracolsep{\fill}}l c r r l l l r@{}}
        \toprule
        \textbf{Dataset} & \textbf{Year} & \textbf{Subj.} & \textbf{Act.}
        & \textbf{Settings} & \textbf{Devices} & \textbf{Modalities} & \textbf{Ch.} \\
        \midrule
        Daphnet~\cite{moore2008daphnet} & 2008 & 10 & 2 & Health, Loc & Node & Acc & 9 \\
        WISDM~\cite{kwapisz2011wisdm} & 2011 & 36 & 6 & ADL, Loc & Phone & Acc & 3 \\
        Opportunity (Loc.)~\cite{roggen2010opportunity} & 2010 & 4 & 4 & ADL, Loc & Node & Acc, Gyr, Mag & 133 \\
        DSADS~\cite{altun2010dsads} & 2010 & 8 & 19 & ADL, Loc, Fit & Node & Acc, Gyr, Mag & 45 \\
        PAMAP2~\cite{reiss2012pamap2} & 2012 & 9 & 12 & ADL, Loc, Fit & Node & Acc, Gyr, Mag, HR & 52 \\
        USC-HAD~\cite{zhang2012uschad} & 2012 & 14 & 12 & ADL, Loc, Fit & Node & Acc, Gyr & 6 \\
        UCI-HAR~\cite{anguita2013public} & 2013 & 30 & 6 & ADL, Loc & Phone & Acc, Gyr & 9 \\
        MHEALTH~\cite{banos2014mhealthdroid} & 2014 & 10 & 12 & ADL, Loc, Fit, Health & Node & Acc, Gyr, Mag, ECG & 23 \\
        SAD~\cite{shoaib2014sad} & 2014 & 10 & 7 & ADL, Loc, Fit & Phone & Acc, Gyr, Mag & 60 \\
        REALDISP~\cite{banos2014realdisp} & 2014 & 17 & 33 & ADL, Loc, Fit & Node & Acc, Gyr, Mag & 117 \\
        HHAR~\cite{stisen2015hhar} & 2015 & 9 & 6 & ADL, Loc, Fit & Phone, Watch & Acc, Gyr & 6 \\
        UTD-MHAD~\cite{chen2015utdmhad} & 2015 & 8 & 27 & ADL, Loc, Fit & Node & Acc, Gyr & 6 \\
        HAPT~\cite{reyesortiz2016hapt} & 2016 & 30 & 12 & ADL, Loc & Phone & Acc, Gyr & 6 \\
        UniMiB-SHAR~\cite{micucci2017unimib} & 2017 & 30 & 17 & ADL, Fall & Phone & Acc & 3 \\
        UMAFall~\cite{casilari2017umafall} & 2017 & 19 & 15 & ADL, Fall & Node, Phone & Acc, Gyr, Mag & 39 \\
        HuGaDB~\cite{chereshnev2017hugadb} & 2017 & 18 & 10 & ADL, Loc, Fit & Node & Acc, Gyr, EMG & 38 \\
        UP-Fall~\cite{martinez2019upfall} & 2019 & 17 & 11 & ADL, Fall & Node, Helmet & Acc, Gyr, ALS, EEG & 42 \\
        MotionSense~\cite{malekzadeh2019motionsense} & 2019 & 24 & 6 & ADL, Loc & Phone & Acc, Gyr & 18 \\
        WISDM-19 (Phone)~\cite{weiss2019wisdm19} & 2019 & 51 & 18 & ADL, Loc & Phone & Acc, Gyr & 6 \\
        WISDM-19 (Watch)~\cite{weiss2019wisdm19} & 2019 & 51 & 18 & ADL, Loc & Watch & Acc, Gyr & 6 \\
        RealLifeHAR~\cite{garcia2020reallifehar} & 2020 & 17 & 4 & ADL, Loc & Phone & Acc, Gyr, Mag, GPS & 15 \\
        w-HAR~\cite{bhat2020whar} & 2020 & 22 & 8 & ADL, Loc & Node & Acc, Gyr, Mag, Stretch & 7 \\
        HARTH~\cite{logacjov2021harth} & 2021 & 22 & 12 & ADL, Loc, Fit & Node & Acc & 6 \\
        HAR70+~\cite{bach2022har70} & 2021 & 18 & 7 & ADL, Loc, Health & Activity tracker & Acc & 6 \\
        UCA-EHAR~\cite{novac2022ucaehar} & 2022 & 20 & 11 & ADL, Loc, Fit & Glasses & Acc, Gyr, Baro & 7 \\
        GOTOV~\cite{paraschiakos2022gotov} & 2020 & 30 & 16 & ADL, Loc, Fit, Health & Node, others & Acc, ECG, Resp, HR & 28 \\
        Hang-Time~\cite{hoelzemann2023hangtime} & 2023 & 24 & 4 & Fit, Sports & Watch & Acc & 3 \\
        WEAR~\cite{bock2024wear} & 2024 & 24 & 18 & Fit, Sports & Watch & Acc & 12 \\
        AICOS-HAR~\cite{barandas2026human} & 2026 & 94 & 11 & ADL, Loc & Phone, wearable & Acc, Gyr, Mag, Baro & 10 \\
        \bottomrule
    \end{tabular*}
\end{table*}

HAR datasets are neither introduced nor indexed consistently in the literature, which prevents search queries from isolating them reliably.
We therefore collected all datasets used in the original publications of the architectures in the model suite and added datasets referenced in the surveys and benchmark studies from our literature analysis, which yielded 81 candidate datasets.

We then applied seven selection criteria.
\emph{Modality}: datasets must contain time-series signals of human motion or physiology, which excludes vision-, audio-, and skeleton-based datasets.
\emph{Subjects}: datasets must provide subject identifiers and more than one subject, which excludes datasets such as Skoda \cite{stiefmeier2008skoda}.
\emph{Availability}: datasets must be publicly accessible without manual approval or registration, which excludes, among others, HASC2010 \cite{kawaguchi2011hasc} and MobiAct \cite{vavoulas2016mobiact}, as well as RealWorld \cite{sztyler2016realworld}, whose download was not available at the time of this study.
\emph{Size}: datasets whose size would disproportionately constrain the number of datasets that can be evaluated, such as ExtraSensory \cite{vaizman2017extrasensory} and Capture-24 \cite{chan2024capture24}, are excluded.
\emph{Quality}: datasets with incomplete releases, corrupted files, or severe annotation inconsistencies, such as SHL \cite{gjoreski2018shl} and KU-HAR \cite{sikder2021ku}, are excluded.
\emph{Redundancy}: extensions that only add modalities outside our scope, such as Opportunity++ \cite{ciliberto2021opportunitypp}, are not included separately.
\emph{Adoption}: datasets must have been cited at least 30 times, as a proxy for their use in HAR research.
Since citations lag behind publication and a living benchmark should not exclude the newest releases for that reason alone, we also admit newly published, peer-reviewed datasets that substantially extend the suite, which applies to AICOS-HAR \cite{barandas2026human}, with by far the most participants and recordings across multiple devices and body positions.
None of these criteria depends on the performance of any architecture.

The resulting suite comprises \NumDatasets{} datasets (\autoref{tab:dataset-suite}), from early ones such as WISDM \cite{kwapisz2011wisdm} and Opportunity \cite{roggen2010opportunity} to recent ones such as WEAR \cite{bock2024wear} and AICOS-HAR \cite{barandas2026human}.
It ranges from 4 to 94 participants, from \ClassMin{} to \ClassMax{} activity classes, and from 3 to more than 100 channels, recorded with smartphones, smartwatches, dedicated sensor nodes, and other wearable devices.
This heterogeneity is deliberate: the benchmark does not represent one application scenario but tests whether architectural advantages persist across sensing configurations, activity spaces, and populations.

\subsection{WHAR Datasets Library}
\label{sec:method-library}

\begin{figure}[t]
    \centering
    \begin{tikzpicture}[
        font=\footnotesize,
        proc/.style={draw, rounded corners=2pt, align=center, inner sep=3pt,
                     minimum height=0.95cm, text width=1.85cm},
        data/.style={draw, fill=black!8, align=center, inner sep=3pt,
                     minimum height=0.95cm, text width=1.85cm},
        arr/.style={-{Stealth[length=4pt]}, thick},
        cfgarr/.style={-{Stealth[length=3.5pt]}, dashed, black!60},
        group/.style={draw=black!45, dashed, rounded corners=4pt, inner sep=6pt},
    ]
    \node[data] (raw) at (0,0) {Raw dataset\\files};
    \node[proc] (dl) at (2.5,0) {Download and\\extract};
    \node[proc] (parse) at (5.0,0) {Parse and\\validate};
    \node[data] (std) at (7.5,0) {Sessions and\\session index};
    \node[proc] (win) at (10.0,0) {Resample and\\window};
    \draw[arr] (raw) -- (dl);
    \draw[arr] (dl) -- (parse);
    \draw[arr] (parse) -- (std);
    \draw[arr] (std) -- (win);

    \node[data] (wins) at (12.5,-2.0) {Windows and\\window index};
    \draw[arr] (win.east) -| (wins.north);

    \node[proc] (split) at (10.0,-4.0) {Split by\\subject};
    \node[proc] (post) at (7.5,-4.0) {Normalize and\\transform};
    \node[data] (samples) at (5.0,-4.0) {Samples\\per fold};
    \node[proc] (load) at (2.5,-4.0) {Loader and\\adapters};
    \node[data] (fw) at (0,-4.0) {PyTorch or\\TensorFlow};
    \draw[arr] (wins.south) |- (split.east);
    \draw[arr] (split) -- (post);
    \draw[arr] (post) -- (samples);
    \draw[arr] (samples) -- (load);
    \draw[arr] (load) -- (fw);

    \node[group, fit=(dl)(parse)(std)(win)] (pre) {};
    \node[anchor=south west, font=\footnotesize\itshape]
        at (pre.north west) {Preprocessing (once per configuration)};
    \node[group, fit=(split)(post)(samples)(load)] (postg) {};
    \node[anchor=north west, font=\footnotesize\itshape]
        at (postg.south west) {Postprocessing (once per fold)};

    \path (dl.west) -- (win.east) coordinate[pos=0.5] (cfgc);
    \node[proc, minimum height=0.6cm, text width=9.2cm]
        (cfg) at (cfgc |- wins) {Dataset configuration (metadata, parameters, and parser)};
    \foreach \n in {dl, parse, win}
        \draw[cfgarr] (\n.south |- cfg.north) -- (\n.south);
    \foreach \n in {split, post, load}
        \draw[cfgarr] (\n.north |- cfg.south) -- (\n.north);
    \end{tikzpicture}
    \caption{Data preparation in the WHAR Datasets Library. A dataset-specific
    configuration, consisting of metadata, preparation parameters, and a
    parser into the standardized session-based format, controls every step, whereas the
    implementation of all steps is dataset-independent. Preprocessing is
    executed once per configuration and produces windows, from which
    postprocessing derives and loads the samples of each fold. Grey boxes denote cached artifacts.}
    \label{fig:whar-pipeline}
\end{figure}

To prepare these recordings consistently, all datasets are integrated into the WHAR Datasets Library\footnote{\WharDatasetsURL{}} \cite{burzer2025whar}, which we extended from nine to 39 datasets and improved in its data preparation, splitting, and caching.

The library separates what is specific to a dataset from what is common to all datasets.
Each dataset is integrated through a typed configuration of metadata (its source, sampling rate, subjects, activities, and channels), parameters specifying its preparation, and a parser that converts the raw recordings into a standardized format.
All subsequent operations are implemented once in a dataset-independent core and are controlled solely by the configuration, so that a dataset is prepared identically whenever the same configuration is used.

The standardized format is session-centric: a session is a continuous, timestamped multivariate recording of one subject performing one activity.
Sensor data are stored separately from index tables that map activities to their names, sessions to their subjects and activities, and windows to their session and sample range, so that every window remains traceable to its subject and its position in the original recording.
Parsed data are validated against the format, including consistent identifiers and the declared numbers of subjects and activities.

Data preparation is divided into a fold-independent preprocessing and a fold-specific postprocessing pipeline (\autoref{fig:whar-pipeline}).
Preprocessing downloads and parses the raw data, selects activities and channels, optionally resamples each session, and segments it into fixed-length sliding windows.
Parsers split recordings into separate sessions at every change of activity and at timestamp gaps above a configurable threshold, align sensors recorded at different rates to a common timeline, and interpolate missing values.
Sessions that still contain missing or non-finite values are rejected during validation, and sessions shorter than one window are dropped.
Problems specific to a dataset are resolved in its parser.
UCI-HAR \cite{anguita2013public}, for example, is distributed as windows with 50\,\% overlap, of which the parser keeps only the non-overlapping half to restore a continuous signal, the one HARTH \cite{logacjov2021harth} recording stored at 100\,Hz is filtered and downsampled to the published 50\,Hz, HHAR \cite{stisen2015hhar} treats each device stream as a separate source, and acquisitions of AICOS-HAR \cite{barandas2026human} without all four sensors are left out.
Resampling first maps a session onto a regular grid at its source rate, which removes timing jitter, and then converts it with a polyphase filter.
Since windows are extracted within sessions, every window carries exactly one activity label, and none spans an activity transition or a recording gap.
Splitters partition the windows into folds of training, validation, and test data and support LOSO, subject-group, and $k$-fold protocols.
The subject-group splitter shuffles the subjects with a fixed seed and assigns them in turn to $\min(k, n)$ groups for $n$ subjects, so that group sizes differ by at most one and the protocol reduces to LOSO for $n \leq k$.
Because overlapping windows share measurements, validation windows are drawn as up to three contiguous blocks spread over each session of the training subjects, and training windows that overlap them are removed, so that no measurement appears in more than one partition.
Sessions too short to be divided are assigned as a whole.
Postprocessing estimates the per-channel mean and standard deviation on the training data of each fold \cite{jordao2018human}, standardizes all partitions with them, and produces model-ready samples.

Every step caches its output under a hash of the configuration fields and steps it depends on, so preprocessing runs once per dataset and postprocessing once per fold, regardless of how many architectures are evaluated.
Sharing a configuration thus suffices to reproduce the prepared data exactly.

\subsection{Model Suite}
\label{sec:method-models}

\begin{table}[b]
    \centering
    \caption{Architectures of the model suite and their
    building blocks. Attn: attention, CNN: convolution, Dense: fully connected layers, RNN: recurrence,
    Graph: graph-based sensor interaction, Spec: spectral input transformation,
    FE: handcrafted feature extraction.}
    \label{tab:model-suite}
    \footnotesize
    \setlength{\tabcolsep}{3pt}
    \begin{tabular*}{\linewidth}{@{\extracolsep{\fill}}l c c c c c c c c@{}}
        \toprule
        \textbf{Architecture} & \textbf{Year} & \textbf{Attn} & \textbf{CNN}
        & \textbf{Dense} & \textbf{RNN} & \textbf{Graph} & \textbf{Spec} & \textbf{FE} \\
        \midrule
        CNN-HAR~\cite{yang2015deep} & 2015 & & \checkmark & & & & & \\
        DeepConvLSTM~\cite{ordonez2016deep} & 2016 & & \checkmark & \checkmark & \checkmark & & & \\
        Guan-LSTM~\cite{guan2017ensembles} & 2017 & & & \checkmark & \checkmark & & & \\
        DeepConvLSTM-Attn~\cite{murahari2018attention} & 2018 & \checkmark & \checkmark & \checkmark & \checkmark & & & \\
        SA-HAR~\cite{mahmud2020human} & 2020 & \checkmark & \checkmark & \checkmark & & & & \\
        Attend-and-Discriminate~\cite{abedin2021attend} & 2021 & \checkmark & \checkmark & \checkmark & \checkmark & & & \\
        DANA~\cite{malekzadeh2021dana} & 2021 & & \checkmark & \checkmark & \checkmark & & & \\
        DeepConvShallowLSTM~\cite{bock2021improving} & 2021 & & \checkmark & \checkmark & \checkmark & & & \\
        DynamicWHAR~\cite{miao2022towards} & 2022 & & \checkmark & \checkmark & & \checkmark & & \\
        TinyHAR~\cite{zhou2022tinyhar} & 2022 & \checkmark & \checkmark & \checkmark & \checkmark & & & \\
        Triple-Cross-Attn~\cite{tang2022triple} & 2022 & \checkmark & \checkmark & \checkmark & & & & \\
        MLP-Mixer~\cite{ojiako2023mlps} & 2023 & & & \checkmark & & & & \\
        MLP-HAR~\cite{zhou2024mlp} & 2024 & & & \checkmark & & & \checkmark & \\
        TinierHAR~\cite{bian2025tinierhar} & 2025 & \checkmark & \checkmark & \checkmark & \checkmark & & & \\
        LITEWAY~\cite{nshimyimana2026liteway} & 2026 & \checkmark & \checkmark & & & & & \\
        LITEWAY-Light~\cite{nshimyimana2026liteway} & 2026 & \checkmark & \checkmark & & & & & \\
        \midrule
        $k$-NN~\cite{pedregosa2011scikit} & -- & & & & & & & \checkmark \\
        Random Forest~\cite{pedregosa2011scikit} & -- & & & & & & & \checkmark \\
        SVM~\cite{pedregosa2011scikit} & -- & & & & & & & \checkmark \\
        \bottomrule
    \end{tabular*}
\end{table}

The architectures compared on these datasets had to be identified in the literature. Since they are often introduced as part of broader systems or used only as baselines, keyword queries alone do not retrieve them reliably, so we combined a systematic search with backward citation tracking.
Following the PRISMA methodology \cite{page2021prisma}, we queried the ACM Digital Library for publications mentioning human activity recognition in their abstract together with terms such as \emph{model}, \emph{architecture}, or \emph{network}.
Of the 864 results, 83 remained after excluding those outside sensor-based HAR based on their abstracts, and backward citation tracking added influential architectures that the search had not captured.

We selected architectures according to four criteria.
\emph{Availability}: the source code must be public, since key implementation details are often missing from publications, which excludes, among others, SPECTRA \cite{gurung2026spectra}, for which we could not find released code.
\emph{Task fit}: the architecture must address supervised multi-class classification of windowed sensor data.
\emph{Generality}: it must operate on generic multivariate windows with a configurable number of channels, time steps, and classes, without hard-wired assumptions about sensor placement, sensor grouping, or dataset-specific features, since it could not otherwise be evaluated fairly under one protocol.
\emph{Feasibility}: training and inference must fit the budget of a benchmark of this scale and deployment on edge devices.
DeepSense \cite{yao2017deepsense}, AttnSense \cite{ma2019attnsense}, GlobalFusion \cite{liu2020giobalfusion}, and IF-ConvTransformer \cite{zhang2022if} are excluded because they rely on a fixed sensor grouping or preprocessing structure, and AROMA-Joint \cite{peng2018aroma} because it could not be adapted to generic activity recognition without departing from its intended design.

The resulting suite comprises \NumDeepModels{} architectures (\autoref{tab:model-suite}) from the convolutional, recurrent, attention-based, and fully connected families, including multi-sensor and efficient edge designs, with LITEWAY and LITEWAY-Light counted separately.
In addition, three classical baselines, $k$-nearest neighbors, random forests, and support vector machines, are applied to features extracted with tsfresh \cite{christ2018time,pedregosa2011scikit}.
All deep architectures share a common PyTorch \cite{paszke2019pytorch} interface in the WHAR Models Library\footnote{\WharModelsURL{}} that receives a window of sensor data and returns class scores.
The computation of the author-released implementation is preserved wherever possible, and required changes are minimal and documented.
CNN-HAR, for instance, was ported to PyTorch with adaptive temporal pooling to support varying window lengths, Attend-and-Discriminate uses a single dropout setting instead of dataset-specific values, and Guan-LSTM is trained as a single model rather than as an ensemble.
The results therefore characterize these implementations rather than every possible realization of the published designs.

For each deep architecture, we recorded which datasets of the suite its original publication reports (\autoref{fig:dataset-usage}).
These form its \emph{reported} datasets, and all others its \emph{unseen} datasets.
The assignment follows the publication record, since the complete development history cannot be observed, so datasets that informed a design without being reported may count as unseen.

\begin{figure*}[t]
    \centering
    \scriptsize
    \setlength{\tabcolsep}{1.2pt}
    \renewcommand{\arraystretch}{1.05}
    \begin{tabular*}{\linewidth}{@{\extracolsep{\fill}}l c ccccccccccccccccccccccccccccc c@{}}
        \toprule
        \textbf{Architecture} & \textbf{Year} & \rotatebox{90}{WISDM~\cite{kwapisz2011wisdm}} & \rotatebox{90}{UCI-HAR~\cite{anguita2013public}} & \rotatebox{90}{UTD-MHAD~\cite{chen2015utdmhad}} & \rotatebox{90}{HAPT~\cite{reyesortiz2016hapt}} & \rotatebox{90}{USC-HAD~\cite{zhang2012uschad}} & \rotatebox{90}{UniMiB-SHAR~\cite{micucci2017unimib}} & \rotatebox{90}{MotionSense~\cite{malekzadeh2019motionsense}} & \rotatebox{90}{RealLifeHAR~\cite{garcia2020reallifehar}} & \rotatebox{90}{WISDM-19 (P)~\cite{weiss2019wisdm19}} & \rotatebox{90}{WISDM-19 (W)~\cite{weiss2019wisdm19}} & \rotatebox{90}{Hang-Time~\cite{hoelzemann2023hangtime}} & \rotatebox{90}{PAMAP2~\cite{reiss2012pamap2}} & \rotatebox{90}{OPPO~\cite{roggen2010opportunity}} & \rotatebox{90}{HHAR~\cite{stisen2015hhar}} & \rotatebox{90}{MHEALTH~\cite{banos2014mhealthdroid}} & \rotatebox{90}{DSADS~\cite{altun2010dsads}} & \rotatebox{90}{SAD~\cite{shoaib2014sad}} & \rotatebox{90}{Daphnet~\cite{moore2008daphnet}} & \rotatebox{90}{UP-Fall~\cite{martinez2019upfall}} & \rotatebox{90}{UMAFall~\cite{casilari2017umafall}} & \rotatebox{90}{REALDISP~\cite{banos2014realdisp}} & \rotatebox{90}{HuGaDB~\cite{chereshnev2017hugadb}} & \rotatebox{90}{HARTH~\cite{logacjov2021harth}} & \rotatebox{90}{w-HAR~\cite{bhat2020whar}} & \rotatebox{90}{WEAR~\cite{bock2024wear}} & \rotatebox{90}{HAR70+~\cite{bach2022har70}} & \rotatebox{90}{UCA-EHAR~\cite{novac2022ucaehar}} & \rotatebox{90}{GOTOV~\cite{paraschiakos2022gotov}} & \rotatebox{90}{AICOS-HAR~\cite{barandas2026human}} & \rotatebox{90}{\textbf{Reported}}  \\
        \midrule
        CNN-HAR~\cite{yang2015deep} & 2015 &  &  &  &  &  &  &  &  &  &  &  &  & $\bullet$ &  &  &  &  &  &  &  &  &  &  &  &  &  &  &  &  & 1 \\
        DeepConvLSTM~\cite{ordonez2016deep} & 2016 &  &  &  &  &  &  &  &  &  &  &  &  & $\bullet$ &  &  &  &  &  &  &  &  &  &  &  &  &  &  &  &  & 1 \\
        Guan-LSTM~\cite{guan2017ensembles} & 2017 &  &  &  &  &  &  &  &  &  &  &  & $\bullet$ & $\bullet$ &  &  &  &  &  &  &  &  &  &  &  &  &  &  &  &  & 2 \\
        DeepConvLSTM-Attn~\cite{murahari2018attention} & 2018 &  &  &  &  &  &  &  &  &  &  &  & $\bullet$ & $\bullet$ &  &  &  &  &  &  &  &  &  &  &  &  &  &  &  &  & 2 \\
        SA-HAR~\cite{mahmud2020human} & 2020 &  &  &  &  & $\bullet$ &  &  &  &  &  &  & $\bullet$ & $\bullet$ &  &  &  &  &  &  &  &  &  &  &  &  &  &  &  &  & 3 \\
        Attend-and-Discriminate~\cite{abedin2021attend} & 2021 &  &  &  &  &  &  &  &  &  &  &  & $\bullet$ & $\bullet$ &  &  &  &  &  &  &  &  &  &  &  &  &  &  &  &  & 2 \\
        DANA~\cite{malekzadeh2021dana} & 2021 &  & $\bullet$ &  &  &  &  & $\bullet$ &  &  &  &  &  &  &  &  &  &  &  &  &  &  &  &  &  &  &  &  &  &  & 2 \\
        DeepConvShallowLSTM~\cite{bock2021improving} & 2021 &  &  &  & $\bullet$ &  &  &  &  &  &  &  &  & $\bullet$ & $\bullet$ &  &  &  &  &  &  &  &  &  &  &  &  &  &  &  & 3 \\
        DynamicWHAR~\cite{miao2022towards} & 2022 &  &  &  &  &  &  &  &  &  &  &  &  & $\bullet$ &  &  &  &  &  &  &  &  &  &  &  &  &  &  &  &  & 1 \\
        TinyHAR~\cite{zhou2022tinyhar} & 2022 & $\bullet$ &  &  &  &  &  &  &  &  &  &  & $\bullet$ & $\bullet$ &  &  & $\bullet$ &  & $\bullet$ &  &  &  &  &  &  &  &  &  &  &  & 5 \\
        Triple-Cross-Attn~\cite{tang2022triple} & 2022 & $\bullet$ & $\bullet$ &  &  &  & $\bullet$ &  &  &  &  &  & $\bullet$ &  &  &  &  &  &  &  &  &  &  &  &  &  &  &  &  &  & 4 \\
        MLP-Mixer~\cite{ojiako2023mlps} & 2023 &  &  &  &  &  &  &  &  &  &  &  & $\bullet$ & $\bullet$ &  &  &  &  & $\bullet$ &  &  &  &  &  &  &  &  &  &  &  & 3 \\
        MLP-HAR~\cite{zhou2024mlp} & 2024 &  &  &  & $\bullet$ &  &  & $\bullet$ &  &  &  &  & $\bullet$ &  &  & $\bullet$ & $\bullet$ &  & $\bullet$ &  &  &  &  &  &  &  &  &  &  &  & 6 \\
        TinierHAR~\cite{bian2025tinierhar} & 2025 & $\bullet$ & $\bullet$ &  & $\bullet$ & $\bullet$ &  & $\bullet$ &  &  &  &  & $\bullet$ & $\bullet$ &  & $\bullet$ & $\bullet$ & $\bullet$ & $\bullet$ &  &  &  &  &  &  &  &  &  &  &  & 11 \\
        LITEWAY~\cite{nshimyimana2026liteway} & 2026 &  & $\bullet$ &  & $\bullet$ & $\bullet$ &  & $\bullet$ &  &  &  &  & $\bullet$ & $\bullet$ &  & $\bullet$ & $\bullet$ & $\bullet$ & $\bullet$ &  &  & $\bullet$ &  &  &  & $\bullet$ &  &  &  &  & 12 \\
        LITEWAY-Light~\cite{nshimyimana2026liteway} & 2026 &  & $\bullet$ &  & $\bullet$ & $\bullet$ &  & $\bullet$ &  &  &  &  & $\bullet$ & $\bullet$ &  & $\bullet$ & $\bullet$ & $\bullet$ & $\bullet$ &  &  & $\bullet$ &  &  &  & $\bullet$ &  &  &  &  & 12 \\
        \midrule
        \emph{Reported by} & & 3 & 5 &  & 5 & 4 & 1 & 5 &  &  &  &  & 11 & 13 & 1 & 4 & 5 & 3 & 6 &  &  & 2 &  &  &  & 2 &  &  &  &  &  \\
        \bottomrule
    \end{tabular*}
    \caption{Datasets of the benchmark suite reported in the original
    publication of each architecture. Marked datasets form the \emph{reported} datasets of an architecture, and all others its
    \emph{unseen} datasets. The last column counts the reported datasets per architecture, and the
    last row the architectures that report each dataset.}
    \label{fig:dataset-usage}
\end{figure*}

\subsection{Training and Evaluation Protocol}
\label{sec:method-protocol}

\begin{table*}[b]
    \centering
    \caption{Input configuration and size of every dataset after
    preprocessing. Subj.: subjects. Ch.: channels. Cls.: activity classes
    ($^\dagger$: null or unknown segments removed or classes with fewer than ten windows excluded from the evaluation). Native: sampling rate of
    the recordings (var.: device-dependent). Rate and time steps are given per window length after resampling. Windows: number of windows with 50\% overlap. Imb.: ratio between the most and the least frequent class for 3\,s windows.}
    \label{tab:dataset-stats}
    \footnotesize
    \setlength{\tabcolsep}{3pt}
    \begin{tabular*}{\linewidth}{@{\extracolsep{\fill}}l rrr r rr rr rr r@{}}
\toprule
 & & & & & \multicolumn{2}{c}{Rate (Hz)} & \multicolumn{2}{c}{Time steps} & \multicolumn{2}{c}{Windows} & \\
\cmidrule(lr){6-7} \cmidrule(lr){8-9} \cmidrule(lr){10-11}
Dataset & Subj. & Ch. & Cls. & Native & 2\,s & 3\,s & 2\,s & 3\,s & 2\,s & 3\,s & Imb. \\
\midrule
WISDM~\cite{kwapisz2011wisdm} & 36 & 3 & 6 & 20 & 25 & 20 & 50 & 60 & 50\,892 & 33\,724 & 8.8 \\
UCI-HAR~\cite{anguita2013public} & 30 & 9 & 6 & 50 & 50 & 50 & 100 & 150 & 12\,596 & 8\,188 & 1.5 \\
UTD-MHAD~\cite{chen2015utdmhad} & 8 & 6 & 27 & 50 & 50 & 50 & 100 & 150 & 1\,826 & 796 & 3.2 \\
HAPT~\cite{reyesortiz2016hapt} & 30 & 6 & 12 & 50 & 50 & 50 & 100 & 150 & 14\,516 & 9\,070 & 103.8 \\
USC-HAD~\cite{zhang2012uschad} & 14 & 6 & 12 & 100 & 100 & 100 & 200 & 300 & 27\,205 & 17\,604 & 4.0 \\
UniMiB-SHAR~\cite{micucci2017unimib} & 30 & 3 & 17 & 50 & 50 & 50 & 100 & 150 & 12\,153 & 7\,207 & 5.7 \\
MotionSense~\cite{malekzadeh2019motionsense} & 24 & 18 & 6 & 50 & 50 & 50 & 100 & 150 & 27\,720 & 18\,307 & 2.7 \\
RealLifeHAR~\cite{garcia2020reallifehar} & 17 & 15 & 4 & 20 & 25 & 20 & 50 & 60 & 568\,054 & 378\,494 & 2.5 \\
WISDM-19 (Phone)~\cite{weiss2019wisdm19} & 51 & 6 & 18 & 20 & 25 & 20 & 50 & 60 & 233\,808 & 155\,562 & 1.2 \\
WISDM-19 (Watch)~\cite{weiss2019wisdm19} & 51 & 6 & 18 & 20 & 25 & 20 & 50 & 60 & 182\,155 & 121\,156 & 1.1 \\
Hang-Time~\cite{hoelzemann2023hangtime} & 24 & 3 & 4$^\dagger$ & 50 & 50 & 50 & 100 & 150 & 83\,187 & 51\,812 & 4.6 \\
PAMAP2~\cite{reiss2012pamap2} & 9 & 52 & 12$^\dagger$ & 100 & 100 & 100 & 200 & 300 & 19\,266 & 12\,786 & 4.9 \\
Opportunity (Loc.)~\cite{roggen2010opportunity} & 4 & 133 & 4$^\dagger$ & 30 & 30 & 30 & 60 & 90 & 18\,280 & 10\,642 & 10.6 \\
HHAR~\cite{stisen2015hhar} & 9 & 6 & 6 & var. & 25 & 50 & 50 & 150 & 72\,290 & 45\,654 & 1.7 \\
MHEALTH~\cite{banos2014mhealthdroid} & 10 & 23 & 12$^\dagger$ & 50 & 50 & 50 & 100 & 150 & 6\,684 & 4\,348 & 3.2 \\
DSADS~\cite{altun2010dsads} & 8 & 45 & 19 & 25 & 25 & 25 & 50 & 75 & 36\,480 & 18\,240 & 1.0 \\
SAD~\cite{shoaib2014sad} & 10 & 60 & 7 & 50 & 50 & 50 & 100 & 150 & 12\,530 & 8\,330 & 1.0 \\
Daphnet~\cite{moore2008daphnet} & 10 & 9 & 2$^\dagger$ & 64 & 64 & 64 & 128 & 192 & 17\,090 & 11\,165 & 12.4 \\
UP-Fall~\cite{martinez2019upfall} & 17 & 42 & 11 & 20 & 25 & 20 & 50 & 60 & 13\,883 & 9\,006 & 8.0 \\
UMAFall~\cite{casilari2017umafall} & 19 & 39 & 15 & 200 & 200 & 200 & 400 & 600 & 9\,973 & 6\,240 & 3.4 \\
REALDISP~\cite{banos2014realdisp} & 17 & 117 & 33$^\dagger$ & 50 & 50 & 50 & 100 & 150 & 38\,600 & 24\,986 & 10.4 \\
HuGaDB~\cite{chereshnev2017hugadb} & 18 & 38 & 10$^\dagger$ & 60 & 60 & 60 & 120 & 180 & 29\,152 & 17\,746 & 9.9 \\
HARTH~\cite{logacjov2021harth} & 22 & 6 & 12 & 50 & 50 & 50 & 100 & 150 & 116\,007 & 75\,080 & 449.0 \\
w-HAR~\cite{bhat2020whar} & 22 & 7 & 8$^\dagger$ & 250 & 250 & 250 & 500 & 750 & 8\,025 & 4\,919 & 22.8 \\
WEAR~\cite{bock2024wear} & 24 & 12 & 18 & 50 & 50 & 50 & 100 & 150 & 40\,443 & 26\,748 & 1.2 \\
HAR70+~\cite{bach2022har70} & 18 & 6 & 7 & 50 & 50 & 50 & 100 & 150 & 40\,989 & 26\,287 & 387.3 \\
UCA-EHAR~\cite{novac2022ucaehar} & 20 & 7 & 11$^\dagger$ & 26 & 26 & 26 & 52 & 78 & 25\,411 & 15\,922 & 173.2 \\
GOTOV~\cite{paraschiakos2022gotov} & 30 & 28 & 16 & 85.7 & 85.7 & 85.7 & 171 & 257 & 69\,208 & 46\,298 & 24.0 \\
AICOS-HAR~\cite{barandas2026human} & 94 & 10 & 11 & var. & 25 & 50 & 50 & 150 & 215\,205 & 141\,613 & 21.2 \\
\bottomrule
\end{tabular*}

\end{table*}

All architectures are trained and evaluated on each dataset under one fixed protocol.

\paragraph{Data preparation}
All datasets are segmented into windows of 2 and 3\,s with 50\% overlap.
The original publications of the architectures use windows from about 0.5 to 10\,s, mostly between 1 and 5\,s, without a common choice \cite{bock2021improving,tang2022triple,zhou2022tinyhar,bian2025tinierhar}, so we use two lengths within this range that give every dataset enough time steps.
We vary the window length but not the overlap, since the overlap leaves the input shape and thus the cost per inference unchanged, and the gains attributed to overlapping windows disappear under subject-independent evaluation \cite{dehghani2019quantitative}.
Datasets are used at their original sampling rate and upsampled only where necessary, so that every window comprises at least 50 time steps, which several architectures require because they shrink the temporal dimension through unpadded convolutions and pooling.
Datasets recorded below 25\,Hz are therefore upsampled to 25\,Hz for 2\,s windows, and architectures that assumed a fixed input length were adapted to variable lengths (\autoref{sec:method-models}).
Only HHAR and AICOS-HAR, whose devices record at different rates, are brought to a common rate of 50\,Hz (25\,Hz for 2\,s windows).
Activity labels are used as released, without merging or relabeling.
Segments labeled as null or unknown are removed, and activities with too few samples, that is, fewer than ten windows of 3\,s in a dataset, such as jumping in Hang-Time \cite{hoelzemann2023hangtime}, are excluded from the evaluation.
\autoref{tab:dataset-stats} lists the resulting sampling rates, time steps, and numbers of windows, which range from 796 to 378\,494 for 3\,s windows.

\paragraph{Subject-independent splits}
All datasets are evaluated by leaving out subject groups: the subjects are partitioned into at most ten disjoint groups of equal size (\autoref{sec:method-library}), and each group forms the test set once while the architecture is trained on the remaining subjects.
For datasets with at most ten subjects, the protocol thus coincides with leave-one-subject-out evaluation, and every subject is tested exactly once on folds identical for all architectures.
Bounding the number of folds keeps the benchmark feasible: leaving out every subject individually would require \TrainingLosoFitsTotal{} instead of \TrainingFitsTotal{} model training runs, whereas the training data would grow only slightly, since each fold already uses at least 86\,\% of the subjects of every dataset with more than ten subjects, compared with at least 93\,\%.
Within each fold, 20\% of the training windows serve as validation data for early stopping.
They are drawn within the sessions of the training subjects rather than from further held-out subjects, which keeps all training subjects available for training.

\paragraph{Training}
All deep architectures are trained with the same configuration: the AdamW optimizer \cite{loshchilov2017decoupled} with an initial learning rate of $10^{-3}$, cosine annealing, and a weight decay of $10^{-4}$, a batch size of 64, and at most 100 epochs.
Training stops if the validation macro-F1 does not improve for 10 epochs, and the best validation checkpoint is restored.
A cross-entropy loss weighted by the inverse class frequencies of the training data mitigates class imbalance.
Architecture-internal hyperparameters follow the defaults of the original implementations, and each configuration is trained with one random seed.
The classical baselines compute the minimal feature set of tsfresh per channel of the normalized windows and use scikit-learn implementations of $k$-nearest neighbors with $k=5$ and distance weighting, a random forest with 100 trees and a maximum depth of 8, and a support vector machine with a radial basis function kernel \cite{christ2018time,pedregosa2011scikit}.

\paragraph{Evaluation}
Training and evaluation settings were not adapted to individual architectures or datasets, and in contrast to the development setting described in \autoref{sec:related-overfitting}, no architecture or training setting is selected on the folds of HARENA.
A uniform configuration favors comparability over the optimization of individual architectures: it holds influential training choices constant but does not reproduce the architecture-specific tuning of the original publications \cite{huang2023standardizing}.
Recognition performance is measured by the test macro-F1, the macro-F1 score on the test subjects of each fold, which weights all activity classes equally.
Classes that do not occur among the test subjects of a fold are left out of the average, and we report the mean and standard deviation over folds.

\subsection{Deployment Efficiency}
\label{sec:method-efficiency}

Following \autoref{sec:related-efficiency}, HARENA characterizes deployment cost through measured quantities.
\emph{Inference latency} is the time required to classify one window.
With 50\,\% overlap, a new window arrives after half the window length, so recognition keeps pace with the incoming data if the latency stays below this stride, the \emph{real-time budget} of 1\,s for 2\,s and \BudgetS\,s for 3\,s windows.
\emph{Peak memory overhead} captures the intermediate activations that dominate memory use on embedded devices.
On the microcontroller, we also measure the \emph{flash footprint} of the deployed binary, which determines whether an architecture can be stored at all, and the \emph{energy per inference}, which determines the battery cost of continuous recognition.
Parameter and operation counts are reported as supplementary measures, not as substitutes.

\begin{table}[t]
    \centering
    \caption{Deployment platforms and measurement protocol.
    Latency and memory are measured for every architecture, dataset, and window length on each platform.
    PSS: proportional set size of the benchmark process.}
    \label{tab:platforms}
    \footnotesize
    \begin{tabular*}{\linewidth}{@{\extracolsep{\fill}}l p{0.235\linewidth} p{0.235\linewidth} p{0.335\linewidth}@{}}
\toprule
 & \textbf{Phone} & \textbf{Watch} & \textbf{MCU} \\
\midrule
\multicolumn{4}{@{}l}{\textit{Hardware}} \\
\quad Device & Samsung Galaxy S24 Ultra & Samsung Galaxy Watch5 Pro & Nordic nRF5340 DK \\
\quad Processor & Snapdragon 8 Gen 3 (SM8650) & Exynos W920 & Arm Cortex-M33, 128\,MHz \\
\quad Memory & 12\,GB RAM & 1.5\,GB RAM & 512\,KB RAM (500\,KB usable), 1\,MB flash \\
\quad OS & Android 17 & Wear OS 6 (One UI 8 Watch) & None (bare metal) \\
\midrule
\multicolumn{4}{@{}l}{\textit{Neural architectures}} \\
\quad Runtime & \multicolumn{2}{c}{PyTorch Mobile 2.1 (lite interpreter)} & TensorFlow Lite Micro, CMSIS-NN \\
\quad Precision & \multicolumn{2}{c}{FP32} & INT8 weights and activations \\
\midrule
\multicolumn{4}{@{}l}{\textit{Classical baselines}} \\
\quad Runtime & \multicolumn{2}{c}{ONNX Runtime 1.24.3} & Native C++ \\
\quad Precision & \multicolumn{2}{c}{FP32} & FP32 \\
\midrule
\multicolumn{4}{@{}l}{\textit{Measurement}} \\
\quad Runs & \multicolumn{2}{c}{1 warm-up, 50 timed} & 1 warm-up, 1 timed \\
\quad Latency & \multicolumn{2}{c}{Wall-clock time per forward pass} & CPU cycle counter per invocation \\
\quad Memory & \multicolumn{2}{c}{Peak PSS increase} & Peak RAM (static, heap, arena, stack) \\
\quad Energy & \multicolumn{2}{c}{--} & Power Profiler Kit II (3.0\,V, 100\,kHz) \\
\bottomrule
\end{tabular*}

\end{table}
Since these costs depend on the memory architecture, runtime, and numerical support of the hardware, we measure them on a smartphone, a smartwatch, and a microcontroller (\autoref{tab:platforms}).
On the phone and the watch, classical baselines run with ONNX Runtime, with the tsfresh feature extraction reimplemented in PyTorch and exported to ONNX.
On the microcontroller, neural architectures are statically quantized to 8-bit integers after training, as is common for this device class \cite{david2021tensorflow,lai2018cmsis}, and operators without integer support remain in floating point.
Since these exports only serve to measure cost, they are calibrated with 16 synthetic windows, and recognition performance is evaluated on the floating-point models (\autoref{sec:method-protocol}).
Classical baselines run there as native C++ code without an inference runtime, computing the tsfresh features on the device, with the trained classifiers stored in flash as constant arrays.

Since window length and number of channels differ between datasets, every architecture is exported and measured separately for every dataset and window length.
Neural architectures are exported with random weights, since their computation and memory requirements depend only on the architecture and input shape.
Classical baselines are exported after training, since their size and cost depend on the training data, for example through the number and depth of trees or the stored training samples.
On the phone and the watch, latency is the mean over 50 runs, and memory is the peak increase in proportional set size relative to the state before the model is loaded.
On the microcontroller, a single run suffices, since execution without an operating system is deterministic, and memory is the peak RAM use.
We record for which datasets and window lengths an architecture fits into its RAM and flash.
All quantities are specific to the device and runtime but directly comparable across architectures on each device.

\section{Evaluation}
\label{sec:results}

We analyze the measurements descriptively and define each metric together with its results.
We first characterize the datasets (\autoref{sec:results-datasets}) and the drop in recognition performance on unseen subjects (\autoref{sec:results-generalization}).
We then compare the architectures (\autoref{sec:results-overall}, \textbf{RQ1}) and examine how consistently their rankings hold between disjoint subsets of datasets (\autoref{sec:results-stability}), with the number of channels and activities (\autoref{sec:results-consistency}), and between reported and unseen datasets (\autoref{sec:results-unseen}, \textbf{RQ3}).
Finally, we compare the deployment cost of the architectures (\autoref{sec:results-cost}) and relate it to their recognition performance through Pareto fronts (\autoref{sec:results-tradeoffs}, \textbf{RQ2}).
Every analysis covers 2\,s and 3\,s windows, and where the window length does not change the conclusions, we pool both or report one and state which.

\begin{figure*}[t]
    \centering
    \includegraphics[width=\textwidth]{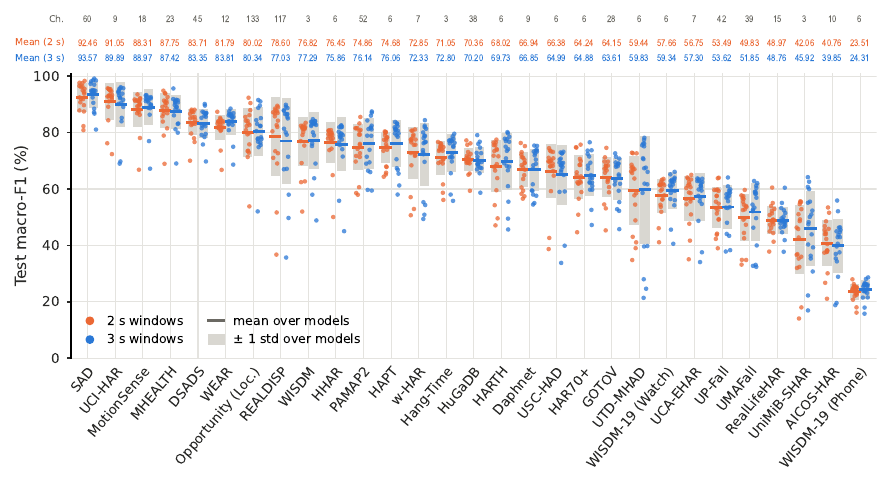}
    \caption{Test macro-F1 of all architectures on each dataset for 2\,s (orange, left bar) and 3\,s windows (blue, right bar), sorted by the mean of the two means over the architectures.
    Each dot is one architecture, the tick in the window color marks the mean over the architectures, and the grey bar spans one standard deviation around it.
    Above each column, the rows give the number of sensor channels and the mean test macro-F1 over the architectures, one line per window size in its color.}
    \label{fig:dataset-difficulty}
\end{figure*}

\subsection{Dataset Difficulty}
\label{sec:results-datasets}

The datasets differ strongly in difficulty (\autoref{fig:dataset-difficulty}).
Pooled over both window sizes, the test macro-F1 over the architectures (mean $\pm$ standard deviation) ranges from $\DsMeanHardestValue \pm \DsMeanHardestStd$\,\% on \DsMeanHardest{} \cite{weiss2019wisdm19} to $\DsMeanEasiestValue \pm \DsMeanEasiestStd$\,\% on \DsMeanEasiest{} \cite{shoaib2014sad}.
\DsNumMeanHigh{} datasets have a mean above \DsHighMedian\,\% and are close to saturation, while \DsNumMeanLow{} remain below \DsLowMedian\,\%.
The choice of architecture also matters to a different degree. The standard deviation over the architectures is $\DsStdMean \pm \DsStdStd$ points on average, but it is much larger on \DsStdLargestOne{} \cite{chen2015utdmhad} ($\DsStdLargestOneMean \pm \DsStdLargestOneValue$\,\%) and \DsStdLargestTwo{} \cite{banos2014realdisp} ($\DsStdLargestTwoMean \pm \DsStdLargestTwoValue$\,\%).
The window size changes the difficulty of most datasets only slightly: on \DsWinNumNear{} of the \DsNumDatasets{} datasets, the mean test macro-F1 for 3\,s windows lies within \GapNear{} points of that for 2\,s windows.

\subsection{Generalization to Unseen Subjects}
\label{sec:results-generalization}

\begin{figure*}[b]
    \centering
    \includegraphics[width=\textwidth]{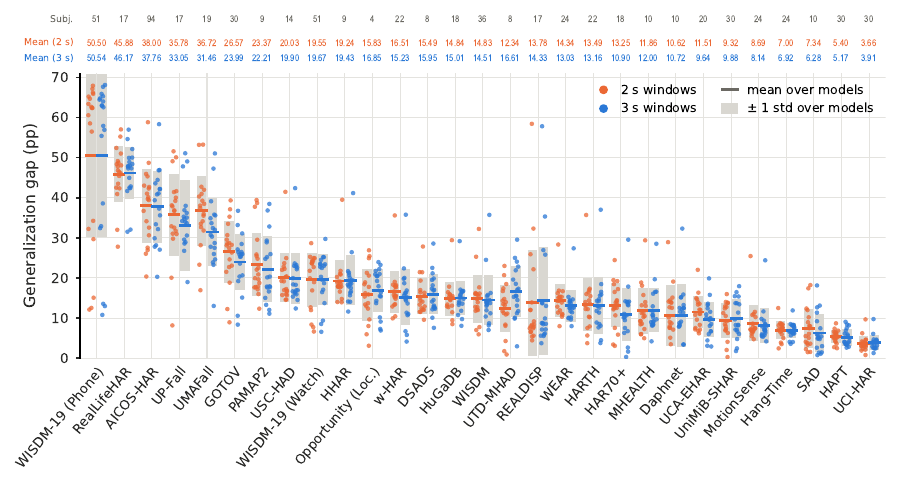}
    \caption{Generalization gap of each dataset: validation minus test macro-F1 in percentage points for 2\,s (orange) and 3\,s (blue) windows.
    Validation windows come from the training subjects, test windows from the held-out subjects, so the gap measures the drop from seen to unseen subjects.
    Each dot is one architecture. The tick marks the mean and the grey bar one standard deviation over the architectures.
    Datasets are sorted by the mean gap over both window sizes. The rows above the columns give the number of subjects and the mean gap, one line per window size in its color.
    The y-axis is cut at 0, which hides three runs with a negative gap (SVM on Daphnet for both window sizes, Random Forest on UP-Fall for 3\,s).}
    \label{fig:generalization-gap}
\end{figure*}

One source of this difficulty is the shift to new subjects. We measure the drop in recognition performance on unseen subjects as the generalization gap (\autoref{fig:generalization-gap}).
For architecture $m$ on dataset $d$, with $V_{m,d}$ the validation macro-F1 on the training subjects and $F_{m,d}$ the test macro-F1 on the held-out subjects, both averaged over the folds, the gap is
\begin{equation}
    \label{eq:generalization-gap}
    G_{m,d} = V_{m,d} - F_{m,d}.
\end{equation}
Training windows that overlap a validation window are removed (\autoref{sec:method-library}), so the gap stems from the change of subjects rather than from leakage.
The validation score also selects the early-stopping checkpoint, which makes it slightly optimistic, but in the same way for every dataset and architecture.

The gap is large. For 3\,s windows, its median over the datasets is \GenGapMedian{} points, about three times the median spread of the middle half of the architectures on a dataset (\InsArchSpread{} points).
It is largest on \GenGapMaxDataset{} (\GenGapMax{} points), \GenGapTopTwoDataset{} \cite{garcia2020reallifehar} (\GenGapTopTwo), and \GenGapTopThreeDataset{} (\GenGapTopThree), and smallest on \GenGapMinDataset{} (\GenGapMin).
The datasets with small gaps were recorded in the laboratory with scripted activities and a fixed device placement, whereas RealLifeHAR was recorded during the daily life of its participants and AICOS-HAR spans several devices and body positions \cite{garcia2020reallifehar,barandas2026human}.
The gap is a property of the dataset rather than of the windowing or the architecture: the gaps for 2\,s and 3\,s windows are almost identical (Spearman's $\rho = \GenGapRhoWindows$), and the architectures on a dataset differ from its mean gap by only \GenGapModelStd{} points on average.

One would expect the gap to shrink with the number of subjects, since more training subjects cover more of the variation between people.
It does not. Across datasets with \GenGapSubjectsMin{} to \GenGapSubjectsMax{} subjects, the gap is unrelated to their number ($\rho = \GenGapRhoSubjects$), and \GenGapMaxDataset{} with \GenGapMaxSubjects{} subjects and \GenGapTopThreeDataset{} with \GenGapTopThreeSubjects{} subjects have two of the three largest gaps.
The number of sensor channels is likewise nearly unrelated to the gap ($\rho = \GenGapRhoChannels$, and $\rho = \GenGapRhoChannelsPartial$ with the number of subjects held fixed), although datasets with more channels reach both a higher validation ($\rho = \GenGapRhoChannelsValFOne$) and a higher test macro-F1 ($\rho = \GenGapRhoChannelsTestFOne$).
Neither more subjects under the same conditions nor a richer sensor setup thus closes the gap. What matters is how varied the recording conditions are.

Overall, much of the difference in difficulty between the datasets stems from the shift between subjects ($\rho = \GenGapRhoTestFOne$ between gap and test macro-F1), which exceeds the differences between architectures by far.

\subsection{Architecture Ranking}
\label{sec:results-overall}

\begin{figure*}[b]
    \centering
    \includegraphics[width=\textwidth]{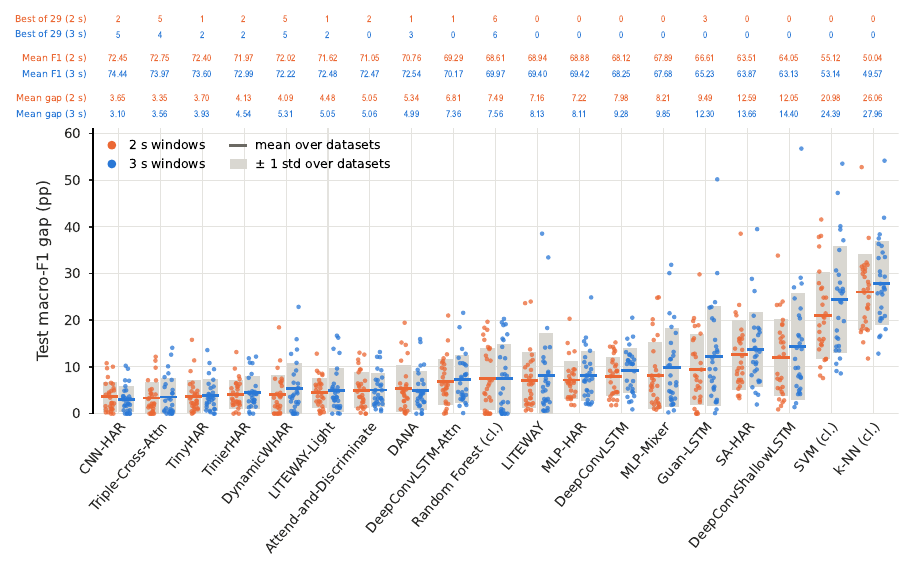}
    \caption{Test macro-F1 gap of each architecture for 2\,s (orange, left bar) and 3\,s windows (blue, right bar): the test macro-F1 of the best architecture on a dataset minus that of the given architecture, in percentage points (0 = best on that dataset).
    Each dot is one dataset, the tick in the window color marks the mean gap over the datasets, and the grey bar spans one standard deviation around it.
    Columns are sorted by the mean test macro-F1 over both window sizes. The rows above give the number of datasets on which the architecture is best, its mean test macro-F1, and its mean gap, one line per window size in its color.
    Classical baselines are marked with (cl.).}
    \label{fig:model-gap}
\end{figure*}

Having characterized the datasets, we now turn to the architectures and ask which of them perform best across the suite and by how much they differ from one another.

\paragraph{Mean rank}
We rank the architectures on each dataset by test macro-F1 (1 = best) and average their ranks over the \RankNumDatasets{} datasets (\autoref{fig:rank-heatmap}).
\RankFirstModel{} achieves the best mean rank (\ModelMeanRankCNNHAR) and the highest mean test macro-F1 (\ModelMeanFOneCNNHAR\,\%), closely followed by \RankSecondModel{} (\ModelMeanRankTripleCrossAttn, \ModelMeanFOneTripleCrossAttn\,\%) and \RankThirdModel{} (\ModelMeanRankTinyHAR, \ModelMeanFOneTinyHAR\,\%).
The leading group is broad and close together. \LeadNum{} architectures (\LeadModels) reach a mean test macro-F1 within \LeadThreshold{} points of the best, and their mean test macro-F1 spans only \LeadFOneSpread{} points.
The compact TinyHAR, TinierHAR, and LITEWAY-Light belong to this group, so a high recognition performance does not require a large model.
Its order is not stable across datasets: \RankSecondModel{} outperforms \RankFirstModel{} on \ModelBeatsFirstTripleCrossAttn{} of the \RankNumDatasets{} datasets, and \RankThirdModel{} on \ModelBeatsFirstTinyHAR.
Random Forest follows with a mean rank of \ModelMeanRankRandomForest{} (\ModelMeanFOneRandomForest\,\%), yet it outperforms \RankFirstModel{} on \ModelBeatsFirstRandomForest{} datasets and ranks among the three best on \ModelTopThreeRandomForest, so a classical model with hand-crafted features is competitive with most deep architectures on many datasets.
SVM and k-NN form the bottom of the ranking and do not outperform \RankFirstModel{} on any dataset.

\paragraph{Gap to the best architecture}
Since ranks do not show how large the differences are, we also measure how many points of test macro-F1 an architecture is behind the best architecture on each dataset (\autoref{fig:model-gap}). Over the set of architectures $\mathcal{A}$, this gap is
\begin{equation}
    \label{eq:gap-to-best}
    \Delta_{m,d} = \max_{m' \in \mathcal{A}} F_{m',d} - F_{m,d}.
\end{equation}
\RankFirstModel{} is the most reliable architecture. It has the smallest gap of all architectures ($\ModelMeanGapCNNHAR \pm \ModelStdGapCNNHAR$ points, mean $\pm$ standard deviation over the datasets), is best on \ModelWinsCNNHAR{} datasets, lies within \GapClose{} points of the best on \ModelWithinCloseCNNHAR{} of the \RankNumDatasets{} datasets, and is never more than \ModelMaxGapCNNHAR{} points behind.
Random Forest is best on \ModelWinsRandomForest{} datasets, but its gap is larger and varies much more ($\ModelMeanGapRandomForest \pm \ModelStdGapRandomForest$ points), since it is either among the best architectures or far behind depending on the number of sensor channels (\autoref{sec:results-consistency}).
SVM and k-NN come within \GapClose{} points of the best on no dataset.

Overall, the top of the ranking is formed by a broad group of architectures that differ by only a few points on average and whose order changes from dataset to dataset, so results on one or a few datasets do not suffice to declare one of them the best.

\subsection{Stability of the Ranking}
\label{sec:results-stability}

\begin{figure}[b]
  \centering
  \includegraphics[width=0.59\textwidth]{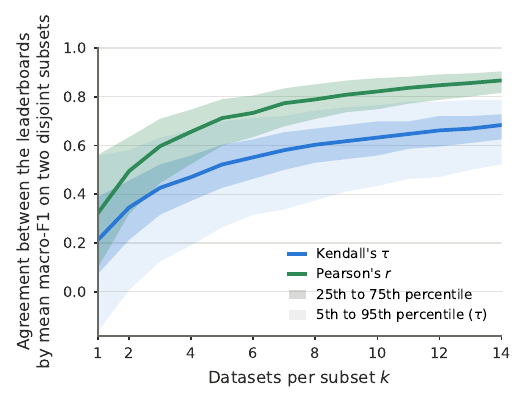}
  \caption{Agreement between two disjoint, randomly drawn subsets of $k$ datasets each, out of \SubsetNumDatasets{} datasets: Kendall's $\tau$ between the rankings (blue) and Pearson's $r$ between the mean test macro-F1 scores (green) of the \SubsetNumModels{} architectures within \GapClose{} points of the best on at least one dataset.
  Both measures are averaged over 2\,s and 3\,s windows, whose curves are nearly identical.
  Lines give the median over \SubsetNumDraws{} draws per subset size, dark bands the 25th to 75th percentile, and the light band the \SubsetLowPercent{}th to \SubsetHighPercent{}th percentile (for $\tau$ only).}
  \label{fig:rank-stability}
\end{figure}

The order of the architectures depends on the dataset. \DsNumModelsWithWin{} different architectures achieve the highest test macro-F1 on at least one dataset, and the per-dataset rankings agree only moderately (Kendall's $W = \KendallW$ \cite{kendall1939problem}, \autoref{fig:rank-heatmap}).
To test how far a ranking carries over to other datasets, we repeatedly draw two disjoint subsets of datasets of equal size, rank the architectures by their mean test macro-F1 $\bar{F}_{m,S}$ on each subset $S$, and compare the two rankings (\autoref{fig:rank-stability}).
We use the \SubsetNumModels{} architectures that come within \GapClose{} points of the best on at least one dataset, since the low scores of \SubsetExcludedModels{} on both subsets would inflate the agreement.
We measure agreement with Kendall's $\tau$: over all pairs of the $M$ architectures, the share of pairs ordered the same way in both rankings minus the share ordered differently,
\begin{equation}
    \label{eq:kendall-tau}
    \tau = \frac{2}{M(M-1)} \sum_{m < m'} \operatorname{sgn}\bigl(\bar{F}_{m,S_1} - \bar{F}_{m',S_1}\bigr) \operatorname{sgn}\bigl(\bar{F}_{m,S_2} - \bar{F}_{m',S_2}\bigr).
\end{equation}
Since $\tau$ counts a swapped pair fully even if the two scores differ by only a fraction of a point, we also report Pearson's $r$ between the mean scores on the two subsets, which takes the size of the differences into account.

Two single datasets agree with a median $\tau$ of only \SubsetMeanTauOne{} ($r = \SubsetMeanROne$), and in \SubsetLowPercent\,\% of the draws $\tau$ is below \SubsetMeanTauOneLow, that is, the rankings are opposed.
The median $\tau$ rises to \SubsetMeanTauFew{} for subsets of \SubsetFewK{} datasets, \SubsetMeanTauSome{} for \SubsetSomeK{}, and \SubsetMeanTauMax{} for \SubsetMeanTauMaxK{}, half of the suite each, where $r$ reaches \SubsetMeanRMax.
This still means that about one in four pairs of architectures swaps its order between two subsets of \SubsetFewK{} datasets and about one in six between two subsets of \SubsetMeanTauMaxK.
Claiming that one architecture is better than another therefore requires many datasets: the few that a typical publication reports do not suffice, and even half of the suite does not yield a stable order.

\begin{figure*}[b]
    \centering
    \begin{subfigure}[t]{0.49\textwidth}
        \centering
        \includegraphics[width=\linewidth]{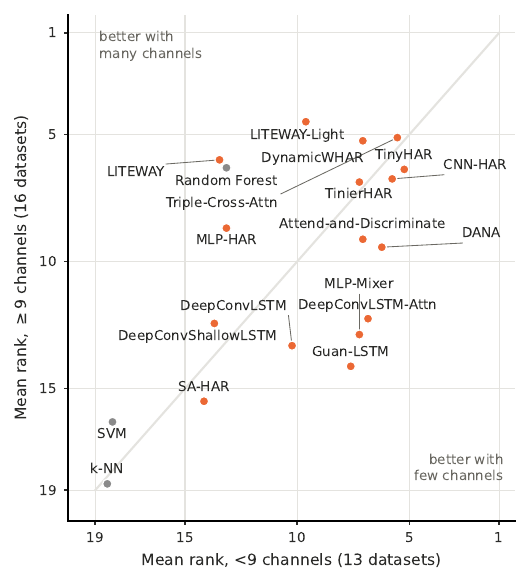}
        \caption{2\,s windows}
        \label{fig:channel-groups-2s}
    \end{subfigure}
    \hfill
    \begin{subfigure}[t]{0.49\textwidth}
        \centering
        \includegraphics[width=\linewidth]{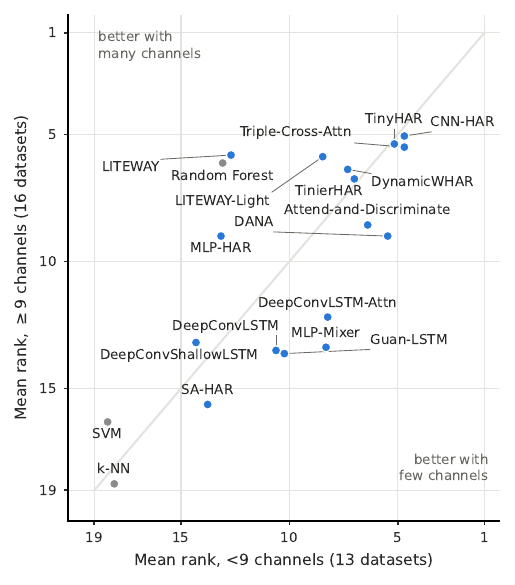}
        \caption{3\,s windows}
        \label{fig:channel-groups-3s}
    \end{subfigure}
    \caption{Mean rank of each architecture on the datasets with fewer than \ChanBorder{} sensor channels (x-axis) and on the datasets with \ChanBorder{} or more channels (y-axis) for 2\,s (a) and 3\,s windows (b) (1 = best). The number of datasets per group is given on the axes.
    Architectures above the diagonal rank better on datasets with many channels, those below it on datasets with few channels.
    Deep architectures are drawn in the window color, classical baselines in grey.}
    \label{fig:channel-groups}
\end{figure*}

\subsection{Architecture Strengths by Dataset Properties}
\label{sec:results-consistency}

Part of this disagreement may follow the properties of the datasets, so we ask which architectures gain or lose with the number of sensor channels and activities by comparing their mean ranks between groups of datasets split at the median.

\paragraph{Number of sensor channels}
We divide the datasets at the median of \ChanBorder{} channels into \ChanNumFew{} datasets with fewer and \ChanNumMany{} with \ChanBorder{} or more channels (\autoref{fig:channel-groups}).
Random Forest moves from rank \ChanRankFewRandomForest{} on datasets with few channels to \ChanRankManyRandomForest{} on datasets with many, close to the best architecture there, \ChanBestMany{} (\ChanRankManyCNNHAR).
LITEWAY (\ChanRankFewLITEWAY{} to \ChanRankManyLITEWAY), MLP-HAR (\ChanRankFewMLPHAR{} to \ChanRankManyMLPHAR), and SVM (\ChanRankFewSVM{} to \ChanRankManySVM) follow the same pattern.
The opposite holds for the recurrent architectures DeepConvLSTM-Attn (\ChanRankFewDeepConvLSTMAttn{} to \ChanRankManyDeepConvLSTMAttn), DeepConvLSTM (\ChanRankFewDeepConvLSTM{} to \ChanRankManyDeepConvLSTM), and DANA (\ChanRankFewDANA{} to \ChanRankManyDANA), as well as for MLP-Mixer (\ChanRankFewMLPMixer{} to \ChanRankManyMLPMixer).
CNN-HAR, Triple-Cross-Attn, and TinyHAR are the only architectures among the five best in both groups, with nearly the same mean rank (\ChanRankFewCNNHAR{} and \ChanRankManyCNNHAR, \ChanRankFewTripleCrossAttn{} and \ChanRankManyTripleCrossAttn, and \ChanRankFewTinyHAR{} and \ChanRankManyTinyHAR).

\begin{figure*}[t]
    \centering
    \begin{subfigure}[t]{0.49\textwidth}
        \centering
        \includegraphics[width=\linewidth]{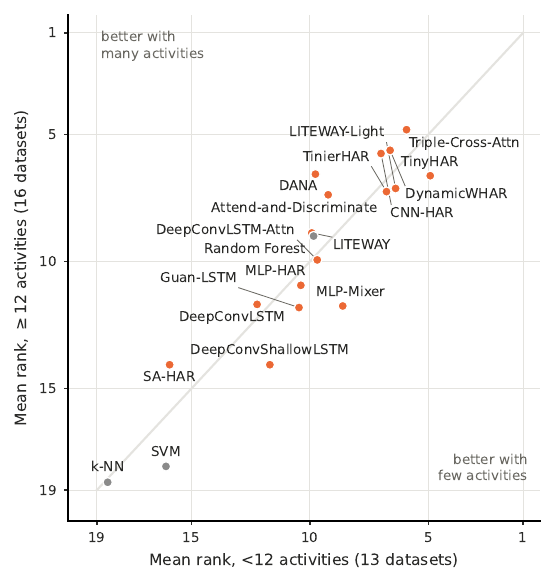}
        \caption{2\,s windows}
        \label{fig:class-groups-2s}
    \end{subfigure}
    \hfill
    \begin{subfigure}[t]{0.49\textwidth}
        \centering
        \includegraphics[width=\linewidth]{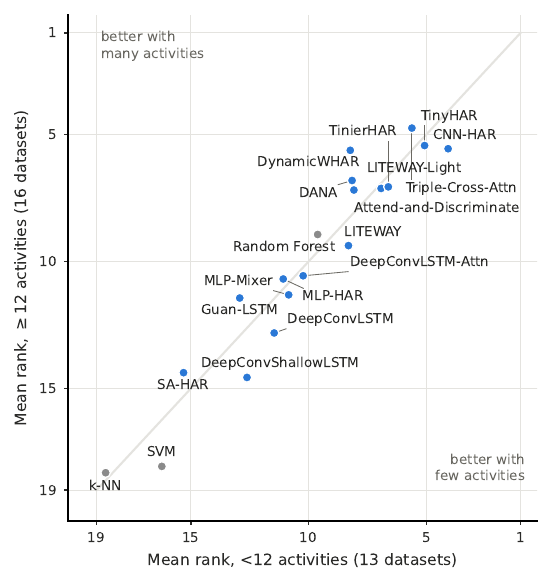}
        \caption{3\,s windows}
        \label{fig:class-groups-3s}
    \end{subfigure}
    \caption{Mean rank of each architecture on the datasets with fewer than \ClassBorder{} activities (x-axis) and on the datasets with \ClassBorder{} or more activities (y-axis) for 2\,s (a) and 3\,s windows (b) (1 = best). The number of datasets per group is given on the axes.
    Architectures above the diagonal rank better on datasets with many activities, those below it on datasets with few activities.
    Deep architectures are drawn in the window color, classical baselines in grey.}
    \label{fig:class-groups}
\end{figure*}

\paragraph{Number of activities}
The activity count is nearly unrelated to the channel count (Spearman's $\rho = \ClassRhoClassesChannels$) and changes the ranks only little.
Split at the median of \ClassBorder{} activities (\autoref{fig:class-groups}), all architectures except DynamicWHAR (\ClassRankFewDynamicWHAR{} to \ClassRankManyDynamicWHAR) change their mean rank by less than two positions.

\paragraph{Architectural components}
The shifts with the channel count follow recurrence rather than attention.
Pooled over both window sizes, the \CompRNNNumWith{} deep architectures with a recurrent layer rank better than the \CompRNNNumWithout{} without one on datasets with few channels (mean rank \CompRNNWithFewChannels{} vs.\ \CompRNNWithoutFewChannels) but worse on datasets with many channels (\CompRNNWithManyChannels{} vs.\ \CompRNNWithoutManyChannels), whereas the deep architectures with attention rank better than those without in both groups (\CompAttnWithFewChannels{} vs.\ \CompAttnWithoutFewChannels{} and \CompAttnWithManyChannels{} vs.\ \CompAttnWithoutManyChannels).
Since each component occurs in only a few architectures and together with other design choices, these comparisons show associations rather than the effect of a component alone.

\paragraph{Consequences for the ranking}
The channel count explains part of the disagreement between the dataset-specific rankings.
The rankings of two datasets from the same group agree better (mean Spearman correlation of \ChanRhoWithinFew{} for few and \ChanRhoWithinMany{} for many channels) than those from different groups (\ChanRhoBetween), and for \PairNumNegative{} of the \PairNumBetween{} pairs from different groups, they are even negatively correlated.
Since datasets with many channels tend to have fewer subjects (Spearman's $\rho = \ChanRhoChannelsSubjects$), the effect could also stem from the amount of training data.
However, it remains when the number of subjects is controlled for: the partial correlation between channel count and rank changes only from $\ChanRhoRandomForest$ to $\ChanPartialRhoRandomForest$ for Random Forest, from $\ChanRhoLITEWAY$ to $\ChanPartialRhoLITEWAY$ for LITEWAY, and from $\ChanRhoDeepConvLSTM$ to $\ChanPartialRhoDeepConvLSTM$ for DeepConvLSTM.
The activity count, in contrast, does not explain the disagreement: two datasets from the same group agree about as well (\ClassRhoWithinFew{} and \ClassRhoWithinMany) as two from different groups (\ClassRhoBetween).

\paragraph{Further properties}
Among further properties, the native sampling rate matters most.
Rankings of datasets recorded at \PropRateBorder\,Hz or more, counting HHAR and AICOS-HAR, which are used at this rate, agree well with each other (mean Spearman correlation of \PropRateWithinHigh), whereas the \PropRateNumLow{} datasets below it agree less, both with each other (\PropRateWithinLow) and with the others (\PropRateBetween).
DeepConvShallowLSTM falls from mean rank \PropRateFallModelLow{} on the low-rate datasets to \PropRateFallModelHigh{} on the others, whereas CNN-HAR (\PropRateGainCNNLow{} to \PropRateGainCNNHigh) and TinyHAR (\PropRateGainTinyLow{} to \PropRateGainTinyHigh) gain.
Whether a dataset uses only phones or watches has a small effect (a difference of \PropDeviceDifference{} between agreement within and between the groups), and class imbalance, the number of windows, inertial-only sensing, the recording setting, and the publication year show none (at most \PropNoEffectMax{} for both window lengths).

\subsection{Benchmark Overfitting}
\label{sec:results-unseen}

\begin{figure*}[tpb]
    \centering
    \includegraphics[width=\textwidth]{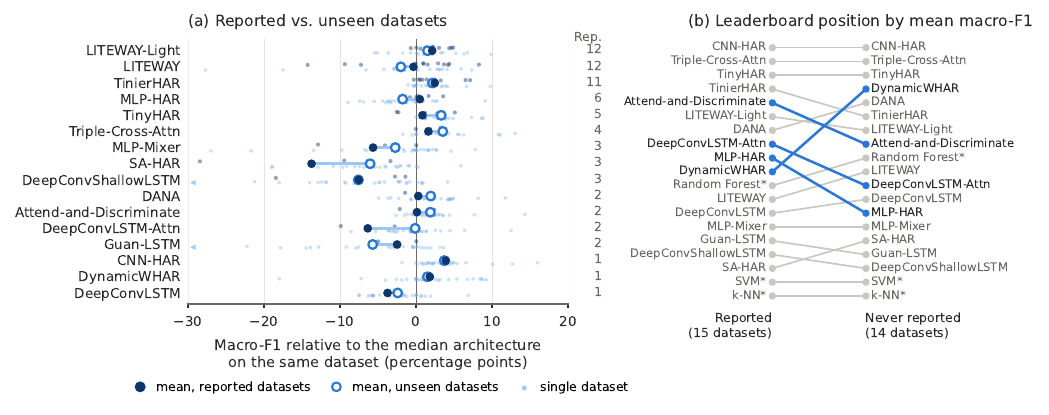}
    \caption{Recognition performance of each architecture on the datasets reported in its original publication and on the remaining, unseen datasets for 3\,s windows, over \RepNumDatasets{} datasets.
    (a) Test macro-F1 relative to the median architecture on the same dataset: mean on the reported (filled) and the unseen (open) datasets of each deep architecture, with the single datasets as faint dots (arrows: below $-30$).
    \emph{Rep.} gives the number of reported datasets.
    (b) Leaderboard position by mean test macro-F1 on the \RepNumFamiliar{} datasets that at least one original publication reports and on the \RepNumNever{} datasets that none reports. Architectures that move by \RepSplitMoveThreshold{} or more positions are highlighted.
    Classical baselines are marked with * and have no reported datasets.
    Results for 2\,s windows are similar and summarized in the text.}
    \label{fig:reported-unseen}
\end{figure*}

Most architectures were evaluated on only a few public datasets, which can lead to benchmark overfitting (\autoref{sec:related-overfitting}).
An overfitted architecture should do better, relative to the others, on the datasets of its original publication than on the others, and the leaderboard on the reported datasets should differ from the one on the datasets that no publication reports.
We test both predictions.
\RepTopDatasetOne{} is reported for \RepTopDatasetOneCount{} and \RepTopDatasetTwo{} for \RepTopDatasetTwoCount{} of the \RepNumModels{} deep architectures, and for \RepNumOnlyTop{} architectures (\RepOnlyTopModels) they are the only reported datasets, whereas \RepNumNever{} of the \RepNumDatasets{} datasets are reported for none (\autoref{fig:dataset-usage}).

Since the reported datasets are on average much easier than the unseen ones, we compare each architecture with the median architecture on the same dataset (\autoref{fig:reported-unseen}a),
\begin{equation}
    \label{eq:relative-performance}
    R_{m,d} = F_{m,d} - \operatorname*{median}_{m' \in \mathcal{A}} F_{m',d},
\end{equation}
and average $R_{m,d}$ over the reported and over the unseen datasets of the architecture.
The architectures have no general advantage on their reported datasets. \RepNumBetter{} of the \RepNumModels{} do better on them and \RepNumWorse{} worse, with a median difference of \RepMedianAbsDiff{} points.
Architectures that report \RepManyThreshold{} or more datasets of the suite lean slightly toward them (\RepManyNumBetter{} of \RepManyNum{}, by at most \RepManyBetterMaxFOne{} points), whereas \RepFewNumWorse{} of the \RepFewNum{} architectures with fewer reported datasets do worse on them.
Without \RepTopDatasetOne{} and \RepTopDatasetTwo, on which the recurrent and attention-based architectures that report them rank lower, the \RepWithoutTopNumModels{} architectures that retain a reported dataset split evenly (\RepWithoutTopNumBetter{} better, \RepWithoutTopNumWorse{} worse).

The leaderboard on the never-reported datasets largely matches the one on the datasets that at least one publication reports (Kendall's $\tau = \RepSplitTau$, \autoref{fig:reported-unseen}b).
This agreement is typical for any selection of datasets: random splits into sets of the same sizes yield a median of $\tau = \RepRandomTauMedian$.
Only \RepSplitNumMovers{} architectures move by \RepSplitMoveThreshold{} or more positions, and \RepSplitLeadInTopNeverText{} architectures of the leading group also take the first \RepSplitLeadNum{} positions on the never-reported datasets.
The results for 2\,s windows lead to the same conclusions: 6 architectures do better on their reported datasets and 10 worse, and the two leaderboards agree with $\tau = 0.78$ (random splits: median 0.73).

Overall, we find no sign of benchmark overfitting: the architectures show no consistent advantage on the datasets of their original evaluation, and the never-reported datasets keep the same leading group.
Since the assignment of datasets follows the publication record, this comparison is observational and cannot rule out overfitting to datasets that guided development without being reported (\autoref{sec:discussion-limitations}).

\subsection{Deployment Cost}
\label{sec:results-cost}

\begin{table*}[b]
    \centering
    \caption{Recognition performance and deployment cost per architecture on a smartphone, a smartwatch, and a microcontroller (MCU).
    Each architecture has one row per window length (\emph{Win.}, orange for 2\,s, blue for 3\,s). F1 is the mean test macro-F1 over the datasets, and latency is the time to classify one window.
    On the phone and the watch, latency and peak memory overhead are means over all datasets, with the best value per column and window length in bold.
    On the MCU, \emph{Fits} gives the number of datasets on which an architecture fits into RAM and flash. Since architectures that fit less often fit only the smaller inputs, MCU costs are given relative to \RelMcuReference{}, which fits every dataset (\autoref{eq:relative-cost}, values below 1 are cheaper), with absolute values in \autoref{sec:appendix-deploy-mcu}.
    Bold marks the lowest ratio among the architectures with at least \RelMcuMinFits{} datasets, and grey ratios rest on fewer datasets and should be read with care.
    $\times$: deployed on the MCU, but no dataset fits into flash. Guan-LSTM exceeds the flash on every dataset because TensorFlow Lite Micro unrolls its LSTM over the time steps.
    Classical baselines have no trainable parameters (-).
    Rows are sorted by mean rank (\autoref{sec:results-overall}).}
    \label{tab:efficiency}
    \scriptsize
    \setlength{\tabcolsep}{2.6pt}
    \definecolor{effwin2}{HTML}{FCEAE3}
\definecolor{effwin3}{HTML}{E1ECF9}
\providecommand{\effnoresult}[1]{\ensuremath{\times}}
\providecommand{\effmcufew}[1]{\textcolor{gray}{#1}}
\begin{tabular*}{\linewidth}{@{\extracolsep{\fill}}l c rr rr rr r rrrr@{}}
\toprule
 & & & & \multicolumn{2}{c}{Phone} & \multicolumn{2}{c}{Watch} & \multicolumn{5}{c}{MCU (relative to LITEWAY-Light)} \\
\cmidrule(lr){5-6} \cmidrule(lr){7-8} \cmidrule(lr){9-13}
Model & Win. & F1 (\%) & Params & Lat.\ (ms) & Mem.\ (MB) & Lat.\ (ms) & Mem.\ (MB) & Fits & Lat. & RAM & Flash & Energy \\
\midrule
\rowcolor{effwin2}CNN-HAR & 2\,s & 72.5 & 216.9k & 10.4 & 6.51 & 91.1 & 4.18 & 20/29 & 23.4 & 5.60 & 1.46 & 25.4 \\
\rowcolor{effwin3} & 3\,s & \textbf{74.4} & 216.9k & 13.6 & 8.32 & 131 & 5.28 & 17/29 & 25.9 & 6.48 & 1.34 & 27.9 \\
\addlinespace[2pt]
\rowcolor{effwin2}Triple-Cross-Attn & 2\,s & \textbf{72.7} & 560.8k & 9.76 & 8.02 & 70.5 & 5.71 & 21/29 & 35.3 & 5.50 & 2.70 & 34.8 \\
\rowcolor{effwin3} & 3\,s & 74.0 & 766.7k & 13.3 & 10.9 & 98.1 & 7.84 & 17/29 & 35.5 & 5.92 & 2.80 & 35.0 \\
\addlinespace[2pt]
\rowcolor{effwin2}TinyHAR & 2\,s & 72.4 & 188.8k & 6.90 & 7.84 & 69.4 & 4.25 & 25/29 & 7.78 & 3.01 & 2.14 & 8.48 \\
\rowcolor{effwin3} & 3\,s & 73.6 & 252.8k & 9.84 & 10.4 & 103 & 5.64 & 22/29 & 8.62 & 3.92 & 2.67 & 9.41 \\
\addlinespace[2pt]
\rowcolor{effwin2}DynamicWHAR & 2\,s & 72.0 & 2.98M & 25.6 & 30.8 & 555 & 25.9 & 3/29 & \effmcufew{8.87} & \effmcufew{2.16} & \effmcufew{5.69} & \effmcufew{9.04} \\
\rowcolor{effwin3} & 3\,s & 72.2 & 2.98M & 27.3 & 31.6 & 568 & 26.0 & 3/29 & \effmcufew{7.06} & \effmcufew{2.28} & \effmcufew{5.69} & \effmcufew{7.22} \\
\addlinespace[2pt]
\rowcolor{effwin2}TinierHAR & 2\,s & 72.0 & 22.0k & 3.96 & 3.06 & 31.1 & 1.51 & 27/29 & 2.30 & 2.86 & 1.88 & 2.45 \\
\rowcolor{effwin3} & 3\,s & 73.0 & 22.0k & 3.80 & 3.24 & 42.7 & 1.62 & 25/29 & 2.33 & 3.28 & 2.19 & 2.47 \\
\addlinespace[2pt]
\rowcolor{effwin2}LITEWAY-Light & 2\,s & 71.6 & \textbf{5.4k} & 1.66 & \textbf{2.31} & 10.9 & \textbf{1.08} & 29/29 & 1 & 1 & 1 & 1 \\
\rowcolor{effwin3} & 3\,s & 72.5 & \textbf{5.4k} & 1.95 & \textbf{2.35} & 12.8 & \textbf{1.11} & 29/29 & 1 & 1 & 1 & 1 \\
\addlinespace[2pt]
\rowcolor{effwin2}DANA & 2\,s & 70.8 & 1.08M & 19.1 & 17.0 & 216 & 12.3 & 17/29 & 63.8 & 7.34 & 4.67 & 70.3 \\
\rowcolor{effwin3} & 3\,s & 72.5 & 1.08M & 22.4 & 18.3 & 288 & 13.2 & 17/29 & 61.3 & 7.29 & 4.65 & 67.5 \\
\addlinespace[2pt]
\rowcolor{effwin2}Attend-and-Discriminate & 2\,s & 71.0 & 836.0k & 41.7 & 30.9 & 633 & 16.6 & 6/29 & \effmcufew{95.7} & \effmcufew{8.06} & \effmcufew{4.65} & \effmcufew{101} \\
\rowcolor{effwin3} & 3\,s & 72.5 & 836.0k & 68.7 & 42.6 & 1\,018 & 22.3 & 3/29 & \effmcufew{112} & \effmcufew{9.93} & \effmcufew{4.97} & \effmcufew{125} \\
\addlinespace[2pt]
\rowcolor{effwin2}LITEWAY & 2\,s & 68.9 & 5.6k & 2.61 & 3.17 & 17.7 & 1.43 & 29/29 & 2.08 & 0.82 & \textbf{1.02} & 2.06 \\
\rowcolor{effwin3} & 3\,s & 69.4 & 5.6k & 2.36 & 3.29 & 20.6 & 1.53 & 29/29 & 2.09 & 0.77 & \textbf{1.02} & 2.07 \\
\addlinespace[2pt]
\rowcolor{effwin2}Random Forest & 2\,s & 68.6 & - & \textbf{0.67} & 31.8 & \textbf{5.60} & 18.8 & 24/29 & \textbf{0.02} & \textbf{0.25} & 2.23 & \textbf{0.03} \\
\rowcolor{effwin3} & 3\,s & 70.0 & - & \textbf{0.76} & 28.9 & \textbf{6.56} & 16.6 & 25/29 & \textbf{0.02} & \textbf{0.25} & 1.89 & \textbf{0.02} \\
\addlinespace[2pt]
\rowcolor{effwin2}DeepConvLSTM-Attn & 2\,s & 69.3 & 1.09M & 43.8 & 15.2 & 561 & 11.0 & 5/29 & \effmcufew{112} & \effmcufew{7.06} & \effmcufew{5.23} & \effmcufew{121} \\
\rowcolor{effwin3} & 3\,s & 70.2 & 1.09M & 66.2 & 17.7 & 821 & 12.1 & 3/28 & \effmcufew{127} & \effmcufew{8.67} & \effmcufew{5.59} & \effmcufew{142} \\
\addlinespace[2pt]
\rowcolor{effwin2}MLP-HAR & 2\,s & 68.9 & 1.16M & 4.66 & 8.43 & 51.7 & 7.13 & 20/29 & 3.19 & 1.54 & 1.71 & 3.24 \\
\rowcolor{effwin3} & 3\,s & 69.4 & 1.43M & 5.66 & 9.96 & 65.5 & 8.36 & 19/29 & 3.42 & 1.52 & 2.16 & 3.48 \\
\addlinespace[2pt]
\rowcolor{effwin2}MLP-Mixer & 2\,s & 67.9 & 733.0k & 13.5 & 10.4 & 109 & 7.00 & 22/29 & 73.6 & 3.94 & 4.97 & 77.6 \\
\rowcolor{effwin3} & 3\,s & 67.7 & 830.9k & 17.4 & 12.6 & 143 & 8.78 & 21/29 & 72.2 & 3.63 & 5.00 & 76.1 \\
\addlinespace[2pt]
\rowcolor{effwin2}Guan-LSTM & 2\,s & 66.6 & 819.1k & 36.4 & 6.29 & 1\,450 & 6.44 & 0/29 & \effnoresult{F} & \effnoresult{F} & \effnoresult{F} & \effnoresult{F} \\
\rowcolor{effwin3} & 3\,s & 65.2 & 819.1k & 55.3 & 6.58 & 2\,187 & 7.26 & 0/29 & \effnoresult{F} & \effnoresult{F} & \effnoresult{F} & \effnoresult{F} \\
\addlinespace[2pt]
\rowcolor{effwin2}DeepConvLSTM & 2\,s & 68.1 & 1.08M & 42.6 & 13.6 & 533 & 10.9 & 6/29 & \effmcufew{103} & \effmcufew{6.85} & \effmcufew{5.09} & \effmcufew{109} \\
\rowcolor{effwin3} & 3\,s & 68.3 & 1.08M & 62.7 & 16.0 & 826 & 11.9 & 3/28 & \effmcufew{120} & \effmcufew{8.47} & \effmcufew{5.33} & \effmcufew{134} \\
\addlinespace[2pt]
\rowcolor{effwin2}DeepConvShallowLSTM & 2\,s & 64.0 & 1.02M & 17.2 & 13.5 & 243 & 10.5 & 18/29 & 82.3 & 3.25 & 3.07 & 86.9 \\
\rowcolor{effwin3} & 3\,s & 63.1 & 1.02M & 27.1 & 16.3 & 384 & 11.7 & 18/29 & 110 & 4.28 & 3.19 & 115 \\
\addlinespace[2pt]
\rowcolor{effwin2}SA-HAR & 2\,s & 63.5 & 424.7k & 11.6 & 8.81 & 76.4 & 6.72 & 20/29 & 82.8 & 7.01 & 3.81 & 84.2 \\
\rowcolor{effwin3} & 3\,s & 63.9 & 424.7k & 15.8 & 12.2 & 118 & 8.65 & 16/29 & 104 & 8.96 & 3.85 & 105 \\
\addlinespace[2pt]
\rowcolor{effwin2}SVM & 2\,s & 55.1 & - & 3.54 & 58.1 & 49.1 & 41.3 & 3/29 & \effmcufew{0.27} & \effmcufew{0.38} & \effmcufew{2.29} & \effmcufew{0.29} \\
\rowcolor{effwin3} & 3\,s & 53.1 & - & 2.94 & 45.3 & 38.3 & 29.2 & 7/29 & \effmcufew{0.33} & \effmcufew{0.47} & \effmcufew{2.58} & \effmcufew{0.36} \\
\addlinespace[2pt]
\rowcolor{effwin2}k-NN & 2\,s & 50.0 & - & 114 & 54.0 & 1\,296 & 24.7 & 2/29 & \effmcufew{0.46} & \effmcufew{0.26} & \effmcufew{2.28} & \effmcufew{0.55} \\
\rowcolor{effwin3} & 3\,s & 49.6 & - & 72.8 & 40.6 & 1\,180 & 22.3 & 4/29 & \effmcufew{0.20} & \effmcufew{0.26} & \effmcufew{2.35} & \effmcufew{0.23} \\
\bottomrule
\end{tabular*}

\end{table*}

Having compared recognition performance, we turn to the cost at which it is achieved (\autoref{tab:efficiency}).

\paragraph{Latency and real-time budget}
On the phone, all architectures stay far below the real-time budget, the window stride of \BudgetS\,s for 3\,s windows (\autoref{sec:method-efficiency}).
On the watch, latency is a median of \WatchSlowdownMedian{} times that on the phone, yet \WatchNumModelsInBudget{} of \NumModels{} architectures still meet the budget with their median latency, and \WatchNumModelsOverBudget{} exceed it on at least one dataset, mainly recurrent architectures and k-NN.
On the MCU, only \McuNumModelsInBudget{} architectures process a window within the budget.

Since architectures that fit on fewer datasets are measured only on smaller inputs, their absolute MCU means are not comparable.
We therefore express each MCU cost $c$ relative to the reference architecture $\mathrm{ref}$, \RelMcuReference, which fits every dataset, as the geometric mean of the ratios over the datasets $\mathcal{D}_m$ on which $m$ fits,
\begin{equation}
    \label{eq:relative-cost}
    \tilde{c}_m = \Biggl( \prod_{d \in \mathcal{D}_m} \frac{c_{m,d}}{c_{\mathrm{ref},d}} \Biggr)^{1 / |\mathcal{D}_m|},
\end{equation}
so that values below 1 mean cheaper than \RelMcuReference{} on the same inputs (\autoref{tab:efficiency}).
For 3\,s windows, TinierHAR needs \RelMcuLatTinierHARThreeS{} times the latency of \RelMcuReference, TinyHAR \RelMcuLatTinyHARThreeS{} times, and CNN-HAR \RelMcuLatCNNHARThreeS{} times, whereas DANA and MLP-Mixer are \RelMcuLatDANAThreeS{} and \RelMcuLatMLPMixerThreeS{} times slower, and Random Forest needs only \RelMcuLatRandomForestThreeS{} times its latency.

\paragraph{Memory on the MCU}
Memory constrains the deployment only on the MCU. On the phone and the watch, the largest peak memory overhead of any architecture is \PhoneMemMaxMb\,MB and \WatchMemMaxMb\,MB, about \PhoneMemMaxShare\,\% and \WatchMemMaxShare\,\% of their RAM.
Flash, not RAM, is what keeps deep architectures off the board. Of the architecture--dataset combinations with 3\,s windows that do not fit, \HwNoFitFlash{} exceed the flash and only \HwNoFitRam{} the RAM.
The number of sensor channels decides on how many datasets an architecture fits.
LITEWAY and LITEWAY-Light fit on \ChanMcuFitsManyLITEWAY{} of the \ChanMcuNumMany{} datasets with \ChanBorder{} or more channels and TinierHAR on \ChanMcuFitsManyTinierHAR, whereas TinyHAR fits on only \ChanMcuFitsManyTinyHAR{} and CNN-HAR and Triple-Cross-Attn on only \ChanMcuFitsManyCNNHAR{} of them, although all three fit on \ChanMcuFitsFewCNNHAR{} of the \ChanMcuNumFew{} datasets with few channels.
Memory also does not necessarily follow the size of an architecture: LITEWAY-Light has slightly fewer parameters than LITEWAY but requires more RAM on the MCU (\EffMcuRamLITEWAYLight{} vs.\ \EffMcuRamLITEWAY\,KB on average), since the quantizer leaves some of its activations in floating point (\autoref{sec:appendix-cost-details}).

\paragraph{Energy and cost proxies}
On the MCU, the energy per inference closely follows latency (Spearman's $\rho = \McuEnergyLatencySpearman$), since the average power of the nRF5340 during inference varies only between \McuPowerMinMw{} and \McuPowerMaxMw\,mW, so on this board measuring it adds little beyond latency.

\begin{figure}[tbp]
    \centering
    \begin{subfigure}[t]{0.49\linewidth}
        \centering
        \includegraphics[width=\linewidth]{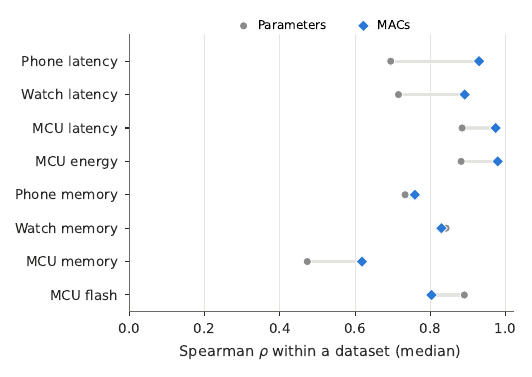}
        \caption{Agreement with cost rankings}
        \label{fig:hw-proxies}
    \end{subfigure}\hfill
    \begin{subfigure}[t]{0.49\linewidth}
        \centering
        \includegraphics[width=\linewidth]{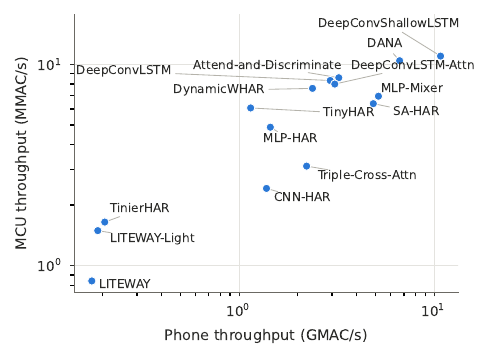}
        \caption{Achieved MAC throughput}
        \label{fig:hw-throughput}
    \end{subfigure}
    \caption{(a) Spearman rank correlation between parameter count (grey circles) or multiply-accumulate (MAC) count (blue diamonds) and each measured cost, computed within one dataset, with each marker showing the median over the datasets. Higher values indicate closer agreement in ordering, not greater accuracy in predicting absolute cost.
    (b) Median achieved MAC throughput over datasets for each architecture on the phone (x-axis, GMAC/s) and MCU (y-axis, MMAC/s). Throughput is MACs per inference divided by measured latency, and MCU medians use only the datasets on which the architecture runs.}
    \label{fig:hw-size-proxies}
\end{figure}

Many HAR architectures are characterized only by their number of parameters.
Within a dataset, the count of multiply-accumulate operations (MACs) agrees more closely than the parameter count with latency rankings on all three platforms (\autoref{fig:hw-proxies}): the median Spearman correlation is $\rho = \HwRhoParamsPhoneLat$ for parameters and $\rho = \HwRhoMacsPhoneLat$ for MACs on the phone, and $\rho = \HwRhoParamsMcuLat$ and $\rho = \HwRhoMacsMcuLat$ on the MCU.
Parameter count, in turn, agrees closely with flash rankings ($\rho = \HwRhoParamsMcuFlash$).
These correlations describe agreement in ordering and do not show how accurately either count predicts absolute cost.
The MACs an architecture executes per second vary by a factor of \HwPhoneGmacsRatio{} on the phone and \HwMcuMmacsRatio{} on the MCU (\autoref{fig:hw-throughput}, \autoref{sec:appendix-cost-details}).

\begin{figure*}[b]
    \centering
    \begin{subfigure}[t]{0.49\textwidth}
        \centering
        \includegraphics[width=\linewidth]{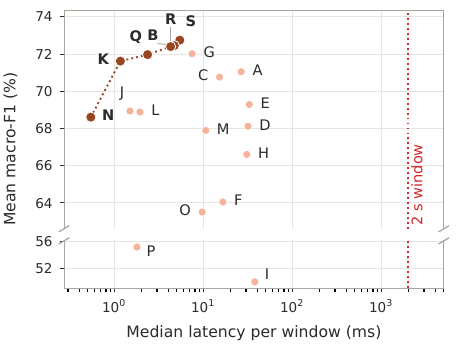}
        \caption{Phone, 2\,s windows}
        \label{fig:pareto-phone-2s}
    \end{subfigure}\hfill
    \begin{subfigure}[t]{0.49\textwidth}
        \centering
        \includegraphics[width=\linewidth]{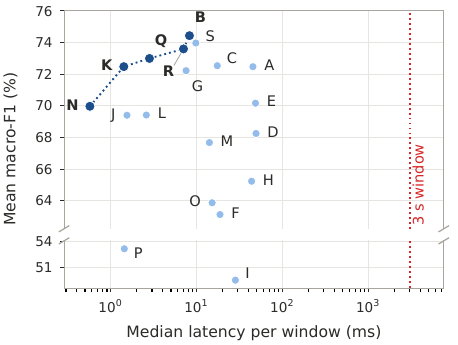}
        \caption{Phone, 3\,s windows}
        \label{fig:pareto-phone}
    \end{subfigure}
    \par\medskip
    \begin{subfigure}[t]{0.49\textwidth}
        \centering
        \includegraphics[width=\linewidth]{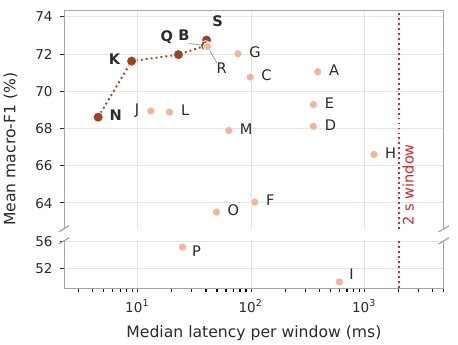}
        \caption{Watch, 2\,s windows}
        \label{fig:pareto-watch-2s}
    \end{subfigure}\hfill
    \begin{subfigure}[t]{0.49\textwidth}
        \centering
        \includegraphics[width=\linewidth]{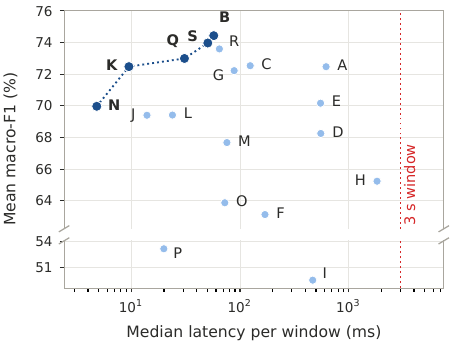}
        \caption{Watch, 3\,s windows}
        \label{fig:pareto-watch}
    \end{subfigure}
    \par\medskip
    \includegraphics[width=\textwidth]{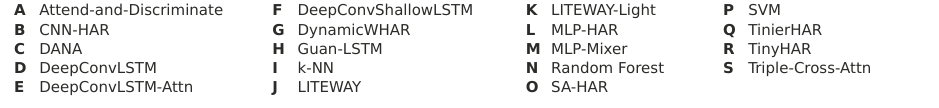}
    \caption{Mean test macro-F1 against inference latency on the phone (top) and the watch (bottom) for 2\,s (left, orange) and 3\,s windows (right, blue).
    Each point gives the mean test macro-F1 of an architecture over the datasets and its median latency.
    Dark points connected by the dotted line form the Pareto front, i.e., no other architecture has both a lower latency and a higher test macro-F1. Their letters, which identify the architectures in the key below the panels, are bold.
    The red dotted line marks the window length, twice the real-time budget.
    The y-axis is broken to also show SVM and k-NN.}
    \label{fig:pareto-devices}
\end{figure*}

\paragraph{Window length}
From 2 to 3\,s windows, the median latency of a deep architecture grows by a mean factor of \PhoneLatGrowth{} on the smartphone, \WatchLatGrowth{} on the smartwatch, and \McuLatGrowth{} on the MCU, while the order of the architectures remains largely the same (\autoref{fig:insight-latency-scaling}).
On the MCU, longer windows also reduce the number of architecture--dataset combinations that fit, from \WinMcuFitsTwoS{} of \WinMcuCombos{} to \WinMcuFitsThreeS.
Flash barely depends on the window length for most architectures, but it grows for \WinBigFlashModels, which fit on \WinBigFitsRangeTwoS{} datasets for 2\,s windows and on only \WinBigFitsRangeThreeS{} for 3\,s windows.

Overall, latency rather than memory separates the architectures on the phone and the watch.
On the MCU, flash decides which architectures fit at all, and latency separates those that do.

\subsection{Trade-offs Between Recognition Performance and Deployment Cost}
\label{sec:results-tradeoffs}

Since the leading architectures differ by only a few points of test macro-F1, deployment cost can decide between them.
We therefore determine the Pareto fronts of test macro-F1 against latency on all three platforms and against memory on the MCU, both over all datasets and per dataset: the architectures for which no other architecture achieves a higher test macro-F1 at equal or lower cost.
These fronts require no weighting between performance and cost and let practitioners choose according to their resource budget.

\begin{figure*}[tpb]
    \centering
    \begin{subfigure}[t]{0.49\textwidth}
        \centering
        \includegraphics[width=\linewidth]{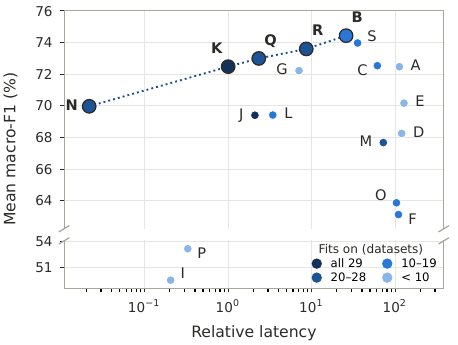}
        \caption{Relative latency}
        \label{fig:pareto-mcu}
    \end{subfigure}\hfill
    \begin{subfigure}[t]{0.49\textwidth}
        \centering
        \includegraphics[width=\linewidth]{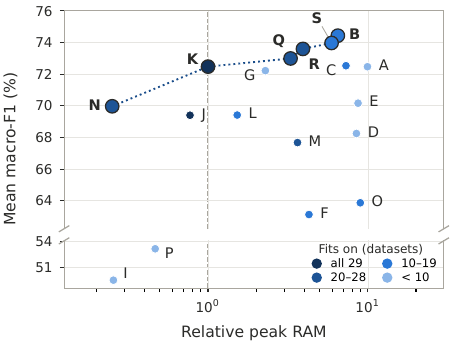}
        \caption{Relative peak RAM}
        \label{fig:pareto-mcu-ram}
    \end{subfigure}
    \par\medskip
    \begin{subfigure}[t]{0.49\textwidth}
        \centering
        \includegraphics[width=\linewidth]{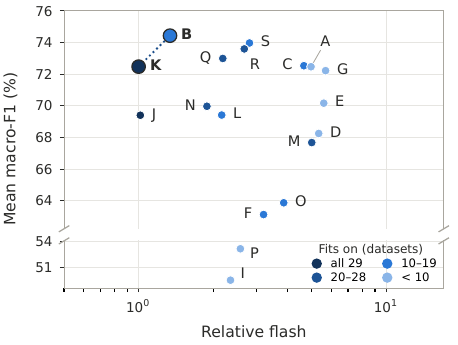}
        \caption{Relative flash}
        \label{fig:pareto-mcu-flash}
    \end{subfigure}\hfill
    \begin{subfigure}[t]{0.49\textwidth}
        \centering
        \includegraphics[width=\linewidth]{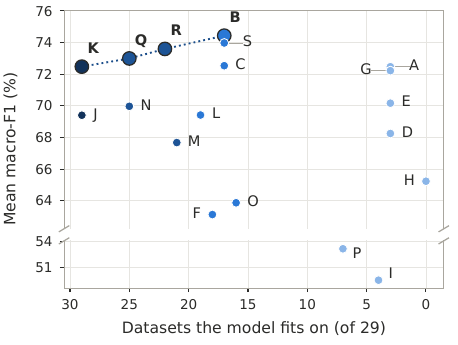}
        \caption{Datasets that fit into RAM and flash}
        \label{fig:pareto-mcu-fits}
    \end{subfigure}
    \par\medskip
    \includegraphics[width=\textwidth]{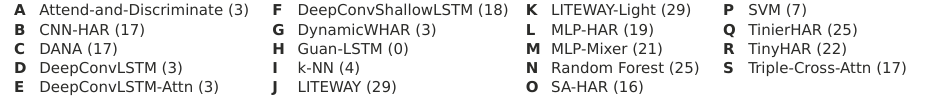}
    \caption{Mean test macro-F1 against deployment cost on the MCU for 3\,s windows: relative latency (a), relative peak RAM (b), relative flash (c), and the number of datasets on which an architecture fits into RAM and flash (d).
    The energy front is not shown, since it nearly equals the latency front.
    Costs in (a)--(c) are relative to \RelMcuReference{} on the datasets on which an architecture fits (\autoref{eq:relative-cost}), with the dashed line marking a ratio of~1.
    Letters identify the architectures in the key below the panels, whose parentheses give the number of datasets on which an architecture fits (darker points = more).
    Outlined points connected by the dotted line form the Pareto front, and their letters are bold.
    The y-axis is broken to also show SVM and k-NN.}
    \label{fig:pareto-mcu-all}
\end{figure*}

\begin{figure*}[tpb]
    \centering
    \begin{subfigure}[t]{0.49\textwidth}
        \centering
        \includegraphics[width=\linewidth]{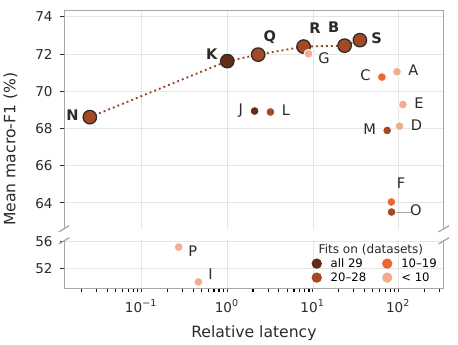}
        \caption{Relative latency}
        \label{fig:pareto-mcu-2s}
    \end{subfigure}\hfill
    \begin{subfigure}[t]{0.49\textwidth}
        \centering
        \includegraphics[width=\linewidth]{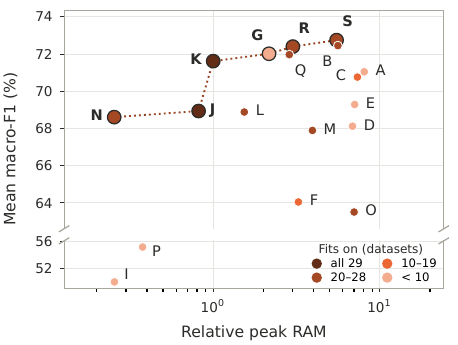}
        \caption{Relative peak RAM}
        \label{fig:pareto-mcu-ram-2s}
    \end{subfigure}
    \par\medskip
    \begin{subfigure}[t]{0.49\textwidth}
        \centering
        \includegraphics[width=\linewidth]{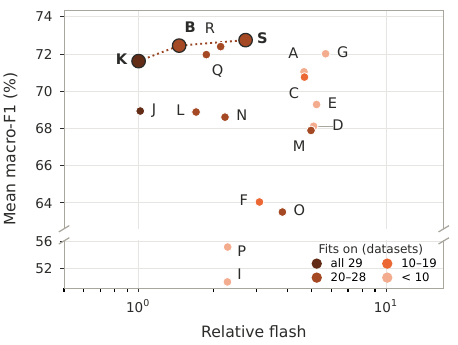}
        \caption{Relative flash}
        \label{fig:pareto-mcu-flash-2s}
    \end{subfigure}\hfill
    \begin{subfigure}[t]{0.49\textwidth}
        \centering
        \includegraphics[width=\linewidth]{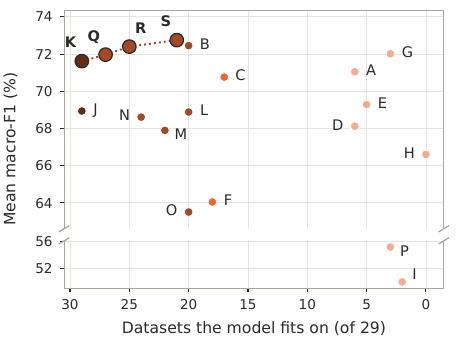}
        \caption{Datasets that fit into RAM and flash}
        \label{fig:pareto-mcu-fits-2s}
    \end{subfigure}
    \par\medskip
    \includegraphics[width=\textwidth]{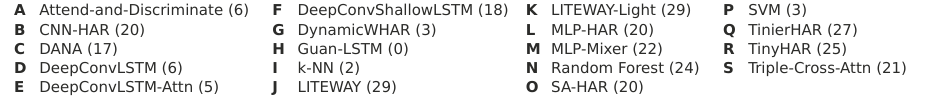}
    \caption{Mean test macro-F1 against deployment cost on the MCU for 2\,s windows: relative latency (a), relative peak RAM (b), relative flash (c), and the number of datasets on which an architecture fits into RAM and flash (d).
    The energy front is not shown, since it nearly equals the latency front.
    Costs in (a)--(c) are relative to \RelMcuReference{} on the datasets on which an architecture fits (\autoref{eq:relative-cost}), with the dashed line marking a ratio of~1.
    Letters identify the architectures in the key below the panels, whose parentheses give the number of datasets on which an architecture fits (darker points = more).
    Outlined points connected by the dotted line form the Pareto front, and their letters are bold.
    The y-axis is broken to also show SVM and k-NN.}
    \label{fig:pareto-mcu-all-2s}
\end{figure*}

\begin{figure*}[tpb]
    \centering
    \includegraphics[width=\textwidth]{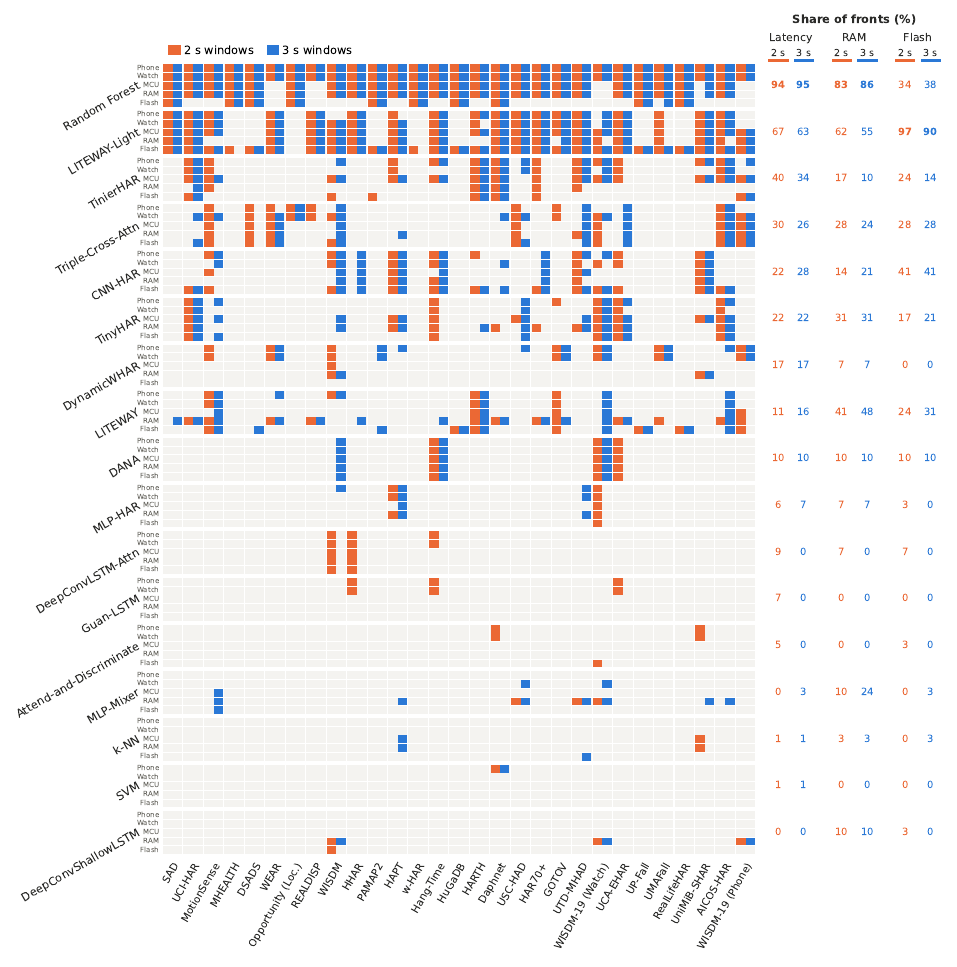}
    \caption{Pareto-optimal architectures per dataset for 2\,s and 3\,s windows.
    Each architecture has one row per front: latency on the phone, the watch, and the MCU, and peak RAM and flash on the MCU.
    Each dataset cell is split into a left half for 2\,s and a right half for 3\,s windows, and a filled half means that the architecture lies on that front of test macro-F1 against the cost, computed on that dataset alone.
    On the MCU, only architectures that fit into memory compete.
    The columns on the right give the share of fronts an architecture lies on per window length, with latency pooled over the three platforms and the best value per column in bold.
    Architectures are sorted by their mean latency share over both window lengths, datasets by their mean test macro-F1.
    DeepConvLSTM and SA-HAR lie on no front and are not shown.}
    \label{fig:pareto-membership}
\end{figure*}

\paragraph{Phone and watch}
On the phone and the watch (\autoref{fig:pareto-devices}), the Pareto front starts at Random Forest and LITEWAY-Light, the fastest architectures, and ends at CNN-HAR, the one with the highest mean test macro-F1, with TinierHAR in between.
The recurrent architectures (DeepConvLSTM, DeepConvLSTM-Attn, Guan-LSTM) and Attend-and-Discriminate are several times to an order of magnitude slower than architectures of similar or higher mean test macro-F1 and lie on none of the fronts on any platform.
Since all architectures stay far below the real-time budget on the phone and \WatchMostAccurateInBudget{} still meets it on the watch, the architecture with the highest mean test macro-F1 can be chosen at no practical cost on both platforms.

\paragraph{Microcontroller}
On the MCU (\autoref{fig:pareto-mcu-all} and \autoref{fig:pareto-mcu-all-2s}), the latency front consists of the same architectures, with TinyHAR joining it for 3\,s windows.
Its high-performance end, however, is available only for part of the suite. Of the architectures on the front, only \RelMcuReference{} fits every dataset, whereas for 3\,s windows CNN-HAR fits \McuFitsCNNHARThreeS{} of the \McuTotalThreeS{} datasets, TinyHAR \McuFitsTinyHARThreeS, TinierHAR \McuFitsTinierHARThreeS, and Random Forest \McuFitsRandomForestThreeS.
Over all datasets, memory and energy lead to nearly the same choice as latency.
For 3\,s windows, the front against relative peak RAM consists of \McuRelRamParetoModelsThreeS{} and the front against the number of datasets that fit of \McuFitParetoModels, whereas the front against relative flash consists only of \McuRelFlashParetoModelsThreeS, since flash differs little between the architectures that fit.
CNN-HAR and Triple-Cross-Attn, the architectures with the highest mean test macro-F1, need a median of \McuMedianLatSCNNHAR{} and \McuMedianLatSTripleCrossAttn\,s per window, so \McuMostAccurateInBudget{} has the highest mean test macro-F1 among the architectures that meet the real-time budget.
LITEWAY and LITEWAY-Light fit on all \ChanMcuNumMany{} datasets with many channels, where LITEWAY also ranks high (\autoref{sec:results-consistency}), so they combine a high recognition performance on these datasets with the ability to run on a microcontroller.

\paragraph{Per-dataset fronts}
The fronts of individual datasets (\autoref{fig:pareto-membership}) show the same leaders.
For 3\,s windows, Random Forest leads on latency (\ParetoShareLatencyRandomForest\,\% of all dataset--platform fronts, followed by \ParetoShareTopTwoModel{} with \ParetoShareTopTwo\,\%) and on peak RAM (\ParetoShareRAMRandomForest\,\%), whereas LITEWAY-Light leads on flash (\ParetoShareFlashLITEWAYLight\,\%), where Random Forest falls to \ParetoShareFlashRandomForest\,\%.
LITEWAY gains most from memory, rising from \ParetoShareLatencyLITEWAY\,\% on latency to \ParetoShareRAMLITEWAY\,\% on RAM.
The platforms largely agree: of the \ParetoNumLatencyCases{} cases in which an architecture lies on a latency front of a dataset, \ParetoNumAllPlatforms{} hold on all three platforms.
On the MCU, fronts against energy nearly equal those against latency (mean Jaccard overlap \ParetoOverlapEnergy), whereas fronts against peak RAM and flash overlap less (\ParetoOverlapRAM{} and \ParetoOverlapFlash) and shift the choice among the leading architectures. Overall, the same few architectures span the trade-off on every platform: Random Forest and LITEWAY-Light at the low-cost end, TinierHAR in between, and CNN-HAR at the high-performance end.
Only the MCU forces a choice.

\section{Living Benchmark Platform}
\label{sec:platform}

The results above are a snapshot of one suite, and they show that rankings depend on the datasets they rest on (\autoref{sec:results-stability}).
HARENA is therefore maintained as a living benchmark, with a public website\footnote{\HarenaURL{}} that brings together the results, model and dataset catalogs, and a contribution workflow.
It can thus grow beyond the initial suite while retaining a common evaluation protocol.

\subsection{Leaderboard}
\label{sec:platform-leaderboard}

The leaderboard lets users compare architectures across datasets and explore trade-offs between recognition performance and deployment metrics, such as inference latency, memory consumption, and model size, which helps identify architectures that suit different sensing conditions and resource budgets.

The dataset catalog can be filtered by properties such as the number of subjects, activities, or sensor channels, as well as sensor modalities and recording settings, and the model catalog by architectural components, implementation framework, parameter count, and computational cost.
Both link to the original publications.

\subsection{Adding Datasets and Architectures}
\label{sec:platform-extension}

The contribution page guides users through proposing a new dataset or architecture by collecting publication details, implementation or data access links, and relevant metadata, and generating an issue draft.
Adding a dataset requires a parser and configuration for the common data-preparation pipeline (\autoref{sec:method-library}), and adding an architecture requires an implementation compatible with the shared model interface (\autoref{sec:method-models}).
Both contributions include tests and are reviewed through issues and pull requests.
Once a contribution is accepted, the benchmark plans its missing combinations, so that a new architecture is evaluated on every dataset and a new dataset with every architecture under the common protocol, and measures its deployment cost on the smartphone, the smartwatch, and the microcontroller before adding the results to the website.
Every planned run has a deterministic fingerprint, so existing results are reused rather than repeated, and results are stored as validated files under version control.
Each added dataset thereby also tests the rankings reported here under further conditions.

\section{Discussion}
\label{sec:discussion}

HARENA does not identify a single best architecture but a broad group of competitive ones (\textbf{RQ1}), within which deployment cost differs by orders of magnitude (\textbf{RQ2}).
The order of the architectures changes from dataset to dataset, partly following the number of sensor channels, and the leading group persists on the datasets that no original publication reports (\textbf{RQ3}).
Beyond these questions, recognition performance drops by a median of \GenGapMedian{} points on unseen subjects, about three times the typical spread between architectures on a dataset.
We discuss what these findings imply, condense them into guidance for choosing an architecture, and describe the limitations of the benchmark and its future extensions.

\subsection{Implications}
\label{sec:discussion-implications}

\subsubsection{Recognition progress has stalled, and deployment cost now separates architectures}
Under the common protocol, a newer architecture is not automatically a better one.
\InsBestModel{}, the oldest deep architecture in the suite (\InsBestModelYear), has the best mean rank, and the three best architectures differ by only \TopThreeFOneSpread{} points of mean test macro-F1.
Larger models do not help either: parameter count is not significantly associated with mean rank ($\rho = \InsParamsRho$, $p = \InsParamsP$, \autoref{fig:insight-progress-params}).
This does not mean that HAR is solved, since \DsNumMeanLow{} datasets remain below \DsLowMedian\,\% mean test macro-F1 and the shared training configuration does not establish what each architecture could reach after individual tuning.
It means that gains of one or two points on a few datasets are weak evidence of general progress.
Where the leading architectures do differ is deployment cost: on identical MCU inputs, DANA needs \RelMcuLatDANAThreeS{} times the latency of \RelMcuReference{}, and Attend-and-Discriminate fits on only \McuFitsAttendAndDiscriminateThreeS{} of \McuTotalThreeS{} datasets.
New architectures should therefore show either that their gains hold across sensing conditions or that they reduce measured deployment cost at comparable recognition performance.

\subsubsection{Generalization to new subjects is the larger problem}
The drop on new subjects exceeds the differences between architectures, and the variability between test folds exceeds the spread between architectures on \InsNumSubjectLarger{} of \InsNumSpreadDatasets{} datasets (\autoref{fig:insight-subjects}).
Since the gap is mainly a property of the dataset and is unrelated to the number of subjects and channels, neither additional subjects recorded under the same conditions nor a richer sensor setup appears to close it, and varied recording conditions seem to matter more.
Two caveats apply: the validation data also select the early-stopping checkpoint, so the gap cannot be attributed entirely to subject shift, and on datasets with more than ten subjects the folds contain subject groups rather than single subjects.
Even so, its magnitude implies that architectures aimed at higher recognition performance under supervised training can recover only a small part of the loss on new users.
Two routes remain, both of which are out of scope for HARENA and left to future work.
The first is to learn more general representations, as foundation models aim to do through large-scale pretraining with self-supervised objectives such as contrastive learning \cite{bian2026foundation}, although their memory and compute demands make deployment on edge devices difficult.
The second is calibration and adaptation to the user, device, and sensor placement \cite{kawsar2026calibration}, for which efficient on-device approaches are already being explored \cite{burzer2026bayahar}.
A benchmark of efficient supervised architectures remains necessary for both routes, since inference, and potentially adaptation, must run on the device.

\subsubsection{Small evaluations cannot establish a general ranking}
Rankings derived from few datasets carry over only partly to others. Evaluating on two datasets, as several publications do with \RepTopDatasetOne{} and \RepTopDatasetTwo{} within our suite \cite{guan2017ensembles,murahari2018attention,abedin2021attend}, is not enough: two disjoint pairs of datasets reach a median Kendall's $\tau$ of only \SubsetMeanTauTwo, so about one in three pairs of architectures swaps its order.
Four datasets are not enough either ($\tau = \SubsetMeanTauFour$, about one in four pairs), and even two halves of the suite reach only \SubsetMeanTauMax, about one in six pairs (\autoref{fig:rank-stability}).
The gain flattens with more datasets, so about \SubsetSomeK{} datasets seem a reasonable minimum for a comparison, although more datasets still improve the agreement.
Recent architecture studies have begun to move in this direction with the multi-dataset evaluation released with TinyHAR \cite{zhou2022tinyhar,bian2025tinierhar,nshimyimana2026liteway}, yet \RepFewNum{} of the \RepNumModels{} deep architectures report fewer than \RepManyThreshold{} datasets of the suite.
Such evaluations support conclusions about the datasets they cover, but not about a generally superior architecture.
The narrow selection does not appear to have produced benchmark overfitting, so the weakness of the published evidence is its narrow scope rather than a measurable bias toward familiar datasets, although the publication record cannot reveal every dataset that informed development.
Since the agreement still rises at the largest subsets the suite allows, no fixed selection of datasets can be considered final.
This is why a living benchmark is needed: every added dataset tests the ranking under further conditions.

\subsubsection{Baselines should represent the current trade-off}
Random Forest remains an informative baseline despite its minimal set of hand-crafted features.
It outperforms \RankFirstModel{} on \ModelBeatsFirstRandomForest{} of \RankNumDatasets{} datasets, ranks near the top on datasets with many channels, and lies on \ParetoShareLatencyRandomForest\,\% of all latency fronts.
Its main weakness on the MCU is flash (\RelMcuFlashRandomForestThreeS{} times that of \RelMcuReference{}), since the trees are compiled into the firmware, and tuning its features and size might make it even more competitive.
A new deep architecture should therefore be compared with Random Forest to establish which recognition benefit justifies its additional cost.
SVM and k-NN come within \GapClose{} points of the best architecture on no dataset and are uninformative as the only classical comparison.
DeepConvLSTM \cite{ordonez2016deep}, the most common deep baseline, is likewise insufficient on its own, since several architectures combine higher recognition performance with lower measured cost.
Comparisons should include an accurate reference such as CNN-HAR and efficient alternatives such as TinierHAR and LITEWAY-Light, so that an improvement is assessed against the current trade-off rather than against an established predecessor.

\subsubsection{Deployment efficiency must be measured, not estimated}
Since deployment cost now separates the leading architectures, it has to be measured: parameter and MAC counts remain useful descriptions, but they do not replace measurements (\autoref{fig:hw-proxies}), as operation counts also predict microcontroller latency less accurately than on-device measurements \cite{king2025micronas}.
On the MCU, the toolchain can even reverse expectations based on model size: Guan-LSTM exceeds the flash on every dataset because TensorFlow Lite Micro unrolls its recurrence over the time steps, and LITEWAY-Light needs more RAM than the larger LITEWAY because the quantizer leaves some of its activations in floating point.
Deployment efficiency is thus a property of the architecture together with its input configuration, export path, runtime, and device.
Latency fronts agree across platforms only partly (\ParetoNumAllPlatforms{} of \ParetoNumLatencyCases{} cases hold on all three), so a result on one device does not establish efficiency on another.
No architecture is cheapest on every resource either, so MCU developers should identify the binding resource and check operators early for integer support.
On the nRF5340, energy per inference follows latency almost exactly, so latency is a good proxy for it on this board, although not necessarily on others.

\subsubsection{Architecture choice depends on the sensing configuration}
The number of sensor channels shifts the ranks of the architectures substantially, whereas the number of activities shifts them little, and low-rate datasets rank the architectures less consistently.
Recurrent layers are therefore worth considering only with few channels, and even then the benefit rarely justifies the cost: DeepConvLSTM, DeepConvLSTM-Attn, and Guan-LSTM are several times to an order of magnitude slower than architectures of similar or higher mean test macro-F1 and lie on none of the overall Pareto fronts, whereas compact designs with a recurrent layer, such as TinyHAR and TinierHAR, do reach them.
Rankings agree more closely within the channel groups than between them, so part of the apparent instability of rankings reflects differences in the sensing configuration.
One possible explanation is that several sensors provide complementary information that per-channel statistics or simple models already capture, whereas a limited sensor setup places greater demands on temporal modeling.
The comparison does not isolate this mechanism, since channel count varies together with sensor modality, placement, and activity content, and the division at \ChanBorder{} channels is a descriptive split of this suite rather than a universal threshold.
Nevertheless, an architectural advantage should be reported together with the sensing conditions under which it holds.

\subsubsection{Window length must be chosen together with deployment cost}
Longer windows provide a modest recognition benefit: from 2 to 3\,s, test macro-F1 changes by a median of \InsWinMedianDiff{} points and improves in \InsWinShareBetter\,\% of \InsWinNumPairs{} architecture--dataset pairs (\autoref{fig:insight-window-channels}).
The ranking hardly changes ($\tau = \InsWinRankTau$), so the conclusions above do not hinge on the window length, although the comparison is not a controlled estimate of temporal context, since some datasets also change their sampling rate and longer windows yield fewer training windows.
For always-on recognition, the energy per hour matters more than the energy per inference.
With 50\,\% overlap, 3\,s windows need a third fewer inferences per hour than 2\,s windows, so they save energy only if a single inference costs less than 1.5 times as much.
On the MCU, 3\,s windows lower the energy per hour by more than 5\,\% for \InsEnergyHourNumSave{} of \InsEnergyHourNumModels{} architectures and raise it for \InsEnergyHourNumCost{}, counting inference only, not sensing or idle power.
Longer windows also delay each prediction.

\subsection{Choosing an Architecture}
\label{sec:discussion-guide}

\begin{table}[b]
    \centering
    \caption{Candidate architectures by device class and channel group, based on the observed mean ranks, Pareto fronts, and MCU fit counts (\autoref{sec:results-consistency} and \autoref{sec:results-tradeoffs}).
    Parentheses indicate alternatives with lower inference cost.
    Candidates must be checked on the target input, since the channel split and aggregate costs do not guarantee feasibility or performance for an individual application.}
    \label{tab:decision-guide}
    \footnotesize
    \begin{tabular}{@{}l p{0.45\linewidth} p{0.45\linewidth}@{}}
        \toprule
         & \textbf{Fewer than \ChanBorder{} channels} & \textbf{\ChanBorder{} or more channels} \\
        \midrule
        Phone & CNN-HAR, TinyHAR, Triple-Cross-Attn & CNN-HAR, LITEWAY (Random Forest) \\
        Watch & CNN-HAR, TinyHAR (TinierHAR, LITEWAY-Light) & CNN-HAR, LITEWAY (Random Forest, LITEWAY-Light) \\
        MCU & TinyHAR (TinierHAR, LITEWAY-Light) & LITEWAY, LITEWAY-Light (Random Forest, TinierHAR) \\
        \bottomrule
    \end{tabular}
\end{table}

For practitioners, \autoref{tab:decision-guide} condenses the results into a shortlist for application-specific evaluation in two steps.
The first step is to check memory and latency for the actual input configuration and prediction interval.
On the phone, all architectures stay below the real-time budget and memory is not restrictive, so recognition performance alone decides.
On the watch, \WatchNumModelsInBudget{} of \NumModels{} architectures meet the budget with their median latency, but \WatchNumModelsOverBudget{} exceed it on at least one dataset.
On the MCU, \McuMostAccurateInBudget{} is the most accurate architecture whose median latency meets the budget, but feasibility depends on the input: TinyHAR fits on only \ChanMcuFitsManyTinyHAR{} of the \ChanMcuNumMany{} datasets with many channels, whereas LITEWAY and LITEWAY-Light fit on all of them, and if flash is the limit, LITEWAY-Light is the first candidate.
The second step is to prioritize the feasible candidates by sensing configuration.
As a consistent default, CNN-HAR stays on average only \ModelMeanGapCNNHAR{} points behind the best architecture on each dataset.
Convolutional and attention-based architectures without recurrence are the safest default on every platform, recurrent layers should be considered only with few channels and in compact designs, and Random Forest should be included whenever its trained model fits into memory.
On the MCU, 2\,s windows are the safer starting point unless energy per hour is the priority.
The final choice should compare the shortlisted architectures on target data, since neither an aggregate rank nor Pareto membership guarantees the best trade-off for a particular application.

\subsection{Limitations}
\label{sec:discussion-limitations}

\subsubsection{Scope}
HARENA deliberately evaluates compact, task-specific architectures trained separately for each sensing configuration, so foundation models and self-supervised pretraining are outside its current scope, and its conclusions about architectural progress apply to this setting and the evaluated implementations.
The selection criteria keep the suite independent of the performance of any architecture but exclude some datasets, including large collections of daily-life recordings.
The publication record identifies reported evaluations, not every dataset used during development, so the comparison of reported and unseen datasets can neither establish nor rule out benchmark overfitting.
A public leaderboard is itself open to the adaptive reuse discussed in \autoref{sec:related-overfitting}, since contributors can tune new architectures against scores on the same datasets \cite{blum2015ladder}.
Datasets added after an architecture was contributed offer a check on this but do not prevent it.

\subsubsection{Training and uncertainty}
Each configuration is trained with one random seed, since one pass through the protocol already comprises \TrainingFitsTotal{} model training runs and every additional seed repeats it in full.
Averaging over test folds and \NumDatasets{} datasets dampens the influence of a single training run but does not remove it, so the leading group should be read as a set of competitive candidates rather than a strict order.
We report no significance tests or confidence intervals over datasets, since the suite is a curated selection rather than a random sample of datasets, and instead measure how the results vary with the choice of datasets through disjoint subsets (\autoref{sec:results-stability}).
For the same reason of cost, datasets with more than ten subjects are evaluated on held-out subject groups, and since each fold already trains on at least 86\,\% instead of at least 93\,\% of the subjects, we expect little effect on the estimates.
The shared training configuration may favor some architectures, and tuned or scaled-down variants might perform better or fit on the MCU.
The classical baselines use the minimal tsfresh feature set and basic hyperparameters, since they serve as simple reference points, so Random Forest in particular might become more competitive after tuning.

\subsubsection{Recognition task}
The task uses fixed windows of known activities with window lengths of 2 and 3\,s and 50\,\% overlap (\autoref{sec:method-protocol}).
Body placement and annotation granularity are not available in a consistent form across the datasets and were not tested. The task also does not assess other window lengths, other overlaps, or continuous recognition with unknown activities and activity boundaries.
Both window lengths lie within the range of the original publications, but these also use shorter and longer windows.

\subsubsection{Deployment measurements}
The measurements describe one phone, one watch, and one MCU with the specified runtimes and export paths, and other runtimes, quantization methods, and devices may change the trade-offs.
The MCU fronts combine floating-point recognition performance with the cost of quantized models and do not establish the recognition performance of the deployed models.
Relative costs that rest on few datasets need not generalize to larger inputs.
Latency on the phone and watch does not establish battery consumption, and none of the measurements include the complete sensing and application workload.

\subsection{Future Work}
\label{sec:discussion-future}

Repeated training with several seeds would enable confidence intervals and significance tests for the aggregate ranking, and tuning the architectures would show whether the leading group and the absence of a trend over publication years persist beyond the shared training configuration.
Controlled variations of window length and overlap, and removing sensor channels within the same dataset, would provide stronger evidence on the role of temporal context and channel count.
Measuring the test macro-F1 of the trained models after conversion would connect recognition performance and cost for the same deployed model, and evaluation on continuous streams would test the efficient candidates in a complete recognition system.
The large generalization gaps motivate extending HARENA to personalization, domain adaptation, cross-dataset transfer, few-shot settings, and pretrained backbones, including wearable foundation models, with explicit rules for the data available during adaptation and the adaptation cost measured on the device.
New datasets should prioritize difficult and underrepresented recording conditions, such as recordings in daily life and with varying devices and placements, since \DsNumMeanHigh{} datasets are close to saturation.
They can test whether earlier conclusions persist, provided they did not inform the evaluated design decisions \cite{dwork2015generalization}.

\section{Conclusion}
\label{sec:conclusion}

New architectures for sensor-based human activity recognition on edge devices are usually evaluated on a few familiar datasets, often without measured deployment cost, which makes it hard to tell whether the field is making general progress.
We therefore introduced HARENA, a living benchmark with a public leaderboard\footnote{\HarenaURL{}} and a contribution workflow, whose initial suite evaluates \NumDeepModels{} neural architectures\footnote{\WharModelsURL{}} and three classical baselines on \NumDatasets{} datasets\footnote{\WharDatasetsURL{}} under one protocol, with costs measured on a smartphone, a smartwatch, and a microcontroller.

There is no single best architecture.
\RankFirstModel{}, the oldest deep architecture in the suite, leads overall, but \LeadNum{} architectures lie within \LeadThreshold{} points of its mean test macro-F1, while their deployment cost differs by orders of magnitude.
On the phone and the watch, the most accurate architecture can be chosen at no practical cost, whereas on the microcontroller, Random Forest leads on latency and peak RAM even without tuning, LITEWAY-Light leads on flash, and \McuMostAccurateInBudget{} is the most accurate within the real-time budget.
The order of the architectures changes from dataset to dataset and shifts with the number of sensor channels, so claims of general superiority require many datasets.
We find no sign of benchmark overfitting, but on unseen subjects, recognition performance drops by a median of \GenGapMedian{} points, about three times the typical spread between architectures on a dataset.

New architectures should therefore be evaluated on many datasets, compared with efficient alternatives and a simple baseline such as Random Forest, and measured on the intended device.
Practitioners can use our shortlist per device and channel group (\autoref{tab:decision-guide}).
Larger gains on new users are more likely to come from better generalization, for example through foundation models, whose memory and compute demands remain a challenge for edge deployment, and from on-device adaptation than from further architectural refinement.
Since rankings depend on the datasets and new architectures keep appearing, these comparisons cannot be final. As a living benchmark, HARENA keeps them current by evaluating every added dataset and architecture under the same protocol.

\begin{acks}
This work was partially funded by the IPAI Foundation gGmbH through the Science Residency Program and by the Helmholtz Association Initiative and Networking Fund through the HAICORE@KIT partition.
Support was also provided by the HammerHAI project, an EU co-funded AI Factory initiative operated by the High-Performance Computing Center Stuttgart.
This project has received funding from the European High Performance Computing Joint Undertaking (EuroHPC JU) under Grant Agreement No.~101234027.
It is jointly co-funded by the EuroHPC JU through the European Union's Digital Europe Programme, the European Commission, the German Federal Ministry of Research, Technology and Space (BMFTR), the Baden-Württemberg Ministry of Science, Research and the Arts, the Bavarian State Ministry of Science and the Arts, and the Lower Saxony Ministry of Science and Culture.
Views and opinions expressed are those of the author(s) only and do not necessarily reflect those of the European Union or the EuroHPC JU.
\end{acks}

\bibliographystyle{ACM-Reference-Format}
\bibliography{references}

\appendix
\newsavebox{\appendixmetadatabox}
\newsavebox{\appendixbodybox}
\newsavebox{\appendixcombinedbox}
\newlength{\appendixtablewidth}
\newlength{\appendixtableheight}
\newlength{\appendixbodyheight}
\newenvironment{appendixtable}[1][0]{%
    \begin{table*}[p]
    \centering
    \captionsetup{skip=4pt}%
    \def\appendixtableangle{#1}%
    \def\tableheaderangle{90}%
    \ifnum#1=90\relax\def\tableheaderangle{-90}\fi%
    \setlength{\appendixtablewidth}{\textwidth}%
    \setlength{\appendixtableheight}{\textheight}%
    \def\fittable##1{\global\setbox\appendixbodybox=\hbox{##1}}%
    \begin{lrbox}{\appendixmetadatabox}%
    \begin{minipage}{\appendixtablewidth}
    \centering
}{%
    \end{minipage}%
    \end{lrbox}%
    \setlength{\appendixbodyheight}{\dimexpr\appendixtableheight
        -\ht\appendixmetadatabox-\dp\appendixmetadatabox-0.1pt\relax}%
    \sbox{\appendixbodybox}{\rotatebox{\appendixtableangle}{\usebox{\appendixbodybox}}}%
    \sbox{\appendixbodybox}{\resizebox{\appendixtablewidth}{!}{\usebox{\appendixbodybox}}}%
    \ifdim\dimexpr\ht\appendixbodybox+\dp\appendixbodybox\relax>\appendixbodyheight
        \sbox{\appendixbodybox}{\resizebox*{!}{\appendixbodyheight}{\usebox{\appendixbodybox}}}%
    \fi
    \setbox\appendixcombinedbox=\vbox{%
        \offinterlineskip
        \hbox{\usebox{\appendixmetadatabox}}%
        \hbox to \appendixtablewidth{\hfil\usebox{\appendixbodybox}\hfil}}%
    \usebox{\appendixcombinedbox}%
    \end{table*}%
}

\newenvironment{appendixfigure}{%
    \par\addvspace{2pt}%
    \noindent\begin{minipage}{\textwidth}%
    \captionsetup{type=figure,skip=4pt}%
    \centering
}{%
    \end{minipage}\par\addvspace{2pt}%
}

\begin{appendixtable}[90]
    \raggedright
    \section{Per-Dataset Results}
    \label{sec:appendix-results}

    \subsection{Macro-F1}
    \label{sec:appendix-f1}

    \noindent Tables~\ref{tab:f1-results-w3s} to~\ref{tab:f1-results-w2s-part2} report the test macro-F1 of every
    architecture on every dataset, with one pair of tables per window length.
    \par\medskip
    \centering
    \caption{Test macro-F1 (\%) of every architecture on every dataset with
    3\,s windows, part 1 of 2 (the \FOneTabPartOneNumDatasetsThreeS{} datasets with the highest mean test macro-F1): mean over the test folds, with the standard deviation over folds below. Bold marks the best architecture
    per dataset. The last column gives the mean $\pm$ standard deviation
    over the architectures.}
    \label{tab:f1-results-w3s}
    \providecommand{\todocell}[1]{\begin{tabular}[c]{@{}c@{}}\textcolor{red}{TODO}\\[-2pt]{\scriptsize\textcolor{red}{#1}}\end{tabular}}
\providecommand{\fcell}[2]{\begin{tabular}[c]{@{}c@{}}#1\\[-2pt]{\scriptsize$\pm$#2}\end{tabular}}
\providecommand{\tableheaderangle}{90}
\providecommand{\tablemaxheight}{0.9\textheight}
\providecommand{\fittable}[1]{\sbox2{#1}%
  \sbox2{\resizebox{\linewidth}{!}{\usebox2}}%
  \dimen2=\ht2\advance\dimen2 by\dp2
  \ifdim\dimen2>\tablemaxheight\resizebox*{!}{\tablemaxheight}{\usebox2}\else\usebox2\fi}

{%
\setlength{\tabcolsep}{1.6pt}
\renewcommand{\arraystretch}{1.1}
\setlength{\aboverulesep}{0pt}\setlength{\belowrulesep}{0pt}
\footnotesize
\fittable{%
\begin{tabular}{@{}lccccccccccccccccccc|c@{}}
Dataset & \rotatebox{\tableheaderangle}{CNN-HAR\hspace{4pt}} & \rotatebox{\tableheaderangle}{Triple-Cross-Attn\hspace{4pt}} & \rotatebox{\tableheaderangle}{TinyHAR\hspace{4pt}} & \rotatebox{\tableheaderangle}{TinierHAR\hspace{4pt}} & \rotatebox{\tableheaderangle}{DANA\hspace{4pt}} & \rotatebox{\tableheaderangle}{LITEWAY-Light\hspace{4pt}} & \rotatebox{\tableheaderangle}{Attend-and-Discriminate\hspace{4pt}} & \rotatebox{\tableheaderangle}{DynamicWHAR\hspace{4pt}} & \rotatebox{\tableheaderangle}{DeepConvLSTM-Attn\hspace{4pt}} & \rotatebox{\tableheaderangle}{Random Forest\hspace{4pt}} & \rotatebox{\tableheaderangle}{MLP-HAR\hspace{4pt}} & \rotatebox{\tableheaderangle}{LITEWAY\hspace{4pt}} & \rotatebox{\tableheaderangle}{DeepConvLSTM\hspace{4pt}} & \rotatebox{\tableheaderangle}{MLP-Mixer\hspace{4pt}} & \rotatebox{\tableheaderangle}{Guan-LSTM\hspace{4pt}} & \rotatebox{\tableheaderangle}{SA-HAR\hspace{4pt}} & \rotatebox{\tableheaderangle}{DeepConvShallowLSTM\hspace{4pt}} & \rotatebox{\tableheaderangle}{SVM\hspace{4pt}} & \rotatebox{\tableheaderangle}{k-NN\hspace{4pt}} & \rotatebox{\tableheaderangle}{\textbf{Mean}\hspace{4pt}} \\
\midrule
\rowcolor{black!5}
SAD & \fcell{97.3}{4.1} & \fcell{94.8}{6.7} & \fcell{98.6}{2.3} & \fcell{94.5}{8.6} & \fcell{94.8}{8.5} & \fcell{\textbf{99.2}}{1.7} & \fcell{94.2}{7.3} & \fcell{95.1}{6.7} & \fcell{94.0}{7.7} & \fcell{98.2}{1.5} & \fcell{97.0}{4.0} & \fcell{98.5}{2.2} & \fcell{88.5}{16.6} & \fcell{92.5}{9.3} & \fcell{87.3}{12.6} & \fcell{86.7}{11.5} & \fcell{95.0}{9.6} & \fcell{90.6}{11.5} & \fcell{81.1}{12.0} & \fcell{93.6}{4.8} \\
UCI-HAR & \fcell{95.5}{2.5} & \fcell{95.8}{2.0} & \fcell{\textbf{96.1}}{3.2} & \fcell{95.7}{3.1} & \fcell{94.5}{3.8} & \fcell{94.8}{2.8} & \fcell{93.8}{4.0} & \fcell{90.6}{3.7} & \fcell{93.1}{4.1} & \fcell{84.0}{2.5} & \fcell{92.2}{4.5} & \fcell{85.2}{8.7} & \fcell{92.5}{4.5} & \fcell{94.1}{3.4} & \fcell{89.1}{11.2} & \fcell{91.7}{5.0} & \fcell{90.2}{4.4} & \fcell{69.8}{3.5} & \fcell{69.0}{6.0} & \fcell{89.9}{8.0} \\
\rowcolor{black!5}
MotionSense & \fcell{\textbf{93.3}}{3.7} & \fcell{93.2}{4.1} & \fcell{92.3}{4.7} & \fcell{92.0}{5.1} & \fcell{89.7}{6.1} & \fcell{91.7}{4.1} & \fcell{90.0}{4.7} & \fcell{92.8}{4.5} & \fcell{88.7}{5.8} & \fcell{91.2}{3.6} & \fcell{91.3}{4.3} & \fcell{92.1}{4.9} & \fcell{89.5}{6.2} & \fcell{92.6}{3.9} & \fcell{90.7}{5.8} & \fcell{86.4}{6.1} & \fcell{88.8}{5.1} & \fcell{77.0}{10.7} & \fcell{67.2}{5.9} & \fcell{89.0}{6.4} \\
MHEALTH & \fcell{89.8}{10.3} & \fcell{90.6}{7.5} & \fcell{91.9}{10.1} & \fcell{91.5}{8.1} & \fcell{91.0}{5.1} & \fcell{90.7}{7.7} & \fcell{91.7}{7.7} & \fcell{91.3}{8.7} & \fcell{89.2}{7.8} & \fcell{\textbf{95.7}}{4.5} & \fcell{88.3}{10.2} & \fcell{89.1}{12.4} & \fcell{83.1}{10.1} & \fcell{83.2}{8.9} & \fcell{86.1}{5.6} & \fcell{80.0}{14.0} & \fcell{87.8}{6.4} & \fcell{81.2}{10.7} & \fcell{69.1}{17.1} & \fcell{87.4}{6.0} \\
\rowcolor{black!5}
WEAR & \fcell{86.7}{8.6} & \fcell{87.6}{8.1} & \fcell{86.0}{8.8} & \fcell{85.6}{8.7} & \fcell{86.1}{9.8} & \fcell{86.9}{10.2} & \fcell{85.3}{9.9} & \fcell{\textbf{88.5}}{6.4} & \fcell{83.9}{11.0} & \fcell{86.1}{6.9} & \fcell{85.6}{7.6} & \fcell{86.6}{8.9} & \fcell{83.2}{11.3} & \fcell{84.1}{9.6} & \fcell{82.8}{12.7} & \fcell{82.9}{10.2} & \fcell{80.9}{8.8} & \fcell{75.0}{11.9} & \fcell{68.7}{12.3} & \fcell{83.8}{4.7} \\
DSADS & \fcell{87.4}{5.9} & \fcell{87.2}{7.0} & \fcell{82.9}{7.5} & \fcell{85.6}{7.6} & \fcell{81.5}{6.0} & \fcell{86.6}{6.7} & \fcell{87.0}{5.9} & \fcell{86.8}{6.5} & \fcell{82.3}{6.7} & \fcell{\textbf{90.2}}{4.5} & \fcell{85.2}{8.1} & \fcell{89.6}{5.6} & \fcell{79.5}{8.0} & \fcell{78.4}{6.4} & \fcell{82.5}{10.7} & \fcell{79.2}{10.4} & \fcell{86.0}{5.4} & \fcell{75.8}{6.2} & \fcell{69.8}{4.2} & \fcell{83.3}{5.1} \\
\rowcolor{black!5}
Opportunity (Loc.) & \fcell{84.4}{11.9} & \fcell{\textbf{91.6}}{4.0} & \fcell{78.0}{33.3} & \fcell{83.1}{18.0} & \fcell{75.6}{25.0} & \fcell{85.1}{19.1} & \fcell{78.4}{26.0} & \fcell{82.3}{22.9} & \fcell{77.7}{24.1} & \fcell{91.1}{9.2} & \fcell{89.6}{11.5} & \fcell{88.8}{10.1} & \fcell{76.8}{22.2} & \fcell{76.4}{18.1} & \fcell{80.5}{12.7} & \fcell{52.1}{44.4} & \fcell{77.7}{17.9} & \fcell{82.4}{12.8} & \fcell{74.9}{14.6} & \fcell{80.3}{8.7} \\
WISDM & \fcell{84.1}{8.8} & \fcell{84.1}{7.4} & \fcell{83.9}{8.9} & \fcell{83.2}{7.8} & \fcell{\textbf{84.8}}{10.0} & \fcell{79.0}{7.1} & \fcell{82.8}{9.6} & \fcell{80.1}{10.0} & \fcell{83.1}{10.2} & \fcell{69.0}{9.1} & \fcell{75.3}{8.4} & \fcell{76.1}{10.5} & \fcell{83.9}{10.3} & \fcell{82.7}{7.9} & \fcell{82.4}{10.4} & \fcell{63.8}{6.3} & \fcell{83.4}{7.9} & \fcell{58.0}{5.8} & \fcell{48.9}{9.2} & \fcell{77.3}{10.2} \\
\rowcolor{black!5}
REALDISP & \fcell{86.6}{10.7} & \fcell{88.7}{9.8} & \fcell{87.9}{10.7} & \fcell{86.5}{10.6} & \fcell{85.0}{10.4} & \fcell{\textbf{89.9}}{10.5} & \fcell{86.6}{9.9} & \fcell{67.0}{24.1} & \fcell{81.4}{12.1} & \fcell{85.6}{9.8} & \fcell{87.8}{10.3} & \fcell{89.2}{9.9} & \fcell{79.3}{13.1} & \fcell{58.0}{15.5} & \fcell{67.2}{15.4} & \fcell{71.4}{26.9} & \fcell{80.1}{8.7} & \fcell{49.7}{13.3} & \fcell{35.7}{8.0} & \fcell{77.0}{15.2} \\
PAMAP2 & \fcell{84.3}{13.6} & \fcell{78.9}{15.4} & \fcell{81.7}{12.0} & \fcell{86.0}{11.5} & \fcell{81.1}{12.4} & \fcell{78.9}{13.2} & \fcell{81.3}{14.0} & \fcell{\textbf{87.5}}{11.1} & \fcell{69.0}{25.2} & \fcell{87.1}{9.6} & \fcell{82.5}{13.5} & \fcell{81.8}{10.7} & \fcell{71.5}{15.7} & \fcell{66.0}{25.0} & \fcell{73.9}{19.0} & \fcell{69.5}{19.0} & \fcell{66.5}{20.0} & \fcell{59.5}{32.6} & \fcell{59.7}{30.1} & \fcell{76.1}{9.0} \\
\rowcolor{black!5}
HAPT & \fcell{\textbf{83.8}}{9.6} & \fcell{83.3}{7.2} & \fcell{83.1}{8.9} & \fcell{81.1}{7.4} & \fcell{80.8}{6.4} & \fcell{79.0}{8.7} & \fcell{80.9}{8.8} & \fcell{82.3}{8.3} & \fcell{79.7}{10.0} & \fcell{66.8}{6.8} & \fcell{81.4}{7.7} & \fcell{65.5}{11.3} & \fcell{78.5}{8.7} & \fcell{81.9}{11.0} & \fcell{70.6}{14.4} & \fcell{79.7}{10.7} & \fcell{61.3}{10.6} & \fcell{58.0}{7.9} & \fcell{67.4}{10.5} & \fcell{76.1}{8.2} \\
HHAR & \fcell{\textbf{82.6}}{7.1} & \fcell{81.9}{8.3} & \fcell{79.6}{8.7} & \fcell{78.6}{8.0} & \fcell{78.2}{12.2} & \fcell{82.5}{9.4} & \fcell{82.0}{9.7} & \fcell{79.0}{12.6} & \fcell{80.3}{9.3} & \fcell{74.1}{11.2} & \fcell{75.6}{11.0} & \fcell{81.9}{7.7} & \fcell{76.5}{8.9} & \fcell{79.1}{8.3} & \fcell{79.6}{13.9} & \fcell{71.5}{11.5} & \fcell{77.6}{11.1} & \fcell{55.8}{9.7} & \fcell{45.0}{10.6} & \fcell{75.9}{9.6} \\
\rowcolor{black!5}
Hang-Time & \fcell{77.4}{3.3} & \fcell{75.3}{4.1} & \fcell{76.4}{3.6} & \fcell{77.0}{3.3} & \fcell{\textbf{79.6}}{2.7} & \fcell{76.2}{3.6} & \fcell{78.7}{2.8} & \fcell{71.2}{3.2} & \fcell{77.7}{3.0} & \fcell{71.2}{4.2} & \fcell{71.5}{3.2} & \fcell{72.7}{5.2} & \fcell{77.1}{3.6} & \fcell{74.9}{3.2} & \fcell{77.2}{3.3} & \fcell{65.6}{4.8} & \fcell{69.8}{3.8} & \fcell{55.8}{3.6} & \fcell{58.0}{3.7} & \fcell{72.8}{6.6} \\
w-HAR & \fcell{81.4}{14.6} & \fcell{76.6}{16.0} & \fcell{79.0}{15.9} & \fcell{80.1}{10.1} & \fcell{78.1}{13.9} & \fcell{78.8}{15.4} & \fcell{76.7}{12.6} & \fcell{81.9}{14.6} & \fcell{73.8}{13.6} & \fcell{\textbf{84.4}}{14.4} & \fcell{69.6}{17.2} & \fcell{72.3}{14.3} & \fcell{73.0}{9.3} & \fcell{76.2}{9.2} & \fcell{54.3}{11.0} & \fcell{82.5}{11.1} & \fcell{55.3}{13.4} & \fcell{49.4}{15.2} & \fcell{50.9}{17.0} & \fcell{72.3}{11.2} \\
\rowcolor{black!5}
HuGaDB & \fcell{76.0}{14.0} & \fcell{72.8}{14.2} & \fcell{72.7}{15.5} & \fcell{71.4}{10.8} & \fcell{68.8}{13.5} & \fcell{74.7}{14.9} & \fcell{69.5}{15.1} & \fcell{68.2}{12.6} & \fcell{68.8}{13.0} & \fcell{\textbf{79.1}}{14.4} & \fcell{70.4}{15.5} & \fcell{76.8}{17.0} & \fcell{67.5}{13.0} & \fcell{68.9}{14.5} & \fcell{69.6}{14.8} & \fcell{67.0}{14.2} & \fcell{67.4}{12.6} & \fcell{65.6}{15.8} & \fcell{58.7}{12.3} & \fcell{70.2}{4.6} \\
\bottomrule
\end{tabular}}
}

\end{appendixtable}

\begin{appendixtable}[90]
    \centering
    \caption{Test macro-F1 (\%) of every architecture on every dataset with
    3\,s windows, part 2 of 2 (the remaining \FOneTabPartTwoNumDatasetsThreeS{} datasets, continued from
    \autoref{tab:f1-results-w3s}). The last row gives the mean $\pm$
    standard deviation over all \FOneTabNumDatasetsThreeS{} datasets.}
    \label{tab:f1-results-w3s-part2}
    \providecommand{\todocell}[1]{\begin{tabular}[c]{@{}c@{}}\textcolor{red}{TODO}\\[-2pt]{\scriptsize\textcolor{red}{#1}}\end{tabular}}
\providecommand{\fcell}[2]{\begin{tabular}[c]{@{}c@{}}#1\\[-2pt]{\scriptsize$\pm$#2}\end{tabular}}
\providecommand{\tableheaderangle}{90}
\providecommand{\tablemaxheight}{0.9\textheight}
\providecommand{\fittable}[1]{\sbox2{#1}%
  \sbox2{\resizebox{\linewidth}{!}{\usebox2}}%
  \dimen2=\ht2\advance\dimen2 by\dp2
  \ifdim\dimen2>\tablemaxheight\resizebox*{!}{\tablemaxheight}{\usebox2}\else\usebox2\fi}

{%
\setlength{\tabcolsep}{1.6pt}
\renewcommand{\arraystretch}{1.1}
\setlength{\aboverulesep}{0pt}\setlength{\belowrulesep}{0pt}
\footnotesize
\fittable{%
\begin{tabular}{@{}lccccccccccccccccccc|c@{}}
Dataset & \rotatebox{\tableheaderangle}{CNN-HAR\hspace{4pt}} & \rotatebox{\tableheaderangle}{Triple-Cross-Attn\hspace{4pt}} & \rotatebox{\tableheaderangle}{TinyHAR\hspace{4pt}} & \rotatebox{\tableheaderangle}{TinierHAR\hspace{4pt}} & \rotatebox{\tableheaderangle}{DANA\hspace{4pt}} & \rotatebox{\tableheaderangle}{LITEWAY-Light\hspace{4pt}} & \rotatebox{\tableheaderangle}{Attend-and-Discriminate\hspace{4pt}} & \rotatebox{\tableheaderangle}{DynamicWHAR\hspace{4pt}} & \rotatebox{\tableheaderangle}{DeepConvLSTM-Attn\hspace{4pt}} & \rotatebox{\tableheaderangle}{Random Forest\hspace{4pt}} & \rotatebox{\tableheaderangle}{MLP-HAR\hspace{4pt}} & \rotatebox{\tableheaderangle}{LITEWAY\hspace{4pt}} & \rotatebox{\tableheaderangle}{DeepConvLSTM\hspace{4pt}} & \rotatebox{\tableheaderangle}{MLP-Mixer\hspace{4pt}} & \rotatebox{\tableheaderangle}{Guan-LSTM\hspace{4pt}} & \rotatebox{\tableheaderangle}{SA-HAR\hspace{4pt}} & \rotatebox{\tableheaderangle}{DeepConvShallowLSTM\hspace{4pt}} & \rotatebox{\tableheaderangle}{SVM\hspace{4pt}} & \rotatebox{\tableheaderangle}{k-NN\hspace{4pt}} & \rotatebox{\tableheaderangle}{\textbf{Mean}\hspace{4pt}} \\
\midrule
\rowcolor{black!5}
HARTH & \fcell{78.5}{7.1} & \fcell{79.8}{7.0} & \fcell{77.8}{8.8} & \fcell{\textbf{80.2}}{6.7} & \fcell{77.7}{5.9} & \fcell{76.8}{7.4} & \fcell{70.9}{7.9} & \fcell{73.5}{6.6} & \fcell{68.2}{3.7} & \fcell{61.0}{5.9} & \fcell{69.1}{8.0} & \fcell{77.0}{5.7} & \fcell{67.9}{8.6} & \fcell{74.1}{7.9} & \fcell{75.0}{8.3} & \fcell{66.6}{10.4} & \fcell{55.5}{6.8} & \fcell{49.5}{7.7} & \fcell{45.7}{5.8} & \fcell{69.7}{10.2} \\
Daphnet & \fcell{73.3}{17.6} & \fcell{72.7}{16.7} & \fcell{74.0}{16.7} & \fcell{\textbf{75.4}}{15.3} & \fcell{73.7}{16.4} & \fcell{71.9}{18.1} & \fcell{73.3}{18.6} & \fcell{61.7}{21.9} & \fcell{75.3}{14.7} & \fcell{59.3}{22.2} & \fcell{67.2}{18.6} & \fcell{70.4}{23.3} & \fcell{68.4}{18.8} & \fcell{68.9}{17.6} & \fcell{55.0}{14.6} & \fcell{58.0}{25.0} & \fcell{59.5}{17.3} & \fcell{57.6}{22.4} & \fcell{54.5}{20.9} & \fcell{66.9}{7.4} \\
\rowcolor{black!5}
USC-HAD & \fcell{68.3}{13.2} & \fcell{72.7}{10.5} & \fcell{\textbf{73.2}}{13.5} & \fcell{70.8}{10.9} & \fcell{68.4}{13.9} & \fcell{70.5}{13.2} & \fcell{66.8}{14.2} & \fcell{72.6}{11.0} & \fcell{69.2}{15.1} & \fcell{67.8}{9.7} & \fcell{69.2}{12.5} & \fcell{59.0}{16.3} & \fcell{66.2}{14.6} & \fcell{73.0}{12.8} & \fcell{69.4}{14.9} & \fcell{65.0}{17.6} & \fcell{59.1}{11.5} & \fcell{33.8}{6.2} & \fcell{39.9}{9.2} & \fcell{65.0}{10.7} \\
HAR70+ & \fcell{\textbf{76.6}}{8.4} & \fcell{69.3}{7.3} & \fcell{72.5}{10.9} & \fcell{73.6}{6.6} & \fcell{67.2}{8.1} & \fcell{73.7}{9.6} & \fcell{71.5}{7.0} & \fcell{63.8}{8.4} & \fcell{68.4}{8.5} & \fcell{62.3}{8.7} & \fcell{61.1}{9.1} & \fcell{62.4}{12.7} & \fcell{63.9}{7.0} & \fcell{64.9}{8.9} & \fcell{62.2}{14.4} & \fcell{59.8}{8.4} & \fcell{59.7}{12.1} & \fcell{52.6}{10.0} & \fcell{47.2}{11.4} & \fcell{64.9}{7.4} \\
\rowcolor{black!5}
GOTOV & \fcell{69.0}{6.0} & \fcell{70.8}{4.9} & \fcell{69.0}{6.2} & \fcell{68.5}{6.6} & \fcell{69.4}{8.8} & \fcell{71.4}{7.3} & \fcell{68.1}{6.5} & \fcell{\textbf{72.7}}{5.4} & \fcell{61.8}{6.0} & \fcell{62.9}{5.0} & \fcell{65.2}{6.2} & \fcell{67.9}{8.3} & \fcell{61.6}{7.3} & \fcell{55.4}{7.1} & \fcell{64.7}{8.1} & \fcell{59.4}{8.0} & \fcell{52.2}{4.7} & \fcell{52.8}{5.8} & \fcell{45.8}{4.2} & \fcell{63.6}{7.5} \\
UTD-MHAD & \fcell{77.4}{7.6} & \fcell{\textbf{78.2}}{12.1} & \fcell{76.2}{9.2} & \fcell{67.4}{6.7} & \fcell{75.5}{7.4} & \fcell{61.9}{11.3} & \fcell{70.9}{11.2} & \fcell{76.1}{8.4} & \fcell{70.6}{8.7} & \fcell{58.6}{5.5} & \fcell{66.1}{9.7} & \fcell{39.6}{5.4} & \fcell{61.7}{7.1} & \fcell{75.1}{8.9} & \fcell{28.0}{11.8} & \fcell{67.6}{8.5} & \fcell{21.4}{4.1} & \fcell{24.6}{4.1} & \fcell{39.8}{5.4} & \fcell{59.8}{19.1} \\
\rowcolor{black!5}
WISDM-19 (Watch) & \fcell{62.4}{8.7} & \fcell{62.8}{8.5} & \fcell{62.9}{8.8} & \fcell{61.3}{8.4} & \fcell{\textbf{66.1}}{9.2} & \fcell{56.4}{6.9} & \fcell{64.0}{8.9} & \fcell{63.5}{8.9} & \fcell{62.9}{8.8} & \fcell{54.2}{8.6} & \fcell{56.0}{8.6} & \fcell{58.6}{7.1} & \fcell{61.6}{9.3} & \fcell{63.3}{9.2} & \fcell{65.2}{9.8} & \fcell{59.3}{9.5} & \fcell{60.7}{9.8} & \fcell{45.6}{8.1} & \fcell{40.6}{6.6} & \fcell{59.3}{6.6} \\
UCA-EHAR & \fcell{62.1}{8.6} & \fcell{\textbf{63.8}}{9.5} & \fcell{63.4}{7.0} & \fcell{59.4}{8.4} & \fcell{63.0}{8.2} & \fcell{62.5}{8.9} & \fcell{62.9}{9.3} & \fcell{61.3}{9.7} & \fcell{59.8}{10.2} & \fcell{55.6}{9.3} & \fcell{56.9}{7.8} & \fcell{57.6}{8.3} & \fcell{61.1}{9.6} & \fcell{60.5}{9.4} & \fcell{63.7}{11.1} & \fcell{54.7}{8.2} & \fcell{48.7}{6.8} & \fcell{37.6}{9.1} & \fcell{34.1}{8.3} & \fcell{57.3}{8.5} \\
\rowcolor{black!5}
UP-Fall & \fcell{57.0}{7.3} & \fcell{58.4}{5.4} & \fcell{59.9}{7.1} & \fcell{58.2}{7.7} & \fcell{53.6}{7.0} & \fcell{58.7}{6.0} & \fcell{56.9}{8.5} & \fcell{57.6}{3.3} & \fcell{53.3}{7.6} & \fcell{\textbf{64.1}}{5.6} & \fcell{59.4}{4.1} & \fcell{61.0}{7.0} & \fcell{53.4}{7.1} & \fcell{43.4}{8.0} & \fcell{44.1}{5.8} & \fcell{37.8}{19.9} & \fcell{55.7}{5.8} & \fcell{48.0}{9.8} & \fcell{38.3}{9.7} & \fcell{53.6}{7.7} \\
UMAFall & \fcell{59.1}{10.8} & \fcell{52.7}{9.8} & \fcell{58.4}{11.6} & \fcell{58.3}{9.9} & \fcell{59.0}{10.4} & \fcell{61.2}{13.0} & \fcell{56.9}{14.8} & \fcell{\textbf{62.8}}{11.4} & \fcell{54.5}{10.7} & \fcell{62.1}{12.1} & \fcell{52.2}{13.2} & \fcell{60.4}{11.8} & \fcell{49.7}{9.1} & \fcell{32.8}{10.3} & \fcell{40.0}{11.2} & \fcell{53.4}{10.0} & \fcell{46.3}{9.4} & \fcell{33.1}{12.8} & \fcell{32.4}{11.7} & \fcell{51.8}{10.2} \\
\rowcolor{black!5}
RealLifeHAR & \fcell{50.2}{19.9} & \fcell{47.8}{18.8} & \fcell{49.6}{20.2} & \fcell{48.2}{20.9} & \fcell{53.2}{22.9} & \fcell{46.6}{19.5} & \fcell{48.4}{19.1} & \fcell{55.1}{22.1} & \fcell{47.6}{20.8} & \fcell{\textbf{60.5}}{25.3} & \fcell{47.0}{19.2} & \fcell{51.5}{22.5} & \fcell{49.1}{20.4} & \fcell{41.9}{14.1} & \fcell{44.7}{19.6} & \fcell{46.3}{18.8} & \fcell{52.5}{21.3} & \fcell{48.6}{20.7} & \fcell{37.7}{19.3} & \fcell{48.8}{4.8} \\
UniMiB-SHAR & \fcell{\textbf{64.2}}{5.6} & \fcell{50.1}{3.7} & \fcell{55.5}{4.4} & \fcell{52.4}{6.8} & \fcell{60.0}{5.4} & \fcell{47.5}{5.6} & \fcell{62.6}{4.9} & \fcell{48.3}{5.9} & \fcell{60.7}{6.9} & \fcell{43.9}{3.3} & \fcell{39.3}{3.9} & \fcell{30.7}{8.7} & \fcell{55.1}{4.9} & \fcell{51.0}{4.6} & \fcell{40.3}{16.4} & \fcell{35.3}{5.4} & \fcell{36.3}{4.0} & \fcell{16.9}{3.0} & \fcell{22.2}{2.5} & \fcell{45.9}{13.4} \\
\rowcolor{black!5}
AICOS-HAR & \fcell{46.3}{6.4} & \fcell{\textbf{55.9}}{9.6} & \fcell{46.4}{7.3} & \fcell{45.0}{8.1} & \fcell{40.6}{6.4} & \fcell{42.9}{7.1} & \fcell{43.7}{8.2} & \fcell{52.1}{6.5} & \fcell{34.3}{6.0} & \fcell{37.0}{5.8} & \fcell{39.6}{7.0} & \fcell{44.7}{7.0} & \fcell{35.4}{6.8} & \fcell{45.5}{8.8} & \fcell{42.9}{8.6} & \fcell{37.5}{4.4} & \fcell{28.9}{7.0} & \fcell{18.8}{5.6} & \fcell{19.6}{6.0} & \fcell{39.9}{9.6} \\
WISDM-19 (Phone) & \fcell{23.6}{3.0} & \fcell{28.1}{2.7} & \fcell{25.5}{2.1} & \fcell{26.2}{3.1} & \fcell{26.6}{1.8} & \fcell{25.8}{2.2} & \fcell{26.1}{3.3} & \fcell{\textbf{28.6}}{2.4} & \fcell{25.5}{2.9} & \fcell{26.1}{2.6} & \fcell{21.5}{1.5} & \fcell{25.8}{2.2} & \fcell{23.9}{1.7} & \fcell{23.8}{2.8} & \fcell{23.1}{3.0} & \fcell{21.5}{2.6} & \fcell{26.4}{4.1} & \fcell{18.0}{3.1} & \fcell{15.8}{2.2} & \fcell{24.3}{3.3} \\
\midrule
\textbf{Mean} & \fcell{74.4}{16.4} & \fcell{74.0}{16.3} & \fcell{73.6}{16.1} & \fcell{73.0}{16.5} & \fcell{72.5}{15.7} & \fcell{72.5}{17.1} & \fcell{72.5}{15.8} & \fcell{72.2}{15.4} & \fcell{70.2}{16.3} & \fcell{70.0}{17.8} & \fcell{69.4}{18.1} & \fcell{69.4}{18.7} & \fcell{68.3}{15.9} & \fcell{67.7}{17.9} & \fcell{65.2}{18.8} & \fcell{63.9}{17.1} & \fcell{63.1}{19.8} & \fcell{53.1}{19.7} & \fcell{49.6}{17.1} & \fcell{68.1}{6.8} \\
\bottomrule
\end{tabular}}
}

\end{appendixtable}

\begin{appendixtable}[90]
    \centering
    \caption{Test macro-F1 (\%) of every architecture on every dataset with
    2\,s windows, part 1 of 2 (the \FOneTabPartOneNumDatasetsTwoS{} datasets with the highest mean test macro-F1): mean over the test folds, with the standard deviation over folds below. Bold marks the best architecture
    per dataset. The last column gives the mean $\pm$ standard deviation
    over the architectures.}
    \label{tab:f1-results-w2s}
    \providecommand{\todocell}[1]{\begin{tabular}[c]{@{}c@{}}\textcolor{red}{TODO}\\[-2pt]{\scriptsize\textcolor{red}{#1}}\end{tabular}}
\providecommand{\fcell}[2]{\begin{tabular}[c]{@{}c@{}}#1\\[-2pt]{\scriptsize$\pm$#2}\end{tabular}}
\providecommand{\tableheaderangle}{90}
\providecommand{\tablemaxheight}{0.9\textheight}
\providecommand{\fittable}[1]{\sbox2{#1}%
  \sbox2{\resizebox{\linewidth}{!}{\usebox2}}%
  \dimen2=\ht2\advance\dimen2 by\dp2
  \ifdim\dimen2>\tablemaxheight\resizebox*{!}{\tablemaxheight}{\usebox2}\else\usebox2\fi}

{%
\setlength{\tabcolsep}{1.6pt}
\renewcommand{\arraystretch}{1.1}
\setlength{\aboverulesep}{0pt}\setlength{\belowrulesep}{0pt}
\footnotesize
\fittable{%
\begin{tabular}{@{}lccccccccccccccccccc|c@{}}
Dataset & \rotatebox{\tableheaderangle}{Triple-Cross-Attn\hspace{4pt}} & \rotatebox{\tableheaderangle}{CNN-HAR\hspace{4pt}} & \rotatebox{\tableheaderangle}{TinyHAR\hspace{4pt}} & \rotatebox{\tableheaderangle}{DynamicWHAR\hspace{4pt}} & \rotatebox{\tableheaderangle}{TinierHAR\hspace{4pt}} & \rotatebox{\tableheaderangle}{LITEWAY-Light\hspace{4pt}} & \rotatebox{\tableheaderangle}{Attend-and-Discriminate\hspace{4pt}} & \rotatebox{\tableheaderangle}{DANA\hspace{4pt}} & \rotatebox{\tableheaderangle}{DeepConvLSTM-Attn\hspace{4pt}} & \rotatebox{\tableheaderangle}{LITEWAY\hspace{4pt}} & \rotatebox{\tableheaderangle}{MLP-HAR\hspace{4pt}} & \rotatebox{\tableheaderangle}{Random Forest\hspace{4pt}} & \rotatebox{\tableheaderangle}{DeepConvLSTM\hspace{4pt}} & \rotatebox{\tableheaderangle}{MLP-Mixer\hspace{4pt}} & \rotatebox{\tableheaderangle}{Guan-LSTM\hspace{4pt}} & \rotatebox{\tableheaderangle}{DeepConvShallowLSTM\hspace{4pt}} & \rotatebox{\tableheaderangle}{SA-HAR\hspace{4pt}} & \rotatebox{\tableheaderangle}{SVM\hspace{4pt}} & \rotatebox{\tableheaderangle}{k-NN\hspace{4pt}} & \rotatebox{\tableheaderangle}{\textbf{Mean}\hspace{4pt}} \\
\midrule
\rowcolor{black!5}
SAD & \fcell{96.2}{5.5} & \fcell{92.0}{10.9} & \fcell{96.4}{5.6} & \fcell{94.9}{5.9} & \fcell{93.8}{6.8} & \fcell{\textbf{98.4}}{3.0} & \fcell{91.9}{7.5} & \fcell{92.5}{13.9} & \fcell{89.7}{13.6} & \fcell{97.6}{5.2} & \fcell{96.6}{4.8} & \fcell{97.7}{2.1} & \fcell{87.9}{17.3} & \fcell{94.3}{7.0} & \fcell{87.5}{8.5} & \fcell{95.5}{5.7} & \fcell{82.3}{27.3} & \fcell{90.7}{11.8} & \fcell{80.9}{11.6} & \fcell{92.5}{5.0} \\
UCI-HAR & \fcell{95.3}{2.7} & \fcell{95.4}{2.8} & \fcell{\textbf{95.7}}{3.1} & \fcell{93.3}{3.1} & \fcell{95.5}{3.2} & \fcell{95.4}{2.2} & \fcell{93.3}{4.1} & \fcell{94.9}{2.7} & \fcell{93.1}{4.2} & \fcell{89.2}{9.5} & \fcell{92.6}{4.0} & \fcell{84.4}{1.9} & \fcell{92.9}{3.9} & \fcell{94.7}{3.7} & \fcell{92.8}{3.9} & \fcell{91.1}{3.6} & \fcell{91.9}{3.8} & \fcell{76.2}{4.1} & \fcell{72.4}{4.4} & \fcell{91.1}{6.5} \\
\rowcolor{black!5}
MotionSense & \fcell{92.2}{4.0} & \fcell{92.0}{3.8} & \fcell{91.2}{5.0} & \fcell{\textbf{93.6}}{3.7} & \fcell{91.8}{3.8} & \fcell{91.0}{4.9} & \fcell{87.8}{6.6} & \fcell{90.2}{5.3} & \fcell{87.4}{6.0} & \fcell{91.5}{5.1} & \fcell{90.0}{4.3} & \fcell{89.4}{4.1} & \fcell{86.9}{7.5} & \fcell{91.2}{4.6} & \fcell{91.1}{6.2} & \fcell{89.4}{4.0} & \fcell{86.3}{8.4} & \fcell{78.1}{10.0} & \fcell{66.8}{6.7} & \fcell{88.3}{6.2} \\
MHEALTH & \fcell{90.6}{6.9} & \fcell{93.9}{5.2} & \fcell{90.7}{7.8} & \fcell{91.4}{5.6} & \fcell{89.7}{6.4} & \fcell{93.9}{5.5} & \fcell{91.3}{7.1} & \fcell{86.9}{9.1} & \fcell{88.6}{8.4} & \fcell{92.6}{5.6} & \fcell{87.8}{8.2} & \fcell{\textbf{94.9}}{5.4} & \fcell{85.8}{7.4} & \fcell{81.5}{8.3} & \fcell{86.7}{5.1} & \fcell{89.2}{5.0} & \fcell{83.9}{10.4} & \fcell{79.0}{10.3} & \fcell{69.0}{14.5} & \fcell{87.8}{6.2} \\
\rowcolor{black!5}
DSADS & \fcell{\textbf{88.1}}{6.4} & \fcell{87.3}{6.0} & \fcell{85.4}{4.6} & \fcell{88.1}{6.5} & \fcell{85.6}{6.9} & \fcell{87.5}{5.1} & \fcell{86.7}{5.3} & \fcell{82.7}{9.4} & \fcell{83.1}{7.4} & \fcell{87.5}{5.5} & \fcell{84.6}{6.6} & \fcell{87.7}{5.0} & \fcell{83.6}{7.2} & \fcell{78.0}{6.7} & \fcell{83.0}{6.8} & \fcell{85.1}{6.6} & \fcell{79.7}{8.4} & \fcell{76.7}{6.0} & \fcell{70.0}{3.9} & \fcell{83.7}{4.7} \\
WEAR & \fcell{85.2}{7.7} & \fcell{82.9}{8.1} & \fcell{83.3}{8.7} & \fcell{\textbf{87.4}}{6.6} & \fcell{83.6}{8.5} & \fcell{84.8}{8.8} & \fcell{82.0}{10.8} & \fcell{83.6}{8.4} & \fcell{81.8}{10.0} & \fcell{84.1}{9.0} & \fcell{84.0}{8.0} & \fcell{83.9}{7.4} & \fcell{81.8}{10.4} & \fcell{80.9}{12.1} & \fcell{81.5}{10.3} & \fcell{82.1}{8.0} & \fcell{80.6}{10.9} & \fcell{73.6}{11.7} & \fcell{66.8}{12.0} & \fcell{81.8}{4.5} \\
\rowcolor{black!5}
Opportunity (Loc.) & \fcell{\textbf{92.4}}{4.4} & \fcell{82.3}{18.4} & \fcell{83.3}{21.9} & \fcell{85.6}{19.3} & \fcell{82.7}{23.1} & \fcell{86.2}{12.3} & \fcell{79.7}{20.8} & \fcell{72.9}{23.4} & \fcell{71.4}{30.9} & \fcell{80.9}{19.1} & \fcell{90.5}{5.0} & \fcell{88.0}{8.1} & \fcell{79.4}{16.3} & \fcell{84.2}{13.3} & \fcell{73.5}{21.3} & \fcell{75.8}{21.7} & \fcell{53.8}{41.8} & \fcell{84.2}{11.8} & \fcell{73.6}{14.8} & \fcell{80.0}{8.7} \\
REALDISP & \fcell{\textbf{89.5}}{10.1} & \fcell{87.2}{10.2} & \fcell{88.5}{11.1} & \fcell{71.0}{27.6} & \fcell{86.9}{10.4} & \fcell{89.0}{11.0} & \fcell{87.9}{10.1} & \fcell{86.8}{10.7} & \fcell{82.7}{11.8} & \fcell{87.6}{11.4} & \fcell{85.8}{10.5} & \fcell{84.3}{10.0} & \fcell{80.5}{12.6} & \fcell{72.7}{12.8} & \fcell{69.1}{14.9} & \fcell{82.0}{11.4} & \fcell{73.4}{27.4} & \fcell{51.7}{13.4} & \fcell{36.7}{9.2} & \fcell{78.6}{14.0} \\
\rowcolor{black!5}
WISDM & \fcell{82.0}{8.4} & \fcell{81.3}{9.5} & \fcell{81.3}{8.4} & \fcell{82.4}{7.6} & \fcell{81.5}{8.7} & \fcell{79.1}{5.9} & \fcell{80.2}{9.9} & \fcell{81.9}{9.7} & \fcell{\textbf{83.5}}{8.4} & \fcell{75.4}{3.6} & \fcell{77.0}{7.5} & \fcell{69.2}{9.2} & \fcell{81.7}{8.0} & \fcell{81.6}{9.8} & \fcell{82.5}{8.4} & \fcell{81.9}{7.7} & \fcell{62.4}{6.5} & \fcell{62.7}{6.2} & \fcell{52.0}{9.4} & \fcell{76.8}{8.8} \\
HHAR & \fcell{79.2}{7.8} & \fcell{79.3}{9.4} & \fcell{80.4}{6.9} & \fcell{79.3}{10.0} & \fcell{78.7}{11.3} & \fcell{80.7}{9.2} & \fcell{79.9}{9.2} & \fcell{77.7}{11.5} & \fcell{80.8}{6.8} & \fcell{78.0}{9.8} & \fcell{74.3}{8.6} & \fcell{73.8}{10.3} & \fcell{77.9}{6.7} & \fcell{79.5}{7.7} & \fcell{\textbf{80.9}}{8.0} & \fcell{80.0}{10.3} & \fcell{77.0}{9.4} & \fcell{64.9}{10.7} & \fcell{50.1}{12.6} & \fcell{76.4}{7.4} \\
\rowcolor{black!5}
PAMAP2 & \fcell{74.2}{16.7} & \fcell{81.5}{15.7} & \fcell{69.7}{28.4} & \fcell{81.6}{12.5} & \fcell{83.3}{10.1} & \fcell{79.3}{15.0} & \fcell{81.6}{11.0} & \fcell{81.0}{12.1} & \fcell{76.0}{18.2} & \fcell{79.3}{15.2} & \fcell{80.6}{13.7} & \fcell{\textbf{85.6}}{9.5} & \fcell{71.7}{14.9} & \fcell{60.6}{21.8} & \fcell{72.7}{20.8} & \fcell{71.4}{14.4} & \fcell{75.7}{16.5} & \fcell{58.6}{31.7} & \fcell{58.0}{29.6} & \fcell{74.9}{8.3} \\
HAPT & \fcell{79.9}{5.6} & \fcell{\textbf{80.4}}{6.3} & \fcell{80.2}{6.6} & \fcell{75.9}{6.6} & \fcell{78.3}{5.4} & \fcell{73.9}{8.8} & \fcell{77.9}{6.3} & \fcell{80.2}{5.9} & \fcell{78.3}{7.8} & \fcell{67.8}{12.7} & \fcell{77.3}{4.5} & \fcell{68.0}{6.4} & \fcell{77.5}{7.8} & \fcell{79.1}{5.8} & \fcell{73.6}{12.0} & \fcell{64.4}{9.7} & \fcell{75.1}{6.8} & \fcell{65.8}{8.9} & \fcell{65.1}{7.4} & \fcell{74.7}{5.6} \\
\rowcolor{black!5}
w-HAR & \fcell{77.5}{13.0} & \fcell{80.7}{9.2} & \fcell{80.7}{10.0} & \fcell{77.2}{12.9} & \fcell{79.3}{12.1} & \fcell{75.7}{9.4} & \fcell{77.5}{10.4} & \fcell{76.6}{10.8} & \fcell{75.7}{12.2} & \fcell{71.8}{14.7} & \fcell{72.7}{13.4} & \fcell{\textbf{81.1}}{11.8} & \fcell{71.9}{14.4} & \fcell{78.7}{10.6} & \fcell{68.4}{17.3} & \fcell{57.1}{12.7} & \fcell{77.8}{13.1} & \fcell{50.7}{16.5} & \fcell{52.9}{16.7} & \fcell{72.9}{9.3} \\
Hang-Time & \fcell{73.6}{3.5} & \fcell{75.0}{3.8} & \fcell{75.1}{2.8} & \fcell{69.2}{4.0} & \fcell{74.2}{3.5} & \fcell{74.0}{4.5} & \fcell{75.4}{3.3} & \fcell{76.1}{3.2} & \fcell{76.2}{2.9} & \fcell{70.4}{4.5} & \fcell{69.7}{3.6} & \fcell{70.7}{4.0} & \fcell{74.9}{3.7} & \fcell{73.5}{3.3} & \fcell{\textbf{76.7}}{3.5} & \fcell{69.0}{3.0} & \fcell{61.6}{6.1} & \fcell{56.1}{3.6} & \fcell{58.7}{3.0} & \fcell{71.1}{6.0} \\
\rowcolor{black!5}
HuGaDB & \fcell{73.1}{13.0} & \fcell{71.8}{14.3} & \fcell{72.5}{14.0} & \fcell{69.7}{13.5} & \fcell{71.5}{12.7} & \fcell{75.7}{14.6} & \fcell{70.2}{14.1} & \fcell{71.0}{14.6} & \fcell{69.4}{14.6} & \fcell{76.2}{13.1} & \fcell{70.5}{12.8} & \fcell{\textbf{77.2}}{13.5} & \fcell{69.2}{15.0} & \fcell{69.1}{12.3} & \fcell{69.3}{11.6} & \fcell{65.1}{12.4} & \fcell{70.4}{12.5} & \fcell{65.5}{15.1} & \fcell{59.4}{11.1} & \fcell{70.4}{4.1} \\
\bottomrule
\end{tabular}}
}

\end{appendixtable}

\begin{appendixtable}[90]
    \centering
    \caption{Test macro-F1 (\%) of every architecture on every dataset with
    2\,s windows, part 2 of 2 (the remaining \FOneTabPartTwoNumDatasetsTwoS{} datasets, continued from
    \autoref{tab:f1-results-w2s}). The last row gives the mean $\pm$
    standard deviation over all \FOneTabNumDatasetsTwoS{} datasets.}
    \label{tab:f1-results-w2s-part2}
    \providecommand{\todocell}[1]{\begin{tabular}[c]{@{}c@{}}\textcolor{red}{TODO}\\[-2pt]{\scriptsize\textcolor{red}{#1}}\end{tabular}}
\providecommand{\fcell}[2]{\begin{tabular}[c]{@{}c@{}}#1\\[-2pt]{\scriptsize$\pm$#2}\end{tabular}}
\providecommand{\tableheaderangle}{90}
\providecommand{\tablemaxheight}{0.9\textheight}
\providecommand{\fittable}[1]{\sbox2{#1}%
  \sbox2{\resizebox{\linewidth}{!}{\usebox2}}%
  \dimen2=\ht2\advance\dimen2 by\dp2
  \ifdim\dimen2>\tablemaxheight\resizebox*{!}{\tablemaxheight}{\usebox2}\else\usebox2\fi}

{%
\setlength{\tabcolsep}{1.6pt}
\renewcommand{\arraystretch}{1.1}
\setlength{\aboverulesep}{0pt}\setlength{\belowrulesep}{0pt}
\footnotesize
\fittable{%
\begin{tabular}{@{}lccccccccccccccccccc|c@{}}
Dataset & \rotatebox{\tableheaderangle}{Triple-Cross-Attn\hspace{4pt}} & \rotatebox{\tableheaderangle}{CNN-HAR\hspace{4pt}} & \rotatebox{\tableheaderangle}{TinyHAR\hspace{4pt}} & \rotatebox{\tableheaderangle}{DynamicWHAR\hspace{4pt}} & \rotatebox{\tableheaderangle}{TinierHAR\hspace{4pt}} & \rotatebox{\tableheaderangle}{LITEWAY-Light\hspace{4pt}} & \rotatebox{\tableheaderangle}{Attend-and-Discriminate\hspace{4pt}} & \rotatebox{\tableheaderangle}{DANA\hspace{4pt}} & \rotatebox{\tableheaderangle}{DeepConvLSTM-Attn\hspace{4pt}} & \rotatebox{\tableheaderangle}{LITEWAY\hspace{4pt}} & \rotatebox{\tableheaderangle}{MLP-HAR\hspace{4pt}} & \rotatebox{\tableheaderangle}{Random Forest\hspace{4pt}} & \rotatebox{\tableheaderangle}{DeepConvLSTM\hspace{4pt}} & \rotatebox{\tableheaderangle}{MLP-Mixer\hspace{4pt}} & \rotatebox{\tableheaderangle}{Guan-LSTM\hspace{4pt}} & \rotatebox{\tableheaderangle}{DeepConvShallowLSTM\hspace{4pt}} & \rotatebox{\tableheaderangle}{SA-HAR\hspace{4pt}} & \rotatebox{\tableheaderangle}{SVM\hspace{4pt}} & \rotatebox{\tableheaderangle}{k-NN\hspace{4pt}} & \rotatebox{\tableheaderangle}{\textbf{Mean}\hspace{4pt}} \\
\midrule
\rowcolor{black!5}
HARTH & \fcell{76.5}{6.2} & \fcell{76.5}{7.9} & \fcell{75.6}{7.2} & \fcell{71.6}{4.3} & \fcell{\textbf{77.5}}{7.2} & \fcell{74.2}{8.0} & \fcell{68.1}{7.0} & \fcell{73.5}{8.7} & \fcell{67.7}{8.2} & \fcell{75.6}{5.2} & \fcell{68.8}{7.6} & \fcell{59.6}{5.5} & \fcell{66.5}{7.6} & \fcell{71.8}{9.7} & \fcell{73.9}{5.8} & \fcell{54.3}{7.3} & \fcell{63.9}{5.4} & \fcell{49.6}{7.8} & \fcell{47.1}{5.9} & \fcell{68.0}{9.2} \\
Daphnet & \fcell{69.2}{17.1} & \fcell{68.5}{18.8} & \fcell{73.1}{16.1} & \fcell{64.9}{20.2} & \fcell{75.2}{14.2} & \fcell{71.3}{19.0} & \fcell{\textbf{76.3}}{12.8} & \fcell{71.1}{16.3} & \fcell{72.2}{17.3} & \fcell{69.4}{20.8} & \fcell{66.8}{17.9} & \fcell{59.4}{22.0} & \fcell{67.7}{19.2} & \fcell{70.4}{17.4} & \fcell{60.6}{23.4} & \fcell{63.5}{18.5} & \fcell{58.3}{19.3} & \fcell{57.4}{22.5} & \fcell{56.6}{21.0} & \fcell{66.9}{6.1} \\
\rowcolor{black!5}
USC-HAD & \fcell{\textbf{74.5}}{9.5} & \fcell{69.6}{13.0} & \fcell{73.4}{12.2} & \fcell{73.2}{11.9} & \fcell{69.1}{12.1} & \fcell{70.9}{11.5} & \fcell{68.1}{14.1} & \fcell{68.0}{14.6} & \fcell{70.2}{13.3} & \fcell{66.4}{13.2} & \fcell{69.8}{12.2} & \fcell{68.0}{10.4} & \fcell{69.9}{14.3} & \fcell{73.6}{11.6} & \fcell{71.1}{14.1} & \fcell{59.3}{12.4} & \fcell{64.7}{16.7} & \fcell{38.7}{7.4} & \fcell{42.6}{11.0} & \fcell{66.4}{9.7} \\
HAR70+ & \fcell{70.7}{8.0} & \fcell{74.7}{6.7} & \fcell{74.8}{5.9} & \fcell{65.9}{11.2} & \fcell{\textbf{75.7}}{7.5} & \fcell{69.9}{9.5} & \fcell{64.9}{6.1} & \fcell{62.8}{9.6} & \fcell{64.5}{8.2} & \fcell{62.0}{7.7} & \fcell{62.4}{8.3} & \fcell{61.9}{7.3} & \fcell{62.3}{7.2} & \fcell{68.5}{8.1} & \fcell{68.5}{9.8} & \fcell{56.6}{6.0} & \fcell{58.9}{7.0} & \fcell{50.9}{7.9} & \fcell{44.7}{10.4} & \fcell{64.2}{7.9} \\
\rowcolor{black!5}
GOTOV & \fcell{71.3}{4.8} & \fcell{68.2}{6.2} & \fcell{68.3}{7.9} & \fcell{\textbf{74.6}}{6.1} & \fcell{67.2}{6.4} & \fcell{68.3}{4.8} & \fcell{65.8}{7.6} & \fcell{69.7}{7.1} & \fcell{63.7}{7.1} & \fcell{69.7}{7.0} & \fcell{66.6}{6.0} & \fcell{67.4}{6.1} & \fcell{63.1}{6.4} & \fcell{58.8}{8.5} & \fcell{62.5}{10.5} & \fcell{54.5}{4.9} & \fcell{60.7}{9.4} & \fcell{53.1}{6.3} & \fcell{45.3}{4.8} & \fcell{64.1}{7.1} \\
UTD-MHAD & \fcell{67.6}{7.6} & \fcell{\textbf{72.9}}{6.9} & \fcell{66.2}{7.0} & \fcell{68.6}{7.6} & \fcell{67.0}{6.6} & \fcell{65.7}{4.6} & \fcell{72.7}{6.9} & \fcell{72.7}{5.0} & \fcell{67.7}{6.8} & \fcell{48.9}{5.2} & \fcell{57.7}{5.9} & \fcell{54.5}{7.0} & \fcell{58.5}{4.2} & \fcell{66.9}{5.8} & \fcell{43.0}{18.1} & \fcell{39.0}{3.3} & \fcell{63.7}{4.2} & \fcell{34.8}{3.9} & \fcell{41.1}{3.1} & \fcell{59.4}{12.3} \\
\rowcolor{black!5}
WISDM-19 (Watch) & \fcell{61.5}{8.0} & \fcell{61.4}{8.5} & \fcell{61.4}{8.6} & \fcell{61.9}{8.8} & \fcell{59.4}{7.7} & \fcell{54.4}{6.6} & \fcell{62.5}{8.5} & \fcell{\textbf{63.6}}{9.3} & \fcell{61.4}{8.7} & \fcell{53.7}{6.5} & \fcell{56.2}{8.1} & \fcell{51.9}{8.3} & \fcell{59.9}{8.8} & \fcell{61.6}{9.0} & \fcell{63.3}{9.3} & \fcell{56.4}{8.6} & \fcell{57.7}{9.5} & \fcell{46.1}{8.0} & \fcell{41.1}{6.4} & \fcell{57.7}{6.0} \\
UCA-EHAR & \fcell{63.4}{6.1} & \fcell{61.6}{6.0} & \fcell{63.7}{6.0} & \fcell{61.3}{8.3} & \fcell{58.6}{7.0} & \fcell{56.5}{7.5} & \fcell{60.3}{7.8} & \fcell{64.1}{6.8} & \fcell{58.8}{7.5} & \fcell{54.6}{6.1} & \fcell{55.6}{7.3} & \fcell{51.0}{7.8} & \fcell{60.6}{6.2} & \fcell{62.4}{6.3} & \fcell{\textbf{65.0}}{7.9} & \fcell{47.7}{5.2} & \fcell{56.4}{5.2} & \fcell{41.7}{8.9} & \fcell{35.1}{6.7} & \fcell{56.7}{8.0} \\
\rowcolor{black!5}
UP-Fall & \fcell{59.7}{6.6} & \fcell{53.2}{9.3} & \fcell{59.7}{5.5} & \fcell{63.8}{5.5} & \fcell{57.5}{6.8} & \fcell{59.3}{7.5} & \fcell{52.4}{9.2} & \fcell{49.3}{8.6} & \fcell{51.3}{7.3} & \fcell{59.8}{5.6} & \fcell{57.6}{8.0} & \fcell{\textbf{64.0}}{5.2} & \fcell{54.8}{7.8} & \fcell{43.8}{8.4} & \fcell{44.2}{3.9} & \fcell{55.5}{6.1} & \fcell{42.3}{17.7} & \fcell{49.2}{8.6} & \fcell{39.0}{6.7} & \fcell{53.5}{7.3} \\
UMAFall & \fcell{52.5}{10.2} & \fcell{55.4}{10.2} & \fcell{54.0}{10.2} & \fcell{\textbf{59.7}}{11.0} & \fcell{56.5}{9.6} & \fcell{57.7}{11.4} & \fcell{53.6}{11.8} & \fcell{54.5}{9.6} & \fcell{49.1}{10.6} & \fcell{57.5}{11.0} & \fcell{52.8}{11.4} & \fcell{56.8}{11.8} & \fcell{49.5}{10.5} & \fcell{34.9}{11.3} & \fcell{42.5}{8.7} & \fcell{44.4}{8.2} & \fcell{47.3}{7.1} & \fcell{34.8}{12.7} & \fcell{33.2}{10.6} & \fcell{49.8}{8.3} \\
\rowcolor{black!5}
RealLifeHAR & \fcell{48.3}{20.0} & \fcell{50.8}{19.7} & \fcell{49.1}{20.2} & \fcell{53.9}{21.9} & \fcell{47.3}{20.6} & \fcell{52.3}{18.9} & \fcell{47.5}{20.5} & \fcell{49.2}{21.2} & \fcell{48.7}{18.0} & \fcell{52.9}{21.3} & \fcell{51.3}{20.7} & \fcell{\textbf{60.5}}{24.6} & \fcell{46.2}{21.0} & \fcell{41.3}{18.2} & \fcell{44.4}{20.8} & \fcell{55.0}{22.9} & \fcell{45.0}{21.3} & \fcell{48.9}{20.7} & \fcell{37.8}{19.0} & \fcell{49.0}{5.1} \\
UniMiB-SHAR & \fcell{45.8}{2.7} & \fcell{55.4}{4.2} & \fcell{52.9}{3.8} & \fcell{48.6}{4.5} & \fcell{49.3}{3.7} & \fcell{42.8}{2.8} & \fcell{\textbf{55.7}}{4.2} & \fcell{55.3}{4.3} & \fcell{54.7}{4.6} & \fcell{32.0}{5.0} & \fcell{35.3}{2.5} & \fcell{36.0}{2.7} & \fcell{51.1}{5.4} & \fcell{46.3}{3.8} & \fcell{36.9}{20.5} & \fcell{36.7}{3.3} & \fcell{32.4}{3.7} & \fcell{14.1}{1.8} & \fcell{18.0}{1.7} & \fcell{42.1}{12.3} \\
\rowcolor{black!5}
AICOS-HAR & \fcell{\textbf{53.5}}{9.9} & \fcell{46.9}{8.2} & \fcell{48.6}{8.1} & \fcell{51.6}{8.2} & \fcell{45.5}{8.0} & \fcell{44.7}{7.4} & \fcell{44.0}{6.7} & \fcell{38.2}{6.7} & \fcell{37.8}{5.3} & \fcell{42.3}{4.8} & \fcell{40.5}{6.0} & \fcell{37.7}{6.0} & \fcell{38.2}{4.2} & \fcell{45.9}{7.6} & \fcell{44.0}{5.4} & \fcell{30.1}{6.6} & \fcell{37.4}{5.0} & \fcell{26.7}{7.7} & \fcell{21.1}{5.0} & \fcell{40.8}{8.2} \\
WISDM-19 (Phone) & \fcell{26.3}{2.5} & \fcell{23.0}{2.6} & \fcell{24.5}{2.2} & \fcell{\textbf{28.0}}{1.7} & \fcell{25.1}{2.2} & \fcell{24.3}{4.3} & \fcell{25.0}{2.3} & \fcell{25.3}{2.7} & \fcell{23.8}{2.7} & \fcell{24.6}{2.2} & \fcell{22.1}{1.5} & \fcell{25.2}{2.4} & \fcell{23.7}{2.5} & \fcell{22.9}{3.0} & \fcell{22.4}{3.5} & \fcell{25.6}{2.8} & \fcell{20.7}{2.3} & \fcell{18.0}{2.6} & \fcell{16.2}{2.6} & \fcell{23.5}{2.8} \\
\midrule
\textbf{Mean} & \fcell{72.7}{16.6} & \fcell{72.5}{16.4} & \fcell{72.4}{16.2} & \fcell{72.0}{15.3} & \fcell{72.0}{16.6} & \fcell{71.6}{17.4} & \fcell{71.0}{16.2} & \fcell{70.8}{16.3} & \fcell{69.3}{16.3} & \fcell{68.9}{18.0} & \fcell{68.9}{17.9} & \fcell{68.6}{18.1} & \fcell{68.1}{16.1} & \fcell{67.9}{17.8} & \fcell{66.6}{17.9} & \fcell{64.0}{19.0} & \fcell{63.5}{17.2} & \fcell{55.1}{19.1} & \fcell{50.0}{17.0} & \fcell{67.7}{6.0} \\
\bottomrule
\end{tabular}}
}

\end{appendixtable}

\clearpage
\begin{figure*}[p]
    \raggedright
    \subsection{Rank per Dataset}
    \label{sec:appendix-ranks}

    \noindent Figures~\ref{fig:rank-heatmap} and~\ref{fig:rank-heatmap-2s} show the rank of every architecture on every dataset for 3\,s and 2\,s windows, respectively.
    \par\medskip
    \centering
    \includegraphics[width=\textwidth]{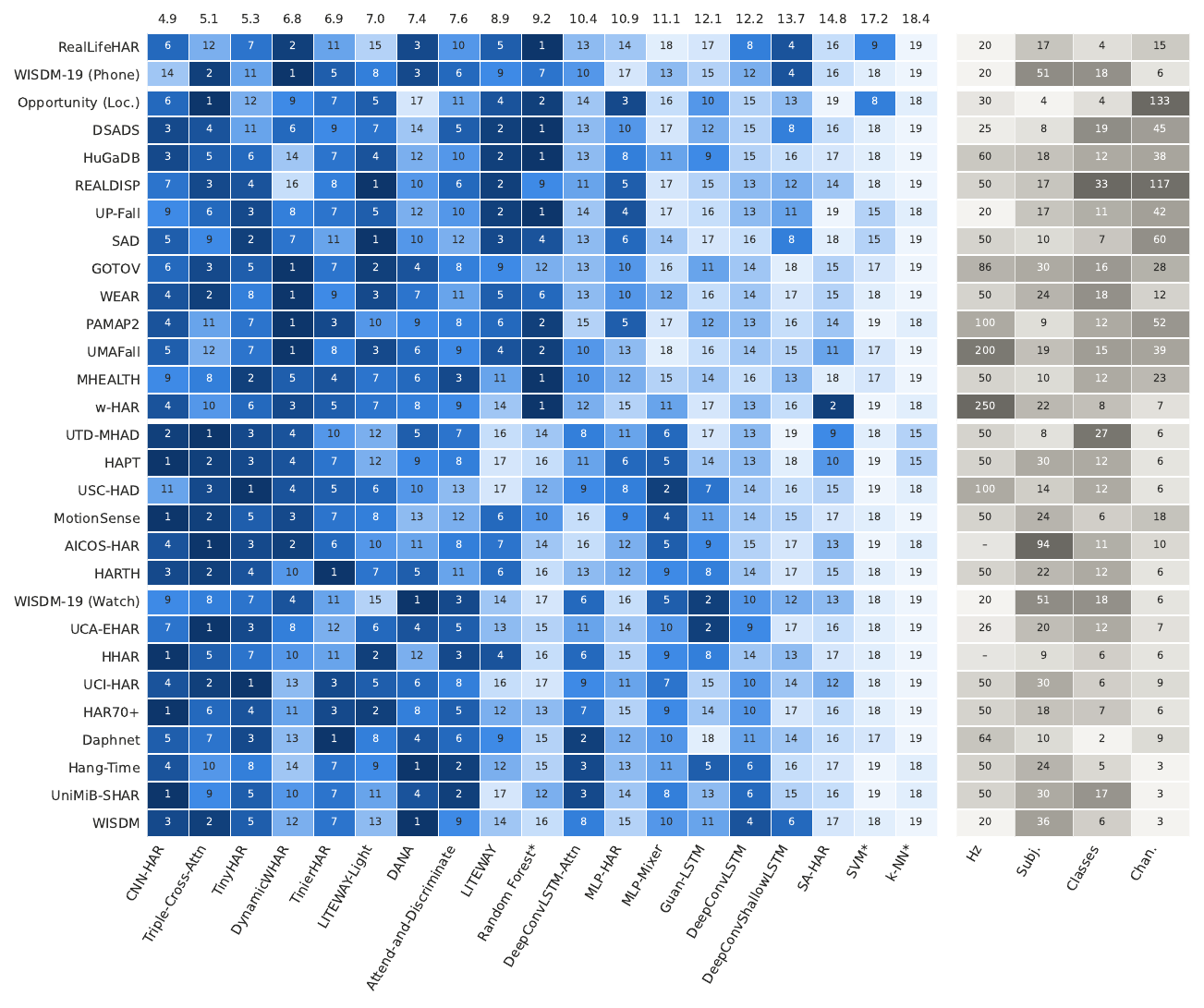}
    \caption{Rank of each architecture (columns) on each dataset (rows) for 3\,s windows, by test macro-F1 (1 = best, darker = better).
    Columns are sorted by mean rank over the \RankNumDatasets{} datasets, given above each column.
    Rows are grouped by how similarly the datasets rank the architectures (average-linkage clustering on 1 $-$ Spearman correlation), with white gaps separating the \RankNumClusters{} groups.
    The grey columns on the right give the sampling rate, the number of subjects, classes, and sensor channels.
    Classical baselines are marked with *.}
    \label{fig:rank-heatmap}
\end{figure*}

\begin{figure*}[p]
    \centering
    \includegraphics[width=\textwidth]{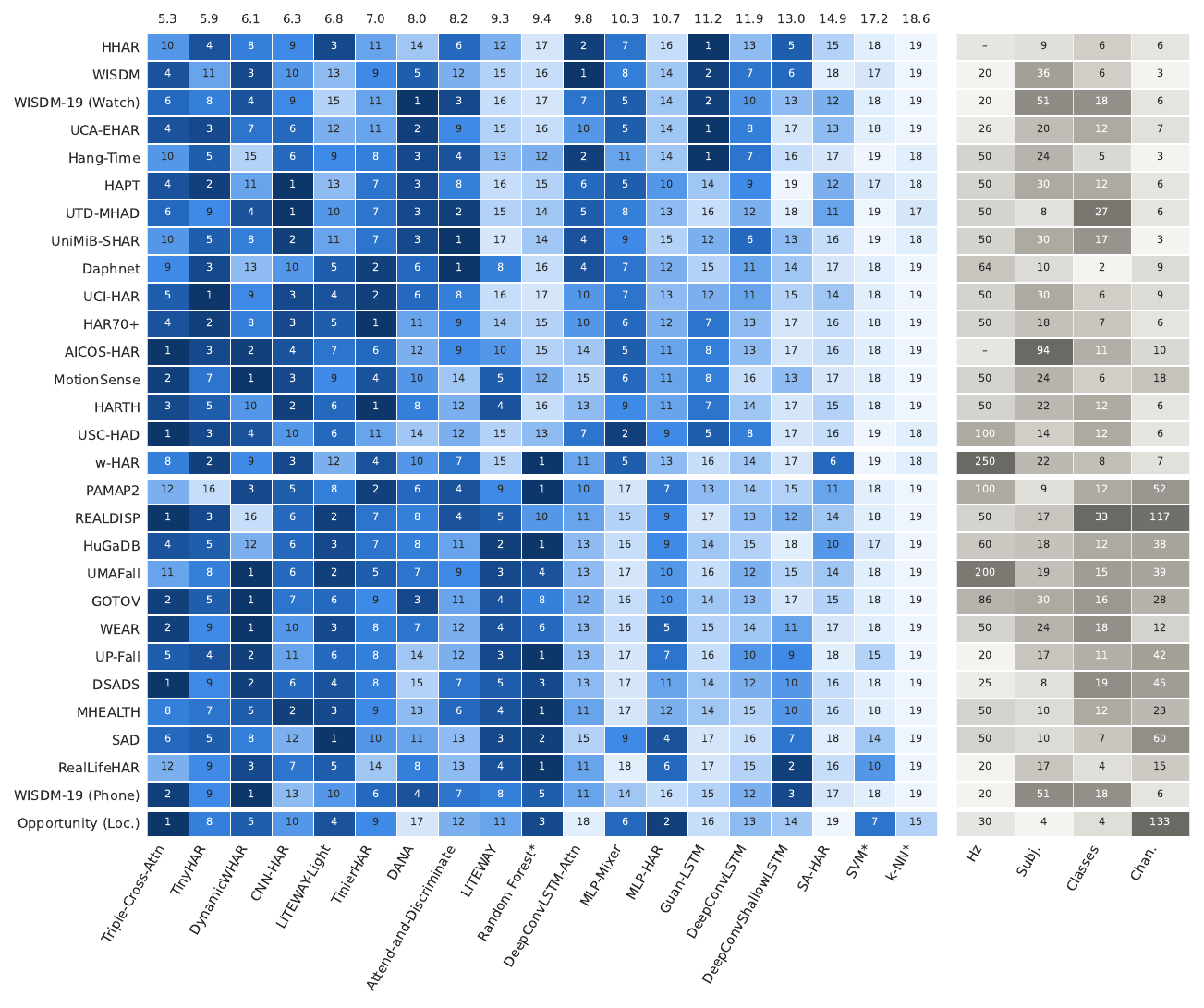}
    \caption{Rank of each architecture (columns) on each dataset (rows) for 2\,s windows, by test macro-F1 (1 = best, darker = better).
    Columns are sorted by mean rank over the \RankNumDatasets{} datasets, given above each column.
    Rows are grouped by how similarly the datasets rank the architectures (average-linkage clustering on 1 $-$ Spearman correlation), with white gaps separating the \RankNumClusters{} groups.
    The grey columns on the right give the sampling rate, the number of subjects, classes, and sensor channels.
    Classical baselines are marked with *.}
    \label{fig:rank-heatmap-2s}
\end{figure*}

\clearpage

\begin{appendixtable}
    \raggedright
    \subsection{Deployment Cost}
    \label{sec:appendix-deployment}

    \noindent Tables~\ref{tab:deploy-phone-latency-w2s} to~\ref{tab:deploy-mcu-energy-w3s}
    report every deployment measurement for every architecture, dataset, and
    window length on the smartphone, the smartwatch, and the MCU. On the
    smartphone and the smartwatch, latency is the mean over all timed inferences,
    with the standard deviation below. On the MCU, repeated runs differ by only a
    few CPU cycles, so we report a single measurement without a standard deviation.
    On the MCU, F and R mark architectures that do not fit in flash or RAM,
    respectively, and Failed marks an export or execution failure.
    The last column and the last row give the mean $\pm$
    standard deviation over the architectures and over the datasets; both leave
    out cells without a value (F, R, Failed, and unmeasured cells).

    \autoref{tab:efficiency} summarizes these measurements per architecture and window length.

    \subsubsection{Smartphone}
    \label{sec:appendix-deploy-phone}

    \noindent Tables~\ref{tab:deploy-phone-latency-w2s} to~\ref{tab:deploy-phone-memory-w3s} report latency and peak memory overhead on the smartphone.
    \par\medskip
    \centering
    \caption{Latency (ms) to classify one window on the smartphone with 2\,s windows, part 1 of 2: mean over all timed inferences, with the standard deviation below. Bold marks the lowest mean per dataset.}
    \label{tab:deploy-phone-latency-w2s}
    \providecommand{\nofit}[1]{\textcolor{gray}{#1}}
\providecommand{\faileddeployment}{\textcolor{gray}{Failed}}
\providecommand{\todomissing}{\textcolor{red}{TODO}}
\providecommand{\unmeasured}{\textcolor{gray}{--}}
\providecommand{\fcell}[2]{\begin{tabular}[c]{@{}c@{}}#1\\[-2pt]{\scriptsize$\pm$#2}\end{tabular}}
\providecommand{\tableheaderangle}{90}
\providecommand{\tablemaxheight}{0.9\textheight}
\providecommand{\fittable}[1]{\sbox2{#1}%
  \sbox2{\resizebox{\linewidth}{!}{\usebox2}}%
  \dimen2=\ht2\advance\dimen2 by\dp2
  \ifdim\dimen2>\tablemaxheight\resizebox*{!}{\tablemaxheight}{\usebox2}\else\usebox2\fi}

{%
\setlength{\tabcolsep}{2pt}
\setlength{\aboverulesep}{0pt}\setlength{\belowrulesep}{0pt}
\renewcommand{\arraystretch}{1.15}
\footnotesize
\fittable{%
\begin{tabular}{@{}lccccccccccccccccccc|c@{}}
Dataset & \rotatebox{\tableheaderangle}{CNN-HAR\hspace{4pt}} & \rotatebox{\tableheaderangle}{Triple-Cross-Attn\hspace{4pt}} & \rotatebox{\tableheaderangle}{TinyHAR\hspace{4pt}} & \rotatebox{\tableheaderangle}{TinierHAR\hspace{4pt}} & \rotatebox{\tableheaderangle}{DANA\hspace{4pt}} & \rotatebox{\tableheaderangle}{LITEWAY-Light\hspace{4pt}} & \rotatebox{\tableheaderangle}{Attend-and-Discriminate\hspace{4pt}} & \rotatebox{\tableheaderangle}{DynamicWHAR\hspace{4pt}} & \rotatebox{\tableheaderangle}{DeepConvLSTM-Attn\hspace{4pt}} & \rotatebox{\tableheaderangle}{Random Forest\hspace{4pt}} & \rotatebox{\tableheaderangle}{MLP-HAR\hspace{4pt}} & \rotatebox{\tableheaderangle}{LITEWAY\hspace{4pt}} & \rotatebox{\tableheaderangle}{DeepConvLSTM\hspace{4pt}} & \rotatebox{\tableheaderangle}{MLP-Mixer\hspace{4pt}} & \rotatebox{\tableheaderangle}{Guan-LSTM\hspace{4pt}} & \rotatebox{\tableheaderangle}{SA-HAR\hspace{4pt}} & \rotatebox{\tableheaderangle}{DeepConvShallowLSTM\hspace{4pt}} & \rotatebox{\tableheaderangle}{SVM\hspace{4pt}} & \rotatebox{\tableheaderangle}{k-NN\hspace{4pt}} & \rotatebox{\tableheaderangle}{\textbf{Mean}\hspace{4pt}} \\
\midrule
\rowcolor{black!5}
SAD & \fcell{19.0}{3.90} & \fcell{21.9}{3.33} & \fcell{11.5}{1.60} & \fcell{11.4}{11.5} & \fcell{31.8}{8.87} & \fcell{2.27}{0.79} & \fcell{49.7}{2.27} & \fcell{36.7}{5.90} & \fcell{50.7}{4.25} & \fcell{\textbf{0.94}}{1.25} & \fcell{7.74}{1.24} & \fcell{2.09}{0.50} & \fcell{45.8}{5.28} & \fcell{13.5}{5.96} & \fcell{29.9}{0.88} & \fcell{12.9}{1.52} & \fcell{23.7}{0.19} & \fcell{1.97}{0.14} & \fcell{27.2}{2.62} & \fcell{21.1}{16.3} \\
UCI-HAR & \fcell{4.25}{0.68} & \fcell{5.26}{0.96} & \fcell{3.80}{0.84} & \fcell{2.21}{0.57} & \fcell{12.6}{2.24} & \fcell{1.01}{0.54} & \fcell{26.5}{1.56} & \fcell{6.00}{0.96} & \fcell{27.0}{1.21} & \fcell{\textbf{0.51}}{0.34} & \fcell{1.71}{0.67} & \fcell{1.66}{2.25} & \fcell{27.1}{0.29} & \fcell{10.6}{1.85} & \fcell{28.5}{0.68} & \fcell{9.09}{1.58} & \fcell{17.7}{3.15} & \fcell{0.85}{0.11} & \fcell{18.3}{2.20} & \fcell{10.8}{10.2} \\
\rowcolor{black!5}
MotionSense & \fcell{8.91}{4.08} & \fcell{9.43}{1.30} & \fcell{5.47}{0.98} & \fcell{3.81}{2.56} & \fcell{22.4}{3.09} & \fcell{0.96}{0.34} & \fcell{33.3}{2.08} & \fcell{12.3}{1.61} & \fcell{34.7}{3.59} & \fcell{\textbf{0.56}}{0.24} & \fcell{1.93}{0.47} & \fcell{1.42}{0.44} & \fcell{34.9}{1.27} & \fcell{17.5}{2.46} & \fcell{30.6}{1.08} & \fcell{10.9}{1.47} & \fcell{9.34}{4.01} & \fcell{1.76}{0.15} & \fcell{46.0}{3.62} & \fcell{15.1}{14.3} \\
MHEALTH & \fcell{8.16}{1.64} & \fcell{10.0}{1.73} & \fcell{7.58}{1.79} & \fcell{2.35}{0.48} & \fcell{21.8}{7.90} & \fcell{1.49}{2.57} & \fcell{39.6}{7.99} & \fcell{14.7}{1.03} & \fcell{34.1}{3.67} & \fcell{\textbf{0.59}}{0.51} & \fcell{2.51}{0.41} & \fcell{1.59}{0.52} & \fcell{32.9}{0.86} & \fcell{18.5}{4.82} & \fcell{41.2}{0.85} & \fcell{11.2}{1.17} & \fcell{16.5}{0.94} & \fcell{1.14}{0.14} & \fcell{11.7}{1.92} & \fcell{14.6}{13.5} \\
\rowcolor{black!5}
WEAR & \fcell{4.69}{0.60} & \fcell{6.99}{1.59} & \fcell{5.49}{2.78} & \fcell{2.43}{0.82} & \fcell{9.81}{3.01} & \fcell{1.04}{0.35} & \fcell{26.1}{0.70} & \fcell{7.79}{0.51} & \fcell{39.5}{1.62} & \fcell{\textbf{0.54}}{0.38} & \fcell{1.89}{1.52} & \fcell{1.46}{0.54} & \fcell{30.5}{3.50} & \fcell{12.7}{1.96} & \fcell{31.5}{1.02} & \fcell{9.66}{1.74} & \fcell{21.1}{2.09} & \fcell{3.00}{0.36} & \fcell{70.1}{9.44} & \fcell{15.1}{17.9} \\
DSADS & \fcell{8.50}{2.36} & \fcell{11.2}{3.29} & \fcell{4.78}{0.67} & \fcell{1.74}{0.40} & \fcell{26.8}{4.07} & \fcell{1.17}{0.48} & \fcell{24.4}{2.07} & \fcell{32.8}{2.21} & \fcell{21.2}{2.02} & \fcell{\textbf{0.63}}{0.53} & \fcell{4.86}{4.10} & \fcell{1.62}{1.27} & \fcell{23.1}{4.85} & \fcell{20.0}{2.37} & \fcell{16.4}{0.60} & \fcell{7.82}{2.03} & \fcell{21.4}{4.34} & \fcell{2.88}{0.41} & \fcell{23.9}{2.04} & \fcell{13.4}{10.4} \\
\rowcolor{black!5}
Opportunity (Loc.) & \fcell{17.5}{3.27} & \fcell{15.0}{3.48} & \fcell{13.7}{3.91} & \fcell{3.27}{0.67} & \fcell{36.3}{1.42} & \fcell{2.48}{0.44} & \fcell{55.0}{1.68} & \fcell{216}{16.7} & \fcell{46.9}{2.71} & \fcell{\textbf{0.98}}{0.59} & \fcell{27.4}{3.40} & \fcell{9.36}{7.90} & \fcell{51.7}{6.64} & \fcell{39.7}{7.33} & \fcell{20.2}{1.88} & \fcell{9.72}{1.45} & \fcell{25.9}{2.46} & \fcell{3.16}{0.80} & \fcell{31.3}{1.73} & \fcell{33.0}{47.6} \\
WISDM & \fcell{1.47}{0.40} & \fcell{5.12}{4.63} & \fcell{2.52}{2.12} & \fcell{2.60}{2.68} & \fcell{5.84}{1.10} & \fcell{1.54}{4.19} & \fcell{14.3}{1.15} & \fcell{2.51}{0.42} & \fcell{12.8}{1.46} & \fcell{\textbf{0.45}}{0.38} & \fcell{1.50}{2.14} & \fcell{1.32}{0.76} & \fcell{12.4}{2.43} & \fcell{5.98}{0.63} & \fcell{16.2}{0.45} & \fcell{5.29}{1.18} & \fcell{4.35}{1.33} & \fcell{1.01}{0.14} & \fcell{81.2}{6.91} & \fcell{9.39}{18.1} \\
\rowcolor{black!5}
REALDISP & \fcell{32.3}{3.06} & \fcell{39.7}{4.59} & \fcell{19.3}{4.93} & \fcell{5.57}{3.95} & \fcell{56.1}{5.60} & \fcell{3.27}{0.52} & \fcell{80.3}{1.28} & \fcell{171}{12.7} & \fcell{52.9}{1.08} & \fcell{\textbf{1.33}}{1.50} & \fcell{21.2}{0.96} & \fcell{4.14}{1.11} & \fcell{54.5}{1.00} & \fcell{37.4}{4.52} & \fcell{32.1}{2.31} & \fcell{17.8}{1.86} & \fcell{40.9}{2.35} & \fcell{19.6}{1.94} & \fcell{83.5}{11.5} & \fcell{40.7}{39.7} \\
PAMAP2 & \fcell{36.2}{3.44} & \fcell{18.3}{4.20} & \fcell{14.0}{2.40} & \fcell{4.92}{0.53} & \fcell{29.0}{1.82} & \fcell{2.97}{0.74} & \fcell{93.0}{0.34} & \fcell{51.2}{2.52} & \fcell{99.3}{1.21} & \fcell{\textbf{1.49}}{1.87} & \fcell{7.86}{0.52} & \fcell{3.91}{0.53} & \fcell{76.2}{2.18} & \fcell{16.9}{1.95} & \fcell{61.8}{3.68} & \fcell{26.1}{2.51} & \fcell{40.1}{4.08} & \fcell{3.44}{0.47} & \fcell{37.6}{2.52} & \fcell{32.9}{30.8} \\
\rowcolor{black!5}
HAPT & \fcell{3.17}{0.72} & \fcell{5.21}{0.97} & \fcell{4.12}{0.83} & \fcell{1.83}{0.42} & \fcell{10.9}{1.38} & \fcell{0.95}{0.33} & \fcell{32.4}{0.68} & \fcell{4.35}{1.16} & \fcell{33.6}{0.84} & \fcell{\textbf{0.50}}{0.43} & \fcell{1.39}{0.76} & \fcell{1.19}{0.52} & \fcell{28.0}{1.48} & \fcell{10.1}{3.06} & \fcell{31.7}{0.92} & \fcell{8.33}{1.04} & \fcell{12.6}{3.66} & \fcell{1.00}{0.16} & \fcell{16.1}{2.23} & \fcell{10.9}{11.8} \\
HHAR & \fcell{2.22}{0.50} & \fcell{3.79}{0.89} & \fcell{2.64}{1.33} & \fcell{21.4}{10.9} & \fcell{7.87}{2.48} & \fcell{0.70}{0.29} & \fcell{16.0}{4.10} & \fcell{3.89}{0.61} & \fcell{14.7}{0.96} & \fcell{\textbf{0.48}}{0.49} & \fcell{1.33}{1.08} & \fcell{1.32}{0.73} & \fcell{13.5}{2.29} & \fcell{7.28}{5.40} & \fcell{15.5}{0.64} & \fcell{5.78}{0.79} & \fcell{5.03}{0.92} & \fcell{1.72}{0.24} & \fcell{93.4}{9.68} & \fcell{11.5}{20.8} \\
\rowcolor{black!5}
Hang-Time & \fcell{2.83}{2.20} & \fcell{4.11}{0.98} & \fcell{3.83}{1.62} & \fcell{2.22}{2.08} & \fcell{7.40}{1.70} & \fcell{1.00}{0.41} & \fcell{25.9}{1.02} & \fcell{2.79}{0.51} & \fcell{26.5}{1.44} & \fcell{\textbf{0.70}}{0.97} & \fcell{1.27}{0.36} & \fcell{1.22}{0.55} & \fcell{31.9}{1.94} & \fcell{7.54}{1.14} & \fcell{28.9}{1.41} & \fcell{9.83}{1.98} & \fcell{8.25}{2.13} & \fcell{1.26}{0.28} & \fcell{112}{14.5} & \fcell{14.7}{25.8} \\
w-HAR & \fcell{19.4}{3.42} & \fcell{13.3}{1.56} & \fcell{14.8}{1.97} & \fcell{6.60}{1.81} & \fcell{31.2}{4.28} & \fcell{5.27}{4.50} & \fcell{140}{5.80} & \fcell{11.2}{1.63} & \fcell{152}{1.71} & \fcell{\textbf{0.65}}{0.31} & \fcell{7.41}{1.01} & \fcell{1.94}{0.71} & \fcell{141}{1.99} & \fcell{14.1}{3.06} & \fcell{133}{2.25} & \fcell{42.0}{20.4} & \fcell{16.6}{1.67} & \fcell{0.83}{0.12} & \fcell{10.2}{0.78} & \fcell{40.1}{54.8} \\
\rowcolor{black!5}
HuGaDB & \fcell{17.6}{3.16} & \fcell{17.3}{1.80} & \fcell{7.45}{3.15} & \fcell{2.89}{0.61} & \fcell{27.8}{8.23} & \fcell{1.62}{1.00} & \fcell{43.5}{1.07} & \fcell{31.3}{1.92} & \fcell{55.0}{4.06} & \fcell{\textbf{0.80}}{0.79} & \fcell{4.62}{0.63} & \fcell{2.45}{1.53} & \fcell{45.2}{1.15} & \fcell{17.0}{5.02} & \fcell{35.0}{0.82} & \fcell{15.3}{2.05} & \fcell{17.5}{1.51} & \fcell{2.18}{0.24} & \fcell{46.4}{6.76} & \fcell{20.6}{17.6} \\
\bottomrule
\end{tabular}}
}

\end{appendixtable}

\begin{appendixtable}
    \centering
    \caption{Latency (ms) to classify one window on the smartphone with 2\,s windows, part 2 of 2 (continued from \autoref{tab:deploy-phone-latency-w2s}): mean over all timed inferences, with the standard deviation below. Bold marks the lowest mean per dataset. The last row summarizes all datasets across both parts.}
    \label{tab:deploy-phone-latency-w2s-part2}
    \providecommand{\nofit}[1]{\textcolor{gray}{#1}}
\providecommand{\faileddeployment}{\textcolor{gray}{Failed}}
\providecommand{\todomissing}{\textcolor{red}{TODO}}
\providecommand{\unmeasured}{\textcolor{gray}{--}}
\providecommand{\fcell}[2]{\begin{tabular}[c]{@{}c@{}}#1\\[-2pt]{\scriptsize$\pm$#2}\end{tabular}}
\providecommand{\tableheaderangle}{90}
\providecommand{\tablemaxheight}{0.9\textheight}
\providecommand{\fittable}[1]{\sbox2{#1}%
  \sbox2{\resizebox{\linewidth}{!}{\usebox2}}%
  \dimen2=\ht2\advance\dimen2 by\dp2
  \ifdim\dimen2>\tablemaxheight\resizebox*{!}{\tablemaxheight}{\usebox2}\else\usebox2\fi}

{%
\setlength{\tabcolsep}{2pt}
\setlength{\aboverulesep}{0pt}\setlength{\belowrulesep}{0pt}
\renewcommand{\arraystretch}{1.15}
\footnotesize
\fittable{%
\begin{tabular}{@{}lccccccccccccccccccc|c@{}}
Dataset & \rotatebox{\tableheaderangle}{CNN-HAR\hspace{4pt}} & \rotatebox{\tableheaderangle}{Triple-Cross-Attn\hspace{4pt}} & \rotatebox{\tableheaderangle}{TinyHAR\hspace{4pt}} & \rotatebox{\tableheaderangle}{TinierHAR\hspace{4pt}} & \rotatebox{\tableheaderangle}{DANA\hspace{4pt}} & \rotatebox{\tableheaderangle}{LITEWAY-Light\hspace{4pt}} & \rotatebox{\tableheaderangle}{Attend-and-Discriminate\hspace{4pt}} & \rotatebox{\tableheaderangle}{DynamicWHAR\hspace{4pt}} & \rotatebox{\tableheaderangle}{DeepConvLSTM-Attn\hspace{4pt}} & \rotatebox{\tableheaderangle}{Random Forest\hspace{4pt}} & \rotatebox{\tableheaderangle}{MLP-HAR\hspace{4pt}} & \rotatebox{\tableheaderangle}{LITEWAY\hspace{4pt}} & \rotatebox{\tableheaderangle}{DeepConvLSTM\hspace{4pt}} & \rotatebox{\tableheaderangle}{MLP-Mixer\hspace{4pt}} & \rotatebox{\tableheaderangle}{Guan-LSTM\hspace{4pt}} & \rotatebox{\tableheaderangle}{SA-HAR\hspace{4pt}} & \rotatebox{\tableheaderangle}{DeepConvShallowLSTM\hspace{4pt}} & \rotatebox{\tableheaderangle}{SVM\hspace{4pt}} & \rotatebox{\tableheaderangle}{k-NN\hspace{4pt}} & \rotatebox{\tableheaderangle}{\textbf{Mean}\hspace{4pt}} \\
\midrule
\rowcolor{black!5}
HARTH & \fcell{3.78}{1.58} & \fcell{5.39}{1.97} & \fcell{3.92}{0.80} & \fcell{4.19}{7.07} & \fcell{15.1}{7.60} & \fcell{1.08}{0.55} & \fcell{25.5}{0.87} & \fcell{5.20}{3.82} & \fcell{28.7}{0.48} & \fcell{\textbf{0.49}}{0.48} & \fcell{5.56}{5.45} & \fcell{1.34}{0.69} & \fcell{34.7}{2.38} & \fcell{8.97}{0.92} & \fcell{31.9}{1.90} & \fcell{9.74}{1.39} & \fcell{12.5}{2.48} & \fcell{1.51}{0.28} & \fcell{176}{20.2} & \fcell{19.8}{39.5} \\
Daphnet & \fcell{5.03}{0.82} & \fcell{5.26}{0.98} & \fcell{4.92}{0.61} & \fcell{2.23}{0.70} & \fcell{17.3}{2.10} & \fcell{1.16}{0.35} & \fcell{41.9}{1.53} & \fcell{6.36}{0.86} & \fcell{41.1}{1.55} & \fcell{0.51}{0.55} & \fcell{2.12}{3.76} & \fcell{1.36}{0.38} & \fcell{39.0}{1.01} & \fcell{11.1}{3.54} & \fcell{36.3}{0.80} & \fcell{9.30}{2.40} & \fcell{16.6}{6.18} & \fcell{\textbf{0.49}}{0.08} & \fcell{26.1}{2.08} & \fcell{14.1}{15.1} \\
\rowcolor{black!5}
USC-HAD & \fcell{5.53}{1.03} & \fcell{5.64}{0.75} & \fcell{6.58}{2.15} & \fcell{2.89}{0.45} & \fcell{17.1}{3.92} & \fcell{1.56}{2.67} & \fcell{53.7}{1.73} & \fcell{7.43}{4.12} & \fcell{60.2}{0.77} & \fcell{\textbf{0.57}}{0.46} & \fcell{2.27}{2.23} & \fcell{1.49}{0.88} & \fcell{60.1}{1.25} & \fcell{12.6}{1.26} & \fcell{78.5}{1.75} & \fcell{19.6}{2.32} & \fcell{18.0}{2.40} & \fcell{1.28}{0.11} & \fcell{37.3}{3.70} & \fcell{20.7}{24.6} \\
HAR70+ & \fcell{3.30}{1.55} & \fcell{5.02}{1.41} & \fcell{4.27}{1.10} & \fcell{2.12}{0.84} & \fcell{11.3}{1.42} & \fcell{1.14}{0.58} & \fcell{26.2}{1.13} & \fcell{4.29}{0.47} & \fcell{32.6}{2.32} & \fcell{\textbf{0.46}}{0.15} & \fcell{1.49}{0.44} & \fcell{1.25}{0.53} & \fcell{28.8}{2.32} & \fcell{8.95}{1.06} & \fcell{31.5}{0.96} & \fcell{9.65}{1.50} & \fcell{10.7}{2.36} & \fcell{0.72}{0.08} & \fcell{62.2}{7.02} & \fcell{12.9}{16.3} \\
\rowcolor{black!5}
GOTOV & \fcell{20.5}{2.96} & \fcell{18.9}{2.53} & \fcell{13.3}{2.89} & \fcell{3.56}{0.39} & \fcell{37.1}{4.31} & \fcell{1.99}{1.66} & \fcell{57.8}{1.57} & \fcell{22.6}{2.52} & \fcell{56.4}{2.83} & \fcell{\textbf{0.79}}{0.48} & \fcell{4.34}{1.26} & \fcell{16.5}{7.29} & \fcell{61.6}{0.31} & \fcell{14.1}{3.58} & \fcell{53.4}{0.68} & \fcell{17.2}{1.56} & \fcell{22.8}{2.31} & \fcell{7.22}{1.57} & \fcell{121}{8.36} & \fcell{29.0}{29.8} \\
UTD-MHAD & \fcell{3.04}{0.57} & \fcell{4.59}{0.70} & \fcell{3.74}{0.48} & \fcell{2.27}{0.96} & \fcell{11.3}{2.81} & \fcell{1.01}{0.73} & \fcell{25.3}{2.11} & \fcell{4.14}{0.43} & \fcell{28.3}{4.54} & \fcell{\textbf{0.47}}{0.09} & \fcell{1.53}{0.76} & \fcell{1.38}{1.09} & \fcell{30.7}{2.87} & \fcell{10.0}{3.32} & \fcell{34.9}{1.14} & \fcell{9.52}{1.54} & \fcell{13.7}{1.82} & \fcell{2.63}{0.71} & \fcell{2.92}{0.38} & \fcell{10.1}{11.2} \\
\rowcolor{black!5}
WISDM-19 (Watch) & \fcell{2.68}{3.49} & \fcell{4.26}{1.13} & \fcell{2.39}{1.16} & \fcell{1.43}{0.43} & \fcell{8.12}{1.23} & \fcell{1.87}{4.26} & \fcell{12.8}{0.64} & \fcell{3.91}{1.15} & \fcell{16.7}{1.64} & \fcell{\textbf{0.49}}{0.17} & \fcell{1.29}{0.53} & \fcell{1.79}{4.34} & \fcell{16.1}{1.71} & \fcell{6.77}{1.05} & \fcell{20.1}{2.28} & \fcell{5.87}{0.83} & \fcell{4.95}{0.75} & \fcell{8.60}{0.77} & \fcell{318}{26.5} & \fcell{23.0}{71.6} \\
UCA-EHAR & \fcell{2.28}{0.57} & \fcell{3.57}{0.86} & \fcell{2.36}{0.57} & \fcell{2.27}{3.94} & \fcell{7.64}{0.97} & \fcell{1.01}{0.34} & \fcell{16.1}{1.43} & \fcell{4.03}{0.60} & \fcell{17.2}{1.46} & \fcell{\textbf{0.49}}{0.24} & \fcell{1.37}{0.43} & \fcell{1.42}{0.70} & \fcell{16.1}{1.24} & \fcell{7.62}{1.14} & \fcell{18.3}{0.73} & \fcell{6.14}{0.71} & \fcell{5.51}{2.98} & \fcell{1.50}{0.20} & \fcell{36.4}{5.03} & \fcell{7.96}{9.14} \\
\rowcolor{black!5}
UP-Fall & \fcell{9.41}{1.32} & \fcell{9.01}{1.45} & \fcell{4.14}{0.53} & \fcell{2.07}{0.73} & \fcell{15.8}{2.50} & \fcell{1.15}{0.55} & \fcell{19.0}{2.98} & \fcell{32.6}{1.55} & \fcell{30.2}{3.72} & \fcell{\textbf{0.54}}{0.24} & \fcell{3.55}{0.48} & \fcell{1.58}{0.48} & \fcell{18.5}{2.97} & \fcell{10.5}{2.36} & \fcell{16.7}{0.83} & \fcell{6.27}{0.79} & \fcell{20.8}{3.36} & \fcell{1.78}{0.24} & \fcell{22.5}{1.46} & \fcell{11.9}{10.0} \\
UMAFall & \fcell{48.4}{5.18} & \fcell{16.1}{5.79} & \fcell{21.0}{2.46} & \fcell{9.13}{0.80} & \fcell{38.6}{7.73} & \fcell{3.87}{1.38} & \fcell{155}{1.62} & \fcell{26.9}{2.21} & \fcell{177}{6.55} & \fcell{\textbf{1.49}}{0.90} & \fcell{11.6}{1.46} & \fcell{5.10}{0.84} & \fcell{199}{5.00} & \fcell{23.6}{4.76} & \fcell{108}{1.04} & \fcell{15.4}{5.44} & \fcell{62.5}{5.86} & \fcell{2.88}{0.50} & \fcell{9.75}{1.36} & \fcell{49.3}{62.9} \\
\rowcolor{black!5}
RealLifeHAR & \fcell{3.44}{0.72} & \fcell{5.41}{2.53} & \fcell{3.42}{2.74} & \fcell{2.32}{3.76} & \fcell{13.1}{1.70} & \fcell{1.58}{1.89} & \fcell{14.3}{0.70} & \fcell{9.04}{0.72} & \fcell{20.7}{2.68} & \fcell{\textbf{0.51}}{0.40} & \fcell{1.56}{0.44} & \fcell{1.45}{2.07} & \fcell{19.3}{2.38} & \fcell{6.24}{2.57} & \fcell{15.1}{0.60} & \fcell{6.51}{2.30} & \fcell{9.52}{1.28} & \fcell{10.4}{0.75} & \fcell{991}{165} & \fcell{59.7}{226} \\
UniMiB-SHAR & \fcell{2.21}{0.69} & \fcell{4.57}{2.41} & \fcell{4.03}{3.13} & \fcell{1.79}{0.54} & \fcell{6.76}{0.65} & \fcell{0.98}{0.75} & \fcell{28.8}{1.82} & \fcell{3.45}{2.19} & \fcell{29.9}{1.30} & \fcell{\textbf{0.51}}{0.49} & \fcell{1.38}{1.16} & \fcell{1.17}{0.45} & \fcell{31.6}{1.56} & \fcell{7.89}{4.57} & \fcell{30.2}{1.06} & \fcell{8.63}{1.29} & \fcell{8.36}{1.23} & \fcell{1.22}{0.12} & \fcell{13.7}{0.65} & \fcell{9.85}{11.3} \\
\rowcolor{black!5}
AICOS-HAR & \fcell{2.71}{1.46} & \fcell{4.52}{0.76} & \fcell{2.53}{0.63} & \fcell{1.75}{1.03} & \fcell{10.0}{4.29} & \fcell{1.01}{0.44} & \fcell{16.2}{1.71} & \fcell{4.80}{0.47} & \fcell{17.0}{1.21} & \fcell{\textbf{0.48}}{0.22} & \fcell{1.26}{0.66} & \fcell{1.83}{3.01} & \fcell{15.1}{2.25} & \fcell{8.04}{1.16} & \fcell{14.7}{0.83} & \fcell{6.32}{0.71} & \fcell{6.79}{1.29} & \fcell{6.24}{0.52} & \fcell{350}{52.3} & \fcell{24.8}{78.9} \\
WISDM-19 (Phone) & \fcell{2.06}{1.64} & \fcell{4.13}{0.74} & \fcell{2.34}{0.91} & \fcell{1.69}{0.45} & \fcell{7.52}{1.48} & \fcell{0.90}{0.36} & \fcell{15.6}{1.96} & \fcell{3.45}{0.44} & \fcell{14.4}{1.53} & \fcell{\textbf{0.50}}{0.37} & \fcell{1.32}{0.48} & \fcell{1.44}{0.79} & \fcell{16.1}{1.18} & \fcell{6.81}{2.40} & \fcell{14.5}{0.87} & \fcell{5.38}{1.47} & \fcell{5.70}{0.96} & \fcell{10.4}{0.38} & \fcell{419}{58.4} & \fcell{28.0}{94.8} \\
\midrule
\textbf{Mean} & \fcell{10.4}{11.7} & \fcell{9.76}{7.92} & \fcell{6.90}{5.30} & \fcell{3.96}{4.07} & \fcell{19.1}{12.5} & \fcell{1.66}{1.04} & \fcell{41.7}{35.4} & \fcell{25.6}{48.6} & \fcell{43.8}{38.3} & \fcell{0.67}{0.30} & \fcell{4.66}{6.09} & \fcell{2.61}{3.16} & \fcell{42.6}{39.6} & \fcell{13.5}{8.30} & \fcell{36.4}{27.6} & \fcell{11.6}{7.61} & \fcell{17.2}{12.8} & \fcell{3.54}{4.18} & \fcell{114}{199} & \fcell{21.5}{26.7} \\
\bottomrule
\end{tabular}}
}

\end{appendixtable}

\begin{appendixtable}
    \centering
    \caption{Latency (ms) to classify one window on the smartphone with 3\,s windows, part 1 of 2: mean over all timed inferences, with the standard deviation below. Bold marks the lowest mean per dataset.}
    \label{tab:deploy-phone-latency-w3s}
    \providecommand{\nofit}[1]{\textcolor{gray}{#1}}
\providecommand{\faileddeployment}{\textcolor{gray}{Failed}}
\providecommand{\todomissing}{\textcolor{red}{TODO}}
\providecommand{\unmeasured}{\textcolor{gray}{--}}
\providecommand{\fcell}[2]{\begin{tabular}[c]{@{}c@{}}#1\\[-2pt]{\scriptsize$\pm$#2}\end{tabular}}
\providecommand{\tableheaderangle}{90}
\providecommand{\tablemaxheight}{0.9\textheight}
\providecommand{\fittable}[1]{\sbox2{#1}%
  \sbox2{\resizebox{\linewidth}{!}{\usebox2}}%
  \dimen2=\ht2\advance\dimen2 by\dp2
  \ifdim\dimen2>\tablemaxheight\resizebox*{!}{\tablemaxheight}{\usebox2}\else\usebox2\fi}

{%
\setlength{\tabcolsep}{2pt}
\setlength{\aboverulesep}{0pt}\setlength{\belowrulesep}{0pt}
\renewcommand{\arraystretch}{1.15}
\footnotesize
\fittable{%
\begin{tabular}{@{}lccccccccccccccccccc|c@{}}
Dataset & \rotatebox{\tableheaderangle}{CNN-HAR\hspace{4pt}} & \rotatebox{\tableheaderangle}{Triple-Cross-Attn\hspace{4pt}} & \rotatebox{\tableheaderangle}{TinyHAR\hspace{4pt}} & \rotatebox{\tableheaderangle}{TinierHAR\hspace{4pt}} & \rotatebox{\tableheaderangle}{DANA\hspace{4pt}} & \rotatebox{\tableheaderangle}{LITEWAY-Light\hspace{4pt}} & \rotatebox{\tableheaderangle}{Attend-and-Discriminate\hspace{4pt}} & \rotatebox{\tableheaderangle}{DynamicWHAR\hspace{4pt}} & \rotatebox{\tableheaderangle}{DeepConvLSTM-Attn\hspace{4pt}} & \rotatebox{\tableheaderangle}{Random Forest\hspace{4pt}} & \rotatebox{\tableheaderangle}{MLP-HAR\hspace{4pt}} & \rotatebox{\tableheaderangle}{LITEWAY\hspace{4pt}} & \rotatebox{\tableheaderangle}{DeepConvLSTM\hspace{4pt}} & \rotatebox{\tableheaderangle}{MLP-Mixer\hspace{4pt}} & \rotatebox{\tableheaderangle}{Guan-LSTM\hspace{4pt}} & \rotatebox{\tableheaderangle}{SA-HAR\hspace{4pt}} & \rotatebox{\tableheaderangle}{DeepConvShallowLSTM\hspace{4pt}} & \rotatebox{\tableheaderangle}{SVM\hspace{4pt}} & \rotatebox{\tableheaderangle}{k-NN\hspace{4pt}} & \rotatebox{\tableheaderangle}{\textbf{Mean}\hspace{4pt}} \\
\midrule
\rowcolor{black!5}
SAD & \fcell{35.2}{11.6} & \fcell{23.4}{6.12} & \fcell{16.9}{2.44} & \fcell{4.59}{1.34} & \fcell{28.9}{1.58} & \fcell{2.55}{0.54} & \fcell{66.6}{0.36} & \fcell{56.7}{3.42} & \fcell{61.1}{7.32} & \fcell{\textbf{1.28}}{1.57} & \fcell{9.50}{0.97} & \fcell{3.42}{0.64} & \fcell{63.6}{2.80} & \fcell{37.8}{7.08} & \fcell{46.5}{0.88} & \fcell{19.5}{2.08} & \fcell{39.1}{7.02} & \fcell{2.38}{2.81} & \fcell{18.7}{2.20} & \fcell{28.3}{22.5} \\
UCI-HAR & \fcell{6.76}{0.79} & \fcell{6.98}{1.37} & \fcell{6.69}{1.03} & \fcell{2.71}{0.47} & \fcell{17.3}{2.06} & \fcell{1.44}{0.49} & \fcell{49.8}{1.25} & \fcell{6.20}{0.83} & \fcell{46.8}{2.23} & \fcell{\textbf{0.53}}{0.36} & \fcell{1.81}{0.43} & \fcell{1.57}{0.55} & \fcell{48.6}{1.13} & \fcell{10.1}{2.70} & \fcell{43.1}{1.56} & \fcell{13.0}{2.08} & \fcell{23.0}{3.71} & \fcell{0.71}{0.06} & \fcell{9.97}{0.62} & \fcell{15.6}{17.7} \\
\rowcolor{black!5}
MotionSense & \fcell{11.4}{2.39} & \fcell{12.4}{3.03} & \fcell{9.51}{3.25} & \fcell{2.81}{0.45} & \fcell{19.7}{4.81} & \fcell{1.25}{0.64} & \fcell{45.0}{1.33} & \fcell{11.8}{0.95} & \fcell{53.8}{2.41} & \fcell{\textbf{0.66}}{0.73} & \fcell{2.91}{3.60} & \fcell{1.84}{1.26} & \fcell{48.9}{1.57} & \fcell{14.2}{1.64} & \fcell{41.4}{1.14} & \fcell{10.3}{3.30} & \fcell{27.4}{7.14} & \fcell{1.38}{0.17} & \fcell{28.2}{2.15} & \fcell{18.1}{17.6} \\
MHEALTH & \fcell{12.1}{1.75} & \fcell{12.7}{1.64} & \fcell{9.66}{1.19} & \fcell{3.28}{1.06} & \fcell{22.4}{6.52} & \fcell{1.62}{1.27} & \fcell{47.5}{1.25} & \fcell{15.2}{0.99} & \fcell{49.2}{1.93} & \fcell{\textbf{0.73}}{0.69} & \fcell{3.53}{0.65} & \fcell{1.90}{1.00} & \fcell{51.6}{2.72} & \fcell{22.0}{7.51} & \fcell{46.2}{0.95} & \fcell{13.3}{2.17} & \fcell{24.2}{7.83} & \fcell{1.16}{0.16} & \fcell{6.93}{0.76} & \fcell{18.2}{17.7} \\
\rowcolor{black!5}
WEAR & \fcell{8.22}{1.44} & \fcell{9.75}{1.49} & \fcell{8.08}{1.25} & \fcell{3.16}{3.08} & \fcell{25.3}{3.85} & \fcell{1.73}{2.57} & \fcell{45.7}{1.01} & \fcell{9.74}{3.03} & \fcell{45.6}{0.66} & \fcell{\textbf{0.55}}{0.45} & \fcell{2.59}{1.72} & \fcell{1.51}{0.53} & \fcell{46.7}{1.24} & \fcell{14.3}{2.54} & \fcell{42.0}{1.13} & \fcell{12.1}{1.76} & \fcell{22.8}{3.54} & \fcell{2.34}{0.22} & \fcell{40.4}{2.62} & \fcell{18.0}{17.4} \\
DSADS & \fcell{11.5}{2.40} & \fcell{12.2}{3.01} & \fcell{7.56}{1.38} & \fcell{2.94}{4.30} & \fcell{37.8}{4.24} & \fcell{1.68}{0.40} & \fcell{30.1}{3.62} & \fcell{32.7}{2.18} & \fcell{37.9}{8.46} & \fcell{\textbf{0.62}}{0.54} & \fcell{5.99}{3.00} & \fcell{1.79}{0.48} & \fcell{30.5}{4.77} & \fcell{23.3}{2.14} & \fcell{22.2}{0.57} & \fcell{9.57}{3.06} & \fcell{16.4}{4.51} & \fcell{3.32}{0.31} & \fcell{38.3}{2.25} & \fcell{17.2}{13.8} \\
\rowcolor{black!5}
Opportunity (Loc.) & \fcell{26.8}{4.51} & \fcell{31.0}{3.40} & \fcell{16.3}{3.63} & \fcell{5.47}{0.72} & \fcell{74.3}{12.5} & \fcell{3.38}{0.37} & \fcell{73.6}{0.93} & \fcell{209}{16.8} & \fcell{55.5}{2.55} & \fcell{\textbf{1.18}}{0.73} & \fcell{27.9}{0.63} & \fcell{4.66}{3.03} & \fcell{58.9}{8.28} & \fcell{41.2}{5.67} & \fcell{27.2}{0.82} & \fcell{13.2}{1.83} & \fcell{49.2}{8.46} & \fcell{2.30}{0.14} & \fcell{18.7}{0.94} & \fcell{38.9}{47.4} \\
WISDM & \fcell{1.81}{0.99} & \fcell{3.97}{3.06} & \fcell{2.30}{0.79} & \fcell{1.43}{0.65} & \fcell{6.51}{1.51} & \fcell{1.53}{2.69} & \fcell{16.9}{3.82} & \fcell{3.46}{4.69} & \fcell{17.0}{1.07} & \fcell{\textbf{0.49}}{0.44} & \fcell{1.23}{0.52} & \fcell{1.41}{2.17} & \fcell{16.4}{4.37} & \fcell{5.31}{0.85} & \fcell{19.9}{1.18} & \fcell{5.04}{0.66} & \fcell{5.31}{0.87} & \fcell{0.81}{0.10} & \fcell{53.8}{5.68} & \fcell{8.67}{12.6} \\
\rowcolor{black!5}
REALDISP & \fcell{44.3}{9.63} & \fcell{53.8}{4.13} & \fcell{22.6}{1.26} & \fcell{8.45}{3.34} & \fcell{49.5}{3.38} & \fcell{6.24}{1.14} & \fcell{125}{4.20} & \fcell{193}{13.9} & \fcell{92.2}{6.89} & \fcell{\textbf{1.57}}{1.06} & \fcell{24.4}{4.20} & \fcell{7.73}{1.35} & \fcell{84.8}{0.95} & \fcell{30.5}{2.61} & \fcell{47.7}{0.80} & \fcell{25.9}{2.56} & \fcell{65.3}{7.01} & \fcell{15.6}{1.48} & \fcell{46.8}{3.91} & \fcell{49.7}{47.5} \\
PAMAP2 & \fcell{49.2}{3.94} & \fcell{23.3}{2.92} & \fcell{28.9}{3.55} & \fcell{7.34}{0.61} & \fcell{50.1}{10.7} & \fcell{4.09}{1.26} & \fcell{172}{5.35} & \fcell{51.2}{2.82} & \fcell{127}{2.56} & \fcell{\textbf{1.33}}{0.69} & \fcell{11.3}{0.52} & \fcell{4.94}{0.71} & \fcell{130}{2.05} & \fcell{28.6}{5.58} & \fcell{88.8}{0.70} & \fcell{38.0}{5.90} & \fcell{68.8}{3.51} & \fcell{3.05}{0.25} & \fcell{27.1}{3.24} & \fcell{48.1}{49.0} \\
\rowcolor{black!5}
HAPT & \fcell{4.76}{1.75} & \fcell{5.97}{0.93} & \fcell{6.61}{0.69} & \fcell{2.69}{0.48} & \fcell{7.82}{1.64} & \fcell{2.19}{3.66} & \fcell{39.4}{1.99} & \fcell{4.58}{0.89} & \fcell{41.3}{1.25} & \fcell{\textbf{0.50}}{0.18} & \fcell{1.59}{0.47} & \fcell{1.44}{0.43} & \fcell{41.5}{1.50} & \fcell{9.82}{1.35} & \fcell{41.7}{2.13} & \fcell{15.5}{2.42} & \fcell{16.8}{3.59} & \fcell{0.89}{0.12} & \fcell{9.19}{1.14} & \fcell{13.4}{15.3} \\
HHAR & \fcell{4.75}{1.28} & \fcell{5.64}{1.08} & \fcell{6.01}{1.55} & \fcell{2.47}{0.76} & \fcell{13.7}{4.14} & \fcell{1.12}{0.56} & \fcell{43.1}{2.22} & \fcell{4.93}{0.80} & \fcell{48.2}{1.06} & \fcell{\textbf{0.49}}{0.53} & \fcell{1.83}{0.93} & \fcell{1.36}{0.44} & \fcell{51.1}{2.43} & \fcell{12.7}{3.75} & \fcell{43.2}{2.62} & \fcell{15.1}{2.11} & \fcell{18.6}{5.23} & \fcell{1.43}{0.27} & \fcell{57.3}{4.48} & \fcell{17.5}{19.9} \\
\rowcolor{black!5}
Hang-Time & \fcell{3.16}{0.62} & \fcell{4.59}{0.90} & \fcell{4.97}{0.79} & \fcell{2.33}{0.56} & \fcell{9.33}{1.30} & \fcell{1.27}{2.02} & \fcell{49.9}{1.61} & \fcell{3.91}{1.19} & \fcell{40.6}{1.65} & \fcell{\textbf{0.53}}{0.39} & \fcell{1.88}{0.65} & \fcell{1.35}{0.65} & \fcell{38.7}{0.37} & \fcell{9.35}{1.77} & \fcell{43.1}{2.67} & \fcell{15.1}{3.34} & \fcell{11.5}{1.06} & \fcell{0.94}{0.14} & \fcell{63.3}{6.00} & \fcell{16.1}{20.0} \\
w-HAR & \fcell{19.2}{3.06} & \fcell{12.4}{2.42} & \fcell{21.7}{1.83} & \fcell{7.77}{0.70} & \fcell{26.5}{5.90} & \fcell{1.80}{0.42} & \fcell{304}{6.01} & \fcell{11.5}{2.05} & \fcell{284}{24.8} & \fcell{\textbf{0.93}}{1.13} & \fcell{13.0}{0.99} & \fcell{2.78}{0.53} & \fcell{223}{11.9} & \fcell{14.0}{6.15} & \fcell{262}{2.44} & \fcell{21.5}{2.66} & \fcell{23.8}{4.87} & \fcell{0.97}{0.18} & \fcell{6.33}{0.50} & \fcell{66.2}{109} \\
\rowcolor{black!5}
HuGaDB & \fcell{19.8}{4.17} & \fcell{22.4}{2.32} & \fcell{10.2}{2.24} & \fcell{4.42}{2.16} & \fcell{35.0}{12.2} & \fcell{2.24}{0.71} & \fcell{72.9}{0.83} & \fcell{23.7}{12.7} & \fcell{81.0}{0.49} & \fcell{\textbf{1.15}}{1.56} & \fcell{6.48}{2.75} & \fcell{3.11}{4.31} & \fcell{74.3}{1.64} & \fcell{17.2}{2.23} & \fcell{56.8}{0.77} & \fcell{22.1}{3.38} & \fcell{41.8}{7.05} & \fcell{1.99}{0.18} & \fcell{28.3}{3.95} & \fcell{27.6}{26.2} \\
\bottomrule
\end{tabular}}
}

\end{appendixtable}

\begin{appendixtable}
    \centering
    \caption{Latency (ms) to classify one window on the smartphone with 3\,s windows, part 2 of 2 (continued from \autoref{tab:deploy-phone-latency-w3s}): mean over all timed inferences, with the standard deviation below. Bold marks the lowest mean per dataset. The last row summarizes all datasets across both parts.}
    \label{tab:deploy-phone-latency-w3s-part2}
    \providecommand{\nofit}[1]{\textcolor{gray}{#1}}
\providecommand{\faileddeployment}{\textcolor{gray}{Failed}}
\providecommand{\todomissing}{\textcolor{red}{TODO}}
\providecommand{\unmeasured}{\textcolor{gray}{--}}
\providecommand{\fcell}[2]{\begin{tabular}[c]{@{}c@{}}#1\\[-2pt]{\scriptsize$\pm$#2}\end{tabular}}
\providecommand{\tableheaderangle}{90}
\providecommand{\tablemaxheight}{0.9\textheight}
\providecommand{\fittable}[1]{\sbox2{#1}%
  \sbox2{\resizebox{\linewidth}{!}{\usebox2}}%
  \dimen2=\ht2\advance\dimen2 by\dp2
  \ifdim\dimen2>\tablemaxheight\resizebox*{!}{\tablemaxheight}{\usebox2}\else\usebox2\fi}

{%
\setlength{\tabcolsep}{2pt}
\setlength{\aboverulesep}{0pt}\setlength{\belowrulesep}{0pt}
\renewcommand{\arraystretch}{1.15}
\footnotesize
\fittable{%
\begin{tabular}{@{}lccccccccccccccccccc|c@{}}
Dataset & \rotatebox{\tableheaderangle}{CNN-HAR\hspace{4pt}} & \rotatebox{\tableheaderangle}{Triple-Cross-Attn\hspace{4pt}} & \rotatebox{\tableheaderangle}{TinyHAR\hspace{4pt}} & \rotatebox{\tableheaderangle}{TinierHAR\hspace{4pt}} & \rotatebox{\tableheaderangle}{DANA\hspace{4pt}} & \rotatebox{\tableheaderangle}{LITEWAY-Light\hspace{4pt}} & \rotatebox{\tableheaderangle}{Attend-and-Discriminate\hspace{4pt}} & \rotatebox{\tableheaderangle}{DynamicWHAR\hspace{4pt}} & \rotatebox{\tableheaderangle}{DeepConvLSTM-Attn\hspace{4pt}} & \rotatebox{\tableheaderangle}{Random Forest\hspace{4pt}} & \rotatebox{\tableheaderangle}{MLP-HAR\hspace{4pt}} & \rotatebox{\tableheaderangle}{LITEWAY\hspace{4pt}} & \rotatebox{\tableheaderangle}{DeepConvLSTM\hspace{4pt}} & \rotatebox{\tableheaderangle}{MLP-Mixer\hspace{4pt}} & \rotatebox{\tableheaderangle}{Guan-LSTM\hspace{4pt}} & \rotatebox{\tableheaderangle}{SA-HAR\hspace{4pt}} & \rotatebox{\tableheaderangle}{DeepConvShallowLSTM\hspace{4pt}} & \rotatebox{\tableheaderangle}{SVM\hspace{4pt}} & \rotatebox{\tableheaderangle}{k-NN\hspace{4pt}} & \rotatebox{\tableheaderangle}{\textbf{Mean}\hspace{4pt}} \\
\midrule
\rowcolor{black!5}
HARTH & \fcell{4.59}{0.63} & \fcell{14.6}{10.6} & \fcell{5.59}{0.65} & \fcell{2.72}{2.29} & \fcell{14.8}{1.35} & \fcell{0.94}{0.61} & \fcell{38.4}{2.25} & \fcell{5.40}{2.50} & \fcell{49.3}{1.92} & \fcell{\textbf{0.57}}{0.54} & \fcell{1.67}{0.47} & \fcell{1.22}{0.32} & \fcell{60.7}{16.7} & \fcell{11.0}{1.72} & \fcell{46.7}{0.86} & \fcell{15.2}{2.25} & \fcell{20.4}{2.39} & \fcell{1.30}{0.19} & \fcell{115}{14.0} & \fcell{21.6}{29.4} \\
Daphnet & \fcell{9.25}{1.54} & \fcell{8.94}{1.46} & \fcell{8.95}{3.40} & \fcell{3.01}{0.75} & \fcell{14.1}{2.53} & \fcell{1.30}{0.38} & \fcell{57.4}{0.90} & \fcell{7.48}{0.77} & \fcell{53.8}{1.58} & \fcell{0.59}{0.68} & \fcell{3.14}{1.27} & \fcell{1.81}{1.40} & \fcell{60.3}{0.78} & \fcell{15.6}{3.39} & \fcell{60.1}{2.58} & \fcell{15.9}{2.41} & \fcell{15.8}{1.67} & \fcell{\textbf{0.47}}{0.06} & \fcell{16.5}{1.25} & \fcell{18.7}{21.6} \\
\rowcolor{black!5}
USC-HAD & \fcell{6.72}{3.36} & \fcell{10.5}{3.10} & \fcell{10.2}{1.23} & \fcell{4.38}{1.54} & \fcell{20.2}{3.28} & \fcell{1.26}{1.05} & \fcell{85.6}{0.96} & \fcell{7.23}{0.81} & \fcell{82.9}{0.98} & \fcell{\textbf{0.60}}{0.58} & \fcell{3.03}{0.39} & \fcell{1.54}{1.00} & \fcell{106}{2.56} & \fcell{19.9}{2.08} & \fcell{95.1}{0.84} & \fcell{31.0}{5.65} & \fcell{15.8}{5.83} & \fcell{1.18}{0.17} & \fcell{22.1}{1.70} & \fcell{27.6}{35.6} \\
HAR70+ & \fcell{4.04}{0.59} & \fcell{5.88}{0.84} & \fcell{6.64}{0.81} & \fcell{2.55}{0.56} & \fcell{11.2}{3.32} & \fcell{1.18}{0.83} & \fcell{43.8}{1.07} & \fcell{4.64}{1.04} & \fcell{53.0}{1.36} & \fcell{\textbf{0.50}}{0.29} & \fcell{1.69}{0.44} & \fcell{1.51}{1.03} & \fcell{50.1}{1.79} & \fcell{11.0}{2.45} & \fcell{44.5}{3.14} & \fcell{14.8}{2.60} & \fcell{13.2}{1.47} & \fcell{0.70}{0.08} & \fcell{36.2}{3.97} & \fcell{16.2}{18.8} \\
\rowcolor{black!5}
GOTOV & \fcell{22.2}{3.38} & \fcell{14.5}{2.52} & \fcell{17.9}{2.08} & \fcell{5.07}{3.20} & \fcell{21.0}{4.44} & \fcell{2.48}{0.56} & \fcell{107}{3.72} & \fcell{24.8}{1.24} & \fcell{93.5}{2.61} & \fcell{\textbf{0.99}}{0.94} & \fcell{5.23}{0.52} & \fcell{2.58}{0.73} & \fcell{86.4}{1.64} & \fcell{15.3}{4.29} & \fcell{80.2}{1.77} & \fcell{28.3}{5.64} & \fcell{32.4}{1.86} & \fcell{5.41}{0.60} & \fcell{77.4}{6.45} & \fcell{33.8}{35.5} \\
UTD-MHAD & \fcell{4.71}{0.52} & \fcell{5.63}{1.43} & \fcell{5.19}{1.09} & \fcell{2.55}{1.45} & \fcell{13.5}{2.69} & \fcell{0.98}{0.52} & \fcell{44.0}{2.96} & \fcell{4.96}{0.79} & \fcell{41.2}{0.64} & \fcell{\textbf{0.49}}{0.18} & \fcell{1.74}{0.39} & \fcell{1.30}{0.47} & \fcell{40.9}{0.57} & \fcell{11.9}{2.11} & \fcell{46.1}{0.76} & \fcell{15.5}{2.19} & \fcell{16.0}{4.14} & \fcell{2.34}{0.30} & \fcell{1.85}{0.19} & \fcell{13.7}{16.3} \\
\rowcolor{black!5}
WISDM-19 (Watch) & \fcell{2.25}{0.59} & \fcell{4.16}{1.10} & \fcell{2.35}{0.69} & \fcell{1.74}{0.44} & \fcell{9.45}{4.53} & \fcell{0.89}{0.42} & \fcell{15.9}{1.01} & \fcell{3.65}{0.66} & \fcell{20.2}{1.00} & \fcell{\textbf{0.52}}{0.73} & \fcell{1.22}{0.38} & \fcell{1.20}{0.40} & \fcell{18.2}{3.36} & \fcell{7.88}{1.14} & \fcell{22.3}{0.82} & \fcell{6.85}{1.52} & \fcell{7.56}{2.71} & \fcell{6.00}{0.52} & \fcell{222}{32.4} & \fcell{18.6}{49.7} \\
UCA-EHAR & \fcell{3.23}{2.55} & \fcell{4.28}{0.57} & \fcell{3.36}{0.44} & \fcell{1.71}{0.39} & \fcell{9.41}{1.34} & \fcell{0.92}{0.97} & \fcell{23.8}{1.79} & \fcell{5.19}{4.80} & \fcell{25.4}{2.24} & \fcell{\textbf{0.57}}{0.65} & \fcell{1.56}{1.40} & \fcell{1.42}{0.41} & \fcell{20.0}{1.91} & \fcell{8.58}{3.30} & \fcell{24.2}{1.21} & \fcell{7.97}{3.25} & \fcell{8.93}{1.32} & \fcell{1.27}{0.24} & \fcell{22.0}{2.66} & \fcell{9.15}{9.05} \\
\rowcolor{black!5}
UP-Fall & \fcell{10.4}{1.82} & \fcell{10.2}{1.29} & \fcell{5.14}{0.80} & \fcell{2.65}{2.00} & \fcell{17.6}{1.82} & \fcell{1.41}{1.28} & \fcell{35.5}{2.50} & \fcell{31.9}{1.53} & \fcell{28.6}{4.13} & \fcell{\textbf{0.54}}{0.07} & \fcell{4.88}{0.46} & \fcell{1.51}{0.35} & \fcell{22.0}{2.48} & \fcell{19.8}{3.21} & \fcell{18.7}{0.38} & \fcell{7.89}{0.75} & \fcell{24.1}{3.82} & \fcell{1.43}{0.11} & \fcell{14.1}{1.87} & \fcell{13.6}{11.1} \\
UMAFall & \fcell{49.9}{8.44} & \fcell{38.1}{6.66} & \fcell{23.9}{1.04} & \fcell{12.1}{1.73} & \fcell{58.1}{3.52} & \fcell{6.74}{1.34} & \fcell{242}{16.0} & \fcell{34.1}{4.64} & \fcell{261}{15.5} & \fcell{\textbf{1.97}}{1.15} & \fcell{16.7}{2.81} & \fcell{7.72}{1.86} & \fcell{220}{13.4} & \fcell{57.8}{0.68} & \fcell{169}{2.09} & \fcell{20.8}{3.34} & \fcell{130}{1.99} & \fcell{3.15}{0.66} & \fcell{6.40}{0.76} & \fcell{71.5}{87.2} \\
\rowcolor{black!5}
RealLifeHAR & \fcell{5.24}{3.56} & \fcell{5.50}{1.14} & \fcell{3.56}{0.82} & \fcell{1.66}{0.46} & \fcell{9.18}{1.95} & \fcell{1.03}{0.40} & \fcell{20.4}{1.12} & \fcell{8.43}{0.59} & \fcell{24.7}{2.29} & \fcell{\textbf{0.55}}{0.63} & \fcell{1.68}{1.79} & \fcell{1.93}{3.06} & \fcell{25.7}{2.31} & \fcell{10.9}{2.10} & \fcell{18.4}{3.08} & \fcell{6.62}{1.29} & \fcell{14.2}{1.93} & \fcell{8.13}{1.05} & \fcell{649}{103} & \fcell{43.0}{147} \\
UniMiB-SHAR & \fcell{3.07}{0.61} & \fcell{5.04}{0.67} & \fcell{4.94}{1.03} & \fcell{2.81}{3.79} & \fcell{8.52}{0.79} & \fcell{1.22}{0.97} & \fcell{38.6}{5.62} & \fcell{3.96}{1.09} & \fcell{40.3}{1.17} & \fcell{\textbf{0.46}}{0.09} & \fcell{1.99}{0.42} & \fcell{1.37}{0.93} & \fcell{40.3}{1.05} & \fcell{8.97}{1.01} & \fcell{44.7}{1.08} & \fcell{15.1}{2.17} & \fcell{12.9}{2.24} & \fcell{1.22}{0.23} & \fcell{8.59}{0.99} & \fcell{12.8}{15.5} \\
\rowcolor{black!5}
AICOS-HAR & \fcell{8.33}{4.11} & \fcell{7.68}{1.44} & \fcell{7.00}{2.93} & \fcell{3.82}{5.60} & \fcell{11.7}{4.01} & \fcell{1.07}{0.43} & \fcell{39.8}{0.96} & \fcell{7.50}{0.71} & \fcell{46.3}{1.35} & \fcell{\textbf{0.63}}{0.64} & \fcell{2.28}{0.41} & \fcell{1.44}{0.61} & \fcell{43.6}{2.17} & \fcell{9.15}{4.35} & \fcell{43.4}{1.34} & \fcell{16.5}{2.14} & \fcell{12.3}{8.00} & \fcell{5.48}{0.92} & \fcell{212}{26.6} & \fcell{25.3}{47.9} \\
WISDM-19 (Phone) & \fcell{2.77}{1.24} & \fcell{9.70}{5.87} & \fcell{2.54}{0.51} & \fcell{1.66}{0.38} & \fcell{7.34}{1.80} & \fcell{1.03}{0.62} & \fcell{19.3}{3.03} & \fcell{3.92}{2.07} & \fcell{18.4}{1.39} & \fcell{\textbf{0.52}}{0.21} & \fcell{1.34}{0.87} & \fcell{1.20}{0.43} & \fcell{16.9}{2.69} & \fcell{7.69}{1.38} & \fcell{19.0}{0.87} & \fcell{6.64}{2.06} & \fcell{8.40}{3.78} & \fcell{7.95}{0.76} & \fcell{254}{20.6} & \fcell{20.5}{56.9} \\
\midrule
\textbf{Mean} & \fcell{13.6}{14.3} & \fcell{13.3}{11.5} & \fcell{9.84}{7.17} & \fcell{3.80}{2.42} & \fcell{22.4}{17.0} & \fcell{1.95}{1.46} & \fcell{68.7}{66.5} & \fcell{27.3}{50.2} & \fcell{66.2}{62.4} & \fcell{0.76}{0.38} & \fcell{5.66}{6.89} & \fcell{2.36}{1.77} & \fcell{62.7}{51.2} & \fcell{17.4}{11.8} & \fcell{55.3}{50.0} & \fcell{15.8}{7.76} & \fcell{27.1}{25.3} & \fcell{2.94}{3.22} & \fcell{72.8}{129} & \fcell{25.8}{25.6} \\
\bottomrule
\end{tabular}}
}

\end{appendixtable}

\begin{appendixtable}[90]
    \centering
    \caption{Peak memory overhead (MB) during inference on the smartphone with 2\,s windows. Bold marks the lowest value per dataset. --: no increase over the baseline memory was measurable, because the memory was already resident when the baseline was taken, and these cases are left out of all summaries.}
    \label{tab:deploy-phone-memory-w2s}
    \providecommand{\nofit}[1]{\textcolor{gray}{#1}}
\providecommand{\faileddeployment}{\textcolor{gray}{Failed}}
\providecommand{\todomissing}{\textcolor{red}{TODO}}
\providecommand{\unmeasured}{\textcolor{gray}{--}}
\providecommand{\fcell}[2]{\begin{tabular}[c]{@{}c@{}}#1\\[-2pt]{\scriptsize$\pm$#2}\end{tabular}}
\providecommand{\tableheaderangle}{90}
\providecommand{\tablemaxheight}{0.9\textheight}
\providecommand{\fittable}[1]{\sbox2{#1}%
  \sbox2{\resizebox{\linewidth}{!}{\usebox2}}%
  \dimen2=\ht2\advance\dimen2 by\dp2
  \ifdim\dimen2>\tablemaxheight\resizebox*{!}{\tablemaxheight}{\usebox2}\else\usebox2\fi}

{%
\setlength{\tabcolsep}{2pt}
\setlength{\aboverulesep}{0pt}\setlength{\belowrulesep}{0pt}
\renewcommand{\arraystretch}{1.15}
\footnotesize
\fittable{%
\begin{tabular}{@{}lccccccccccccccccccc|c@{}}
Dataset & \rotatebox{\tableheaderangle}{CNN-HAR\hspace{4pt}} & \rotatebox{\tableheaderangle}{Triple-Cross-Attn\hspace{4pt}} & \rotatebox{\tableheaderangle}{TinyHAR\hspace{4pt}} & \rotatebox{\tableheaderangle}{TinierHAR\hspace{4pt}} & \rotatebox{\tableheaderangle}{DANA\hspace{4pt}} & \rotatebox{\tableheaderangle}{LITEWAY-Light\hspace{4pt}} & \rotatebox{\tableheaderangle}{Attend-and-Discriminate\hspace{4pt}} & \rotatebox{\tableheaderangle}{DynamicWHAR\hspace{4pt}} & \rotatebox{\tableheaderangle}{DeepConvLSTM-Attn\hspace{4pt}} & \rotatebox{\tableheaderangle}{Random Forest\hspace{4pt}} & \rotatebox{\tableheaderangle}{MLP-HAR\hspace{4pt}} & \rotatebox{\tableheaderangle}{LITEWAY\hspace{4pt}} & \rotatebox{\tableheaderangle}{DeepConvLSTM\hspace{4pt}} & \rotatebox{\tableheaderangle}{MLP-Mixer\hspace{4pt}} & \rotatebox{\tableheaderangle}{Guan-LSTM\hspace{4pt}} & \rotatebox{\tableheaderangle}{SA-HAR\hspace{4pt}} & \rotatebox{\tableheaderangle}{DeepConvShallowLSTM\hspace{4pt}} & \rotatebox{\tableheaderangle}{SVM\hspace{4pt}} & \rotatebox{\tableheaderangle}{k-NN\hspace{4pt}} & \rotatebox{\tableheaderangle}{\textbf{Mean}\hspace{4pt}} \\
\midrule
SAD & 10.8 & 10.4 & 9.53 & 3.24 & 35.4 & \textbf{2.38} & 42.6 & 68.4 & 31.1 & 15.9 & 13.1 & 4.12 & 27.9 & 17.2 & 6.26 & 11.8 & 27.0 & 38.0 & 67.5 & 23.3\,{\scriptsize$\pm$19.9} \\
UCI-HAR & 2.95 & 4.73 & 5.54 & 2.78 & 8.78 & \textbf{2.34} & 21.8 & 11.5 & 7.32 & 16.4 & 3.47 & 2.79 & 6.75 & 5.93 & 6.08 & 6.07 & 6.49 & 13.3 & 7.19 & 7.48\,{\scriptsize$\pm$5.07} \\
MotionSense & 5.12 & 5.19 & 6.15 & 3.19 & 14.7 & \textbf{2.10} & 24.0 & 19.4 & 9.32 & 14.7 & 4.79 & 3.60 & 9.02 & 8.16 & 5.96 & 6.93 & 10.7 & 36.1 & 39.8 & 12.1\,{\scriptsize$\pm$10.8} \\
MHEALTH & 6.04 & 5.30 & 6.57 & 2.75 & 17.1 & \textbf{2.23} & 27.3 & 24.0 & 13.1 & 15.6 & 4.67 & 2.92 & 12.7 & 9.07 & 5.96 & 7.35 & 13.6 & 16.0 & 10.3 & 10.7\,{\scriptsize$\pm$7.07} \\
WEAR & 3.78 & 5.29 & 6.87 & 2.81 & 10.8 & \textbf{2.24} & 23.4 & 13.5 & 8.36 & 60.7 & 3.78 & 2.86 & 7.93 & 6.77 & 6.47 & 6.31 & 8.12 & 45.2 & 39.7 & 13.9\,{\scriptsize$\pm$16.5} \\
DSADS & 6.19 & 6.30 & 4.75 & 3.13 & 26.3 & \textbf{2.23} & 24.0 & 45.5 & 20.1 & 21.7 & 8.16 & 3.15 & 19.0 & 8.86 & 5.82 & 6.29 & 16.9 & 42.2 & 42.5 & 16.5\,{\scriptsize$\pm$14.2} \\
Opportunity (Loc.) & 15.5 & 10.4 & 9.12 & 3.56 & 59.6 & \textbf{2.68} & 51.4 & 183 & 51.1 & 17.6 & 51.6 & 3.60 & 47.7 & 21.5 & 6.25 & 13.2 & 45.1 & 61.6 & 94.5 & 39.4\,{\scriptsize$\pm$43.4} \\
WISDM & \textbf{2.26} & 3.67 & 4.50 & 2.80 & 5.08 & 2.44 & 9.56 & 5.97 & 4.22 & 18.2 & 3.24 & 2.75 & 3.87 & 4.06 & 5.95 & 5.04 & 3.37 & 15.6 & 16.2 & 6.26\,{\scriptsize$\pm$4.95} \\
REALDISP & 20.2 & 37.4 & 11.7 & 4.15 & 64.0 & \textbf{2.78} & 64.3 & 153 & 56.0 & 76.2 & 41.5 & 3.84 & 50.2 & 29.9 & 6.68 & 17.4 & 47.7 & 253 & 150 & 57.4\,{\scriptsize$\pm$64.3} \\
PAMAP2 & 17.7 & 20.0 & 13.8 & 3.94 & 35.2 & \textbf{3.03} & 61.8 & 60.2 & 38.3 & 33.7 & 12.0 & 4.09 & 32.8 & 27.6 & 7.38 & 18.3 & 32.8 & 61.3 & 81.5 & 29.8\,{\scriptsize$\pm$22.7} \\
HAPT & 2.74 & 4.02 & 6.30 & 2.71 & 6.83 & \textbf{2.47} & 21.0 & 8.60 & 5.91 & 23.6 & 3.73 & 2.98 & 5.41 & 5.33 & 6.30 & 6.24 & 5.26 & 10.6 & 5.29 & 7.12\,{\scriptsize$\pm$5.75} \\
HHAR & 2.86 & 3.80 & 4.44 & 2.53 & 6.31 & \textbf{2.24} & 11.3 & 8.50 & 5.51 & 26.1 & 2.97 & 3.19 & 4.83 & 4.59 & 5.49 & 5.00 & 4.10 & 36.6 & 33.2 & 9.13\,{\scriptsize$\pm$10.5} \\
Hang-Time & 2.51 & 3.78 & 5.41 & 2.57 & 5.39 & \textbf{2.11} & 19.9 & 5.88 & 4.50 & 34.2 & 3.57 & 2.76 & 4.15 & 5.02 & 5.70 & 4.98 & 4.60 & 23.0 & 21.5 & 8.50\,{\scriptsize$\pm$9.04} \\
w-HAR & 7.19 & 7.54 & 22.3 & 3.93 & 11.3 & \textbf{2.43} & 86.4 & 11.7 & 10.9 & 17.9 & 14.2 & 3.41 & 9.97 & 12.3 & 8.97 & 20.5 & 12.9 & 10.5 & 5.35 & 14.7\,{\scriptsize$\pm$18.2} \\
HuGaDB & 9.45 & 9.37 & 8.56 & 3.06 & 25.3 & \textbf{2.56} & 38.8 & 41.5 & 23.6 & 26.3 & 6.99 & 3.13 & 20.8 & 14.1 & 6.04 & 7.79 & 20.1 & 39.4 & 74.0 & 20.0\,{\scriptsize$\pm$18.2} \\
HARTH & 2.82 & 3.93 & 6.07 & 3.26 & 6.61 & \textbf{2.08} & 20.9 & 8.82 & 5.93 & 27.6 & 3.13 & 2.89 & 6.02 & 5.27 & 6.41 & 5.75 & 5.09 & 22.5 & 57.6 & 10.7\,{\scriptsize$\pm$13.5} \\
Daphnet & 3.53 & 3.71 & 6.85 & 2.86 & 9.11 & \textbf{2.08} & 24.5 & 11.5 & 7.60 & 9.98 & 3.70 & 3.10 & 6.75 & 6.35 & 6.15 & 6.31 & 7.36 & 7.56 & 10.8 & 7.36\,{\scriptsize$\pm$4.97} \\
USC-HAD & 3.84 & 5.03 & 9.68 & 2.81 & 8.25 & \textbf{2.44} & 31.0 & 9.03 & 7.52 & 27.3 & 4.01 & 3.12 & 6.86 & 6.94 & 6.70 & 8.70 & 7.06 & 22.0 & 9.74 & 9.58\,{\scriptsize$\pm$8.12} \\
HAR70+ & 3.14 & 4.19 & 6.42 & 2.55 & 6.88 & \textbf{2.00} & 20.7 & 8.72 & 6.05 & 14.6 & 3.61 & 2.83 & 5.72 & 5.45 & 5.88 & 5.66 & 5.09 & 11.0 & 22.3 & 7.51\,{\scriptsize$\pm$5.78} \\
GOTOV & 10.4 & 10.9 & 11.4 & 3.12 & 20.7 & \textbf{2.08} & 43.4 & 31.3 & 17.1 & 44.6 & 6.33 & 3.15 & 16.9 & 14.5 & 7.05 & 12.3 & 17.4 & 141 & 107 & 27.4\,{\scriptsize$\pm$36.7} \\
UTD-MHAD & 3.28 & 4.68 & 6.09 & 2.67 & 6.82 & \textbf{2.15} & 21.0 & 8.57 & 6.11 & 22.6 & 3.33 & 2.81 & 5.42 & 5.52 & 6.18 & 5.89 & 5.42 & 12.0 & 3.45 & 7.05\,{\scriptsize$\pm$5.68} \\
WISDM-19 (Watch) & 2.77 & 4.15 & 4.63 & 2.62 & 6.69 & \textbf{2.38} & 10.9 & 8.62 & 5.22 & 80.9 & 3.20 & 2.70 & 4.85 & 4.62 & 5.69 & 4.66 & 4.24 & 160 & 87.5 & 21.4\,{\scriptsize$\pm$41.9} \\
UCA-EHAR & 3.18 & 3.95 & 4.44 & 2.53 & 7.22 & \textbf{2.24} & 12.3 & 9.18 & 5.52 & 30.5 & 3.22 & 3.11 & 5.22 & 5.21 & 5.57 & 5.24 & 4.81 & 24.9 & 10.2 & 7.81\,{\scriptsize$\pm$7.53} \\
UP-Fall & 6.12 & 5.43 & 5.56 & 3.29 & 24.6 & \textbf{2.03} & 23.2 & 44.5 & 18.8 & 24.4 & 7.45 & 2.91 & 17.5 & 8.51 & 6.12 & 6.58 & 15.9 & 33.7 & 39.9 & 15.6\,{\scriptsize$\pm$13.0} \\
UMAFall & 22.7 & 32.4 & 22.6 & 4.48 & 34.5 & \textbf{2.24} & 104 & 48.6 & 47.2 & 27.2 & 14.9 & 4.27 & 32.8 & 39.3 & 8.56 & 30.2 & 39.5 & 34.0 & 23.0 & 30.1\,{\scriptsize$\pm$22.7} \\
RealLifeHAR & 3.45 & 4.26 & 4.06 & 2.94 & 11.2 & \textbf{2.38} & 13.4 & 16.2 & 8.70 & 28.6 & 3.93 & 2.86 & 7.85 & 5.63 & 5.57 & 5.74 & 7.37 & 184 & 294 & 32.2\,{\scriptsize$\pm$75.3} \\
UniMiB-SHAR & 2.30 & 4.81 & 5.59 & 2.51 & 4.99 & \textbf{2.25} & 19.6 & 6.02 & 4.47 & 44.0 & 3.25 & 3.12 & 4.09 & 4.74 & 6.25 & 5.18 & 4.35 & 11.1 & 3.77 & 7.50\,{\scriptsize$\pm$9.68} \\
AICOS-HAR & 3.47 & 4.28 & 3.84 & 2.76 & 8.33 & \textbf{2.12} & 12.0 & 11.7 & 6.63 & 41.1 & 3.45 & 2.75 & 6.19 & 5.19 & 5.43 & 5.41 & 5.83 & 140 & 111 & 20.1\,{\scriptsize$\pm$38.4} \\
WISDM-19 (Phone) & 2.52 & 3.74 & 4.47 & 3.24 & 6.11 & \textbf{2.13} & 10.7 & 8.59 & 5.13 & 80.8 & 3.06 & 3.19 & 5.27 & 4.83 & 5.45 & 4.69 & 4.04 & 179 & 96.2 & 22.8\,{\scriptsize$\pm$46.1} \\
\midrule
\textbf{Mean} & \fcell{6.51}{5.71} & \fcell{8.02}{8.22} & \fcell{7.84}{4.76} & \fcell{3.06}{0.52} & \fcell{17.0}{15.7} & \fcell{2.31}{0.24} & \fcell{30.9}{23.1} & \fcell{30.8}{42.1} & \fcell{15.2}{15.1} & \fcell{31.8}{19.7} & \fcell{8.43}{11.2} & \fcell{3.17}{0.44} & \fcell{13.6}{12.9} & \fcell{10.4}{8.79} & \fcell{6.29}{0.83} & \fcell{8.81}{5.96} & \fcell{13.5}{12.9} & \fcell{58.1}{65.4} & \fcell{54.0}{60.9} & \fcell{17.4}{16.4} \\
\bottomrule
\end{tabular}}
}

\end{appendixtable}

\begin{appendixtable}[90]
    \centering
    \caption{Peak memory overhead (MB) during inference on the smartphone with 3\,s windows. Bold marks the lowest value per dataset. --: no increase over the baseline memory was measurable, because the memory was already resident when the baseline was taken, and these cases are left out of all summaries.}
    \label{tab:deploy-phone-memory-w3s}
    \providecommand{\nofit}[1]{\textcolor{gray}{#1}}
\providecommand{\faileddeployment}{\textcolor{gray}{Failed}}
\providecommand{\todomissing}{\textcolor{red}{TODO}}
\providecommand{\unmeasured}{\textcolor{gray}{--}}
\providecommand{\fcell}[2]{\begin{tabular}[c]{@{}c@{}}#1\\[-2pt]{\scriptsize$\pm$#2}\end{tabular}}
\providecommand{\tableheaderangle}{90}
\providecommand{\tablemaxheight}{0.9\textheight}
\providecommand{\fittable}[1]{\sbox2{#1}%
  \sbox2{\resizebox{\linewidth}{!}{\usebox2}}%
  \dimen2=\ht2\advance\dimen2 by\dp2
  \ifdim\dimen2>\tablemaxheight\resizebox*{!}{\tablemaxheight}{\usebox2}\else\usebox2\fi}

{%
\setlength{\tabcolsep}{2pt}
\setlength{\aboverulesep}{0pt}\setlength{\belowrulesep}{0pt}
\renewcommand{\arraystretch}{1.15}
\footnotesize
\fittable{%
\begin{tabular}{@{}lccccccccccccccccccc|c@{}}
Dataset & \rotatebox{\tableheaderangle}{CNN-HAR\hspace{4pt}} & \rotatebox{\tableheaderangle}{Triple-Cross-Attn\hspace{4pt}} & \rotatebox{\tableheaderangle}{TinyHAR\hspace{4pt}} & \rotatebox{\tableheaderangle}{TinierHAR\hspace{4pt}} & \rotatebox{\tableheaderangle}{DANA\hspace{4pt}} & \rotatebox{\tableheaderangle}{LITEWAY-Light\hspace{4pt}} & \rotatebox{\tableheaderangle}{Attend-and-Discriminate\hspace{4pt}} & \rotatebox{\tableheaderangle}{DynamicWHAR\hspace{4pt}} & \rotatebox{\tableheaderangle}{DeepConvLSTM-Attn\hspace{4pt}} & \rotatebox{\tableheaderangle}{Random Forest\hspace{4pt}} & \rotatebox{\tableheaderangle}{MLP-HAR\hspace{4pt}} & \rotatebox{\tableheaderangle}{LITEWAY\hspace{4pt}} & \rotatebox{\tableheaderangle}{DeepConvLSTM\hspace{4pt}} & \rotatebox{\tableheaderangle}{MLP-Mixer\hspace{4pt}} & \rotatebox{\tableheaderangle}{Guan-LSTM\hspace{4pt}} & \rotatebox{\tableheaderangle}{SA-HAR\hspace{4pt}} & \rotatebox{\tableheaderangle}{DeepConvShallowLSTM\hspace{4pt}} & \rotatebox{\tableheaderangle}{SVM\hspace{4pt}} & \rotatebox{\tableheaderangle}{k-NN\hspace{4pt}} & \rotatebox{\tableheaderangle}{\textbf{Mean}\hspace{4pt}} \\
\midrule
SAD & 15.8 & 14.6 & 10.9 & 3.44 & 38.1 & \textbf{2.64} & 54.7 & 68.7 & 37.0 & 14.9 & 14.2 & 3.36 & 32.8 & 23.8 & 7.00 & 16.2 & 31.9 & 30.8 & 42.4 & 24.4\,{\scriptsize$\pm$18.3} \\
UCI-HAR & 3.48 & 4.86 & 7.63 & 2.91 & 9.38 & \textbf{1.86} & 24.3 & 11.6 & 7.91 & 14.5 & 4.26 & 2.97 & 7.48 & 6.96 & 6.25 & 8.07 & 8.04 & 10.1 & 5.18 & 7.78\,{\scriptsize$\pm$5.12} \\
MotionSense & 6.09 & 6.07 & 9.12 & 3.21 & 14.9 & \textbf{2.07} & 35.0 & 20.4 & 12.2 & 13.7 & 5.56 & 3.10 & 11.8 & 10.1 & 6.21 & 9.21 & 13.1 & 25.7 & 26.9 & 12.3\,{\scriptsize$\pm$8.94} \\
MHEALTH & 7.52 & 8.51 & 9.32 & 3.51 & 17.4 & \textbf{2.14} & 36.9 & 25.3 & 14.4 & 14.4 & 5.10 & 3.42 & 13.6 & 11.6 & 6.16 & 9.98 & 15.3 & 14.8 & 7.84 & 12.0\,{\scriptsize$\pm$8.32} \\
WEAR & 4.31 & 6.61 & 8.04 & 2.72 & 11.9 & \textbf{2.06} & 32.9 & 14.2 & 10.2 & 56.2 & 4.40 & 3.36 & 9.74 & 8.27 & 6.17 & 8.09 & 10.4 & 36.1 & 26.7 & 13.8\,{\scriptsize$\pm$14.2} \\
DSADS & 7.78 & 8.79 & 7.37 & 2.77 & 26.9 & \textbf{2.06} & 30.4 & 46.4 & 22.6 & 26.1 & 8.33 & 2.90 & 20.2 & 11.4 & 5.81 & 8.53 & 19.4 & 49.8 & 71.6 & 20.0\,{\scriptsize$\pm$18.8} \\
Opportunity (Loc.) & 20.5 & 14.9 & 12.0 & 4.62 & 62.8 & \textbf{2.88} & 64.4 & 185 & 59.7 & 17.3 & 52.4 & 4.28 & 54.2 & 30.8 & 6.58 & 17.7 & 52.2 & 44.3 & 75.8 & 41.2\,{\scriptsize$\pm$42.3} \\
WISDM & 2.42 & 3.78 & 4.88 & 3.07 & 4.60 & \textbf{2.27} & 12.0 & 6.07 & 4.04 & 16.1 & 3.34 & 3.19 & 4.06 & 4.33 & 5.49 & 5.36 & 3.56 & 12.1 & 7.78 & 5.71\,{\scriptsize$\pm$3.74} \\
REALDISP & 26.6 & 58.8 & 16.4 & 4.85 & 68.2 & \textbf{2.54} & 77.9 & 156 & 67.4 & 69.4 & 42.2 & 4.68 & 58.8 & 43.3 & 7.44 & 25.2 & 57.2 & 183 & 113 & 57.0\,{\scriptsize$\pm$49.7} \\
PAMAP2 & 24.6 & 29.8 & 18.0 & 4.39 & 39.7 & \textbf{2.90} & 88.4 & 62.6 & 49.4 & 33.0 & 14.8 & 4.41 & 41.5 & 39.0 & 7.95 & 26.3 & 41.3 & 47.1 & 64.4 & 33.7\,{\scriptsize$\pm$22.9} \\
HAPT & 3.31 & 4.62 & 7.13 & 2.47 & 7.66 & \textbf{2.20} & 28.9 & 8.96 & 6.35 & 20.1 & 4.23 & 2.91 & 5.77 & 6.09 & 6.25 & 7.24 & 6.06 & 8.73 & 4.22 & 7.54\,{\scriptsize$\pm$6.47} \\
HHAR & 3.65 & 5.13 & 7.21 & 2.79 & 7.79 & \textbf{2.33} & 28.9 & 9.28 & 6.40 & 23.6 & 4.02 & 2.96 & 5.76 & 6.06 & 6.40 & 6.96 & 6.37 & 26.9 & 21.5 & 9.69\,{\scriptsize$\pm$8.55} \\
Hang-Time & 3.00 & 4.38 & 7.52 & 2.67 & 5.54 & \textbf{2.38} & 27.5 & 6.47 & 4.76 & 27.8 & 4.06 & 2.77 & 4.56 & 5.04 & 6.40 & 6.70 & 4.66 & 15.2 & 11.1 & 8.02\,{\scriptsize$\pm$7.55} \\
w-HAR & 9.65 & 10.1 & 32.0 & 4.38 & 12.5 & \textbf{2.34} & 129 & 13.2 & 13.3 & 14.3 & 28.1 & 3.38 & 12.0 & 16.1 & 5.38 & 34.7 & 16.7 & 10.9 & 5.50 & 19.7\,{\scriptsize$\pm$28.0} \\
HuGaDB & 12.5 & 14.6 & 12.3 & 3.08 & 26.3 & \textbf{2.36} & 47.6 & 42.4 & 27.9 & 23.9 & 8.13 & 2.87 & 24.2 & 19.4 & 7.08 & 14.2 & 23.9 & 29.8 & 44.5 & 20.4\,{\scriptsize$\pm$13.9} \\
HARTH & 3.39 & 4.65 & 7.80 & 2.69 & 7.53 & \textbf{2.13} & 28.9 & 9.02 & 6.25 & 24.0 & 4.08 & 3.03 & 6.29 & 6.05 & 6.08 & 7.60 & 5.95 & 17.0 & 37.5 & 9.99\,{\scriptsize$\pm$9.78} \\
Daphnet & 3.96 & 4.86 & 10.2 & 2.76 & 10.4 & \textbf{2.45} & 30.8 & 12.1 & 6.92 & 9.47 & 4.16 & 2.96 & 6.31 & 7.78 & 6.73 & 9.04 & 9.22 & 7.00 & 7.33 & 8.13\,{\scriptsize$\pm$6.17} \\
USC-HAD & 4.53 & 5.76 & 12.7 & 3.10 & 9.43 & \textbf{2.25} & 49.8 & 9.83 & 8.49 & 25.3 & 6.90 & 3.23 & 8.11 & 8.27 & 7.69 & 12.3 & 9.33 & 14.8 & 6.77 & 11.0\,{\scriptsize$\pm$10.7} \\
HAR70+ & 3.32 & 4.42 & 7.21 & 2.73 & 7.50 & \textbf{2.33} & 28.9 & 8.81 & 6.51 & 11.4 & 4.11 & 3.07 & 5.83 & 6.10 & 6.53 & 7.37 & 5.90 & 8.89 & 9.69 & 7.40\,{\scriptsize$\pm$5.77} \\
GOTOV & 12.5 & 16.7 & 15.2 & 3.07 & 22.6 & \textbf{2.29} & 51.3 & 32.9 & 20.2 & 43.6 & 8.15 & 2.77 & 19.1 & 20.0 & 7.82 & 16.1 & 23.4 & 89.4 & 90.2 & 26.2\,{\scriptsize$\pm$25.9} \\
UTD-MHAD & 3.81 & 5.67 & 7.36 & 3.09 & 7.22 & \textbf{2.66} & 28.8 & 9.07 & 6.55 & 14.7 & 4.45 & 2.86 & 5.84 & 6.09 & 6.67 & 7.41 & 5.95 & 11.2 & 3.48 & 7.52\,{\scriptsize$\pm$5.96} \\
WISDM-19 (Watch) & 2.83 & 3.77 & 4.64 & 2.97 & 6.40 & \textbf{2.00} & 12.7 & 8.48 & 5.20 & 77.4 & 3.50 & 2.95 & 4.97 & 4.78 & 5.98 & 5.15 & 4.45 & 119 & 62.6 & 17.9\,{\scriptsize$\pm$32.0} \\
UCA-EHAR & 2.93 & 4.43 & 5.75 & 2.51 & 6.90 & \textbf{2.28} & 16.6 & 9.58 & 5.65 & 26.5 & 3.52 & 2.86 & 5.27 & 5.26 & 6.18 & 4.76 & 5.17 & 17.2 & 7.22 & 7.40\,{\scriptsize$\pm$6.21} \\
UP-Fall & 6.41 & 6.08 & 5.25 & 3.13 & 25.1 & \textbf{2.11} & 25.6 & 44.4 & 19.7 & 19.6 & 7.79 & 3.38 & 18.4 & 9.79 & 5.78 & 8.55 & 17.5 & 24.7 & 25.5 & 14.7\,{\scriptsize$\pm$11.2} \\
UMAFall & 33.5 & 46.0 & 31.5 & 6.04 & 40.0 & \textbf{3.37} & 155 & 52.6 & 56.9 & 27.9 & 24.2 & 5.71 & 51.5 & 25.0 & 10.4 & 46.5 & 49.5 & 28.9 & 14.6 & 37.3\,{\scriptsize$\pm$33.2} \\
RealLifeHAR & 3.97 & 4.22 & 4.40 & 2.54 & 11.5 & \textbf{2.53} & 15.4 & 16.1 & 8.95 & 22.7 & 3.38 & 3.11 & 8.46 & 5.98 & 5.54 & 5.54 & 7.50 & 151 & 216 & 26.3\,{\scriptsize$\pm$56.7} \\
UniMiB-SHAR & 2.81 & 4.03 & 7.02 & 2.71 & 5.14 & \textbf{2.24} & 27.9 & 6.58 & 4.61 & 34.1 & 3.78 & 3.18 & 4.43 & 5.13 & 6.73 & 7.33 & 4.85 & 8.76 & 3.20 & 7.61\,{\scriptsize$\pm$8.49} \\
AICOS-HAR & 3.77 & 5.81 & 7.85 & 2.87 & 10.5 & \textbf{2.42} & 31.4 & 12.8 & 9.15 & 39.0 & 4.52 & 3.02 & 8.93 & 7.73 & 6.52 & 7.95 & 8.20 & 117 & 87.0 & 19.8\,{\scriptsize$\pm$30.8} \\
WISDM-19 (Phone) & 2.34 & 3.86 & 4.71 & 2.73 & 6.45 & \textbf{2.14} & 12.7 & 8.55 & 5.26 & 77.2 & 3.14 & 2.84 & 4.99 & 5.08 & 5.54 & 5.01 & 4.79 & 155 & 78.4 & 20.6\,{\scriptsize$\pm$39.9} \\
\midrule
\textbf{Mean} & \fcell{8.32}{8.25} & \fcell{10.9}{12.9} & \fcell{10.4}{6.87} & \fcell{3.24}{0.83} & \fcell{18.3}{16.8} & \fcell{2.35}{0.31} & \fcell{42.6}{33.3} & \fcell{31.6}{42.7} & \fcell{17.7}{18.4} & \fcell{28.9}{18.9} & \fcell{9.96}{12.1} & \fcell{3.29}{0.66} & \fcell{16.0}{16.1} & \fcell{12.6}{10.5} & \fcell{6.58}{1.01} & \fcell{12.2}{9.71} & \fcell{16.3}{15.5} & \fcell{45.3}{50.4} & \fcell{40.6}{46.6} & \fcell{17.7}{13.6} \\
\bottomrule
\end{tabular}}
}

\end{appendixtable}

\clearpage
\begin{appendixtable}
    \raggedright
    \subsubsection{Smartwatch}
    \label{sec:appendix-deploy-watch}

    \noindent Tables~\ref{tab:deploy-watch-latency-w2s} to~\ref{tab:deploy-watch-memory-w3s} report latency and peak memory overhead on the smartwatch.
    \par\medskip
    \centering
    \caption{Latency (ms) to classify one window on the smartwatch with 2\,s windows, part 1 of 2: mean over all timed inferences, with the standard deviation below. Bold marks the lowest mean per dataset.}
    \label{tab:deploy-watch-latency-w2s}
    \providecommand{\nofit}[1]{\textcolor{gray}{#1}}
\providecommand{\faileddeployment}{\textcolor{gray}{Failed}}
\providecommand{\todomissing}{\textcolor{red}{TODO}}
\providecommand{\unmeasured}{\textcolor{gray}{--}}
\providecommand{\fcell}[2]{\begin{tabular}[c]{@{}c@{}}#1\\[-2pt]{\scriptsize$\pm$#2}\end{tabular}}
\providecommand{\tableheaderangle}{90}
\providecommand{\tablemaxheight}{0.9\textheight}
\providecommand{\fittable}[1]{\sbox2{#1}%
  \sbox2{\resizebox{\linewidth}{!}{\usebox2}}%
  \dimen2=\ht2\advance\dimen2 by\dp2
  \ifdim\dimen2>\tablemaxheight\resizebox*{!}{\tablemaxheight}{\usebox2}\else\usebox2\fi}

{%
\setlength{\tabcolsep}{2pt}
\setlength{\aboverulesep}{0pt}\setlength{\belowrulesep}{0pt}
\renewcommand{\arraystretch}{1.15}
\footnotesize
\fittable{%
\begin{tabular}{@{}lccccccccccccccccccc|c@{}}
Dataset & \rotatebox{\tableheaderangle}{CNN-HAR\hspace{4pt}} & \rotatebox{\tableheaderangle}{Triple-Cross-Attn\hspace{4pt}} & \rotatebox{\tableheaderangle}{TinyHAR\hspace{4pt}} & \rotatebox{\tableheaderangle}{TinierHAR\hspace{4pt}} & \rotatebox{\tableheaderangle}{DANA\hspace{4pt}} & \rotatebox{\tableheaderangle}{LITEWAY-Light\hspace{4pt}} & \rotatebox{\tableheaderangle}{Attend-and-Discriminate\hspace{4pt}} & \rotatebox{\tableheaderangle}{DynamicWHAR\hspace{4pt}} & \rotatebox{\tableheaderangle}{DeepConvLSTM-Attn\hspace{4pt}} & \rotatebox{\tableheaderangle}{Random Forest\hspace{4pt}} & \rotatebox{\tableheaderangle}{MLP-HAR\hspace{4pt}} & \rotatebox{\tableheaderangle}{LITEWAY\hspace{4pt}} & \rotatebox{\tableheaderangle}{DeepConvLSTM\hspace{4pt}} & \rotatebox{\tableheaderangle}{MLP-Mixer\hspace{4pt}} & \rotatebox{\tableheaderangle}{Guan-LSTM\hspace{4pt}} & \rotatebox{\tableheaderangle}{SA-HAR\hspace{4pt}} & \rotatebox{\tableheaderangle}{DeepConvShallowLSTM\hspace{4pt}} & \rotatebox{\tableheaderangle}{SVM\hspace{4pt}} & \rotatebox{\tableheaderangle}{k-NN\hspace{4pt}} & \rotatebox{\tableheaderangle}{\textbf{Mean}\hspace{4pt}} \\
\midrule
\rowcolor{black!5}
SAD & \fcell{186}{14.9} & \fcell{104}{5.45} & \fcell{110}{24.7} & \fcell{42.1}{6.14} & \fcell{412}{22.0} & \fcell{13.2}{3.30} & \fcell{789}{47.7} & \fcell{1180}{40.3} & \fcell{637}{42.3} & \fcell{\textbf{7.56}}{2.33} & \fcell{93.9}{13.3} & \fcell{20.7}{1.14} & \fcell{641}{39.6} & \fcell{182}{16.0} & \fcell{1238}{51.9} & \fcell{81.2}{21.0} & \fcell{476}{24.9} & \fcell{25.0}{4.48} & \fcell{417}{36.3} & \fcell{350}{388} \\
UCI-HAR & \fcell{35.0}{17.3} & \fcell{30.2}{5.71} & \fcell{39.2}{8.04} & \fcell{22.5}{5.66} & \fcell{98.2}{21.2} & \fcell{7.72}{5.64} & \fcell{392}{51.1} & \fcell{57.1}{8.80} & \fcell{356}{33.2} & \fcell{\textbf{4.65}}{1.70} & \fcell{13.3}{10.0} & \fcell{11.0}{4.30} & \fcell{354}{36.2} & \fcell{52.4}{16.5} & \fcell{1210}{53.9} & \fcell{47.8}{2.68} & \fcell{85.1}{20.1} & \fcell{10.8}{3.24} & \fcell{276}{16.1} & \fcell{163}{286} \\
\rowcolor{black!5}
MotionSense & \fcell{62.5}{7.75} & \fcell{46.1}{8.75} & \fcell{61.0}{7.79} & \fcell{25.0}{5.74} & \fcell{145}{8.20} & \fcell{7.97}{2.01} & \fcell{452}{35.2} & \fcell{126}{14.0} & \fcell{401}{31.4} & \fcell{\textbf{5.14}}{2.21} & \fcell{20.7}{9.88} & \fcell{13.0}{3.09} & \fcell{407}{45.2} & \fcell{85.7}{30.2} & \fcell{1489}{780} & \fcell{54.9}{13.9} & \fcell{148}{10.2} & \fcell{23.0}{4.71} & \fcell{750}{60.1} & \fcell{228}{366} \\
MHEALTH & \fcell{78.0}{13.6} & \fcell{56.1}{11.7} & \fcell{65.2}{7.28} & \fcell{25.1}{5.72} & \fcell{179}{15.8} & \fcell{8.84}{2.19} & \fcell{498}{50.3} & \fcell{196}{13.1} & \fcell{436}{53.2} & \fcell{\textbf{5.10}}{1.67} & \fcell{27.1}{15.6} & \fcell{14.8}{7.02} & \fcell{424}{39.7} & \fcell{103}{43.0} & \fcell{1211}{56.4} & \fcell{57.7}{10.7} & \fcell{184}{7.16} & \fcell{15.4}{3.65} & \fcell{172}{11.7} & \fcell{198}{289} \\
\rowcolor{black!5}
WEAR & \fcell{48.7}{5.60} & \fcell{41.0}{11.8} & \fcell{40.0}{2.53} & \fcell{22.4}{7.56} & \fcell{110}{18.0} & \fcell{12.2}{20.0} & \fcell{401}{38.2} & \fcell{82.8}{4.75} & \fcell{364}{18.9} & \fcell{\textbf{4.48}}{1.72} & \fcell{19.1}{19.1} & \fcell{15.1}{16.3} & \fcell{368}{28.7} & \fcell{63.4}{16.3} & \fcell{1220}{54.3} & \fcell{51.7}{11.9} & \fcell{108}{23.3} & \fcell{40.0}{6.10} & \fcell{1040}{65.6} & \fcell{213}{348} \\
DSADS & \fcell{78.5}{10.0} & \fcell{54.2}{12.8} & \fcell{44.4}{13.9} & \fcell{21.7}{29.0} & \fcell{219}{29.9} & \fcell{16.1}{27.0} & \fcell{308}{22.1} & \fcell{693}{38.6} & \fcell{264}{18.3} & \fcell{\textbf{5.32}}{1.78} & \fcell{53.4}{8.66} & \fcell{19.4}{20.4} & \fcell{267}{24.5} & \fcell{85.8}{8.25} & \fcell{620}{25.9} & \fcell{61.2}{50.8} & \fcell{141}{9.92} & \fcell{35.6}{4.87} & \fcell{361}{26.9} & \fcell{176}{202} \\
\rowcolor{black!5}
Opportunity (Loc.) & \fcell{217}{18.7} & \fcell{123}{16.0} & \fcell{97.1}{19.8} & \fcell{33.3}{5.34} & \fcell{787}{57.0} & \fcell{18.5}{12.5} & \fcell{886}{57.3} & \fcell{4964}{99.3} & \fcell{1224}{355} & \fcell{\textbf{8.81}}{2.38} & \fcell{347}{19.5} & \fcell{34.3}{5.27} & \fcell{625}{33.0} & \fcell{231}{24.2} & \fcell{719}{33.7} & \fcell{78.7}{18.8} & \fcell{521}{47.1} & \fcell{41.8}{6.92} & \fcell{424}{30.0} & \fcell{599}{1114} \\
WISDM & \fcell{11.7}{7.91} & \fcell{15.7}{2.54} & \fcell{20.9}{6.40} & \fcell{15.8}{4.00} & \fcell{47.9}{11.2} & \fcell{5.19}{2.23} & \fcell{143}{16.2} & \fcell{23.4}{7.76} & \fcell{134}{11.9} & \fcell{\textbf{4.12}}{2.17} & \fcell{13.2}{14.4} & \fcell{9.62}{7.11} & \fcell{173}{29.2} & \fcell{38.5}{22.9} & \fcell{600}{30.8} & \fcell{31.1}{9.21} & \fcell{16.2}{2.77} & \fcell{13.7}{3.47} & \fcell{1343}{75.1} & \fcell{140}{322} \\
\rowcolor{black!5}
REALDISP & \fcell{347}{25.9} & \fcell{352}{38.7} & \fcell{155}{18.4} & \fcell{48.2}{16.4} & \fcell{788}{77.4} & \fcell{21.4}{7.88} & \fcell{1437}{52.7} & \fcell{4415}{90.2} & \fcell{962}{70.8} & \fcell{\textbf{11.3}}{3.09} & \fcell{292}{18.7} & \fcell{34.0}{5.19} & \fcell{965}{79.5} & \fcell{328}{40.0} & \fcell{1257}{49.7} & \fcell{115}{37.4} & \fcell{918}{36.8} & \fcell{253}{18.1} & \fcell{1163}{499} & \fcell{730}{1007} \\
PAMAP2 & \fcell{313}{28.4} & \fcell{204}{41.6} & \fcell{162}{20.5} & \fcell{53.7}{10.3} & \fcell{599}{45.7} & \fcell{19.3}{3.67} & \fcell{1560}{95.3} & \fcell{973}{36.6} & \fcell{1539}{108} & \fcell{\textbf{10.1}}{3.17} & \fcell{93.5}{9.38} & \fcell{30.4}{4.21} & \fcell{1227}{131} & \fcell{290}{29.2} & \fcell{2459}{171} & \fcell{138}{13.7} & \fcell{931}{82.5} & \fcell{42.2}{6.01} & \fcell{596}{40.1} & \fcell{592}{688} \\
\rowcolor{black!5}
HAPT & \fcell{24.3}{2.20} & \fcell{27.8}{11.9} & \fcell{41.9}{25.3} & \fcell{23.3}{14.4} & \fcell{73.5}{10.5} & \fcell{7.97}{2.49} & \fcell{386}{69.2} & \fcell{78.2}{138} & \fcell{340}{34.4} & \fcell{\textbf{3.92}}{1.37} & \fcell{14.2}{4.43} & \fcell{10.4}{3.18} & \fcell{340}{45.0} & \fcell{52.8}{17.4} & \fcell{1225}{50.6} & \fcell{46.3}{4.39} & \fcell{62.6}{16.8} & \fcell{13.4}{3.66} & \fcell{255}{13.9} & \fcell{159}{288} \\
HHAR & \fcell{13.6}{2.25} & \fcell{19.7}{11.1} & \fcell{40.2}{66.6} & \fcell{22.6}{19.2} & \fcell{67.5}{31.2} & \fcell{7.46}{12.4} & \fcell{152}{9.61} & \fcell{39.5}{6.30} & \fcell{145}{11.1} & \fcell{\textbf{3.93}}{1.81} & \fcell{8.20}{2.29} & \fcell{8.91}{2.82} & \fcell{143}{12.7} & \fcell{41.0}{15.7} & \fcell{600}{25.5} & \fcell{35.8}{22.0} & \fcell{25.7}{3.06} & \fcell{24.8}{4.76} & \fcell{1538}{86.7} & \fcell{155}{361} \\
\rowcolor{black!5}
Hang-Time & \fcell{17.9}{4.54} & \fcell{19.8}{1.77} & \fcell{37.4}{22.5} & \fcell{22.5}{14.9} & \fcell{70.9}{18.4} & \fcell{5.07}{0.79} & \fcell{362}{76.7} & \fcell{22.3}{2.14} & \fcell{330}{41.6} & \fcell{\textbf{3.71}}{1.30} & \fcell{12.8}{9.51} & \fcell{9.59}{5.52} & \fcell{317}{32.6} & \fcell{42.5}{14.2} & \fcell{1229}{46.8} & \fcell{48.4}{13.4} & \fcell{37.2}{16.2} & \fcell{17.1}{3.61} & \fcell{1732}{90.6} & \fcell{228}{464} \\
w-HAR & \fcell{121}{30.0} & \fcell{75.9}{9.45} & \fcell{216}{47.9} & \fcell{77.5}{9.33} & \fcell{202}{21.7} & \fcell{14.2}{7.02} & \fcell{2116}{110} & \fcell{76.4}{25.1} & \fcell{1895}{87.1} & \fcell{\textbf{5.80}}{1.67} & \fcell{84.2}{14.1} & \fcell{17.4}{3.52} & \fcell{1920}{110} & \fcell{130}{9.04} & \fcell{6024}{230} & \fcell{338}{54.8} & \fcell{353}{35.5} & \fcell{10.3}{3.99} & \fcell{166}{17.3} & \fcell{729}{1460} \\
\rowcolor{black!5}
HuGaDB & \fcell{140}{7.52} & \fcell{81.2}{2.04} & \fcell{78.2}{9.91} & \fcell{32.1}{5.18} & \fcell{314}{41.2} & \fcell{11.6}{5.51} & \fcell{741}{36.7} & \fcell{560}{32.0} & \fcell{623}{46.3} & \fcell{\textbf{6.48}}{2.21} & \fcell{46.1}{4.00} & \fcell{24.6}{28.9} & \fcell{617}{33.1} & \fcell{157}{27.2} & \fcell{1463}{69.1} & \fcell{79.2}{17.9} & \fcell{382}{41.7} & \fcell{28.5}{5.27} & \fcell{781}{57.4} & \fcell{325}{387} \\
\bottomrule
\end{tabular}}
}

\end{appendixtable}

\begin{appendixtable}
    \centering
    \caption{Latency (ms) to classify one window on the smartwatch with 2\,s windows, part 2 of 2 (continued from \autoref{tab:deploy-watch-latency-w2s}): mean over all timed inferences, with the standard deviation below. Bold marks the lowest mean per dataset. The last row summarizes all datasets across both parts.}
    \label{tab:deploy-watch-latency-w2s-part2}
    \providecommand{\nofit}[1]{\textcolor{gray}{#1}}
\providecommand{\faileddeployment}{\textcolor{gray}{Failed}}
\providecommand{\todomissing}{\textcolor{red}{TODO}}
\providecommand{\unmeasured}{\textcolor{gray}{--}}
\providecommand{\fcell}[2]{\begin{tabular}[c]{@{}c@{}}#1\\[-2pt]{\scriptsize$\pm$#2}\end{tabular}}
\providecommand{\tableheaderangle}{90}
\providecommand{\tablemaxheight}{0.9\textheight}
\providecommand{\fittable}[1]{\sbox2{#1}%
  \sbox2{\resizebox{\linewidth}{!}{\usebox2}}%
  \dimen2=\ht2\advance\dimen2 by\dp2
  \ifdim\dimen2>\tablemaxheight\resizebox*{!}{\tablemaxheight}{\usebox2}\else\usebox2\fi}

{%
\setlength{\tabcolsep}{2pt}
\setlength{\aboverulesep}{0pt}\setlength{\belowrulesep}{0pt}
\renewcommand{\arraystretch}{1.15}
\footnotesize
\fittable{%
\begin{tabular}{@{}lccccccccccccccccccc|c@{}}
Dataset & \rotatebox{\tableheaderangle}{CNN-HAR\hspace{4pt}} & \rotatebox{\tableheaderangle}{Triple-Cross-Attn\hspace{4pt}} & \rotatebox{\tableheaderangle}{TinyHAR\hspace{4pt}} & \rotatebox{\tableheaderangle}{TinierHAR\hspace{4pt}} & \rotatebox{\tableheaderangle}{DANA\hspace{4pt}} & \rotatebox{\tableheaderangle}{LITEWAY-Light\hspace{4pt}} & \rotatebox{\tableheaderangle}{Attend-and-Discriminate\hspace{4pt}} & \rotatebox{\tableheaderangle}{DynamicWHAR\hspace{4pt}} & \rotatebox{\tableheaderangle}{DeepConvLSTM-Attn\hspace{4pt}} & \rotatebox{\tableheaderangle}{Random Forest\hspace{4pt}} & \rotatebox{\tableheaderangle}{MLP-HAR\hspace{4pt}} & \rotatebox{\tableheaderangle}{LITEWAY\hspace{4pt}} & \rotatebox{\tableheaderangle}{DeepConvLSTM\hspace{4pt}} & \rotatebox{\tableheaderangle}{MLP-Mixer\hspace{4pt}} & \rotatebox{\tableheaderangle}{Guan-LSTM\hspace{4pt}} & \rotatebox{\tableheaderangle}{SA-HAR\hspace{4pt}} & \rotatebox{\tableheaderangle}{DeepConvShallowLSTM\hspace{4pt}} & \rotatebox{\tableheaderangle}{SVM\hspace{4pt}} & \rotatebox{\tableheaderangle}{k-NN\hspace{4pt}} & \rotatebox{\tableheaderangle}{\textbf{Mean}\hspace{4pt}} \\
\midrule
\rowcolor{black!5}
HARTH & \fcell{25.8}{7.68} & \fcell{35.3}{13.5} & \fcell{38.5}{12.8} & \fcell{21.3}{5.62} & \fcell{75.2}{12.2} & \fcell{6.16}{4.04} & \fcell{368}{46.3} & \fcell{41.6}{6.20} & \fcell{339}{34.0} & \fcell{\textbf{4.00}}{1.61} & \fcell{13.2}{1.90} & \fcell{12.5}{11.4} & \fcell{338}{59.3} & \fcell{51.7}{18.7} & \fcell{1200}{57.4} & \fcell{45.9}{3.40} & \fcell{61.1}{15.3} & \fcell{19.6}{3.81} & \fcell{3095}{317} & \fcell{305}{732} \\
Daphnet & \fcell{39.6}{3.86} & \fcell{40.5}{26.3} & \fcell{53.1}{14.2} & \fcell{25.9}{7.12} & \fcell{102}{12.7} & \fcell{6.64}{1.82} & \fcell{504}{30.4} & \fcell{59.8}{13.7} & \fcell{462}{40.8} & \fcell{\textbf{4.81}}{2.20} & \fcell{20.7}{8.63} & \fcell{10.6}{1.67} & \fcell{471}{59.1} & \fcell{54.6}{9.04} & \fcell{1528}{69.1} & \fcell{57.6}{5.93} & \fcell{111}{20.5} & \fcell{6.13}{2.12} & \fcell{467}{29.4} & \fcell{212}{367} \\
\rowcolor{black!5}
USC-HAD & \fcell{58.3}{48.2} & \fcell{40.5}{12.8} & \fcell{90.1}{40.8} & \fcell{48.1}{22.6} & \fcell{97.9}{6.25} & \fcell{7.02}{2.38} & \fcell{788}{80.2} & \fcell{47.5}{4.60} & \fcell{726}{67.0} & \fcell{\textbf{4.33}}{1.59} & \fcell{23.1}{11.8} & \fcell{15.3}{16.6} & \fcell{714}{73.2} & \fcell{76.8}{26.8} & \fcell{2435}{102} & \fcell{88.9}{6.21} & \fcell{125}{25.8} & \fcell{19.0}{4.20} & \fcell{603}{40.3} & \fcell{316}{582} \\
HAR70+ & \fcell{32.6}{7.81} & \fcell{26.0}{9.43} & \fcell{40.4}{20.0} & \fcell{25.9}{20.0} & \fcell{76.9}{16.3} & \fcell{10.2}{21.3} & \fcell{367}{46.6} & \fcell{40.9}{3.09} & \fcell{347}{39.3} & \fcell{\textbf{4.43}}{2.18} & \fcell{13.8}{2.71} & \fcell{10.5}{7.11} & \fcell{487}{370} & \fcell{47.8}{6.61} & \fcell{1191}{59.1} & \fcell{49.4}{16.6} & \fcell{57.6}{4.96} & \fcell{9.89}{2.71} & \fcell{986}{50.3} & \fcell{201}{344} \\
\rowcolor{black!5}
GOTOV & \fcell{141}{14.0} & \fcell{95.2}{13.7} & \fcell{100}{17.3} & \fcell{36.8}{3.04} & \fcell{295}{13.0} & \fcell{12.4}{8.89} & \fcell{975}{71.2} & \fcell{308}{25.1} & \fcell{814}{47.8} & \fcell{\textbf{6.41}}{2.05} & \fcell{39.5}{2.42} & \fcell{20.0}{8.31} & \fcell{813}{88.4} & \fcell{157}{16.1} & \fcell{2110}{86.3} & \fcell{106}{31.7} & \fcell{421}{25.1} & \fcell{94.1}{7.26} & \fcell{1942}{99.6} & \fcell{447}{630} \\
UTD-MHAD & \fcell{27.9}{13.1} & \fcell{33.1}{18.7} & \fcell{38.8}{16.7} & \fcell{21.8}{10.9} & \fcell{76.3}{26.8} & \fcell{6.69}{7.65} & \fcell{367}{31.2} & \fcell{42.5}{6.73} & \fcell{354}{68.5} & \fcell{\textbf{4.32}}{1.71} & \fcell{10.2}{1.87} & \fcell{11.3}{9.20} & \fcell{334}{28.2} & \fcell{49.4}{8.84} & \fcell{1245}{31.5} & \fcell{48.1}{9.92} & \fcell{92.5}{78.4} & \fcell{32.0}{7.63} & \fcell{48.9}{15.5} & \fcell{150}{291} \\
\rowcolor{black!5}
WISDM-19 (Watch) & \fcell{15.1}{6.75} & \fcell{21.1}{13.5} & \fcell{19.1}{6.91} & \fcell{15.8}{11.5} & \fcell{61.4}{8.43} & \fcell{10.9}{14.3} & \fcell{160}{51.3} & \fcell{37.8}{2.41} & \fcell{149}{25.6} & \fcell{\textbf{3.95}}{1.76} & \fcell{8.62}{3.55} & \fcell{12.5}{14.9} & \fcell{144}{13.7} & \fcell{45.3}{26.2} & \fcell{613}{31.0} & \fcell{33.8}{12.4} & \fcell{29.7}{14.0} & \fcell{123}{9.15} & \fcell{4969}{214} & \fcell{341}{1129} \\
UCA-EHAR & \fcell{15.8}{3.88} & \fcell{19.8}{3.09} & \fcell{18.2}{3.59} & \fcell{15.5}{8.69} & \fcell{70.0}{20.3} & \fcell{8.38}{12.1} & \fcell{167}{15.0} & \fcell{45.1}{8.30} & \fcell{174}{79.3} & \fcell{\textbf{3.94}}{1.87} & \fcell{11.5}{11.6} & \fcell{12.5}{11.1} & \fcell{157}{23.3} & \fcell{42.9}{27.7} & \fcell{629}{34.2} & \fcell{34.6}{12.8} & \fcell{30.8}{3.50} & \fcell{20.5}{3.97} & \fcell{550}{32.1} & \fcell{107}{179} \\
\rowcolor{black!5}
UP-Fall & \fcell{72.3}{6.03} & \fcell{49.3}{7.50} & \fcell{41.2}{7.36} & \fcell{20.2}{14.8} & \fcell{208}{28.5} & \fcell{8.95}{5.61} & \fcell{304}{48.1} & \fcell{1133}{292} & \fcell{257}{25.8} & \fcell{\textbf{5.43}}{2.37} & \fcell{44.4}{2.98} & \fcell{46.6}{57.4} & \fcell{261}{42.3} & \fcell{86.7}{10.5} & \fcell{618}{27.8} & \fcell{42.3}{8.56} & \fcell{132}{10.2} & \fcell{22.9}{4.25} & \fcell{344}{20.2} & \fcell{195}{277} \\
UMAFall & \fcell{438}{24.6} & \fcell{335}{23.8} & \fcell{274}{28.9} & \fcell{110}{19.9} & \fcell{808}{41.3} & \fcell{28.4}{12.0} & \fcell{2858}{134} & \fcell{639}{23.5} & \fcell{2181}{106} & \fcell{\textbf{14.3}}{3.59} & \fcell{117}{5.93} & \fcell{40.6}{6.48} & \fcell{2147}{84.1} & \fcell{434}{37.1} & \fcell{4855}{175} & \fcell{294}{24.9} & \fcell{1438}{55.5} & \fcell{33.3}{5.06} & \fcell{141}{16.4} & \fcell{904}{1278} \\
\rowcolor{black!5}
RealLifeHAR & \fcell{28.5}{18.2} & \fcell{26.0}{9.16} & \fcell{21.6}{4.39} & \fcell{22.9}{39.6} & \fcell{96.7}{12.0} & \fcell{8.41}{2.63} & \fcell{191}{14.5} & \fcell{94.8}{10.3} & \fcell{176}{20.3} & \fcell{\textbf{4.11}}{1.80} & \fcell{18.9}{21.2} & \fcell{13.0}{10.1} & \fcell{172}{11.0} & \fcell{53.9}{15.0} & \fcell{613}{27.1} & \fcell{38.6}{22.3} & \fcell{54.9}{12.0} & \fcell{161}{15.0} & \nofit{M} & \fcell{99.8}{143} \\
UniMiB-SHAR & \fcell{18.0}{5.99} & \fcell{24.7}{12.0} & \fcell{33.2}{4.92} & \fcell{20.8}{8.60} & \fcell{57.2}{17.6} & \fcell{8.10}{11.7} & \fcell{340}{36.5} & \fcell{23.9}{9.85} & \fcell{338}{135} & \fcell{\textbf{3.94}}{1.76} & \fcell{13.3}{28.6} & \fcell{8.73}{1.54} & \fcell{319}{40.6} & \fcell{89.7}{130} & \fcell{1203}{55.3} & \fcell{48.3}{17.3} & \fcell{40.5}{30.5} & \fcell{18.1}{4.21} & \fcell{232}{18.7} & \fcell{149}{282} \\
\rowcolor{black!5}
AICOS-HAR & \fcell{20.2}{9.07} & \fcell{22.6}{8.06} & \fcell{19.0}{2.06} & \fcell{14.4}{6.30} & \fcell{75.0}{5.79} & \fcell{6.54}{3.19} & \fcell{176}{34.5} & \fcell{61.1}{4.49} & \fcell{161}{22.4} & \fcell{\textbf{3.98}}{1.72} & \fcell{16.4}{16.8} & \fcell{11.3}{0.99} & \fcell{159}{24.1} & \fcell{46.0}{16.1} & \fcell{615}{30.3} & \fcell{32.3}{3.46} & \fcell{39.0}{4.49} & \fcell{95.4}{12.4} & \fcell{5411}{678} & \fcell{368}{1229} \\
WISDM-19 (Phone) & \fcell{14.2}{5.27} & \fcell{24.1}{30.4} & \fcell{17.9}{4.55} & \fcell{13.5}{6.74} & \fcell{63.9}{24.2} & \fcell{10.1}{18.9} & \fcell{159}{29.0} & \fcell{39.7}{7.71} & \fcell{145}{13.3} & \fcell{\textbf{3.97}}{1.64} & \fcell{8.03}{1.31} & \fcell{15.4}{16.6} & \fcell{153}{65.6} & \fcell{40.7}{10.7} & \fcell{622}{23.1} & \fcell{30.9}{2.13} & \fcell{27.3}{9.41} & \fcell{173}{13.0} & \fcell{6497}{638} & \fcell{424}{1477} \\
\midrule
\textbf{Mean} & \fcell{91.1}{110} & \fcell{70.5}{85.7} & \fcell{69.4}{62.2} & \fcell{31.1}{20.7} & \fcell{216}{235} & \fcell{10.9}{5.41} & \fcell{633}{634} & \fcell{555}{1199} & \fcell{561}{528} & \fcell{5.60}{2.54} & \fcell{51.7}{80.2} & \fcell{17.7}{10.1} & \fcell{533}{492} & \fcell{109}{98.3} & \fcell{1450}{1230} & \fcell{76.4}{71.7} & \fcell{243}{337} & \fcell{49.1}{59.1} & \fcell{1296}{1681} & \fcell{319}{428} \\
\bottomrule
\end{tabular}}
}

\end{appendixtable}

\begin{appendixtable}
    \centering
    \caption{Latency (ms) to classify one window on the smartwatch with 3\,s windows, part 1 of 2: mean over all timed inferences, with the standard deviation below. Bold marks the lowest mean per dataset.}
    \label{tab:deploy-watch-latency-w3s}
    \providecommand{\nofit}[1]{\textcolor{gray}{#1}}
\providecommand{\faileddeployment}{\textcolor{gray}{Failed}}
\providecommand{\todomissing}{\textcolor{red}{TODO}}
\providecommand{\unmeasured}{\textcolor{gray}{--}}
\providecommand{\fcell}[2]{\begin{tabular}[c]{@{}c@{}}#1\\[-2pt]{\scriptsize$\pm$#2}\end{tabular}}
\providecommand{\tableheaderangle}{90}
\providecommand{\tablemaxheight}{0.9\textheight}
\providecommand{\fittable}[1]{\sbox2{#1}%
  \sbox2{\resizebox{\linewidth}{!}{\usebox2}}%
  \dimen2=\ht2\advance\dimen2 by\dp2
  \ifdim\dimen2>\tablemaxheight\resizebox*{!}{\tablemaxheight}{\usebox2}\else\usebox2\fi}

{%
\setlength{\tabcolsep}{2pt}
\setlength{\aboverulesep}{0pt}\setlength{\belowrulesep}{0pt}
\renewcommand{\arraystretch}{1.15}
\footnotesize
\fittable{%
\begin{tabular}{@{}lccccccccccccccccccc|c@{}}
Dataset & \rotatebox{\tableheaderangle}{CNN-HAR\hspace{4pt}} & \rotatebox{\tableheaderangle}{Triple-Cross-Attn\hspace{4pt}} & \rotatebox{\tableheaderangle}{TinyHAR\hspace{4pt}} & \rotatebox{\tableheaderangle}{TinierHAR\hspace{4pt}} & \rotatebox{\tableheaderangle}{DANA\hspace{4pt}} & \rotatebox{\tableheaderangle}{LITEWAY-Light\hspace{4pt}} & \rotatebox{\tableheaderangle}{Attend-and-Discriminate\hspace{4pt}} & \rotatebox{\tableheaderangle}{DynamicWHAR\hspace{4pt}} & \rotatebox{\tableheaderangle}{DeepConvLSTM-Attn\hspace{4pt}} & \rotatebox{\tableheaderangle}{Random Forest\hspace{4pt}} & \rotatebox{\tableheaderangle}{MLP-HAR\hspace{4pt}} & \rotatebox{\tableheaderangle}{LITEWAY\hspace{4pt}} & \rotatebox{\tableheaderangle}{DeepConvLSTM\hspace{4pt}} & \rotatebox{\tableheaderangle}{MLP-Mixer\hspace{4pt}} & \rotatebox{\tableheaderangle}{Guan-LSTM\hspace{4pt}} & \rotatebox{\tableheaderangle}{SA-HAR\hspace{4pt}} & \rotatebox{\tableheaderangle}{DeepConvShallowLSTM\hspace{4pt}} & \rotatebox{\tableheaderangle}{SVM\hspace{4pt}} & \rotatebox{\tableheaderangle}{k-NN\hspace{4pt}} & \rotatebox{\tableheaderangle}{\textbf{Mean}\hspace{4pt}} \\
\midrule
\rowcolor{black!5}
SAD & \fcell{277}{24.7} & \fcell{165}{21.5} & \fcell{131}{17.3} & \fcell{44.3}{5.23} & \fcell{683}{28.0} & \fcell{20.2}{9.45} & \fcell{1250}{69.3} & \fcell{1217}{54.0} & \fcell{966}{55.6} & \fcell{\textbf{9.37}}{2.83} & \fcell{138}{16.4} & \fcell{39.8}{16.1} & \fcell{963}{51.5} & \fcell{255}{15.5} & \fcell{1854}{162} & \fcell{116}{30.0} & \fcell{765}{34.2} & \fcell{23.4}{4.67} & \fcell{275}{30.6} & \fcell{484}{540} \\
UCI-HAR & \fcell{47.3}{9.15} & \fcell{41.2}{11.9} & \fcell{64.3}{14.0} & \fcell{28.0}{2.76} & \fcell{110}{10.2} & \fcell{9.87}{12.1} & \fcell{609}{79.4} & \fcell{60.8}{5.12} & \fcell{551}{49.9} & \fcell{\textbf{4.81}}{1.87} & \fcell{20.4}{12.0} & \fcell{11.5}{2.31} & \fcell{549}{46.7} & \fcell{64.0}{14.9} & \fcell{1790}{83.3} & \fcell{72.4}{13.5} & \fcell{126}{12.7} & \fcell{9.91}{3.22} & \fcell{158}{11.5} & \fcell{228}{426} \\
\rowcolor{black!5}
MotionSense & \fcell{87.4}{4.12} & \fcell{58.5}{8.66} & \fcell{75.3}{10.7} & \fcell{32.8}{10.5} & \fcell{187}{16.3} & \fcell{9.35}{2.24} & \fcell{718}{71.2} & \fcell{133}{12.8} & \fcell{620}{33.0} & \fcell{\textbf{5.31}}{1.48} & \fcell{27.1}{12.9} & \fcell{17.5}{13.0} & \fcell{618}{38.6} & \fcell{106}{14.9} & \fcell{1829}{85.9} & \fcell{78.3}{8.15} & \fcell{241}{22.1} & \fcell{18.0}{4.18} & \fcell{467}{86.5} & \fcell{280}{441} \\
MHEALTH & \fcell{116}{27.1} & \fcell{75.6}{10.3} & \fcell{80.1}{6.52} & \fcell{41.9}{30.6} & \fcell{236}{31.7} & \fcell{10.4}{2.92} & \fcell{878}{215} & \fcell{207}{26.7} & \fcell{661}{57.2} & \fcell{\textbf{6.06}}{2.30} & \fcell{30.9}{3.22} & \fcell{18.5}{15.3} & \fcell{671}{54.3} & \fcell{128}{27.6} & \fcell{1849}{65.8} & \fcell{80.5}{4.09} & \fcell{304}{27.6} & \fcell{14.3}{3.08} & \fcell{109}{7.70} & \fcell{290}{455} \\
\rowcolor{black!5}
WEAR & \fcell{57.3}{8.02} & \fcell{51.2}{3.73} & \fcell{70.3}{21.2} & \fcell{56.3}{83.5} & \fcell{135}{7.70} & \fcell{7.87}{1.78} & \fcell{620}{45.7} & \fcell{89.0}{23.8} & \fcell{571}{35.9} & \fcell{\textbf{4.95}}{2.16} & \fcell{19.0}{4.13} & \fcell{13.9}{7.85} & \fcell{572}{59.4} & \fcell{76.5}{10.4} & \fcell{1822}{87.4} & \fcell{73.2}{8.77} & \fcell{172}{36.6} & \fcell{31.3}{5.10} & \fcell{678}{57.2} & \fcell{269}{442} \\
DSADS & \fcell{84.2}{4.39} & \fcell{73.0}{11.6} & \fcell{55.5}{24.4} & \fcell{24.0}{9.64} & \fcell{269}{26.4} & \fcell{9.50}{1.59} & \fcell{499}{60.2} & \fcell{703}{21.2} & \fcell{413}{47.0} & \fcell{\textbf{6.10}}{1.98} & \fcell{56.1}{9.23} & \fcell{17.1}{8.97} & \fcell{414}{31.5} & \fcell{118}{13.8} & \fcell{920}{38.2} & \fcell{57.1}{7.04} & \fcell{252}{23.3} & \fcell{41.8}{5.81} & \fcell{594}{40.7} & \fcell{242}{273} \\
\rowcolor{black!5}
Opportunity (Loc.) & \fcell{326}{16.6} & \fcell{188}{21.9} & \fcell{156}{15.7} & \fcell{51.2}{24.5} & \fcell{811}{41.4} & \fcell{23.7}{9.70} & \fcell{1417}{50.7} & \fcell{5086}{107} & \fcell{951}{71.8} & \fcell{\textbf{11.4}}{3.33} & \fcell{379}{15.7} & \fcell{33.0}{1.94} & \fcell{948}{83.2} & \fcell{342}{44.4} & \fcell{1075}{54.9} & \fcell{108}{21.9} & \fcell{905}{41.3} & \fcell{29.9}{4.58} & \fcell{289}{25.8} & \fcell{691}{1149} \\
WISDM & \fcell{14.2}{11.7} & \fcell{16.7}{7.13} & \fcell{20.6}{12.9} & \fcell{17.6}{3.36} & \fcell{51.0}{16.1} & \fcell{6.61}{11.4} & \fcell{195}{47.4} & \fcell{22.7}{9.05} & \fcell{173}{16.9} & \fcell{\textbf{3.81}}{1.73} & \fcell{8.38}{3.67} & \fcell{8.25}{2.55} & \fcell{171}{13.1} & \fcell{37.3}{9.70} & \fcell{730}{32.2} & \fcell{33.0}{3.65} & \fcell{21.2}{4.07} & \fcell{10.8}{2.80} & \fcell{880}{58.7} & \fcell{127}{248} \\
\rowcolor{black!5}
REALDISP & \fcell{501}{23.1} & \fcell{419}{21.5} & \fcell{249}{27.9} & \fcell{80.3}{19.8} & \fcell{1300}{83.7} & \fcell{30.0}{7.00} & \fcell{2256}{63.2} & \fcell{4452}{87.7} & \fcell{1452}{59.9} & \fcell{\textbf{15.4}}{3.73} & \fcell{318}{16.5} & \fcell{50.0}{18.4} & \fcell{1449}{79.3} & \fcell{476}{39.9} & \fcell{1827}{84.2} & \fcell{151}{9.75} & \fcell{1504}{73.9} & \fcell{182}{13.1} & \fcell{653}{49.9} & \fcell{914}{1102} \\
PAMAP2 & \fcell{441}{24.5} & \fcell{326}{67.4} & \fcell{254}{39.5} & \fcell{76.3}{13.4} & \fcell{839}{50.8} & \fcell{38.4}{38.1} & \fcell{2732}{1694} & \fcell{1421}{64.0} & \fcell{1850}{112} & \fcell{\textbf{13.0}}{3.34} & \fcell{121}{9.28} & \fcell{39.8}{3.18} & \fcell{1892}{268} & \fcell{423}{37.8} & \fcell{3605}{143} & \fcell{218}{12.9} & \fcell{1432}{61.4} & \fcell{38.7}{6.02} & \fcell{398}{25.3} & \fcell{850}{1035} \\
\rowcolor{black!5}
HAPT & \fcell{36.0}{8.27} & \fcell{48.2}{35.1} & \fcell{60.3}{11.4} & \fcell{30.0}{12.0} & \fcell{84.4}{6.78} & \fcell{6.76}{2.67} & \fcell{573}{63.4} & \fcell{47.0}{10.3} & \fcell{530}{48.6} & \fcell{\textbf{4.37}}{1.94} & \fcell{17.1}{12.9} & \fcell{37.8}{83.5} & \fcell{525}{30.1} & \fcell{58.0}{8.24} & \fcell{1837}{98.6} & \fcell{66.8}{10.4} & \fcell{90.8}{13.0} & \fcell{11.3}{3.12} & \fcell{154}{13.5} & \fcell{222}{433} \\
HHAR & \fcell{35.2}{4.16} & \fcell{31.0}{5.67} & \fcell{70.6}{45.3} & \fcell{26.8}{4.88} & \fcell{85.9}{10.4} & \fcell{8.39}{8.86} & \fcell{610}{190} & \fcell{45.5}{8.43} & \fcell{523}{46.0} & \fcell{\textbf{4.25}}{1.91} & \fcell{14.5}{1.88} & \fcell{13.2}{1.67} & \fcell{525}{42.6} & \fcell{57.5}{8.08} & \fcell{1856}{67.4} & \fcell{65.9}{6.15} & \fcell{90.0}{11.4} & \fcell{19.9}{3.85} & \fcell{931}{61.9} & \fcell{264}{468} \\
\rowcolor{black!5}
Hang-Time & \fcell{27.8}{14.9} & \fcell{23.8}{3.42} & \fcell{54.9}{15.8} & \fcell{30.1}{27.8} & \fcell{78.7}{17.5} & \fcell{5.26}{0.25} & \fcell{668}{67.3} & \fcell{26.8}{12.6} & \fcell{507}{51.4} & \fcell{\textbf{3.90}}{1.52} & \fcell{14.5}{8.56} & \fcell{11.5}{9.68} & \fcell{500}{45.0} & \fcell{44.6}{14.2} & \fcell{1815}{83.9} & \fcell{65.0}{11.8} & \fcell{66.6}{9.98} & \fcell{13.0}{4.19} & \fcell{1040}{61.5} & \fcell{263}{474} \\
w-HAR & \fcell{181}{24.2} & \fcell{107}{18.0} & \fcell{306}{24.6} & \fcell{116}{23.8} & \fcell{276}{21.4} & \fcell{12.7}{2.74} & \fcell{3249}{162} & \fcell{88.3}{15.1} & \fcell{2911}{147} & \fcell{\textbf{7.06}}{2.23} & \fcell{181}{12.6} & \fcell{27.7}{5.66} & \fcell{2860}{111} & \fcell{175}{13.2} & \fcell{9067}{320} & \fcell{638}{47.1} & \fcell{528}{29.4} & \fcell{10.8}{3.15} & \fcell{102}{9.72} & \fcell{1097}{2205} \\
\rowcolor{black!5}
HuGaDB & \fcell{204}{50.3} & \fcell{130}{18.2} & \fcell{117}{12.6} & \fcell{43.0}{2.49} & \fcell{413}{31.7} & \fcell{16.0}{7.32} & \fcell{1169}{72.4} & \fcell{567}{16.0} & \fcell{959}{72.7} & \fcell{\textbf{8.32}}{2.84} & \fcell{58.2}{9.46} & \fcell{25.7}{14.1} & \fcell{954}{70.4} & \fcell{208}{14.1} & \fcell{2219}{126} & \fcell{114}{11.0} & \fcell{599}{41.6} & \fcell{24.1}{4.69} & \fcell{471}{47.4} & \fcell{437}{563} \\
\bottomrule
\end{tabular}}
}

\end{appendixtable}

\begin{appendixtable}
    \centering
    \caption{Latency (ms) to classify one window on the smartwatch with 3\,s windows, part 2 of 2 (continued from \autoref{tab:deploy-watch-latency-w3s}): mean over all timed inferences, with the standard deviation below. Bold marks the lowest mean per dataset. The last row summarizes all datasets across both parts.}
    \label{tab:deploy-watch-latency-w3s-part2}
    \providecommand{\nofit}[1]{\textcolor{gray}{#1}}
\providecommand{\faileddeployment}{\textcolor{gray}{Failed}}
\providecommand{\todomissing}{\textcolor{red}{TODO}}
\providecommand{\unmeasured}{\textcolor{gray}{--}}
\providecommand{\fcell}[2]{\begin{tabular}[c]{@{}c@{}}#1\\[-2pt]{\scriptsize$\pm$#2}\end{tabular}}
\providecommand{\tableheaderangle}{90}
\providecommand{\tablemaxheight}{0.9\textheight}
\providecommand{\fittable}[1]{\sbox2{#1}%
  \sbox2{\resizebox{\linewidth}{!}{\usebox2}}%
  \dimen2=\ht2\advance\dimen2 by\dp2
  \ifdim\dimen2>\tablemaxheight\resizebox*{!}{\tablemaxheight}{\usebox2}\else\usebox2\fi}

{%
\setlength{\tabcolsep}{2pt}
\setlength{\aboverulesep}{0pt}\setlength{\belowrulesep}{0pt}
\renewcommand{\arraystretch}{1.15}
\footnotesize
\fittable{%
\begin{tabular}{@{}lccccccccccccccccccc|c@{}}
Dataset & \rotatebox{\tableheaderangle}{CNN-HAR\hspace{4pt}} & \rotatebox{\tableheaderangle}{Triple-Cross-Attn\hspace{4pt}} & \rotatebox{\tableheaderangle}{TinyHAR\hspace{4pt}} & \rotatebox{\tableheaderangle}{TinierHAR\hspace{4pt}} & \rotatebox{\tableheaderangle}{DANA\hspace{4pt}} & \rotatebox{\tableheaderangle}{LITEWAY-Light\hspace{4pt}} & \rotatebox{\tableheaderangle}{Attend-and-Discriminate\hspace{4pt}} & \rotatebox{\tableheaderangle}{DynamicWHAR\hspace{4pt}} & \rotatebox{\tableheaderangle}{DeepConvLSTM-Attn\hspace{4pt}} & \rotatebox{\tableheaderangle}{Random Forest\hspace{4pt}} & \rotatebox{\tableheaderangle}{MLP-HAR\hspace{4pt}} & \rotatebox{\tableheaderangle}{LITEWAY\hspace{4pt}} & \rotatebox{\tableheaderangle}{DeepConvLSTM\hspace{4pt}} & \rotatebox{\tableheaderangle}{MLP-Mixer\hspace{4pt}} & \rotatebox{\tableheaderangle}{Guan-LSTM\hspace{4pt}} & \rotatebox{\tableheaderangle}{SA-HAR\hspace{4pt}} & \rotatebox{\tableheaderangle}{DeepConvShallowLSTM\hspace{4pt}} & \rotatebox{\tableheaderangle}{SVM\hspace{4pt}} & \rotatebox{\tableheaderangle}{k-NN\hspace{4pt}} & \rotatebox{\tableheaderangle}{\textbf{Mean}\hspace{4pt}} \\
\midrule
\rowcolor{black!5}
HARTH & \fcell{38.5}{14.8} & \fcell{32.9}{7.88} & \fcell{59.5}{5.22} & \fcell{27.0}{1.82} & \fcell{87.7}{16.5} & \fcell{12.8}{24.3} & \fcell{574}{51.6} & \fcell{44.8}{4.64} & \fcell{533}{60.9} & \fcell{\textbf{4.22}}{1.81} & \fcell{15.0}{4.28} & \fcell{10.2}{1.70} & \fcell{531}{48.2} & \fcell{65.3}{23.6} & \fcell{1841}{76.3} & \fcell{69.8}{15.7} & \fcell{89.1}{11.4} & \fcell{16.5}{3.61} & \fcell{1878}{156} & \fcell{312}{577} \\
Daphnet & \fcell{58.7}{9.34} & \fcell{45.6}{10.9} & \fcell{86.9}{33.4} & \fcell{60.0}{63.9} & \fcell{125}{11.0} & \fcell{10.9}{7.42} & \fcell{798}{65.2} & \fcell{109}{114} & \fcell{728}{89.3} & \fcell{\textbf{4.87}}{2.04} & \fcell{24.7}{13.4} & \fcell{11.4}{0.33} & \fcell{709}{44.0} & \fcell{66.0}{5.80} & \fcell{2318}{97.8} & \fcell{87.2}{5.81} & \fcell{170}{25.4} & \fcell{6.03}{2.22} & \fcell{304}{19.8} & \fcell{301}{553} \\
\rowcolor{black!5}
USC-HAD & \fcell{66.4}{4.07} & \fcell{50.5}{8.62} & \fcell{118}{15.0} & \fcell{50.1}{9.50} & \fcell{124}{10.8} & \fcell{12.0}{14.8} & \fcell{1196}{54.2} & \fcell{54.5}{11.2} & \fcell{1119}{120} & \fcell{\textbf{4.59}}{1.68} & \fcell{36.6}{3.18} & \fcell{13.7}{8.38} & \fcell{1099}{84.4} & \fcell{87.0}{8.30} & \fcell{3657}{210} & \fcell{154}{26.4} & \fcell{187}{16.0} & \fcell{15.2}{3.42} & \fcell{347}{28.7} & \fcell{442}{875} \\
HAR70+ & \fcell{38.3}{12.2} & \fcell{30.5}{2.82} & \fcell{64.6}{21.9} & \fcell{30.8}{16.0} & \fcell{85.8}{11.1} & \fcell{7.14}{6.11} & \fcell{569}{40.7} & \fcell{44.0}{2.53} & \fcell{532}{43.1} & \fcell{\textbf{4.56}}{1.98} & \fcell{19.9}{4.18} & \fcell{11.7}{7.92} & \fcell{553}{114} & \fcell{57.2}{3.72} & \fcell{1824}{72.7} & \fcell{69.8}{19.6} & \fcell{89.2}{8.26} & \fcell{9.47}{3.04} & \fcell{611}{37.2} & \fcell{245}{442} \\
\rowcolor{black!5}
GOTOV & \fcell{219}{25.3} & \fcell{151}{26.4} & \fcell{150}{11.4} & \fcell{58.9}{22.4} & \fcell{407}{24.2} & \fcell{17.8}{13.4} & \fcell{1658}{778} & \fcell{323}{35.7} & \fcell{1229}{64.2} & \fcell{\textbf{7.87}}{2.51} & \fcell{60.3}{21.5} & \fcell{26.7}{12.0} & \fcell{1237}{77.6} & \fcell{211}{17.7} & \fcell{3105}{127} & \fcell{164}{25.9} & \fcell{661}{55.0} & \fcell{67.5}{7.83} & \fcell{1205}{79.2} & \fcell{577}{797} \\
UTD-MHAD & \fcell{43.4}{21.1} & \fcell{38.3}{10.6} & \fcell{60.4}{11.1} & \fcell{28.0}{6.69} & \fcell{87.2}{17.4} & \fcell{7.77}{7.52} & \fcell{571}{46.3} & \fcell{44.4}{4.35} & \fcell{522}{36.4} & \fcell{\textbf{4.34}}{1.59} & \fcell{14.5}{1.51} & \fcell{12.4}{11.1} & \fcell{528}{48.7} & \fcell{58.7}{9.39} & \fcell{1810}{82.8} & \fcell{69.3}{14.0} & \fcell{91.0}{13.7} & \fcell{30.3}{5.09} & \fcell{29.0}{5.12} & \fcell{213}{430} \\
\rowcolor{black!5}
WISDM-19 (Watch) & \fcell{32.8}{48.1} & \fcell{21.1}{6.32} & \fcell{21.6}{9.35} & \fcell{15.7}{5.30} & \fcell{62.6}{8.67} & \fcell{7.08}{2.55} & \fcell{206}{35.8} & \fcell{40.5}{10.1} & \fcell{184}{16.7} & \fcell{\textbf{3.90}}{1.42} & \fcell{8.17}{0.68} & \fcell{11.2}{13.6} & \fcell{187}{32.6} & \fcell{39.7}{2.98} & \fcell{715}{34.4} & \fcell{39.0}{20.3} & \fcell{33.4}{7.91} & \fcell{88.4}{6.92} & \fcell{3344}{274} & \fcell{266}{763} \\
UCA-EHAR & \fcell{22.3}{8.10} & \fcell{24.9}{6.83} & \fcell{30.0}{7.35} & \fcell{17.3}{3.38} & \fcell{76.2}{19.2} & \fcell{5.79}{1.94} & \fcell{284}{48.9} & \fcell{45.4}{4.09} & \fcell{278}{69.1} & \fcell{\textbf{4.02}}{1.93} & \fcell{11.9}{8.31} & \fcell{10.6}{8.82} & \fcell{260}{30.9} & \fcell{45.7}{10.7} & \fcell{958}{38.7} & \fcell{39.8}{5.32} & \fcell{56.8}{30.1} & \fcell{17.2}{4.05} & \fcell{354}{32.9} & \fcell{134}{229} \\
\rowcolor{black!5}
UP-Fall & \fcell{80.9}{28.2} & \fcell{53.7}{12.8} & \fcell{38.3}{4.64} & \fcell{22.6}{14.0} & \fcell{228}{28.2} & \fcell{9.61}{4.04} & \fcell{366}{21.8} & \fcell{623}{33.1} & \fcell{308}{14.9} & \fcell{\textbf{5.66}}{2.00} & \fcell{46.4}{4.30} & \fcell{13.7}{1.49} & \fcell{310}{32.0} & \fcell{96.7}{12.6} & \fcell{737}{57.1} & \fcell{50.8}{18.2} & \fcell{174}{17.7} & \fcell{19.0}{4.11} & \fcell{212}{25.3} & \fcell{179}{211} \\
UMAFall & \fcell{655}{36.7} & \fcell{516}{42.1} & \fcell{423}{65.2} & \fcell{123}{8.02} & \fcell{1175}{68.5} & \fcell{34.2}{5.40} & \fcell{4247}{173} & \fcell{712}{45.5} & \fcell{3255}{112} & \fcell{\textbf{21.3}}{4.72} & \fcell{199}{14.6} & \fcell{58.3}{28.1} & \fcell{3229}{80.8} & \fcell{642}{52.0} & \fcell{7302}{239} & \fcell{518}{26.3} & \fcell{2193}{79.7} & \fcell{35.0}{5.38} & \fcell{90.1}{9.85} & \fcell{1338}{1925} \\
\rowcolor{black!5}
RealLifeHAR & \fcell{30.2}{13.3} & \fcell{28.6}{7.43} & \fcell{27.6}{11.9} & \fcell{20.3}{17.4} & \fcell{107}{19.9} & \fcell{7.04}{6.75} & \fcell{236}{19.4} & \fcell{120}{23.1} & \fcell{219}{26.7} & \fcell{\textbf{4.28}}{1.94} & \fcell{14.1}{1.95} & \fcell{15.9}{19.0} & \fcell{217}{22.6} & \fcell{53.4}{12.4} & \fcell{741}{30.3} & \fcell{38.8}{14.2} & \fcell{70.1}{12.6} & \fcell{113}{13.6} & \fcell{10660}{1772} & \fcell{670}{2425} \\
UniMiB-SHAR & \fcell{25.1}{8.38} & \fcell{26.9}{10.4} & \fcell{59.3}{31.7} & \fcell{26.2}{3.50} & \fcell{59.2}{3.78} & \fcell{8.70}{4.90} & \fcell{541}{76.6} & \fcell{29.1}{30.8} & \fcell{506}{47.3} & \fcell{\textbf{4.10}}{2.04} & \fcell{14.7}{11.1} & \fcell{15.8}{12.1} & \fcell{750}{1027} & \fcell{46.7}{14.9} & \fcell{1820}{74.0} & \fcell{73.2}{32.0} & \fcell{56.7}{18.2} & \fcell{16.2}{3.75} & \fcell{152}{22.2} & \fcell{223}{443} \\
\rowcolor{black!5}
AICOS-HAR & \fcell{51.0}{9.74} & \fcell{50.6}{29.2} & \fcell{63.2}{4.74} & \fcell{37.1}{48.2} & \fcell{119}{12.4} & \fcell{8.14}{5.03} & \fcell{619}{57.7} & \fcell{69.2}{7.49} & \fcell{557}{45.0} & \fcell{\textbf{4.51}}{1.69} & \fcell{23.9}{8.74} & \fcell{13.2}{6.04} & \fcell{554}{44.7} & \fcell{75.7}{13.5} & \fcell{1790}{75.4} & \fcell{71.8}{11.2} & \fcell{137}{12.4} & \fcell{75.5}{8.78} & \fcell{3458}{441} & \fcell{409}{851} \\
WISDM-19 (Phone) & \fcell{15.8}{2.49} & \fcell{21.5}{13.1} & \fcell{21.1}{5.30} & \fcell{23.4}{49.7} & \fcell{65.0}{14.0} & \fcell{6.53}{8.12} & \fcell{201}{22.9} & \fcell{40.5}{14.2} & \fcell{202}{48.9} & \fcell{\textbf{3.91}}{1.72} & \fcell{8.98}{3.22} & \fcell{8.62}{1.42} & \fcell{185}{17.9} & \fcell{41.7}{7.73} & \fcell{722}{34.5} & \fcell{35.1}{9.52} & \fcell{32.9}{8.16} & \fcell{121}{8.37} & \fcell{4374}{471} & \fcell{323}{995} \\
\midrule
\textbf{Mean} & \fcell{131}{163} & \fcell{98.1}{123} & \fcell{103}{94.4} & \fcell{42.7}{27.2} & \fcell{288}{341} & \fcell{12.8}{8.64} & \fcell{1018}{970} & \fcell{568}{1222} & \fcell{821}{738} & \fcell{6.56}{4.06} & \fcell{65.5}{93.6} & \fcell{20.6}{13.2} & \fcell{826}{731} & \fcell{143}{150} & \fcell{2187}{1845} & \fcell{118}{135} & \fcell{384}{526} & \fcell{38.3}{41.3} & \fcell{1180}{2122} & \fcell{424}{567} \\
\bottomrule
\end{tabular}}
}

\end{appendixtable}

\begin{appendixtable}[90]
    \centering
    \caption{Peak memory overhead (MB) during inference on the smartwatch with 2\,s windows. Bold marks the lowest value per dataset. --: no increase over the baseline memory was measurable, because the memory was already resident when the baseline was taken, and these cases are left out of all summaries.}
    \label{tab:deploy-watch-memory-w2s}
    \providecommand{\nofit}[1]{\textcolor{gray}{#1}}
\providecommand{\faileddeployment}{\textcolor{gray}{Failed}}
\providecommand{\todomissing}{\textcolor{red}{TODO}}
\providecommand{\unmeasured}{\textcolor{gray}{--}}
\providecommand{\fcell}[2]{\begin{tabular}[c]{@{}c@{}}#1\\[-2pt]{\scriptsize$\pm$#2}\end{tabular}}
\providecommand{\tableheaderangle}{90}
\providecommand{\tablemaxheight}{0.9\textheight}
\providecommand{\fittable}[1]{\sbox2{#1}%
  \sbox2{\resizebox{\linewidth}{!}{\usebox2}}%
  \dimen2=\ht2\advance\dimen2 by\dp2
  \ifdim\dimen2>\tablemaxheight\resizebox*{!}{\tablemaxheight}{\usebox2}\else\usebox2\fi}

{%
\setlength{\tabcolsep}{2pt}
\setlength{\aboverulesep}{0pt}\setlength{\belowrulesep}{0pt}
\renewcommand{\arraystretch}{1.15}
\footnotesize
\fittable{%
\begin{tabular}{@{}lccccccccccccccccccc|c@{}}
Dataset & \rotatebox{\tableheaderangle}{CNN-HAR\hspace{4pt}} & \rotatebox{\tableheaderangle}{Triple-Cross-Attn\hspace{4pt}} & \rotatebox{\tableheaderangle}{TinyHAR\hspace{4pt}} & \rotatebox{\tableheaderangle}{TinierHAR\hspace{4pt}} & \rotatebox{\tableheaderangle}{DANA\hspace{4pt}} & \rotatebox{\tableheaderangle}{LITEWAY-Light\hspace{4pt}} & \rotatebox{\tableheaderangle}{Attend-and-Discriminate\hspace{4pt}} & \rotatebox{\tableheaderangle}{DynamicWHAR\hspace{4pt}} & \rotatebox{\tableheaderangle}{DeepConvLSTM-Attn\hspace{4pt}} & \rotatebox{\tableheaderangle}{Random Forest\hspace{4pt}} & \rotatebox{\tableheaderangle}{MLP-HAR\hspace{4pt}} & \rotatebox{\tableheaderangle}{LITEWAY\hspace{4pt}} & \rotatebox{\tableheaderangle}{DeepConvLSTM\hspace{4pt}} & \rotatebox{\tableheaderangle}{MLP-Mixer\hspace{4pt}} & \rotatebox{\tableheaderangle}{Guan-LSTM\hspace{4pt}} & \rotatebox{\tableheaderangle}{SA-HAR\hspace{4pt}} & \rotatebox{\tableheaderangle}{DeepConvShallowLSTM\hspace{4pt}} & \rotatebox{\tableheaderangle}{SVM\hspace{4pt}} & \rotatebox{\tableheaderangle}{k-NN\hspace{4pt}} & \rotatebox{\tableheaderangle}{\textbf{Mean}\hspace{4pt}} \\
\midrule
SAD & 8.59 & 6.65 & 5.49 & 1.85 & 24.5 & \textbf{1.31} & 24.8 & 58.1 & 20.6 & 6.05 & 13.5 & 1.63 & 21.8 & 9.99 & 6.24 & 10.3 & 21.1 & 25.0 & 27.9 & 15.5\,{\scriptsize$\pm$13.7} \\
UCI-HAR & 1.93 & 3.00 & 3.30 & 1.29 & 6.73 & \textbf{1.04} & 10.8 & 10.3 & 6.11 & 11.3 & 2.00 & 1.24 & 5.95 & 4.42 & 6.00 & 4.16 & 5.41 & 9.95 & 7.84 & 5.41\,{\scriptsize$\pm$3.39} \\
MotionSense & 3.45 & 3.59 & 3.51 & 1.43 & 10.5 & \textbf{1.02} & 14.1 & 18.6 & 9.87 & 9.34 & 2.69 & 1.34 & 9.20 & 5.85 & 5.66 & 5.51 & 9.39 & 25.7 & 20.8 & 8.50\,{\scriptsize$\pm$6.99} \\
MHEALTH & 4.08 & 4.95 & 3.88 & 1.44 & 12.7 & \textbf{1.08} & 15.4 & 22.8 & 11.1 & 8.61 & 3.36 & 1.37 & 10.7 & 6.35 & 6.20 & 5.41 & 11.0 & 10.3 & 9.14 & 7.89\,{\scriptsize$\pm$5.48} \\
WEAR & 2.65 & 4.16 & 3.29 & 1.35 & 7.76 & \textbf{0.96} & 12.0 & 13.1 & 7.40 & 35.6 & 2.20 & 1.30 & 7.15 & 4.83 & 6.59 & 4.28 & 5.45 & 24.2 & 17.1 & 8.49\,{\scriptsize$\pm$8.86} \\
DSADS & 4.53 & 4.92 & 3.09 & 1.53 & 18.5 & \textbf{1.00} & 15.9 & 45.3 & 16.5 & 15.1 & 7.22 & 1.40 & 16.4 & 5.32 & 6.01 & 5.32 & 15.5 & 21.0 & 18.0 & 11.7\,{\scriptsize$\pm$10.6} \\
Opportunity (Loc.) & 11.2 & 6.74 & 6.16 & 2.33 & 48.6 & \textbf{1.47} & 37.1 & 136 & 39.3 & 6.64 & 50.8 & 1.79 & 40.2 & 13.2 & 6.11 & 10.7 & 38.8 & 37.5 & 59.1 & 29.1\,{\scriptsize$\pm$32.4} \\
WISDM & 1.23 & 2.27 & 1.98 & 1.24 & 3.42 & \textbf{0.92} & 5.39 & 5.44 & 3.39 & 14.2 & 1.41 & 1.24 & 3.10 & 3.45 & 5.64 & 3.50 & 2.76 & 10.1 & 12.9 & 4.40\,{\scriptsize$\pm$3.91} \\
REALDISP & 13.3 & 29.6 & 6.92 & 2.21 & 44.2 & \textbf{1.62} & 39.0 & 118 & 36.3 & 42.2 & 40.4 & 2.13 & 38.6 & 18.8 & 7.16 & 15.7 & 37.3 & 262 & 95.1 & 44.7\,{\scriptsize$\pm$60.7} \\
PAMAP2 & 10.3 & 13.9 & 7.31 & 2.18 & 23.4 & \textbf{1.58} & 33.2 & 52.1 & 20.9 & 21.0 & 11.5 & 1.94 & 21.9 & 17.3 & 7.36 & 15.5 & 22.0 & 34.2 & 38.4 & 18.7\,{\scriptsize$\pm$13.5} \\
HAPT & 1.58 & 2.85 & 3.07 & 1.28 & 5.19 & \textbf{0.95} & 9.90 & 7.96 & 5.06 & 16.6 & 1.83 & 1.16 & 4.48 & 3.93 & 6.07 & 4.00 & 4.30 & 6.29 & 4.54 & 4.79\,{\scriptsize$\pm$3.70} \\
HHAR & 1.55 & 2.46 & 1.83 & 1.24 & 4.28 & \textbf{0.95} & 5.88 & 7.84 & 4.52 & 16.2 & 1.52 & 1.17 & 4.03 & 3.62 & 5.34 & 3.66 & 3.56 & 25.4 & 24.6 & 6.30\,{\scriptsize$\pm$7.44} \\
Hang-Time & 1.44 & 2.41 & 3.01 & 1.31 & 3.42 & \textbf{0.94} & 8.56 & 5.64 & 3.80 & 18.1 & 1.74 & 1.02 & 3.53 & 3.81 & 5.77 & 3.97 & 3.32 & 14.7 & 14.1 & 5.29\,{\scriptsize$\pm$5.01} \\
w-HAR & 4.66 & 5.20 & 11.6 & 1.54 & 8.03 & \textbf{1.13} & 40.9 & 10.8 & 9.12 & 8.80 & 14.6 & 1.54 & 8.51 & 7.94 & 9.80 & 15.0 & 9.34 & 3.63 & 3.81 & 9.26\,{\scriptsize$\pm$8.68} \\
HuGaDB & 6.30 & 6.26 & 4.84 & 1.76 & 18.1 & \textbf{1.39} & 21.1 & 35.3 & 15.7 & 16.7 & 6.43 & 1.60 & 16.0 & 9.18 & 6.44 & 8.47 & 14.9 & 23.3 & 33.8 & 13.0\,{\scriptsize$\pm$10.1} \\
HARTH & 1.66 & 2.74 & 3.13 & 1.07 & 5.21 & \textbf{0.75} & 9.89 & 8.09 & 5.06 & 21.2 & 1.89 & 1.23 & 4.76 & 4.18 & 6.14 & 3.86 & 4.55 & 13.7 & 34.7 & 7.04\,{\scriptsize$\pm$8.35} \\
Daphnet & 2.50 & 3.05 & 3.82 & 1.29 & 7.14 & \textbf{0.97} & 13.2 & 9.62 & 6.60 & 5.19 & 2.24 & 1.29 & 6.21 & 4.83 & 6.21 & 4.55 & 5.75 & 3.07 & 11.4 & 5.20\,{\scriptsize$\pm$3.38} \\
USC-HAD & 2.22 & 3.60 & 5.29 & 1.34 & 5.84 & \textbf{0.96} & 18.0 & 8.66 & 6.55 & 20.3 & 3.05 & 1.28 & 5.94 & 5.20 & 7.12 & 5.49 & 6.15 & 13.5 & 10.4 & 6.89\,{\scriptsize$\pm$5.36} \\
HAR70+ & 1.83 & 2.74 & 3.11 & 1.28 & 5.06 & \textbf{0.95} & 9.80 & 8.03 & 5.08 & 10.1 & 1.89 & 1.28 & 4.73 & 4.17 & 5.98 & 3.69 & 4.32 & 7.53 & 13.1 & 4.98\,{\scriptsize$\pm$3.39} \\
GOTOV & 6.27 & 7.92 & 6.29 & 1.62 & 15.2 & \textbf{1.22} & 23.4 & 27.2 & 12.7 & 23.8 & 4.93 & 1.58 & 13.4 & 9.84 & 7.23 & 8.82 & 13.2 & 79.3 & 55.3 & 16.8\,{\scriptsize$\pm$19.7} \\
UTD-MHAD & 1.74 & 3.58 & 3.12 & 1.29 & 5.08 & \textbf{0.95} & 9.75 & 8.03 & 5.05 & 16.2 & 1.86 & 1.27 & 4.61 & 4.11 & 6.77 & 3.97 & 4.42 & 4.91 & 1.99 & 4.67\,{\scriptsize$\pm$3.66} \\
WISDM-19 (Watch) & 1.63 & 2.53 & 2.09 & 1.22 & 4.53 & \textbf{0.96} & 5.95 & 7.75 & 4.58 & 43.1 & 1.50 & 1.20 & 4.21 & 3.77 & 5.67 & 3.60 & 3.58 & 104 & 33.7 & 12.4\,{\scriptsize$\pm$24.9} \\
UCA-EHAR & 1.62 & 2.32 & 2.11 & 1.26 & 4.65 & \textbf{0.71} & 6.43 & 8.67 & 4.67 & 21.3 & 1.58 & 1.33 & 4.54 & 3.62 & 5.63 & 3.69 & 3.76 & 15.3 & 11.2 & 5.49\,{\scriptsize$\pm$5.33} \\
UP-Fall & 4.44 & 3.80 & 3.49 & 1.59 & 18.1 & \textbf{1.05} & 15.0 & 39.5 & 15.9 & 16.4 & 6.45 & 1.60 & 15.4 & 6.34 & 5.72 & 5.32 & 14.9 & 23.6 & 19.1 & 11.5\,{\scriptsize$\pm$9.77} \\
UMAFall & 13.3 & 23.9 & 12.2 & 2.58 & 24.2 & \textbf{1.73} & 45.6 & 41.9 & 21.8 & 14.4 & 13.3 & 2.61 & 21.7 & 23.6 & 10.7 & 21.2 & 22.6 & 15.7 & 8.83 & 18.0\,{\scriptsize$\pm$11.7} \\
RealLifeHAR & 2.34 & 2.75 & 2.32 & 1.37 & 8.48 & \textbf{0.98} & 8.33 & 15.6 & 6.74 & 16.8 & 1.99 & 1.29 & 6.88 & 4.34 & 5.64 & 3.98 & 5.92 & 147 & \nofit{M} & 13.5\,{\scriptsize$\pm$33.7} \\
UniMiB-SHAR & 1.42 & 2.72 & 2.78 & 1.27 & 3.79 & \textbf{0.93} & 8.81 & 5.60 & 3.94 & 25.4 & 1.70 & 1.24 & 3.65 & 3.55 & 6.15 & 3.74 & 3.54 & 5.25 & 2.84 & 4.65\,{\scriptsize$\pm$5.39} \\
AICOS-HAR & 1.92 & 2.49 & 2.21 & 1.38 & 6.38 & \textbf{0.95} & 7.10 & 7.42 & 5.47 & 21.3 & 1.66 & 1.26 & 5.37 & 3.68 & 5.64 & 3.78 & 4.75 & 79.6 & 58.4 & 11.6\,{\scriptsize$\pm$21.0} \\
WISDM-19 (Phone) & 1.56 & 2.52 & 2.08 & 1.27 & 4.50 & \textbf{0.97} & 5.96 & 7.80 & 4.33 & 42.1 & 1.55 & 1.24 & 4.10 & 3.72 & 5.69 & 3.68 & 3.89 & 150 & 42.5 & 15.3\,{\scriptsize$\pm$35.0} \\
\midrule
\textbf{Mean} & \fcell{4.18}{3.68} & \fcell{5.71}{6.35} & \fcell{4.25}{2.60} & \fcell{1.51}{0.38} & \fcell{12.3}{11.5} & \fcell{1.08}{0.25} & \fcell{16.6}{11.8} & \fcell{25.9}{32.0} & \fcell{11.0}{9.31} & \fcell{18.8}{10.5} & \fcell{7.13}{11.5} & \fcell{1.43}{0.34} & \fcell{10.9}{9.83} & \fcell{7.00}{5.13} & \fcell{6.44}{1.19} & \fcell{6.72}{4.67} & \fcell{10.5}{9.68} & \fcell{41.3}{58.3} & \fcell{24.7}{21.7} & \fcell{11.4}{10.2} \\
\bottomrule
\end{tabular}}
}

\end{appendixtable}

\begin{appendixtable}[90]
    \centering
    \caption{Peak memory overhead (MB) during inference on the smartwatch with 3\,s windows. Bold marks the lowest value per dataset. --: no increase over the baseline memory was measurable, because the memory was already resident when the baseline was taken, and these cases are left out of all summaries.}
    \label{tab:deploy-watch-memory-w3s}
    \providecommand{\nofit}[1]{\textcolor{gray}{#1}}
\providecommand{\faileddeployment}{\textcolor{gray}{Failed}}
\providecommand{\todomissing}{\textcolor{red}{TODO}}
\providecommand{\unmeasured}{\textcolor{gray}{--}}
\providecommand{\fcell}[2]{\begin{tabular}[c]{@{}c@{}}#1\\[-2pt]{\scriptsize$\pm$#2}\end{tabular}}
\providecommand{\tableheaderangle}{90}
\providecommand{\tablemaxheight}{0.9\textheight}
\providecommand{\fittable}[1]{\sbox2{#1}%
  \sbox2{\resizebox{\linewidth}{!}{\usebox2}}%
  \dimen2=\ht2\advance\dimen2 by\dp2
  \ifdim\dimen2>\tablemaxheight\resizebox*{!}{\tablemaxheight}{\usebox2}\else\usebox2\fi}

{%
\setlength{\tabcolsep}{2pt}
\setlength{\aboverulesep}{0pt}\setlength{\belowrulesep}{0pt}
\renewcommand{\arraystretch}{1.15}
\footnotesize
\fittable{%
\begin{tabular}{@{}lccccccccccccccccccc|c@{}}
Dataset & \rotatebox{\tableheaderangle}{CNN-HAR\hspace{4pt}} & \rotatebox{\tableheaderangle}{Triple-Cross-Attn\hspace{4pt}} & \rotatebox{\tableheaderangle}{TinyHAR\hspace{4pt}} & \rotatebox{\tableheaderangle}{TinierHAR\hspace{4pt}} & \rotatebox{\tableheaderangle}{DANA\hspace{4pt}} & \rotatebox{\tableheaderangle}{LITEWAY-Light\hspace{4pt}} & \rotatebox{\tableheaderangle}{Attend-and-Discriminate\hspace{4pt}} & \rotatebox{\tableheaderangle}{DynamicWHAR\hspace{4pt}} & \rotatebox{\tableheaderangle}{DeepConvLSTM-Attn\hspace{4pt}} & \rotatebox{\tableheaderangle}{Random Forest\hspace{4pt}} & \rotatebox{\tableheaderangle}{MLP-HAR\hspace{4pt}} & \rotatebox{\tableheaderangle}{LITEWAY\hspace{4pt}} & \rotatebox{\tableheaderangle}{DeepConvLSTM\hspace{4pt}} & \rotatebox{\tableheaderangle}{MLP-Mixer\hspace{4pt}} & \rotatebox{\tableheaderangle}{Guan-LSTM\hspace{4pt}} & \rotatebox{\tableheaderangle}{SA-HAR\hspace{4pt}} & \rotatebox{\tableheaderangle}{DeepConvShallowLSTM\hspace{4pt}} & \rotatebox{\tableheaderangle}{SVM\hspace{4pt}} & \rotatebox{\tableheaderangle}{k-NN\hspace{4pt}} & \rotatebox{\tableheaderangle}{\textbf{Mean}\hspace{4pt}} \\
\midrule
SAD & 10.1 & 9.55 & 6.53 & 2.02 & 25.8 & \textbf{1.40} & 30.1 & 59.0 & 21.4 & 5.51 & 12.6 & 1.88 & 23.3 & 15.6 & 6.65 & 13.3 & 22.4 & 23.5 & 17.2 & 16.2\,{\scriptsize$\pm$13.6} \\
UCI-HAR & 2.56 & 3.28 & 4.43 & 1.39 & 7.22 & \textbf{1.02} & 15.1 & 7.63 & 7.34 & 9.63 & 2.78 & 1.30 & 6.79 & 4.89 & 6.46 & 5.13 & 6.20 & 6.24 & 4.60 & 5.47\,{\scriptsize$\pm$3.34} \\
MotionSense & 4.45 & 4.20 & 4.96 & 1.52 & 10.4 & \textbf{1.07} & 18.2 & 18.0 & 9.61 & 7.53 & 3.19 & 1.39 & 10.3 & 7.02 & 6.54 & 6.29 & 9.45 & 20.1 & 21.3 & 8.71\,{\scriptsize$\pm$6.38} \\
MHEALTH & 5.55 & 6.32 & 5.46 & 1.54 & 13.0 & \textbf{1.05} & 19.7 & 23.5 & 11.7 & 6.53 & 3.91 & 1.50 & 11.4 & 7.83 & 6.52 & 7.25 & 11.2 & 7.62 & 6.33 & 8.31\,{\scriptsize$\pm$5.80} \\
WEAR & 3.24 & 5.39 & 4.65 & 1.20 & 8.52 & \textbf{1.08} & 16.6 & 13.3 & 8.72 & 32.7 & 2.91 & 1.37 & 8.27 & 4.99 & 6.99 & 5.14 & 7.72 & 17.0 & 14.6 & 8.65\,{\scriptsize$\pm$7.64} \\
DSADS & 5.48 & 6.99 & 4.26 & 1.65 & 19.9 & \textbf{1.11} & 18.5 & 43.0 & 17.0 & 15.6 & 7.39 & 1.54 & 16.9 & 7.53 & 6.01 & 7.77 & 16.3 & 26.6 & 27.0 & 13.2\,{\scriptsize$\pm$10.8} \\
Opportunity (Loc.) & 13.8 & 9.49 & 7.05 & 2.32 & 49.7 & \textbf{1.57} & 42.3 & 131 & 40.2 & 4.88 & 51.1 & 1.94 & 42.4 & 19.6 & 6.39 & 15.6 & 40.6 & 23.6 & 37.9 & 28.5\,{\scriptsize$\pm$30.5} \\
WISDM & 1.30 & 2.33 & 2.12 & 1.27 & 3.21 & \textbf{0.70} & 5.96 & 5.47 & 3.63 & 12.7 & 1.40 & 1.17 & 3.38 & 3.62 & 5.35 & 3.47 & 2.80 & 8.82 & 8.45 & 4.06\,{\scriptsize$\pm$3.13} \\
REALDISP & 17.3 & 48.8 & 8.96 & 2.82 & 46.6 & \textbf{1.96} & 47.7 & 119 & 37.7 & 34.7 & 41.3 & 2.90 & 41.4 & 24.9 & 8.25 & 21.8 & 41.3 & 174 & 56.7 & 41.0\,{\scriptsize$\pm$42.2} \\
PAMAP2 & 13.9 & 22.2 & 10.4 & 2.58 & 27.4 & \textbf{0.74} & 44.1 & 53.3 & 24.2 & 16.6 & 13.3 & 2.42 & 24.6 & 22.5 & 8.84 & 19.2 & 24.8 & 26.1 & 25.9 & 20.2\,{\scriptsize$\pm$13.3} \\
HAPT & 2.19 & 3.17 & 3.98 & 1.35 & 5.60 & \textbf{0.93} & 14.0 & 8.26 & 5.82 & 15.3 & 2.43 & 1.20 & 4.94 & 4.48 & 6.70 & 3.99 & 5.15 & 4.48 & 3.14 & 5.11\,{\scriptsize$\pm$3.86} \\
HHAR & 2.17 & 3.04 & 4.23 & 1.31 & 5.72 & \textbf{0.99} & 13.8 & 8.33 & 5.68 & 16.0 & 2.32 & 1.37 & 5.36 & 4.64 & 6.64 & 4.57 & 4.69 & 20.2 & 14.2 & 6.60\,{\scriptsize$\pm$5.51} \\
Hang-Time & 1.54 & 2.30 & 4.05 & 1.30 & 3.71 & \textbf{0.90} & 12.7 & 5.54 & 4.52 & 16.5 & 2.24 & 1.23 & 4.01 & 3.90 & 6.50 & 4.26 & 3.81 & 10.4 & 9.75 & 5.22\,{\scriptsize$\pm$4.24} \\
w-HAR & 5.91 & 6.47 & 16.6 & 1.85 & 9.41 & \textbf{1.20} & 59.8 & 12.7 & 11.1 & 5.37 & 26.1 & 1.60 & 10.2 & 10.1 & 14.8 & 25.6 & 11.4 & 3.50 & 3.12 & 12.5\,{\scriptsize$\pm$13.5} \\
HuGaDB & 7.68 & 10.2 & 6.01 & 1.82 & 17.8 & \textbf{1.33} & 26.4 & 36.9 & 17.0 & 13.5 & 6.98 & 1.69 & 16.9 & 11.7 & 7.12 & 10.6 & 16.4 & 17.9 & 21.1 & 13.1\,{\scriptsize$\pm$9.00} \\
HARTH & 2.05 & 3.43 & 4.33 & 1.28 & 5.30 & \textbf{1.00} & 14.0 & 8.34 & 5.79 & 19.7 & 2.43 & 1.30 & 5.27 & 4.72 & 6.72 & 4.27 & 5.10 & 10.7 & 20.7 & 6.65\,{\scriptsize$\pm$5.77} \\
Daphnet & 2.99 & 3.25 & 5.35 & 1.27 & 7.88 & \textbf{1.04} & 18.7 & 11.1 & 7.94 & 4.29 & 3.12 & 1.38 & 7.26 & 5.33 & 6.60 & 5.56 & 6.95 & 2.31 & 6.68 & 5.74\,{\scriptsize$\pm$4.14} \\
USC-HAD & 3.23 & 4.12 & 7.46 & 1.37 & 6.85 & \textbf{1.04} & 25.8 & 9.04 & 7.91 & 17.0 & 5.22 & 1.38 & 7.14 & 4.94 & 8.85 & 8.20 & 7.83 & 9.32 & 6.17 & 7.51\,{\scriptsize$\pm$5.74} \\
HAR70+ & 2.06 & 3.05 & 4.11 & 1.33 & 5.36 & \textbf{0.96} & 13.8 & 8.32 & 5.66 & 7.01 & 2.38 & 1.30 & 4.99 & 4.71 & 6.49 & 4.50 & 4.95 & 5.37 & 11.6 & 5.15\,{\scriptsize$\pm$3.36} \\
GOTOV & 8.51 & 12.7 & 7.63 & 1.93 & 16.4 & \textbf{1.30} & 32.0 & 28.2 & 14.6 & 21.7 & 6.58 & 1.70 & 15.1 & 12.8 & 8.87 & 10.5 & 14.5 & 50.6 & 38.1 & 16.0\,{\scriptsize$\pm$13.0} \\
UTD-MHAD & 2.16 & 3.76 & 3.97 & 1.30 & 5.38 & 1.01 & 13.9 & 8.34 & 5.68 & 9.78 & 2.39 & \textbf{1.01} & 5.28 & 4.69 & 7.34 & 4.35 & 5.66 & 4.42 & 1.91 & 4.86\,{\scriptsize$\pm$3.27} \\
WISDM-19 (Watch) & 1.51 & 2.52 & 2.10 & 1.23 & 4.51 & \textbf{0.74} & 6.73 & 7.91 & 4.45 & 41.5 & 1.59 & 0.98 & 4.39 & 3.71 & 5.96 & 3.56 & 3.90 & 66.9 & 29.8 & 10.2\,{\scriptsize$\pm$17.3} \\
UCA-EHAR & 1.80 & 2.79 & 2.75 & 1.26 & 5.56 & \textbf{1.06} & 8.44 & 8.51 & 4.91 & 20.6 & 1.62 & 1.27 & 4.81 & 4.12 & 5.95 & 3.97 & 4.97 & 10.7 & 7.54 & 5.40\,{\scriptsize$\pm$4.59} \\
UP-Fall & 4.74 & 3.81 & 3.51 & 1.61 & 18.6 & \textbf{1.04} & 16.5 & 39.2 & 15.7 & 12.4 & 6.49 & 1.33 & 15.5 & 5.85 & 5.87 & 6.13 & 15.6 & 14.6 & 16.0 & 10.8\,{\scriptsize$\pm$9.16} \\
UMAFall & 18.4 & 32.6 & 15.3 & 3.05 & 28.8 & \textbf{2.06} & 66.8 & 42.8 & 28.3 & 11.1 & 22.2 & 3.16 & 23.0 & 32.6 & 13.8 & 28.3 & 25.2 & 10.4 & 5.90 & 21.8\,{\scriptsize$\pm$15.9} \\
RealLifeHAR & 2.27 & 2.61 & 2.56 & 1.35 & 6.78 & \textbf{1.00} & 9.26 & 15.6 & 7.57 & 15.0 & 2.09 & 1.26 & 7.02 & 4.59 & 5.38 & 4.22 & 6.78 & 102 & 158 & 18.7\,{\scriptsize$\pm$40.6} \\
UniMiB-SHAR & 1.59 & 2.95 & 4.03 & 1.26 & 3.91 & \textbf{0.93} & 12.5 & 5.85 & 4.30 & 24.5 & 2.27 & 1.26 & 3.45 & 3.87 & 6.69 & 4.25 & 3.97 & 3.96 & 2.40 & 4.95\,{\scriptsize$\pm$5.40} \\
AICOS-HAR & 2.91 & 3.53 & 4.55 & 1.43 & 7.61 & \textbf{0.97} & 15.7 & 8.04 & 7.88 & 21.1 & 2.51 & 1.33 & 7.47 & 5.43 & 6.55 & 5.14 & 6.84 & 64.5 & 38.7 & 11.2\,{\scriptsize$\pm$15.7} \\
WISDM-19 (Phone) & 1.58 & 2.51 & 2.11 & 1.27 & 4.82 & \textbf{0.86} & 6.81 & 7.88 & 4.50 & 41.8 & 1.49 & 1.25 & 4.33 & 3.79 & 5.80 & 3.95 & 3.92 & 99.6 & 30.6 & 12.0\,{\scriptsize$\pm$23.7} \\
\midrule
\textbf{Mean} & \fcell{5.28}{4.91} & \fcell{7.84}{10.3} & \fcell{5.64}{3.48} & \fcell{1.62}{0.50} & \fcell{13.2}{12.2} & \fcell{1.11}{0.31} & \fcell{22.3}{15.8} & \fcell{26.0}{31.6} & \fcell{12.1}{9.79} & \fcell{16.6}{10.3} & \fcell{8.36}{12.1} & \fcell{1.53}{0.51} & \fcell{11.9}{10.4} & \fcell{8.78}{7.42} & \fcell{7.26}{2.17} & \fcell{8.65}{6.93} & \fcell{11.7}{10.3} & \fcell{29.2}{38.8} & \fcell{22.3}{29.5} & \fcell{11.6}{8.23} \\
\bottomrule
\end{tabular}}
}

\end{appendixtable}

\clearpage

\begin{appendixtable}[90]
    \raggedright
    \subsubsection{MCU}
    \label{sec:appendix-deploy-mcu}

    \noindent Tables~\ref{tab:deploy-mcu-latency-w2s} to~\ref{tab:deploy-mcu-energy-w3s} report latency, peak RAM, flash, and energy per inference on the MCU.
    \par\medskip
    \centering
    \caption{Latency (ms) to classify one window on the MCU with 2\,s windows. Bold marks the lowest value per dataset. F: does not fit in flash, R: does not fit in RAM.}
    \label{tab:deploy-mcu-latency-w2s}
    \providecommand{\nofit}[1]{\textcolor{gray}{#1}}
\providecommand{\faileddeployment}{\textcolor{gray}{Failed}}
\providecommand{\todomissing}{\textcolor{red}{TODO}}
\providecommand{\unmeasured}{\textcolor{gray}{--}}
\providecommand{\fcell}[2]{\begin{tabular}[c]{@{}c@{}}#1\\[-2pt]{\scriptsize$\pm$#2}\end{tabular}}
\providecommand{\tableheaderangle}{90}
\providecommand{\tablemaxheight}{0.9\textheight}
\providecommand{\fittable}[1]{\sbox2{#1}%
  \sbox2{\resizebox{\linewidth}{!}{\usebox2}}%
  \dimen2=\ht2\advance\dimen2 by\dp2
  \ifdim\dimen2>\tablemaxheight\resizebox*{!}{\tablemaxheight}{\usebox2}\else\usebox2\fi}

{%
\setlength{\tabcolsep}{2pt}
\setlength{\aboverulesep}{0pt}\setlength{\belowrulesep}{0pt}
\renewcommand{\arraystretch}{1.15}
\footnotesize
\fittable{%
\begin{tabular}{@{}lccccccccccccccccccc|c@{}}
Dataset & \rotatebox{\tableheaderangle}{CNN-HAR\hspace{4pt}} & \rotatebox{\tableheaderangle}{Triple-Cross-Attn\hspace{4pt}} & \rotatebox{\tableheaderangle}{TinyHAR\hspace{4pt}} & \rotatebox{\tableheaderangle}{TinierHAR\hspace{4pt}} & \rotatebox{\tableheaderangle}{DANA\hspace{4pt}} & \rotatebox{\tableheaderangle}{LITEWAY-Light\hspace{4pt}} & \rotatebox{\tableheaderangle}{Attend-and-Discriminate\hspace{4pt}} & \rotatebox{\tableheaderangle}{DynamicWHAR\hspace{4pt}} & \rotatebox{\tableheaderangle}{DeepConvLSTM-Attn\hspace{4pt}} & \rotatebox{\tableheaderangle}{Random Forest\hspace{4pt}} & \rotatebox{\tableheaderangle}{MLP-HAR\hspace{4pt}} & \rotatebox{\tableheaderangle}{LITEWAY\hspace{4pt}} & \rotatebox{\tableheaderangle}{DeepConvLSTM\hspace{4pt}} & \rotatebox{\tableheaderangle}{MLP-Mixer\hspace{4pt}} & \rotatebox{\tableheaderangle}{Guan-LSTM\hspace{4pt}} & \rotatebox{\tableheaderangle}{SA-HAR\hspace{4pt}} & \rotatebox{\tableheaderangle}{DeepConvShallowLSTM\hspace{4pt}} & \rotatebox{\tableheaderangle}{SVM\hspace{4pt}} & \rotatebox{\tableheaderangle}{k-NN\hspace{4pt}} & \rotatebox{\tableheaderangle}{\textbf{Mean}\hspace{4pt}} \\
\midrule
SAD & \nofit{R} & \nofit{R} & 3732 & 766 & \nofit{F} & 498 & \nofit{F} & \nofit{F} & \nofit{F} & \textbf{10.4} & \nofit{F} & 1043 & \nofit{F} & \nofit{F} & \nofit{F} & \nofit{R} & \nofit{F} & \nofit{F} & \nofit{F} & 1210\,{\scriptsize$\pm$1460} \\
UCI-HAR & 2230 & 3288 & 726 & 217 & 5601 & 87.0 & \nofit{F} & \nofit{F} & \nofit{F} & \textbf{2.00} & 301 & 183 & \nofit{F} & 5086 & \nofit{F} & 7383 & 9453 & \nofit{F} & \nofit{F} & 2880\,{\scriptsize$\pm$3276} \\
MotionSense & 4115 & 6354 & 1257 & 318 & \nofit{F} & 160 & \nofit{F} & \nofit{F} & \nofit{F} & \textbf{3.47} & 712 & 334 & \nofit{F} & 12559 & \nofit{F} & 8386 & 18034 & \nofit{F} & \nofit{F} & 4748\,{\scriptsize$\pm$6041} \\
MHEALTH & \nofit{R} & 8100 & 1593 & 377 & \nofit{F} & 200 & \nofit{F} & \nofit{F} & \nofit{F} & \textbf{4.51} & 950 & 420 & \nofit{F} & 14916 & \nofit{F} & 9026 & \nofit{F} & \nofit{F} & \nofit{F} & 3954\,{\scriptsize$\pm$5393} \\
WEAR & 2860 & 4318 & 903 & 251 & 7332 & \textbf{111} & \nofit{F} & \nofit{F} & \nofit{F} & \nofit{F} & 426 & 234 & \nofit{F} & 7609 & \nofit{F} & 7716 & 12318 & \nofit{F} & \nofit{F} & 4007\,{\scriptsize$\pm$4174} \\
DSADS & 4198 & 7305 & 1217 & 300 & \nofit{F} & 194 & \nofit{F} & \nofit{F} & \nofit{F} & \textbf{4.45} & \nofit{F} & 385 & \nofit{F} & 14328 & \nofit{F} & 5346 & \nofit{F} & \nofit{F} & \nofit{F} & 3697\,{\scriptsize$\pm$4790} \\
Opportunity (Loc.) & \nofit{F} & \nofit{R} & 4445 & 947 & \nofit{F} & 652 & \nofit{F} & \nofit{F} & \nofit{F} & \textbf{12.7} & \nofit{F} & 1363 & \nofit{F} & \nofit{F} & \nofit{F} & \nofit{R} & \nofit{F} & \nofit{F} & \nofit{F} & 1484\,{\scriptsize$\pm$1727} \\
WISDM & 392 & 584 & 149 & 72.5 & 1308 & 20.3 & 2378 & 253 & 2743 & \textbf{0.95} & 41.5 & 41.7 & 2554 & 1416 & \nofit{F} & 3027 & 909 & \nofit{F} & \nofit{F} & 993\,{\scriptsize$\pm$1102} \\
REALDISP & \nofit{F} & \nofit{F} & \nofit{R} & 1395 & \nofit{F} & \textbf{957} & \nofit{F} & \nofit{F} & \nofit{F} & \nofit{F} & \nofit{F} & 2002 & \nofit{F} & \nofit{F} & \nofit{F} & \nofit{R} & \nofit{F} & \nofit{F} & \nofit{F} & 1451\,{\scriptsize$\pm$525} \\
PAMAP2 & \nofit{R} & \nofit{F} & \nofit{R} & 1383 & \nofit{F} & 865 & \nofit{F} & \nofit{F} & \nofit{F} & \textbf{16.8} & \nofit{F} & 1814 & \nofit{F} & \nofit{F} & \nofit{F} & \nofit{R} & \nofit{F} & \nofit{F} & \nofit{F} & 1020\,{\scriptsize$\pm$773} \\
HAPT & 1603 & 2264 & 549 & 183 & 3873 & 62.7 & \nofit{F} & \nofit{F} & \nofit{F} & \textbf{1.80} & 190 & 133 & \nofit{F} & 4994 & \nofit{F} & 7050 & 6596 & \nofit{F} & \nofit{F} & 2292\,{\scriptsize$\pm$2667} \\
HHAR & 663 & 1056 & 224 & 88.4 & 2196 & 32.7 & 3088 & \nofit{F} & 3530 & \textbf{1.21} & 85.5 & 66.0 & 3336 & 2668 & \nofit{F} & 3191 & 1671 & \nofit{F} & \nofit{F} & 1460\,{\scriptsize$\pm$1407} \\
Hang-Time & 971 & 1241 & 370 & 148 & 2146 & 38.5 & \nofit{F} & 288 & \nofit{F} & \textbf{1.15} & 90.7 & 83.3 & \nofit{F} & 2572 & \nofit{F} & 6718 & 3735 & \nofit{F} & \nofit{F} & 1416\,{\scriptsize$\pm$1979} \\
w-HAR & \nofit{R} & \nofit{R} & \nofit{F} & \nofit{R} & \nofit{R} & 348 & \nofit{F} & \nofit{F} & \nofit{F} & \textbf{6.34} & \nofit{F} & 744 & \nofit{F} & 24593 & \nofit{F} & \nofit{R} & \nofit{R} & 47.8 & \nofit{F} & 5148\,{\scriptsize$\pm$10874} \\
HuGaDB & \nofit{R} & \nofit{R} & 3001 & 645 & \nofit{F} & 384 & \nofit{F} & \nofit{F} & \nofit{F} & \textbf{7.89} & \nofit{F} & 806 & \nofit{F} & \nofit{F} & \nofit{F} & \nofit{R} & \nofit{F} & \nofit{F} & \nofit{F} & 969\,{\scriptsize$\pm$1175} \\
HARTH & 1603 & 2264 & 549 & 183 & 3873 & 62.7 & \nofit{F} & \nofit{F} & \nofit{F} & \textbf{1.78} & 190 & 133 & \nofit{F} & 4994 & \nofit{F} & 7050 & 6596 & \nofit{F} & \nofit{F} & 2292\,{\scriptsize$\pm$2667} \\
Daphnet & 2959 & 4203 & 988 & 279 & 7265 & 111 & \nofit{F} & \nofit{F} & \nofit{F} & \textbf{2.29} & 398 & 234 & \nofit{F} & 6158 & \nofit{F} & 10112 & 14160 & 33.3 & \nofit{F} & 3608\,{\scriptsize$\pm$4582} \\
USC-HAD & 3515 & 4469 & 1213 & 365 & 7228 & 124 & \nofit{F} & \nofit{F} & \nofit{F} & \textbf{2.85} & 474 & 264 & \nofit{F} & 9377 & \nofit{F} & \nofit{R} & 16443 & \nofit{F} & \nofit{F} & 3952\,{\scriptsize$\pm$5225} \\
HAR70+ & 1602 & 2261 & 549 & 183 & 3873 & 62.6 & \nofit{F} & \nofit{F} & \nofit{F} & \textbf{1.60} & 190 & 133 & \nofit{F} & 4994 & \nofit{F} & 7050 & 6596 & \nofit{F} & \nofit{F} & 2291\,{\scriptsize$\pm$2667} \\
GOTOV & \nofit{R} & \nofit{R} & 3494 & 729 & \nofit{F} & 411 & \nofit{F} & \nofit{F} & \nofit{F} & \textbf{7.20} & 2457 & 849 & \nofit{F} & \nofit{F} & \nofit{F} & \nofit{R} & \nofit{F} & \nofit{F} & \nofit{F} & 1324\,{\scriptsize$\pm$1352} \\
UTD-MHAD & 1605 & 2271 & 549 & 183 & 3874 & 62.8 & \nofit{F} & \nofit{F} & \nofit{F} & \textbf{2.19} & 190 & 133 & \nofit{F} & 4994 & \nofit{F} & 7051 & 6604 & 30.3 & 13.5 & 1969\,{\scriptsize$\pm$2589} \\
WISDM-19 (Watch) & 666 & 1058 & 225 & 89.1 & 2196 & \textbf{32.8} & 3113 & \nofit{F} & 3506 & \nofit{F} & 85.6 & 66.0 & 3295 & 2668 & \nofit{F} & 3192 & 1672 & \nofit{F} & \nofit{F} & 1562\,{\scriptsize$\pm$1394} \\
UCA-EHAR & 778 & 1251 & 256 & 101 & 2571 & 37.4 & 3512 & \nofit{F} & 3988 & \textbf{1.38} & 102 & 79.3 & 3790 & 2679 & \nofit{F} & 3386 & 2151 & \nofit{F} & \nofit{F} & 1645\,{\scriptsize$\pm$1561} \\
UP-Fall & 3925 & 6807 & 1140 & 286 & \nofit{F} & 182 & \nofit{F} & \nofit{F} & \nofit{F} & \textbf{4.11} & \nofit{F} & 360 & \nofit{F} & 14707 & \nofit{F} & 5185 & \nofit{F} & \nofit{F} & \nofit{F} & 3622\,{\scriptsize$\pm$4853} \\
UMAFall & \nofit{R} & \nofit{F} & \nofit{F} & \nofit{R} & \nofit{F} & 1310 & \nofit{F} & \nofit{F} & \nofit{F} & \textbf{24.9} & \nofit{F} & 2751 & \nofit{F} & \nofit{F} & \nofit{F} & \nofit{R} & \nofit{F} & \nofit{F} & \nofit{F} & 1362\,{\scriptsize$\pm$1364} \\
RealLifeHAR & 1478 & 2500 & 453 & 138 & 5019 & 69.9 & \nofit{F} & \nofit{F} & \nofit{F} & \textbf{1.87} & 247 & 140 & \nofit{F} & 5007 & \nofit{F} & 3744 & 4174 & \nofit{F} & \nofit{F} & 1914\,{\scriptsize$\pm$2055} \\
UniMiB-SHAR & 976 & 1244 & 372 & 150 & 2146 & 38.5 & \nofit{F} & 288 & \nofit{F} & \nofit{F} & 90.8 & 83.3 & \nofit{F} & 2573 & \nofit{F} & 6719 & 3740 & \nofit{F} & \textbf{38.1} & 1420\,{\scriptsize$\pm$1977} \\
AICOS-HAR & 1026 & 1694 & 326 & 111 & 3381 & 49.3 & 4038 & \nofit{F} & \nofit{F} & \textbf{1.52} & 154 & 98.4 & 4353 & 4009 & \nofit{F} & 3412 & 2689 & \nofit{F} & \nofit{F} & 1810\,{\scriptsize$\pm$1751} \\
WISDM-19 (Phone) & 666 & 1058 & 225 & 89.1 & 2196 & \textbf{32.8} & 3113 & \nofit{F} & 3506 & \nofit{F} & 85.6 & 66.0 & 3295 & 2668 & \nofit{F} & 3192 & 1672 & \nofit{F} & \nofit{F} & 1562\,{\scriptsize$\pm$1394} \\
\midrule
\textbf{Mean} & \fcell{1892}{1260} & \fcell{3123}{2301} & \fcell{1140}{1210} & \fcell{370}{371} & \fcell{3887}{1968} & \fcell{248}{323} & \fcell{3207}{548} & \fcell{276}{20.4} & \fcell{3455}{448} & \fcell{5.22}{5.83} & \fcell{373}{543} & \fcell{519}{677} & \fcell{3437}{599} & \fcell{7071}{5777} & \unmeasured & \fcell{5897}{2225} & \fcell{6623}{5359} & \fcell{37.1}{9.36} & \fcell{25.8}{17.4} & \fcell{2310}{2389} \\
\bottomrule
\end{tabular}}
}

\end{appendixtable}

\begin{appendixtable}[90]
    \centering
    \caption{Latency (ms) to classify one window on the MCU with 3\,s windows. Bold marks the lowest value per dataset. F: does not fit in flash, R: does not fit in RAM.}
    \label{tab:deploy-mcu-latency-w3s}
    \providecommand{\nofit}[1]{\textcolor{gray}{#1}}
\providecommand{\faileddeployment}{\textcolor{gray}{Failed}}
\providecommand{\todomissing}{\textcolor{red}{TODO}}
\providecommand{\unmeasured}{\textcolor{gray}{--}}
\providecommand{\fcell}[2]{\begin{tabular}[c]{@{}c@{}}#1\\[-2pt]{\scriptsize$\pm$#2}\end{tabular}}
\providecommand{\tableheaderangle}{90}
\providecommand{\tablemaxheight}{0.9\textheight}
\providecommand{\fittable}[1]{\sbox2{#1}%
  \sbox2{\resizebox{\linewidth}{!}{\usebox2}}%
  \dimen2=\ht2\advance\dimen2 by\dp2
  \ifdim\dimen2>\tablemaxheight\resizebox*{!}{\tablemaxheight}{\usebox2}\else\usebox2\fi}

{%
\setlength{\tabcolsep}{2pt}
\setlength{\aboverulesep}{0pt}\setlength{\belowrulesep}{0pt}
\renewcommand{\arraystretch}{1.15}
\footnotesize
\fittable{%
\begin{tabular}{@{}lccccccccccccccccccc|c@{}}
Dataset & \rotatebox{\tableheaderangle}{CNN-HAR\hspace{4pt}} & \rotatebox{\tableheaderangle}{Triple-Cross-Attn\hspace{4pt}} & \rotatebox{\tableheaderangle}{TinyHAR\hspace{4pt}} & \rotatebox{\tableheaderangle}{TinierHAR\hspace{4pt}} & \rotatebox{\tableheaderangle}{DANA\hspace{4pt}} & \rotatebox{\tableheaderangle}{LITEWAY-Light\hspace{4pt}} & \rotatebox{\tableheaderangle}{Attend-and-Discriminate\hspace{4pt}} & \rotatebox{\tableheaderangle}{DynamicWHAR\hspace{4pt}} & \rotatebox{\tableheaderangle}{DeepConvLSTM-Attn\hspace{4pt}} & \rotatebox{\tableheaderangle}{Random Forest\hspace{4pt}} & \rotatebox{\tableheaderangle}{MLP-HAR\hspace{4pt}} & \rotatebox{\tableheaderangle}{LITEWAY\hspace{4pt}} & \rotatebox{\tableheaderangle}{DeepConvLSTM\hspace{4pt}} & \rotatebox{\tableheaderangle}{MLP-Mixer\hspace{4pt}} & \rotatebox{\tableheaderangle}{Guan-LSTM\hspace{4pt}} & \rotatebox{\tableheaderangle}{SA-HAR\hspace{4pt}} & \rotatebox{\tableheaderangle}{DeepConvShallowLSTM\hspace{4pt}} & \rotatebox{\tableheaderangle}{SVM\hspace{4pt}} & \rotatebox{\tableheaderangle}{k-NN\hspace{4pt}} & \rotatebox{\tableheaderangle}{\textbf{Mean}\hspace{4pt}} \\
\midrule
SAD & \nofit{R} & \nofit{R} & \nofit{R} & 1151 & \nofit{F} & 752 & \nofit{F} & \nofit{F} & \nofit{F} & \textbf{14.3} & \nofit{F} & 1548 & \nofit{F} & \nofit{F} & \nofit{F} & \nofit{R} & \nofit{F} & \nofit{F} & \nofit{F} & 867\,{\scriptsize$\pm$655} \\
UCI-HAR & 3559 & 5070 & 1166 & 320 & 8412 & 130 & \nofit{F} & \nofit{F} & \nofit{F} & \textbf{2.68} & 487 & 273 & \nofit{F} & 7273 & \nofit{F} & 12330 & 17433 & 83.7 & \nofit{F} & 4349\,{\scriptsize$\pm$5582} \\
MotionSense & \nofit{R} & \nofit{R} & 2007 & 470 & \nofit{F} & 240 & \nofit{F} & \nofit{F} & \nofit{F} & \textbf{4.67} & 1127 & 498 & \nofit{F} & 18307 & \nofit{F} & \nofit{R} & 33567 & \nofit{F} & \nofit{F} & 7027\,{\scriptsize$\pm$12379} \\
MHEALTH & \nofit{R} & \nofit{R} & 2543 & 559 & \nofit{F} & 301 & \nofit{F} & \nofit{F} & \nofit{F} & \textbf{5.90} & 1552 & 623 & \nofit{F} & 22037 & \nofit{F} & \nofit{R} & \nofit{F} & \nofit{F} & \nofit{F} & 3946\,{\scriptsize$\pm$8024} \\
WEAR & 4565 & 6652 & 1450 & 371 & 10686 & \textbf{167} & \nofit{F} & \nofit{F} & \nofit{F} & \nofit{F} & 709 & 346 & \nofit{F} & 11424 & \nofit{F} & 12633 & 21439 & \nofit{F} & \nofit{F} & 6404\,{\scriptsize$\pm$6934} \\
DSADS & \nofit{R} & \nofit{R} & 2052 & 448 & \nofit{F} & 286 & \nofit{F} & \nofit{F} & \nofit{F} & \textbf{5.59} & \nofit{F} & 575 & \nofit{F} & 21204 & \nofit{F} & \nofit{R} & \nofit{F} & \nofit{F} & \nofit{F} & 4095\,{\scriptsize$\pm$8412} \\
Opportunity (Loc.) & \nofit{F} & \nofit{R} & \nofit{R} & 1397 & \nofit{F} & 991 & \nofit{F} & \nofit{F} & \nofit{F} & \textbf{18.4} & \nofit{F} & 2011 & \nofit{F} & \nofit{F} & \nofit{F} & \nofit{R} & \nofit{F} & \nofit{F} & \nofit{F} & 1105\,{\scriptsize$\pm$837} \\
WISDM & 498 & 665 & 188 & 90.1 & 1475 & 23.6 & 3058 & 259 & 3522 & \textbf{0.99} & 48.8 & 51.0 & 3285 & 1625 & \nofit{F} & 3706 & 1474 & \nofit{F} & \nofit{F} & 1248\,{\scriptsize$\pm$1394} \\
REALDISP & \nofit{F} & \nofit{F} & \nofit{R} & 2071 & \nofit{F} & \textbf{1448} & \nofit{F} & \nofit{F} & \nofit{F} & \nofit{F} & \nofit{F} & 2975 & \nofit{F} & \nofit{F} & \nofit{F} & \nofit{R} & \nofit{F} & \nofit{F} & \nofit{F} & 2165\,{\scriptsize$\pm$768} \\
PAMAP2 & \nofit{R} & \nofit{F} & \nofit{R} & \nofit{R} & \nofit{F} & 1297 & \nofit{F} & \nofit{F} & \nofit{F} & \textbf{25.2} & \nofit{F} & 2720 & \nofit{F} & \nofit{F} & \nofit{F} & \nofit{R} & \nofit{F} & \nofit{F} & \nofit{F} & 1347\,{\scriptsize$\pm$1348} \\
HAPT & 2578 & 3486 & 889 & 271 & 5550 & 94.0 & \nofit{F} & \nofit{F} & \nofit{F} & \textbf{2.20} & 319 & 197 & \nofit{F} & 7172 & \nofit{F} & 11633 & 11519 & 59.7 & 47.3 & 3130\,{\scriptsize$\pm$4233} \\
HHAR & 2569 & 3481 & 886 & 269 & 5550 & 94.0 & \nofit{F} & \nofit{F} & \nofit{F} & \textbf{2.10} & 319 & 197 & \nofit{F} & 7172 & \nofit{F} & 11632 & 11515 & \nofit{F} & \nofit{F} & 3640\,{\scriptsize$\pm$4378} \\
Hang-Time & 1579 & 1909 & 606 & 219 & 2985 & 57.4 & \nofit{F} & 325 & \nofit{F} & \textbf{1.33} & 153 & 122 & \nofit{F} & 3801 & \nofit{F} & 11134 & 6561 & \nofit{F} & \nofit{F} & 2266\,{\scriptsize$\pm$3296} \\
w-HAR & \nofit{R} & \nofit{R} & \nofit{F} & \nofit{F} & \nofit{R} & 522 & \nofit{F} & \nofit{F} & \faileddeployment & \textbf{8.39} & \nofit{F} & 1112 & \nofit{F} & \nofit{F} & \nofit{F} & \nofit{R} & \nofit{R} & 26.4 & 50.3 & 344\,{\scriptsize$\pm$480} \\
HuGaDB & \nofit{R} & \nofit{F} & 4705 & 971 & \nofit{F} & 575 & \nofit{F} & \nofit{F} & \nofit{F} & \textbf{11.0} & \nofit{F} & 1208 & \nofit{F} & \nofit{F} & \nofit{F} & \nofit{R} & \nofit{F} & \nofit{F} & \nofit{F} & 1494\,{\scriptsize$\pm$1851} \\
HARTH & 2578 & 3486 & 889 & 271 & 5550 & 94.0 & \nofit{F} & \nofit{F} & \nofit{F} & \textbf{2.19} & 319 & 197 & \nofit{F} & 7172 & \nofit{F} & 11633 & 11519 & \nofit{F} & \nofit{F} & 3642\,{\scriptsize$\pm$4378} \\
Daphnet & 4638 & 6261 & 1520 & 417 & 10608 & 166 & \nofit{F} & \nofit{F} & \nofit{F} & \textbf{3.05} & 703 & 350 & \nofit{F} & 9178 & \nofit{F} & \nofit{R} & 23674 & 21.2 & \nofit{F} & 4795\,{\scriptsize$\pm$7040} \\
USC-HAD & 5426 & 6868 & 1878 & \nofit{R} & 10582 & 186 & \nofit{F} & \nofit{F} & \nofit{F} & \textbf{3.77} & 896 & 395 & \nofit{F} & 14696 & \nofit{F} & \nofit{R} & 26291 & \nofit{F} & \nofit{F} & 6722\,{\scriptsize$\pm$8488} \\
HAR70+ & 2577 & 3482 & 889 & 271 & 5551 & 94.0 & \nofit{F} & \nofit{F} & \nofit{F} & \textbf{1.99} & 319 & 197 & \nofit{F} & 7171 & \nofit{F} & 11632 & 11521 & 73.1 & \nofit{F} & 3368\,{\scriptsize$\pm$4307} \\
GOTOV & \nofit{R} & \nofit{F} & \nofit{R} & 1107 & \nofit{F} & 619 & \nofit{F} & \nofit{F} & \nofit{F} & \textbf{10.4} & \nofit{F} & 1288 & \faileddeployment & \nofit{F} & \nofit{F} & \nofit{R} & \nofit{F} & \nofit{F} & \nofit{F} & 756\,{\scriptsize$\pm$572} \\
UTD-MHAD & 2579 & 3498 & 889 & 271 & 5551 & 94.2 & \nofit{F} & \nofit{F} & \nofit{F} & \textbf{2.57} & 319 & 197 & \nofit{F} & 7171 & \nofit{F} & 11634 & 11535 & 18.2 & 8.41 & 3126\,{\scriptsize$\pm$4240} \\
WISDM-19 (Watch) & 839 & 1206 & 283 & 110 & 2532 & \textbf{38.1} & 3987 & \nofit{F} & 4485 & \nofit{F} & 101 & 81.1 & 4221 & 3053 & \nofit{F} & 3904 & 2658 & \nofit{F} & \nofit{F} & 1964\,{\scriptsize$\pm$1757} \\
UCA-EHAR & 1332 & 1987 & 464 & 150 & 3714 & 56.2 & \nofit{F} & \nofit{F} & \nofit{F} & \textbf{1.66} & 156 & 116 & \nofit{F} & 3820 & \nofit{F} & 5470 & 5363 & \nofit{F} & \nofit{F} & 1886\,{\scriptsize$\pm$2140} \\
UP-Fall & \nofit{R} & \nofit{R} & 1424 & 347 & \nofit{F} & 212 & \nofit{F} & \nofit{F} & \nofit{F} & \textbf{4.71} & \nofit{F} & 445 & \nofit{F} & 16708 & \nofit{F} & 6305 & \nofit{F} & \nofit{F} & \nofit{F} & 3635\,{\scriptsize$\pm$6175} \\
UMAFall & \nofit{R} & \nofit{F} & \nofit{F} & \nofit{F} & \nofit{F} & 1966 & \nofit{F} & \nofit{F} & \nofit{F} & \textbf{35.5} & \nofit{F} & 4126 & \nofit{F} & \nofit{F} & \nofit{F} & \nofit{R} & \nofit{F} & \nofit{F} & \nofit{F} & 2042\,{\scriptsize$\pm$2046} \\
RealLifeHAR & 1858 & 2837 & 569 & 171 & 5704 & 81.7 & \nofit{F} & \nofit{F} & \nofit{F} & \textbf{2.07} & 302 & 171 & \nofit{F} & 5796 & \nofit{F} & 4502 & 6202 & \nofit{F} & \nofit{F} & 2350\,{\scriptsize$\pm$2533} \\
UniMiB-SHAR & 1586 & 1914 & 609 & 221 & 2985 & 57.5 & \nofit{F} & 325 & \nofit{F} & \textbf{1.63} & 153 & 122 & \nofit{F} & 3802 & \nofit{F} & 11135 & 6570 & 60.2 & 23.3 & 1971\,{\scriptsize$\pm$3151} \\
AICOS-HAR & 3817 & 5595 & 1289 & 341 & 9299 & 143 & \nofit{F} & \nofit{F} & \nofit{F} & \textbf{2.99} & 550 & 298 & \nofit{F} & 11554 & \nofit{F} & 12486 & 19193 & \nofit{F} & \nofit{F} & 5381\,{\scriptsize$\pm$6360} \\
WISDM-19 (Phone) & 839 & 1206 & 283 & 110 & 2532 & \textbf{38.1} & 3987 & \nofit{F} & 4485 & \nofit{F} & 101 & 81.1 & 4221 & 3053 & \nofit{F} & 3904 & 2658 & \nofit{F} & \nofit{F} & 1964\,{\scriptsize$\pm$1757} \\
\midrule
\textbf{Mean} & \fcell{2554}{1441} & \fcell{3506}{1970} & \fcell{1249}{998} & \fcell{496}{480} & \fcell{5839}{3057} & \fcell{373}{486} & \fcell{3677}{536} & \fcell{303}{37.8} & \fcell{4164}{556} & \fcell{7.01}{8.41} & \fcell{454}{394} & \fcell{776}{1010} & \fcell{3909}{540} & \fcell{9200}{6103} & \unmeasured & \fcell{9105}{3651} & \fcell{12816}{9033} & \fcell{48.9}{26.6} & \fcell{32.3}{20.0} & \fcell{3251}{3785} \\
\bottomrule
\end{tabular}}
}

\end{appendixtable}

\begin{appendixtable}[90]
    \centering
    \caption{Peak RAM (KB) during inference on the MCU with 2\,s windows. Bold marks the lowest value per dataset. F: does not fit in flash, R: does not fit in RAM.}
    \label{tab:deploy-mcu-ram-w2s}
    \providecommand{\nofit}[1]{\textcolor{gray}{#1}}
\providecommand{\faileddeployment}{\textcolor{gray}{Failed}}
\providecommand{\todomissing}{\textcolor{red}{TODO}}
\providecommand{\unmeasured}{\textcolor{gray}{--}}
\providecommand{\fcell}[2]{\begin{tabular}[c]{@{}c@{}}#1\\[-2pt]{\scriptsize$\pm$#2}\end{tabular}}
\providecommand{\tableheaderangle}{90}
\providecommand{\tablemaxheight}{0.9\textheight}
\providecommand{\fittable}[1]{\sbox2{#1}%
  \sbox2{\resizebox{\linewidth}{!}{\usebox2}}%
  \dimen2=\ht2\advance\dimen2 by\dp2
  \ifdim\dimen2>\tablemaxheight\resizebox*{!}{\tablemaxheight}{\usebox2}\else\usebox2\fi}

{%
\setlength{\tabcolsep}{2pt}
\setlength{\aboverulesep}{0pt}\setlength{\belowrulesep}{0pt}
\renewcommand{\arraystretch}{1.15}
\footnotesize
\fittable{%
\begin{tabular}{@{}lccccccccccccccccccc|c@{}}
Dataset & \rotatebox{\tableheaderangle}{CNN-HAR\hspace{4pt}} & \rotatebox{\tableheaderangle}{Triple-Cross-Attn\hspace{4pt}} & \rotatebox{\tableheaderangle}{TinyHAR\hspace{4pt}} & \rotatebox{\tableheaderangle}{TinierHAR\hspace{4pt}} & \rotatebox{\tableheaderangle}{DANA\hspace{4pt}} & \rotatebox{\tableheaderangle}{LITEWAY-Light\hspace{4pt}} & \rotatebox{\tableheaderangle}{Attend-and-Discriminate\hspace{4pt}} & \rotatebox{\tableheaderangle}{DynamicWHAR\hspace{4pt}} & \rotatebox{\tableheaderangle}{DeepConvLSTM-Attn\hspace{4pt}} & \rotatebox{\tableheaderangle}{Random Forest\hspace{4pt}} & \rotatebox{\tableheaderangle}{MLP-HAR\hspace{4pt}} & \rotatebox{\tableheaderangle}{LITEWAY\hspace{4pt}} & \rotatebox{\tableheaderangle}{DeepConvLSTM\hspace{4pt}} & \rotatebox{\tableheaderangle}{MLP-Mixer\hspace{4pt}} & \rotatebox{\tableheaderangle}{Guan-LSTM\hspace{4pt}} & \rotatebox{\tableheaderangle}{SA-HAR\hspace{4pt}} & \rotatebox{\tableheaderangle}{DeepConvShallowLSTM\hspace{4pt}} & \rotatebox{\tableheaderangle}{SVM\hspace{4pt}} & \rotatebox{\tableheaderangle}{k-NN\hspace{4pt}} & \rotatebox{\tableheaderangle}{\textbf{Mean}\hspace{4pt}} \\
\midrule
SAD & \nofit{R} & \nofit{R} & 319 & 166 & \nofit{F} & 123 & \nofit{F} & \nofit{F} & \nofit{F} & \textbf{31.5} & \nofit{F} & 78.6 & \nofit{F} & \nofit{F} & \nofit{F} & \nofit{R} & \nofit{F} & \nofit{F} & \nofit{F} & 143\,{\scriptsize$\pm$110} \\
UCI-HAR & 205 & 191 & 108 & 117 & 229 & 29.8 & \nofit{F} & \nofit{F} & \nofit{F} & \textbf{7.58} & 44.8 & 25.6 & \nofit{F} & 107 & \nofit{F} & 207 & 116 & \nofit{F} & \nofit{F} & 116\,{\scriptsize$\pm$78.2} \\
MotionSense & 391 & 358 & 146 & 125 & \nofit{F} & 46.2 & \nofit{F} & \nofit{F} & \nofit{F} & \textbf{11.8} & 62.3 & 35.0 & \nofit{F} & 155 & \nofit{F} & 322 & 217 & \nofit{F} & \nofit{F} & 170\,{\scriptsize$\pm$135} \\
MHEALTH & \nofit{R} & 451 & 166 & 129 & \nofit{F} & 55.3 & \nofit{F} & \nofit{F} & \nofit{F} & \textbf{14.2} & 72.1 & 40.2 & \nofit{F} & 172 & \nofit{F} & 385 & \nofit{F} & \nofit{F} & \nofit{F} & 165\,{\scriptsize$\pm$154} \\
WEAR & 267 & 247 & 121 & 119 & 209 & 35.4 & \nofit{F} & \nofit{F} & \nofit{F} & \nofit{F} & 50.7 & \textbf{28.8} & \nofit{F} & 120 & \nofit{F} & 246 & 150 & \nofit{F} & \nofit{F} & 145\,{\scriptsize$\pm$87.2} \\
DSADS & 454 & 429 & 114 & 83.7 & \nofit{F} & 57.2 & \nofit{F} & \nofit{F} & \nofit{F} & \textbf{15.5} & \nofit{F} & 41.2 & \nofit{F} & 167 & \nofit{F} & 354 & \nofit{F} & \nofit{F} & \nofit{F} & 191\,{\scriptsize$\pm$174} \\
Opportunity (Loc.) & \nofit{F} & \nofit{R} & 339 & 147 & \nofit{F} & 162 & \nofit{F} & \nofit{F} & \nofit{F} & \textbf{44.8} & \nofit{F} & 102 & \nofit{F} & \nofit{F} & \nofit{F} & \nofit{R} & \nofit{F} & \nofit{F} & \nofit{F} & 159\,{\scriptsize$\pm$110} \\
WISDM & 48.2 & 51.6 & 38.7 & 63.0 & 121 & 16.3 & 144 & 33.0 & 130 & \textbf{3.98} & 27.6 & 17.9 & 127 & 87.3 & \nofit{F} & 112 & 30.8 & \nofit{F} & \nofit{F} & 65.8\,{\scriptsize$\pm$47.0} \\
REALDISP & \nofit{F} & \nofit{F} & \nofit{R} & 221 & \nofit{F} & 227 & \nofit{F} & \nofit{F} & \nofit{F} & \nofit{F} & \nofit{F} & \textbf{138} & \nofit{F} & \nofit{F} & \nofit{F} & \nofit{R} & \nofit{F} & \nofit{F} & \nofit{F} & 195\,{\scriptsize$\pm$49.7} \\
PAMAP2 & \nofit{R} & \nofit{F} & \nofit{R} & 302 & \nofit{F} & 200 & \nofit{F} & \nofit{F} & \nofit{F} & \textbf{48.5} & \nofit{F} & 121 & \nofit{F} & \nofit{F} & \nofit{F} & \nofit{R} & \nofit{F} & \nofit{F} & \nofit{F} & 168\,{\scriptsize$\pm$109} \\
HAPT & 143 & 136 & 96.1 & 117 & 184 & 24.4 & \nofit{F} & \nofit{F} & \nofit{F} & \textbf{6.20} & 39.0 & 22.5 & \nofit{F} & 107 & \nofit{F} & 196 & 81.5 & \nofit{F} & \nofit{F} & 96.0\,{\scriptsize$\pm$63.4} \\
HHAR & 77.2 & 77.6 & 44.1 & 63.0 & 147 & 19.2 & 153 & \nofit{F} & 130 & \textbf{4.80} & 31.1 & 19.6 & 127 & 89.7 & \nofit{F} & 112 & 43.9 & \nofit{F} & \nofit{F} & 75.9\,{\scriptsize$\pm$49.0} \\
Hang-Time & 82.4 & 81.9 & 83.7 & 117 & 139 & 18.9 & \nofit{F} & 42.3 & \nofit{F} & \textbf{4.76} & 33.1 & 19.4 & \nofit{F} & 89.5 & \nofit{F} & 196 & 74.9 & \nofit{F} & \nofit{F} & 75.6\,{\scriptsize$\pm$54.0} \\
w-HAR & \nofit{R} & \nofit{R} & \nofit{F} & \nofit{R} & \nofit{R} & 75.4 & \nofit{F} & \nofit{F} & \nofit{F} & \textbf{19.2} & \nofit{F} & 50.9 & \nofit{F} & 197 & \nofit{F} & \nofit{R} & \nofit{R} & 25.4 & \nofit{F} & 73.5\,{\scriptsize$\pm$72.5} \\
HuGaDB & \nofit{R} & \nofit{R} & 280 & 170 & \nofit{F} & 96.0 & \nofit{F} & \nofit{F} & \nofit{F} & \textbf{24.2} & \nofit{F} & 63.2 & \nofit{F} & \nofit{F} & \nofit{F} & \nofit{R} & \nofit{F} & \nofit{F} & \nofit{F} & 127\,{\scriptsize$\pm$101} \\
HARTH & 143 & 136 & 96.1 & 117 & 184 & 24.4 & \nofit{F} & \nofit{F} & \nofit{F} & \textbf{6.20} & 39.0 & 22.5 & \nofit{F} & 107 & \nofit{F} & 196 & 81.5 & \nofit{F} & \nofit{F} & 96.0\,{\scriptsize$\pm$63.4} \\
Daphnet & 260 & 237 & 142 & 146 & 260 & 34.2 & \nofit{F} & \nofit{F} & \nofit{F} & \textbf{8.66} & 49.8 & 28.0 & \nofit{F} & 113 & \nofit{F} & 267 & 163 & 13.2 & \nofit{F} & 132\,{\scriptsize$\pm$100} \\
USC-HAD & 275 & 247 & 203 & 220 & 259 & 34.9 & \nofit{F} & \nofit{F} & \nofit{F} & \textbf{8.93} & 54.6 & 28.4 & \nofit{F} & 137 & \nofit{F} & \nofit{R} & 175 & \nofit{F} & \nofit{F} & 149\,{\scriptsize$\pm$101} \\
HAR70+ & 143 & 135 & 96.0 & 117 & 184 & 24.4 & \nofit{F} & \nofit{F} & \nofit{F} & \textbf{6.18} & 39.0 & 22.5 & \nofit{F} & 107 & \nofit{F} & 196 & 81.4 & \nofit{F} & \nofit{F} & 96.0\,{\scriptsize$\pm$63.4} \\
GOTOV & \nofit{R} & \nofit{R} & 338 & 227 & \nofit{F} & 99.8 & \nofit{F} & \nofit{F} & \nofit{F} & \textbf{24.6} & 119 & 73.8 & \nofit{F} & \nofit{F} & \nofit{F} & \nofit{R} & \nofit{F} & \nofit{F} & \nofit{F} & 147\,{\scriptsize$\pm$115} \\
UTD-MHAD & 143 & 136 & 96.2 & 117 & 184 & 24.5 & \nofit{F} & \nofit{F} & \nofit{F} & \textbf{6.25} & 39.1 & 22.6 & \nofit{F} & 107 & \nofit{F} & 197 & 81.6 & 10.1 & 6.29 & 83.6\,{\scriptsize$\pm$66.6} \\
WISDM-19 (Watch) & 77.3 & 77.7 & 44.2 & 63.1 & 147 & \textbf{19.3} & 153 & \nofit{F} & 130 & \nofit{F} & 31.2 & 19.7 & 127 & 89.8 & \nofit{F} & 112 & 44.0 & \nofit{F} & \nofit{F} & 81.1\,{\scriptsize$\pm$46.6} \\
UCA-EHAR & 90.0 & 90.6 & 46.5 & 67.1 & 157 & 20.3 & 166 & \nofit{F} & 142 & \textbf{5.17} & 32.3 & 20.3 & 139 & 89.9 & \nofit{F} & 116 & 50.1 & \nofit{F} & \nofit{F} & 82.2\,{\scriptsize$\pm$53.1} \\
UP-Fall & 425 & 402 & 108 & 82.0 & \nofit{F} & 54.2 & \nofit{F} & \nofit{F} & \nofit{F} & \textbf{14.7} & \nofit{F} & 39.4 & \nofit{F} & 167 & \nofit{F} & 335 & \nofit{F} & \nofit{F} & \nofit{F} & 181\,{\scriptsize$\pm$162} \\
UMAFall & \nofit{R} & \nofit{F} & \nofit{F} & \nofit{R} & \nofit{F} & 290 & \nofit{F} & \nofit{F} & \nofit{F} & \textbf{68.6} & \nofit{F} & 171 & \nofit{F} & \nofit{F} & \nofit{F} & \nofit{R} & \nofit{F} & \nofit{F} & \nofit{F} & 177\,{\scriptsize$\pm$111} \\
RealLifeHAR & 164 & 159 & 60.1 & 66.7 & 161 & 27.9 & \nofit{F} & \nofit{F} & \nofit{F} & \textbf{7.25} & 41.5 & 24.5 & \nofit{F} & 107 & \nofit{F} & 163 & 83.3 & \nofit{F} & \nofit{F} & 88.8\,{\scriptsize$\pm$60.0} \\
UniMiB-SHAR & 82.5 & 82.0 & 83.8 & 117 & 140 & 19.0 & \nofit{F} & 42.4 & \nofit{F} & \nofit{F} & 33.2 & 19.5 & \nofit{F} & 89.6 & \nofit{F} & 197 & 75.0 & \nofit{F} & \textbf{4.84} & 75.8\,{\scriptsize$\pm$54.1} \\
AICOS-HAR & 116 & 114 & 51.2 & 64.0 & 181 & 23.1 & 175 & \nofit{F} & \nofit{F} & \textbf{5.92} & 35.8 & 21.8 & 156 & 100 & \nofit{F} & 132 & 61.5 & \nofit{F} & \nofit{F} & 88.4\,{\scriptsize$\pm$58.9} \\
WISDM-19 (Phone) & 77.3 & 77.7 & 44.2 & 63.1 & 147 & \textbf{19.3} & 153 & \nofit{F} & 130 & \nofit{F} & 31.2 & 19.7 & 127 & 89.8 & \nofit{F} & 112 & 44.0 & \nofit{F} & \nofit{F} & 81.1\,{\scriptsize$\pm$46.6} \\
\midrule
\textbf{Mean} & \fcell{183}{123} & \fcell{186}{126} & \fcell{131}{93.7} & \fcell{126}{60.3} & \fcell{178}{41.1} & \fcell{65.5}{70.5} & \fcell{157}{11.2} & \fcell{39.2}{5.42} & \fcell{133}{5.29} & \fcell{16.6}{16.5} & \fcell{45.3}{20.9} & \fcell{46.2}{40.0} & \fcell{134}{12.1} & \fcell{118}{32.7} & \unmeasured & \fcell{208}{86.1} & \fcell{92.0}{52.1} & \fcell{16.2}{8.10} & \fcell{5.56}{1.02} & \fcell{105}{65.5} \\
\bottomrule
\end{tabular}}
}

\end{appendixtable}

\begin{appendixtable}[90]
    \centering
    \caption{Peak RAM (KB) during inference on the MCU with 3\,s windows. Bold marks the lowest value per dataset. F: does not fit in flash, R: does not fit in RAM.}
    \label{tab:deploy-mcu-ram-w3s}
    \providecommand{\nofit}[1]{\textcolor{gray}{#1}}
\providecommand{\faileddeployment}{\textcolor{gray}{Failed}}
\providecommand{\todomissing}{\textcolor{red}{TODO}}
\providecommand{\unmeasured}{\textcolor{gray}{--}}
\providecommand{\fcell}[2]{\begin{tabular}[c]{@{}c@{}}#1\\[-2pt]{\scriptsize$\pm$#2}\end{tabular}}
\providecommand{\tableheaderangle}{90}
\providecommand{\tablemaxheight}{0.9\textheight}
\providecommand{\fittable}[1]{\sbox2{#1}%
  \sbox2{\resizebox{\linewidth}{!}{\usebox2}}%
  \dimen2=\ht2\advance\dimen2 by\dp2
  \ifdim\dimen2>\tablemaxheight\resizebox*{!}{\tablemaxheight}{\usebox2}\else\usebox2\fi}

{%
\setlength{\tabcolsep}{2pt}
\setlength{\aboverulesep}{0pt}\setlength{\belowrulesep}{0pt}
\renewcommand{\arraystretch}{1.15}
\footnotesize
\fittable{%
\begin{tabular}{@{}lccccccccccccccccccc|c@{}}
Dataset & \rotatebox{\tableheaderangle}{CNN-HAR\hspace{4pt}} & \rotatebox{\tableheaderangle}{Triple-Cross-Attn\hspace{4pt}} & \rotatebox{\tableheaderangle}{TinyHAR\hspace{4pt}} & \rotatebox{\tableheaderangle}{TinierHAR\hspace{4pt}} & \rotatebox{\tableheaderangle}{DANA\hspace{4pt}} & \rotatebox{\tableheaderangle}{LITEWAY-Light\hspace{4pt}} & \rotatebox{\tableheaderangle}{Attend-and-Discriminate\hspace{4pt}} & \rotatebox{\tableheaderangle}{DynamicWHAR\hspace{4pt}} & \rotatebox{\tableheaderangle}{DeepConvLSTM-Attn\hspace{4pt}} & \rotatebox{\tableheaderangle}{Random Forest\hspace{4pt}} & \rotatebox{\tableheaderangle}{MLP-HAR\hspace{4pt}} & \rotatebox{\tableheaderangle}{LITEWAY\hspace{4pt}} & \rotatebox{\tableheaderangle}{DeepConvLSTM\hspace{4pt}} & \rotatebox{\tableheaderangle}{MLP-Mixer\hspace{4pt}} & \rotatebox{\tableheaderangle}{Guan-LSTM\hspace{4pt}} & \rotatebox{\tableheaderangle}{SA-HAR\hspace{4pt}} & \rotatebox{\tableheaderangle}{DeepConvShallowLSTM\hspace{4pt}} & \rotatebox{\tableheaderangle}{SVM\hspace{4pt}} & \rotatebox{\tableheaderangle}{k-NN\hspace{4pt}} & \rotatebox{\tableheaderangle}{\textbf{Mean}\hspace{4pt}} \\
\midrule
SAD & \nofit{R} & \nofit{R} & \nofit{R} & 256 & \nofit{F} & 177 & \nofit{F} & \nofit{F} & \nofit{F} & \textbf{43.4} & \nofit{F} & 108 & \nofit{F} & \nofit{F} & \nofit{F} & \nofit{R} & \nofit{F} & \nofit{F} & \nofit{F} & 146\,{\scriptsize$\pm$91.4} \\
UCI-HAR & 304 & 272 & 171 & 168 & 285 & 37.9 & \nofit{F} & \nofit{F} & \nofit{F} & \textbf{9.53} & 54.9 & 30.0 & \nofit{F} & 121 & \nofit{F} & 331 & 201 & 19.6 & \nofit{F} & 154\,{\scriptsize$\pm$118} \\
MotionSense & \nofit{R} & \nofit{R} & 229 & 183 & \nofit{F} & 62.5 & \nofit{F} & \nofit{F} & \nofit{F} & \textbf{15.5} & 79.5 & 43.8 & \nofit{F} & 159 & \nofit{F} & \nofit{R} & 386 & \nofit{F} & \nofit{F} & 145\,{\scriptsize$\pm$122} \\
MHEALTH & \nofit{R} & \nofit{R} & 261 & 192 & \nofit{F} & 76.2 & \nofit{F} & \nofit{F} & \nofit{F} & \textbf{18.9} & 93.2 & 51.5 & \nofit{F} & 183 & \nofit{F} & \nofit{R} & \nofit{F} & \nofit{F} & \nofit{F} & 125\,{\scriptsize$\pm$88.0} \\
WEAR & 399 & 355 & 190 & 173 & 284 & 46.2 & \nofit{F} & \nofit{F} & \nofit{F} & \nofit{F} & 63.2 & \textbf{34.7} & \nofit{F} & 147 & \nofit{F} & 347 & 263 & \nofit{F} & \nofit{F} & 209\,{\scriptsize$\pm$130} \\
DSADS & \nofit{R} & \nofit{R} & 188 & 120 & \nofit{F} & 76.4 & \nofit{F} & \nofit{F} & \nofit{F} & \textbf{20.0} & \nofit{F} & 58.2 & \nofit{F} & 171 & \nofit{F} & \nofit{R} & \nofit{F} & \nofit{F} & \nofit{F} & 106\,{\scriptsize$\pm$65.8} \\
Opportunity (Loc.) & \nofit{F} & \nofit{R} & \nofit{R} & 233 & \nofit{F} & 236 & \nofit{F} & \nofit{F} & \nofit{F} & \textbf{60.5} & \nofit{F} & 141 & \nofit{F} & \nofit{F} & \nofit{F} & \nofit{R} & \nofit{F} & \nofit{F} & \nofit{F} & 168\,{\scriptsize$\pm$83.9} \\
WISDM & 54.8 & 56.9 & 46.4 & 75.4 & 124 & 16.8 & 183 & 34.8 & 165 & \textbf{4.14} & 28.5 & 18.2 & 161 & 87.5 & \nofit{F} & 126 & 34.9 & \nofit{F} & \nofit{F} & 76.1\,{\scriptsize$\pm$58.2} \\
REALDISP & \nofit{F} & \nofit{F} & \nofit{R} & 355 & \nofit{F} & 333 & \nofit{F} & \nofit{F} & \nofit{F} & \nofit{F} & \nofit{F} & \textbf{195} & \nofit{F} & \nofit{F} & \nofit{F} & \nofit{R} & \nofit{F} & \nofit{F} & \nofit{F} & 294\,{\scriptsize$\pm$86.7} \\
PAMAP2 & \nofit{R} & \nofit{F} & \nofit{R} & \nofit{R} & \nofit{F} & 291 & \nofit{F} & \nofit{F} & \nofit{F} & \textbf{69.2} & \nofit{F} & 172 & \nofit{F} & \nofit{F} & \nofit{F} & \nofit{R} & \nofit{F} & \nofit{F} & \nofit{F} & 177\,{\scriptsize$\pm$111} \\
HAPT & 209 & 190 & 151 & 166 & 222 & 29.8 & \nofit{F} & \nofit{F} & \nofit{F} & \textbf{7.56} & 46.7 & 25.5 & \nofit{F} & 121 & \nofit{F} & 332 & 125 & 16.7 & 7.59 & 118\,{\scriptsize$\pm$99.7} \\
HHAR & 209 & 190 & 151 & 166 & 222 & 29.7 & \nofit{F} & \nofit{F} & \nofit{F} & \textbf{7.54} & 46.7 & 25.4 & \nofit{F} & 121 & \nofit{F} & 331 & 125 & \nofit{F} & \nofit{F} & 135\,{\scriptsize$\pm$96.9} \\
Hang-Time & 128 & 110 & 132 & 166 & 158 & 21.5 & \nofit{F} & 51.7 & \nofit{F} & \textbf{5.54} & 38.5 & 20.9 & \nofit{F} & 99.6 & \nofit{F} & 331 & 125 & \nofit{F} & \nofit{F} & 107\,{\scriptsize$\pm$86.9} \\
w-HAR & \nofit{R} & \nofit{R} & \nofit{F} & \nofit{F} & \nofit{R} & 106 & \nofit{F} & \nofit{F} & \faileddeployment & \textbf{27.0} & \nofit{F} & 68.0 & \nofit{F} & \nofit{F} & \nofit{F} & \nofit{R} & \nofit{R} & 30.5 & 27.0 & 51.8\,{\scriptsize$\pm$35.1} \\
HuGaDB & \nofit{R} & \nofit{F} & 433 & 249 & \nofit{F} & 136 & \nofit{F} & \nofit{F} & \nofit{F} & \textbf{33.4} & \nofit{F} & 85.5 & \nofit{F} & \nofit{F} & \nofit{F} & \nofit{R} & \nofit{F} & \nofit{F} & \nofit{F} & 187\,{\scriptsize$\pm$159} \\
HARTH & 209 & 190 & 151 & 166 & 222 & 29.8 & \nofit{F} & \nofit{F} & \nofit{F} & \textbf{7.56} & 46.7 & 25.5 & \nofit{F} & 121 & \nofit{F} & 332 & 125 & \nofit{F} & \nofit{F} & 135\,{\scriptsize$\pm$96.9} \\
Daphnet & 387 & 343 & 220 & 212 & 332 & 44.3 & \nofit{F} & \nofit{F} & \nofit{F} & \textbf{11.2} & 63.4 & 33.7 & \nofit{F} & 135 & \nofit{F} & \nofit{R} & 271 & 13.8 & \nofit{F} & 172\,{\scriptsize$\pm$140} \\
USC-HAD & 407 & 358 & 310 & \nofit{R} & 334 & 45.5 & \nofit{F} & \nofit{F} & \nofit{F} & \textbf{11.7} & 71.4 & 34.2 & \nofit{F} & 172 & \nofit{F} & \nofit{R} & 275 & \nofit{F} & \nofit{F} & 202\,{\scriptsize$\pm$152} \\
HAR70+ & 209 & 190 & 151 & 166 & 222 & 29.7 & \nofit{F} & \nofit{F} & \nofit{F} & \textbf{7.54} & 46.7 & 25.4 & \nofit{F} & 121 & \nofit{F} & 331 & 125 & 19.7 & \nofit{F} & 126\,{\scriptsize$\pm$98.1} \\
GOTOV & \nofit{R} & \nofit{F} & \nofit{R} & 337 & \nofit{F} & 143 & \nofit{F} & \nofit{F} & \nofit{F} & \textbf{34.3} & \nofit{F} & 102 & \faileddeployment & \nofit{F} & \nofit{F} & \nofit{R} & \nofit{F} & \nofit{F} & \nofit{F} & 154\,{\scriptsize$\pm$130} \\
UTD-MHAD & 209 & 190 & 151 & 167 & 222 & 29.9 & \nofit{F} & \nofit{F} & \nofit{F} & \textbf{7.62} & 46.9 & 25.6 & \nofit{F} & 121 & \nofit{F} & 332 & 125 & 9.76 & 7.65 & 117\,{\scriptsize$\pm$100} \\
WISDM-19 (Watch) & 90.5 & 87.8 & 53.3 & 75.5 & 154 & \textbf{20.2} & 192 & \nofit{F} & 165 & \nofit{F} & 32.5 & \textbf{20.2} & 161 & 93.2 & \nofit{F} & 126 & 51.5 & \nofit{F} & \nofit{F} & 94.5\,{\scriptsize$\pm$57.0} \\
UCA-EHAR & 130 & 123 & 77.3 & 91.9 & 180 & 23.6 & \nofit{F} & \nofit{F} & \nofit{F} & \textbf{5.98} & 36.5 & 22.1 & \nofit{F} & 101 & \nofit{F} & 159 & 72.8 & \nofit{F} & \nofit{F} & 85.2\,{\scriptsize$\pm$56.2} \\
UP-Fall & \nofit{R} & \nofit{R} & 134 & 94.2 & \nofit{F} & 60.4 & \nofit{F} & \nofit{F} & \nofit{F} & \textbf{16.3} & \nofit{F} & 43.5 & \nofit{F} & 144 & \nofit{F} & 393 & \nofit{F} & \nofit{F} & \nofit{F} & 127\,{\scriptsize$\pm$126} \\
UMAFall & \nofit{R} & \nofit{F} & \nofit{F} & \nofit{F} & \nofit{F} & 427 & \nofit{F} & \nofit{F} & \nofit{F} & \textbf{99.8} & \nofit{F} & 247 & \nofit{F} & \nofit{F} & \nofit{F} & \nofit{R} & \nofit{F} & \nofit{F} & \nofit{F} & 258\,{\scriptsize$\pm$164} \\
RealLifeHAR & 197 & 184 & 73.4 & 78.3 & 179 & 30.2 & \nofit{F} & \nofit{F} & \nofit{F} & \textbf{7.88} & 44.0 & 26.0 & \nofit{F} & 111 & \nofit{F} & 187 & 102 & \nofit{F} & \nofit{F} & 102\,{\scriptsize$\pm$69.8} \\
UniMiB-SHAR & 128 & 110 & 132 & 167 & 158 & 21.6 & \nofit{F} & 51.8 & \nofit{F} & \textbf{5.59} & 38.6 & 21.0 & \nofit{F} & 99.7 & \nofit{F} & 332 & 125 & 18.8 & 5.62 & 94.3\,{\scriptsize$\pm$87.2} \\
AICOS-HAR & 336 & 300 & 177 & 169 & 306 & 40.7 & \nofit{F} & \nofit{F} & \nofit{F} & \textbf{10.2} & 57.7 & 31.6 & \nofit{F} & 147 & \nofit{F} & 332 & 221 & \nofit{F} & \nofit{F} & 177\,{\scriptsize$\pm$123} \\
WISDM-19 (Phone) & 90.5 & 87.8 & 53.3 & 75.5 & 154 & \textbf{20.2} & 192 & \nofit{F} & 165 & \nofit{F} & 32.5 & \textbf{20.2} & 161 & 93.2 & \nofit{F} & 126 & 51.5 & \nofit{F} & \nofit{F} & 94.5\,{\scriptsize$\pm$57.0} \\
\midrule
\textbf{Mean} & \fcell{217}{112} & \fcell{196}{97.8} & \fcell{165}{89.9} & \fcell{172}{74.1} & \fcell{221}{66.2} & \fcell{91.1}{105} & \fcell{189}{5.05} & \fcell{46.1}{9.79} & \fcell{165}{0} & \fcell{21.9}{23.7} & \fcell{50.9}{17.0} & \fcell{60.6}{59.2} & \fcell{161}{0} & \fcell{127}{28.1} & \unmeasured & \fcell{278}{95.1} & \fcell{156}{94.3} & \fcell{18.4}{6.43} & \fcell{12.0}{10.1} & \fcell{130}{80.2} \\
\bottomrule
\end{tabular}}
}

\end{appendixtable}

\begin{appendixtable}[90]
    \centering
    \caption{Flash (KB) required on the MCU with 2\,s windows, also where it exceeds the available 1024\,KB. Bold marks the lowest value per dataset.}
    \label{tab:deploy-mcu-flash-w2s}
    \providecommand{\nofit}[1]{\textcolor{gray}{#1}}
\providecommand{\faileddeployment}{\textcolor{gray}{Failed}}
\providecommand{\todomissing}{\textcolor{red}{TODO}}
\providecommand{\unmeasured}{\textcolor{gray}{--}}
\providecommand{\fcell}[2]{\begin{tabular}[c]{@{}c@{}}#1\\[-2pt]{\scriptsize$\pm$#2}\end{tabular}}
\providecommand{\tableheaderangle}{90}
\providecommand{\tablemaxheight}{0.9\textheight}
\providecommand{\fittable}[1]{\sbox2{#1}%
  \sbox2{\resizebox{\linewidth}{!}{\usebox2}}%
  \dimen2=\ht2\advance\dimen2 by\dp2
  \ifdim\dimen2>\tablemaxheight\resizebox*{!}{\tablemaxheight}{\usebox2}\else\usebox2\fi}

{%
\setlength{\tabcolsep}{2pt}
\setlength{\aboverulesep}{0pt}\setlength{\belowrulesep}{0pt}
\renewcommand{\arraystretch}{1.15}
\footnotesize
\fittable{%
\begin{tabular}{@{}lccccccccccccccccccc|c@{}}
Dataset & \rotatebox{\tableheaderangle}{CNN-HAR\hspace{4pt}} & \rotatebox{\tableheaderangle}{Triple-Cross-Attn\hspace{4pt}} & \rotatebox{\tableheaderangle}{TinyHAR\hspace{4pt}} & \rotatebox{\tableheaderangle}{TinierHAR\hspace{4pt}} & \rotatebox{\tableheaderangle}{DANA\hspace{4pt}} & \rotatebox{\tableheaderangle}{LITEWAY-Light\hspace{4pt}} & \rotatebox{\tableheaderangle}{Attend-and-Discriminate\hspace{4pt}} & \rotatebox{\tableheaderangle}{DynamicWHAR\hspace{4pt}} & \rotatebox{\tableheaderangle}{DeepConvLSTM-Attn\hspace{4pt}} & \rotatebox{\tableheaderangle}{Random Forest\hspace{4pt}} & \rotatebox{\tableheaderangle}{MLP-HAR\hspace{4pt}} & \rotatebox{\tableheaderangle}{LITEWAY\hspace{4pt}} & \rotatebox{\tableheaderangle}{DeepConvLSTM\hspace{4pt}} & \rotatebox{\tableheaderangle}{MLP-Mixer\hspace{4pt}} & \rotatebox{\tableheaderangle}{Guan-LSTM\hspace{4pt}} & \rotatebox{\tableheaderangle}{SA-HAR\hspace{4pt}} & \rotatebox{\tableheaderangle}{DeepConvShallowLSTM\hspace{4pt}} & \rotatebox{\tableheaderangle}{SVM\hspace{4pt}} & \rotatebox{\tableheaderangle}{k-NN\hspace{4pt}} & \rotatebox{\tableheaderangle}{\textbf{Mean}\hspace{4pt}} \\
\midrule
SAD & 672 & 759 & 533 & 417 & 2624 & 218 & 2561 & 6768 & 3104 & \textbf{178} & 2310 & 221 & 3059 & 1137 & 1390 & 710 & 2249 & 5205 & 20267 & 2862\,{\scriptsize$\pm$4572} \\
UCI-HAR & 253 & 465 & 411 & 359 & 918 & \textbf{179} & 1317 & 1541 & 1452 & 322 & 307 & 182 & 1407 & 877 & 1318 & 683 & 609 & 1547 & 2482 & 875\,{\scriptsize$\pm$634} \\
MotionSense & 327 & 509 & 432 & 369 & 1115 & \textbf{186} & 1537 & 2416 & 1744 & 255 & 462 & 188 & 1699 & 972 & 1331 & 688 & 901 & 5309 & 12709 & 1745\,{\scriptsize$\pm$2909} \\
MHEALTH & 370 & 637 & 445 & 375 & 1296 & \textbf{190} & 1659 & 2908 & 1906 & 232 & 589 & 193 & 1862 & 1005 & 1339 & 695 & 1051 & 1727 & 3738 & 1169\,{\scriptsize$\pm$964} \\
WEAR & 283 & 588 & 419 & 363 & 901 & \textbf{182} & 1392 & 1833 & 1551 & 1473 & 350 & 185 & 1506 & 911 & 1326 & 693 & 708 & 7304 & 12593 & 1819\,{\scriptsize$\pm$3042} \\
DSADS & 545 & 643 & 368 & 303 & 2078 & \textbf{198} & 1752 & 5132 & 2184 & 408 & 1342 & 201 & 2139 & 998 & 1144 & 696 & 1731 & 6144 & 14050 & 2214\,{\scriptsize$\pm$3282} \\
Opportunity (Loc.) & 1237 & 584 & 585 & 411 & 5227 & 254 & 3972 & 16891 & 5105 & \textbf{215} & 10222 & 257 & 5061 & 1268 & 1295 & 720 & 4573 & 9176 & 38724 & 5567\,{\scriptsize$\pm$9149} \\
WISDM & 203 & 425 & 275 & 263 & 701 & \textbf{173} & 735 & 987 & 830 & 429 & 181 & 176 & 785 & 829 & 1091 & 673 & 389 & 2301 & 4076 & 817\,{\scriptsize$\pm$933} \\
REALDISP & 1138 & 3352 & 672 & 483 & 4676 & \textbf{264} & 3955 & 14280 & 4954 & 1714 & 8075 & 267 & 4909 & 1523 & 1475 & 760 & 4099 & 68612 & 93770 & 11525\,{\scriptsize$\pm$25133} \\
PAMAP2 & 629 & 1387 & 786 & 599 & 2358 & \textbf{233} & 3259 & 5912 & 3721 & 641 & 2073 & 235 & 3677 & 1438 & 1837 & 742 & 2061 & 8948 & 26737 & 3541\,{\scriptsize$\pm$6026} \\
HAPT & 231 & 477 & 404 & 356 & 811 & \textbf{177} & 1245 & 1252 & 1356 & 512 & 276 & 180 & 1311 & 877 & 1316 & 686 & 513 & 1097 & 1490 & 767\,{\scriptsize$\pm$459} \\
HHAR & 227 & 430 & 281 & 265 & 809 & \textbf{175} & 807 & 1248 & 926 & 603 & 202 & 178 & 881 & 842 & 1095 & 674 & 486 & 5763 & 9229 & 1322\,{\scriptsize$\pm$2275} \\
Hang-Time & 203 & 433 & 396 & 352 & 702 & \textbf{174} & 1171 & 990 & 1257 & 728 & 254 & 177 & 1213 & 844 & 1310 & 679 & 415 & 3417 & 5179 & 1047\,{\scriptsize$\pm$1243} \\
w-HAR & 248 & 687 & 1147 & 909 & 857 & \textbf{188} & 4766 & 1375 & 4819 & 239 & 2388 & 191 & 4775 & 998 & 3076 & 744 & 756 & 501 & 1156 & 1569\,{\scriptsize$\pm$1603} \\
HuGaDB & 497 & 781 & 534 & 429 & 1837 & \textbf{204} & 2201 & 4419 & 2566 & 489 & 1112 & 207 & 2522 & 1054 & 1451 & 708 & 1550 & 5528 & 23627 & 2722\,{\scriptsize$\pm$5258} \\
HARTH & 231 & 477 & 404 & 356 & 811 & \textbf{177} & 1245 & 1252 & 1356 & 642 & 276 & 180 & 1311 & 877 & 1316 & 686 & 513 & 3172 & 17545 & 1728\,{\scriptsize$\pm$3894} \\
Daphnet & 252 & 443 & 482 & 408 & 918 & 179 & 1562 & 1543 & 1692 & \textbf{176} & 358 & 182 & 1648 & 895 & 1441 & 685 & 624 & 426 & 3855 & 935\,{\scriptsize$\pm$891} \\
USC-HAD & 233 & 534 & 658 & 529 & 813 & \textbf{179} & 2119 & 1259 & 2213 & 602 & 543 & 182 & 2167 & 940 & 1756 & 701 & 565 & 2791 & 3592 & 1178\,{\scriptsize$\pm$978} \\
HAR70+ & 229 & 455 & 404 & 356 & 810 & \textbf{176} & 1244 & 1252 & 1355 & 269 & 275 & 179 & 1310 & 876 & 1314 & 682 & 512 & 1210 & 5868 & 988\,{\scriptsize$\pm$1262} \\
GOTOV & 421 & 1000 & 646 & 505 & 1484 & \textbf{202} & 2409 & 3415 & 2684 & 933 & 936 & 205 & 2639 & 1057 & 1666 & 717 & 1257 & 25877 & 51765 & 5253\,{\scriptsize$\pm$12624} \\
UTD-MHAD & 237 & 545 & 406 & 357 & 813 & \textbf{178} & 1247 & 1254 & 1358 & 456 & 277 & 181 & 1313 & 879 & 1320 & 698 & 515 & 337 & 238 & 664\,{\scriptsize$\pm$436} \\
WISDM-19 (Watch) & 232 & 449 & 283 & 266 & 810 & \textbf{176} & 809 & 1250 & 928 & 2010 & 203 & 179 & 883 & 844 & 1098 & 683 & 488 & 32318 & 29681 & 3873\,{\scriptsize$\pm$9580} \\
UCA-EHAR & 238 & 443 & 284 & 274 & 845 & \textbf{176} & 849 & 1346 & 976 & 717 & 212 & 179 & 932 & 847 & 1106 & 679 & 520 & 3527 & 3890 & 949\,{\scriptsize$\pm$1033} \\
UP-Fall & 518 & 544 & 361 & 300 & 1970 & \textbf{196} & 1679 & 4818 & 2086 & 461 & 1189 & 199 & 2041 & 992 & 1139 & 689 & 1634 & 4433 & 12771 & 2001\,{\scriptsize$\pm$2916} \\
UMAFall & 549 & 2519 & 1063 & 950 & 1917 & \textbf{248} & 4711 & 4575 & 5036 & 431 & 2464 & 251 & 4990 & 1820 & 2720 & 788 & 1766 & 4059 & 4538 & 2389\,{\scriptsize$\pm$1751} \\
RealLifeHAR & 299 & 438 & 301 & 274 & 1004 & \textbf{180} & 1025 & 2118 & 1216 & 617 & 325 & 183 & 1171 & 879 & 1105 & 676 & 775 & 49657 & 238571 & 15832\,{\scriptsize$\pm$55092} \\
UniMiB-SHAR & 209 & 460 & 398 & 353 & 704 & \textbf{175} & 1172 & 992 & 1259 & 1044 & 254 & 178 & 1214 & 846 & 1313 & 688 & 417 & 1011 & 680 & 703\,{\scriptsize$\pm$397} \\
AICOS-HAR & 261 & 450 & 291 & 270 & 952 & \textbf{178} & 904 & 1635 & 1056 & 910 & 246 & 181 & 1011 & 859 & 1101 & 679 & 615 & 26403 & 54914 & 4890\,{\scriptsize$\pm$13476} \\
WISDM-19 (Phone) & 232 & 449 & 283 & 266 & 810 & \textbf{176} & 809 & 1250 & 928 & 2015 & 203 & 179 & 883 & 844 & 1098 & 683 & 488 & 46590 & 38497 & 5088\,{\scriptsize$\pm$13276} \\
\midrule
\textbf{Mean} & \fcell{386}{261} & \fcell{737}{652} & \fcell{481}{218} & \fcell{404}{168} & \fcell{1433}{1118} & \fcell{193}{25.9} & \fcell{1866}{1178} & \fcell{3307}{3795} & \fcell{2125}{1349} & \fcell{680}{517} & \fcell{1307}{2306} & \fcell{196}{25.8} & \fcell{2080}{1349} & \fcell{1001}{235} & \fcell{1424}{454} & \fcell{699}{27.8} & \fcell{1130}{1040} & \fcell{11531}{17286} & \fcell{25387}{46097} & \fcell{2967}{5980} \\
\bottomrule
\end{tabular}}
}

\end{appendixtable}

\begin{appendixtable}[90]
    \centering
    \caption{Flash (KB) required on the MCU with 3\,s windows, also where it exceeds the available 1024\,KB. Bold marks the lowest value per dataset.}
    \label{tab:deploy-mcu-flash-w3s}
    \providecommand{\nofit}[1]{\textcolor{gray}{#1}}
\providecommand{\faileddeployment}{\textcolor{gray}{Failed}}
\providecommand{\todomissing}{\textcolor{red}{TODO}}
\providecommand{\unmeasured}{\textcolor{gray}{--}}
\providecommand{\fcell}[2]{\begin{tabular}[c]{@{}c@{}}#1\\[-2pt]{\scriptsize$\pm$#2}\end{tabular}}
\providecommand{\tableheaderangle}{90}
\providecommand{\tablemaxheight}{0.9\textheight}
\providecommand{\fittable}[1]{\sbox2{#1}%
  \sbox2{\resizebox{\linewidth}{!}{\usebox2}}%
  \dimen2=\ht2\advance\dimen2 by\dp2
  \ifdim\dimen2>\tablemaxheight\resizebox*{!}{\tablemaxheight}{\usebox2}\else\usebox2\fi}

{%
\setlength{\tabcolsep}{2pt}
\setlength{\aboverulesep}{0pt}\setlength{\belowrulesep}{0pt}
\renewcommand{\arraystretch}{1.15}
\footnotesize
\fittable{%
\begin{tabular}{@{}lccccccccccccccccccc|c@{}}
Dataset & \rotatebox{\tableheaderangle}{CNN-HAR\hspace{4pt}} & \rotatebox{\tableheaderangle}{Triple-Cross-Attn\hspace{4pt}} & \rotatebox{\tableheaderangle}{TinyHAR\hspace{4pt}} & \rotatebox{\tableheaderangle}{TinierHAR\hspace{4pt}} & \rotatebox{\tableheaderangle}{DANA\hspace{4pt}} & \rotatebox{\tableheaderangle}{LITEWAY-Light\hspace{4pt}} & \rotatebox{\tableheaderangle}{Attend-and-Discriminate\hspace{4pt}} & \rotatebox{\tableheaderangle}{DynamicWHAR\hspace{4pt}} & \rotatebox{\tableheaderangle}{DeepConvLSTM-Attn\hspace{4pt}} & \rotatebox{\tableheaderangle}{Random Forest\hspace{4pt}} & \rotatebox{\tableheaderangle}{MLP-HAR\hspace{4pt}} & \rotatebox{\tableheaderangle}{LITEWAY\hspace{4pt}} & \rotatebox{\tableheaderangle}{DeepConvLSTM\hspace{4pt}} & \rotatebox{\tableheaderangle}{MLP-Mixer\hspace{4pt}} & \rotatebox{\tableheaderangle}{Guan-LSTM\hspace{4pt}} & \rotatebox{\tableheaderangle}{SA-HAR\hspace{4pt}} & \rotatebox{\tableheaderangle}{DeepConvShallowLSTM\hspace{4pt}} & \rotatebox{\tableheaderangle}{SVM\hspace{4pt}} & \rotatebox{\tableheaderangle}{k-NN\hspace{4pt}} & \rotatebox{\tableheaderangle}{\textbf{Mean}\hspace{4pt}} \\
\midrule
SAD & 684 & 981 & 676 & 511 & 2636 & 230 & 3009 & 6781 & 3544 & \textbf{168} & 2427 & 233 & 3497 & 1349 & 1620 & 728 & 2286 & 4012 & 13188 & 2556\,{\scriptsize$\pm$3083} \\
UCI-HAR & 255 & 494 & 544 & 443 & 919 & \textbf{180} & 1755 & 1545 & 1882 & 255 & 420 & 183 & 1837 & 911 & 1539 & 691 & 636 & 965 & 1397 & 887\,{\scriptsize$\pm$594} \\
MotionSense & 330 & 566 & 567 & 455 & 1119 & \textbf{189} & 1976 & 2421 & 2175 & 214 & 577 & 192 & 2130 & 929 & 1553 & 697 & 929 & 3556 & 7964 & 1502\,{\scriptsize$\pm$1819} \\
MHEALTH & 375 & 780 & 580 & 462 & 1300 & 194 & 2100 & 2914 & 2339 & \textbf{186} & 706 & 197 & 2294 & 976 & 1563 & 705 & 1081 & 1226 & 2244 & 1170\,{\scriptsize$\pm$845} \\
WEAR & 285 & 698 & 553 & 447 & 903 & \textbf{184} & 1830 & 1838 & 1981 & 1298 & 465 & 187 & 1936 & 961 & 1547 & 702 & 736 & 4849 & 7962 & 1545\,{\scriptsize$\pm$1888} \\
DSADS & 550 & 862 & 434 & 348 & 2083 & \textbf{203} & 1974 & 5137 & 2402 & 420 & 1374 & 206 & 2357 & 957 & 1258 & 704 & 1748 & 7876 & 22476 & 2809\,{\scriptsize$\pm$5125} \\
Opportunity (Loc.) & 1252 & 734 & 682 & 475 & 5242 & 269 & 4249 & 16907 & 5377 & \textbf{176} & 10281 & 272 & 5333 & 1555 & 1441 & 739 & 4603 & 6028 & 23555 & 4693\,{\scriptsize$\pm$6220} \\
WISDM & 203 & 425 & 295 & 283 & 702 & \textbf{174} & 822 & 988 & 915 & 368 & 191 & 176 & 871 & 832 & 1135 & 674 & 394 & 1594 & 2622 & 719\,{\scriptsize$\pm$603} \\
REALDISP & 1161 & 5305 & 825 & 588 & 4699 & \textbf{287} & 4414 & 14305 & 5405 & 1452 & 8203 & 290 & 5360 & 1935 & 1717 & 789 & 4147 & 44147 & 54145 & 8378\,{\scriptsize$\pm$14862} \\
PAMAP2 & 649 & 2031 & 893 & 791 & 2378 & \textbf{253} & 4148 & 5936 & 4595 & 578 & 2548 & 256 & 4552 & 1793 & 2295 & 775 & 2131 & 6538 & 17402 & 3186\,{\scriptsize$\pm$3938} \\
HAPT & 232 & 515 & 537 & 439 & 812 & \textbf{178} & 1682 & 1255 & 1785 & 412 & 389 & 181 & 1740 & 906 & 1536 & 693 & 539 & 660 & 860 & 808\,{\scriptsize$\pm$537} \\
HHAR & 230 & 469 & 536 & 439 & 811 & \textbf{177} & 1681 & 1255 & 1784 & 530 & 388 & 180 & 1739 & 905 & 1534 & 689 & 538 & 4303 & 5482 & 1246\,{\scriptsize$\pm$1405} \\
Hang-Time & 204 & 441 & 528 & 435 & 702 & \textbf{174} & 1607 & 992 & 1686 & 649 & 366 & 177 & 1639 & 861 & 1529 & 686 & 440 & 2100 & 3172 & 968\,{\scriptsize$\pm$793} \\
w-HAR & 255 & 819 & 1647 & 1271 & 864 & 195 & 6952 & 1391 & \faileddeployment & \textbf{148} & 5119 & 198 & 6915 & 1135 & 4176 & 782 & 888 & 300 & 675 & 1874\,{\scriptsize$\pm$2271} \\
HuGaDB & 505 & 1121 & 694 & 541 & 1846 & \textbf{213} & 2735 & 4430 & 3088 & 409 & 1288 & 216 & 3042 & 1218 & 1722 & 724 & 1589 & 3826 & 14456 & 2298\,{\scriptsize$\pm$3200} \\
HARTH & 232 & 515 & 537 & 439 & 812 & \textbf{178} & 1682 & 1255 & 1785 & 557 & 389 & 181 & 1740 & 906 & 1536 & 693 & 539 & 2179 & 11242 & 1442\,{\scriptsize$\pm$2451} \\
Daphnet & 255 & 454 & 645 & 519 & 920 & 182 & 2122 & 1548 & 2242 & \textbf{138} & 546 & 185 & 2197 & 936 & 1723 & 695 & 658 & 257 & 2467 & 984\,{\scriptsize$\pm$798} \\
USC-HAD & 236 & 608 & 912 & 702 & 815 & \textbf{181} & 2993 & 1265 & 3069 & 518 & 1014 & 184 & 3024 & 1003 & 2195 & 715 & 618 & 1660 & 2077 & 1252\,{\scriptsize$\pm$969} \\
HAR70+ & 230 & 477 & 536 & 439 & 811 & \textbf{177} & 1681 & 1255 & 1784 & 190 & 388 & 180 & 1739 & 905 & 1535 & 690 & 538 & 808 & 3661 & 949\,{\scriptsize$\pm$855} \\
GOTOV & 430 & 1446 & 877 & 665 & 1493 & \textbf{207} & 3168 & 3427 & 3429 & 872 & 1224 & 214 & \faileddeployment & 1218 & 2051 & 737 & 1310 & 16231 & 31437 & 3913\,{\scriptsize$\pm$7767} \\
UTD-MHAD & 239 & 628 & 538 & 440 & 814 & 179 & 1684 & 1257 & 1787 & 268 & 390 & 182 & 1742 & 908 & 1540 & 705 & 542 & 194 & \textbf{138} & 746\,{\scriptsize$\pm$577} \\
WISDM-19 (Watch) & 233 & 449 & 303 & 287 & 811 & \textbf{176} & 896 & 1251 & 1014 & 1898 & 213 & 179 & 969 & 853 & 1142 & 684 & 493 & 21528 & 19351 & 2775\,{\scriptsize$\pm$6251} \\
UCA-EHAR & 239 & 465 & 356 & 315 & 846 & \textbf{177} & 1077 & 1347 & 1199 & 610 & 243 & 180 & 1155 & 864 & 1220 & 683 & 534 & 2342 & 2429 & 857\,{\scriptsize$\pm$662} \\
UP-Fall & 519 & 546 & 383 & 322 & 1972 & \textbf{198} & 1768 & 4820 & 2173 & 358 & 1198 & 201 & 2128 & 907 & 1184 & 692 & 1640 & 3043 & 7481 & 1660\,{\scriptsize$\pm$1830} \\
UMAFall & 579 & 3573 & 1485 & 1154 & 1947 & \textbf{279} & 6483 & 4614 & 6775 & 405 & 4241 & 281 & 6730 & 2366 & 3625 & 844 & 1896 & 2358 & 2518 & 2745\,{\scriptsize$\pm$2174} \\
RealLifeHAR & 299 & 439 & 322 & 295 & 1005 & \textbf{181} & 1112 & 2119 & 1302 & 572 & 335 & 184 & 1257 & 889 & 1149 & 677 & 781 & 34433 & 158473 & 10833\,{\scriptsize$\pm$36578} \\
UniMiB-SHAR & 209 & 487 & 529 & 435 & 704 & \textbf{175} & 1609 & 994 & 1686 & 794 & 367 & 178 & 1641 & 862 & 1532 & 695 & 442 & 614 & 412 & 756\,{\scriptsize$\pm$508} \\
AICOS-HAR & 265 & 564 & 547 & 444 & 956 & \textbf{182} & 1780 & 1643 & 1915 & 854 & 434 & 185 & 1870 & 954 & 1542 & 695 & 669 & 20591 & 35010 & 3742\,{\scriptsize$\pm$8831} \\
WISDM-19 (Phone) & 233 & 449 & 303 & 287 & 811 & \textbf{176} & 896 & 1251 & 1014 & 1921 & 213 & 179 & 969 & 853 & 1142 & 684 & 493 & 30739 & 25196 & 3569\,{\scriptsize$\pm$8659} \\
\midrule
\textbf{Mean} & \fcell{392}{267} & \fcell{943}{1055} & \fcell{630}{311} & \fcell{506}{231} & \fcell{1439}{1124} & \fcell{198}{33.1} & \fcell{2410}{1531} & \fcell{3315}{3800} & \fcell{2505}{1456} & \fcell{594}{485} & \fcell{1584}{2445} & \fcell{201}{33.1} & \fcell{2586}{1676} & \fcell{1091}{373} & \fcell{1699}{678} & \fcell{713}{39.4} & \fcell{1167}{1046} & \fcell{7895}{11477} & \fcell{16534}{30170} & \fcell{2442}{3836} \\
\bottomrule
\end{tabular}}
}

\end{appendixtable}

\begin{appendixtable}[90]
    \centering
    \caption{Energy per inference (mJ) on the MCU with 2\,s windows. Bold marks the lowest value per dataset. F: does not fit in flash, R: does not fit in RAM.}
    \label{tab:deploy-mcu-energy-w2s}
    \providecommand{\nofit}[1]{\textcolor{gray}{#1}}
\providecommand{\faileddeployment}{\textcolor{gray}{Failed}}
\providecommand{\todomissing}{\textcolor{red}{TODO}}
\providecommand{\unmeasured}{\textcolor{gray}{--}}
\providecommand{\fcell}[2]{\begin{tabular}[c]{@{}c@{}}#1\\[-2pt]{\scriptsize$\pm$#2}\end{tabular}}
\providecommand{\tableheaderangle}{90}
\providecommand{\tablemaxheight}{0.9\textheight}
\providecommand{\fittable}[1]{\sbox2{#1}%
  \sbox2{\resizebox{\linewidth}{!}{\usebox2}}%
  \dimen2=\ht2\advance\dimen2 by\dp2
  \ifdim\dimen2>\tablemaxheight\resizebox*{!}{\tablemaxheight}{\usebox2}\else\usebox2\fi}

{%
\setlength{\tabcolsep}{2pt}
\setlength{\aboverulesep}{0pt}\setlength{\belowrulesep}{0pt}
\renewcommand{\arraystretch}{1.15}
\footnotesize
\fittable{%
\begin{tabular}{@{}lccccccccccccccccccc|c@{}}
Dataset & \rotatebox{\tableheaderangle}{CNN-HAR\hspace{4pt}} & \rotatebox{\tableheaderangle}{Triple-Cross-Attn\hspace{4pt}} & \rotatebox{\tableheaderangle}{TinyHAR\hspace{4pt}} & \rotatebox{\tableheaderangle}{TinierHAR\hspace{4pt}} & \rotatebox{\tableheaderangle}{DANA\hspace{4pt}} & \rotatebox{\tableheaderangle}{LITEWAY-Light\hspace{4pt}} & \rotatebox{\tableheaderangle}{Attend-and-Discriminate\hspace{4pt}} & \rotatebox{\tableheaderangle}{DynamicWHAR\hspace{4pt}} & \rotatebox{\tableheaderangle}{DeepConvLSTM-Attn\hspace{4pt}} & \rotatebox{\tableheaderangle}{Random Forest\hspace{4pt}} & \rotatebox{\tableheaderangle}{MLP-HAR\hspace{4pt}} & \rotatebox{\tableheaderangle}{LITEWAY\hspace{4pt}} & \rotatebox{\tableheaderangle}{DeepConvLSTM\hspace{4pt}} & \rotatebox{\tableheaderangle}{MLP-Mixer\hspace{4pt}} & \rotatebox{\tableheaderangle}{Guan-LSTM\hspace{4pt}} & \rotatebox{\tableheaderangle}{SA-HAR\hspace{4pt}} & \rotatebox{\tableheaderangle}{DeepConvShallowLSTM\hspace{4pt}} & \rotatebox{\tableheaderangle}{SVM\hspace{4pt}} & \rotatebox{\tableheaderangle}{k-NN\hspace{4pt}} & \rotatebox{\tableheaderangle}{\textbf{Mean}\hspace{4pt}} \\
\midrule
SAD & \nofit{R} & \nofit{R} & 107 & 21.5 & \nofit{F} & 12.6 & \nofit{F} & \nofit{F} & \nofit{F} & \textbf{0.30} & \nofit{F} & 26.0 & \nofit{F} & \nofit{F} & \nofit{F} & \nofit{R} & \nofit{F} & \nofit{F} & \nofit{F} & 33.4\,{\scriptsize$\pm$42.1} \\
UCI-HAR & 62.5 & 81.7 & 20.6 & 6.05 & 161 & 2.19 & \nofit{F} & \nofit{F} & \nofit{F} & \textbf{0.06} & 7.78 & 4.60 & \nofit{F} & 143 & \nofit{F} & 190 & 250 & \nofit{F} & \nofit{F} & 77.5\,{\scriptsize$\pm$87.5} \\
MotionSense & 116 & 158 & 35.7 & 8.89 & \nofit{F} & 4.02 & \nofit{F} & \nofit{F} & \nofit{F} & \textbf{0.10} & 18.5 & 8.34 & \nofit{F} & 352 & \nofit{F} & 216 & 483 & \nofit{F} & \nofit{F} & 127\,{\scriptsize$\pm$163} \\
MHEALTH & \nofit{R} & 225 & 45.4 & 10.6 & \nofit{F} & 5.62 & \nofit{F} & \nofit{F} & \nofit{F} & \textbf{0.13} & 27.1 & 11.7 & \nofit{F} & 427 & \nofit{F} & 258 & \nofit{F} & \nofit{F} & \nofit{F} & 112\,{\scriptsize$\pm$154} \\
WEAR & 80.3 & 108 & 25.7 & 6.99 & 206 & \textbf{2.81} & \nofit{F} & \nofit{F} & \nofit{F} & \nofit{F} & 10.9 & 5.84 & \nofit{F} & 195 & \nofit{F} & 198 & 345 & \nofit{F} & \nofit{F} & 108\,{\scriptsize$\pm$114} \\
DSADS & 118 & 183 & 34.7 & 8.46 & \nofit{F} & 4.93 & \nofit{F} & \nofit{F} & \nofit{F} & \textbf{0.13} & \nofit{F} & 9.61 & \nofit{F} & 383 & \nofit{F} & 138 & \nofit{F} & \nofit{F} & \nofit{F} & 97.8\,{\scriptsize$\pm$127} \\
Opportunity (Loc.) & \nofit{F} & \nofit{R} & 127 & 26.7 & \nofit{F} & 18.3 & \nofit{F} & \nofit{F} & \nofit{F} & \textbf{0.37} & \nofit{F} & 37.7 & \nofit{F} & \nofit{F} & \nofit{F} & \nofit{R} & \nofit{F} & \nofit{F} & \nofit{F} & 42.0\,{\scriptsize$\pm$49.3} \\
WISDM & 11.0 & 14.5 & 4.20 & 2.01 & 37.3 & 0.51 & 66.2 & 6.40 & 76.9 & \textbf{0.03} & 1.06 & 1.03 & 72.3 & 39.5 & \nofit{F} & 77.8 & 24.0 & \nofit{F} & \nofit{F} & 27.2\,{\scriptsize$\pm$30.2} \\
REALDISP & \nofit{F} & \nofit{F} & \nofit{R} & 39.5 & \nofit{F} & \textbf{26.8} & \nofit{F} & \nofit{F} & \nofit{F} & \nofit{F} & \nofit{F} & 55.4 & \nofit{F} & \nofit{F} & \nofit{F} & \nofit{R} & \nofit{F} & \nofit{F} & \nofit{F} & 40.6\,{\scriptsize$\pm$14.4} \\
PAMAP2 & \nofit{R} & \nofit{F} & \nofit{R} & 39.0 & \nofit{F} & 24.2 & \nofit{F} & \nofit{F} & \nofit{F} & \textbf{0.48} & \nofit{F} & 50.1 & \nofit{F} & \nofit{F} & \nofit{F} & \nofit{R} & \nofit{F} & \nofit{F} & \nofit{F} & 28.5\,{\scriptsize$\pm$21.5} \\
HAPT & 44.9 & 56.3 & 15.6 & 5.09 & 111 & 1.58 & \nofit{F} & \nofit{F} & \nofit{F} & \textbf{0.05} & 4.88 & 3.33 & \nofit{F} & 140 & \nofit{F} & 181 & 175 & \nofit{F} & \nofit{F} & 61.6\,{\scriptsize$\pm$70.9} \\
HHAR & 18.7 & 29.2 & 6.42 & 2.49 & 63.5 & 0.92 & 87.8 & \nofit{F} & 101 & \textbf{0.03} & 2.42 & 1.84 & 94.7 & 75.8 & \nofit{F} & 91.1 & 49.0 & \nofit{F} & \nofit{F} & 41.7\,{\scriptsize$\pm$40.2} \\
Hang-Time & 27.3 & 34.3 & 10.5 & 4.15 & 62.1 & 1.08 & \nofit{F} & 8.23 & \nofit{F} & \textbf{0.03} & 2.58 & 2.31 & \nofit{F} & 73.1 & \nofit{F} & 191 & 109 & \nofit{F} & \nofit{F} & 40.4\,{\scriptsize$\pm$56.6} \\
w-HAR & \nofit{R} & \nofit{R} & \nofit{F} & \nofit{R} & \nofit{R} & 9.80 & \nofit{F} & \nofit{F} & \nofit{F} & \textbf{0.18} & \nofit{F} & 20.7 & \nofit{F} & 698 & \nofit{F} & \nofit{R} & \nofit{R} & 1.41 & \nofit{F} & 146\,{\scriptsize$\pm$309} \\
HuGaDB & \nofit{R} & \nofit{R} & 85.9 & 18.0 & \nofit{F} & 9.73 & \nofit{F} & \nofit{F} & \nofit{F} & \textbf{0.23} & \nofit{F} & 20.1 & \nofit{F} & \nofit{F} & \nofit{F} & \nofit{R} & \nofit{F} & \nofit{F} & \nofit{F} & 26.8\,{\scriptsize$\pm$34.0} \\
HARTH & 44.9 & 56.5 & 15.6 & 5.09 & 108 & 1.57 & \nofit{F} & \nofit{F} & \nofit{F} & \textbf{0.05} & 4.87 & 3.33 & \nofit{F} & 128 & \nofit{F} & 181 & 176 & \nofit{F} & \nofit{F} & 60.3\,{\scriptsize$\pm$69.6} \\
Daphnet & 83.4 & 116 & 28.1 & 7.78 & 209 & 3.12 & \nofit{F} & \nofit{F} & \nofit{F} & \textbf{0.07} & 11.4 & 6.52 & \nofit{F} & 175 & \nofit{F} & 289 & 409 & 0.96 & \nofit{F} & 103\,{\scriptsize$\pm$132} \\
USC-HAD & 98.4 & 111 & 34.2 & 10.1 & 209 & 3.13 & \nofit{F} & \nofit{F} & \nofit{F} & \textbf{0.08} & 12.3 & 6.59 & \nofit{F} & 267 & \nofit{F} & \nofit{R} & 438 & \nofit{F} & \nofit{F} & 108\,{\scriptsize$\pm$142} \\
HAR70+ & 44.8 & 56.4 & 15.6 & 5.08 & 110 & 1.57 & \nofit{F} & \nofit{F} & \nofit{F} & \textbf{0.05} & 4.84 & 3.33 & \nofit{F} & 127 & \nofit{F} & 181 & 185 & \nofit{F} & \nofit{F} & 61.2\,{\scriptsize$\pm$71.1} \\
GOTOV & \nofit{R} & \nofit{R} & 99.6 & 20.5 & \nofit{F} & 11.6 & \nofit{F} & \nofit{F} & \nofit{F} & \textbf{0.21} & 68.8 & 23.5 & \nofit{F} & \nofit{F} & \nofit{F} & \nofit{R} & \nofit{F} & \nofit{F} & \nofit{F} & 37.4\,{\scriptsize$\pm$38.4} \\
UTD-MHAD & 44.9 & 56.4 & 15.6 & 5.10 & 111 & 1.59 & \nofit{F} & \nofit{F} & \nofit{F} & \textbf{0.06} & 4.86 & 3.33 & \nofit{F} & 140 & \nofit{F} & 181 & 175 & 0.86 & 0.40 & 52.9\,{\scriptsize$\pm$68.9} \\
WISDM-19 (Watch) & 18.8 & 26.3 & 6.40 & 2.49 & 63.3 & \textbf{0.84} & 87.3 & \nofit{F} & 98.9 & \nofit{F} & 2.17 & 1.64 & 93.8 & 72.0 & \nofit{F} & 82.1 & 45.2 & \nofit{F} & \nofit{F} & 42.9\,{\scriptsize$\pm$38.7} \\
UCA-EHAR & 22.0 & 34.6 & 7.35 & 2.85 & 74.5 & 1.05 & 100 & \nofit{F} & 114 & \textbf{0.04} & 2.88 & 2.21 & 108 & 76.1 & \nofit{F} & 96.6 & 63.1 & \nofit{F} & \nofit{F} & 47.1\,{\scriptsize$\pm$44.7} \\
UP-Fall & 110 & 170 & 32.6 & 8.04 & \nofit{F} & 4.60 & \nofit{F} & \nofit{F} & \nofit{F} & \textbf{0.12} & \nofit{F} & 9.01 & \nofit{F} & 377 & \nofit{F} & 134 & \nofit{F} & \nofit{F} & \nofit{F} & 94.0\,{\scriptsize$\pm$124} \\
UMAFall & \nofit{R} & \nofit{F} & \nofit{F} & \nofit{R} & \nofit{F} & 33.0 & \nofit{F} & \nofit{F} & \nofit{F} & \textbf{0.72} & \nofit{F} & 68.5 & \nofit{F} & \nofit{F} & \nofit{F} & \nofit{R} & \nofit{F} & \nofit{F} & \nofit{F} & 34.1\,{\scriptsize$\pm$33.9} \\
RealLifeHAR & 41.5 & 62.3 & 13.0 & 3.88 & 143 & 1.77 & \nofit{F} & \nofit{F} & \nofit{F} & \textbf{0.05} & 6.30 & 3.50 & \nofit{F} & 137 & \nofit{F} & 96.6 & 109 & \nofit{F} & \nofit{F} & 51.5\,{\scriptsize$\pm$56.0} \\
UniMiB-SHAR & 27.3 & 30.9 & 10.5 & 4.15 & 61.7 & \textbf{0.97} & \nofit{F} & 7.41 & \nofit{F} & \nofit{F} & 2.32 & 2.08 & \nofit{F} & 73.2 & \nofit{F} & 173 & 98.7 & \nofit{F} & 1.14 & 37.9\,{\scriptsize$\pm$51.6} \\
AICOS-HAR & 28.9 & 46.8 & 9.36 & 3.12 & 98.1 & 1.39 & 116 & \nofit{F} & \nofit{F} & \textbf{0.04} & 4.38 & 2.75 & 124 & 114 & \nofit{F} & 97.4 & 79.0 & \nofit{F} & \nofit{F} & 51.8\,{\scriptsize$\pm$50.2} \\
WISDM-19 (Phone) & 18.7 & 26.3 & 6.40 & 2.48 & 63.2 & \textbf{0.83} & 87.4 & \nofit{F} & 99.0 & \nofit{F} & 2.19 & 1.64 & 93.9 & 72.0 & \nofit{F} & 82.2 & 45.2 & \nofit{F} & \nofit{F} & 43.0\,{\scriptsize$\pm$38.7} \\
\midrule
\textbf{Mean} & \fcell{53.1}{35.4} & \fcell{80.2}{59.8} & \fcell{32.5}{34.5} & \fcell{10.4}{10.5} & \fcell{111}{56.1} & \fcell{6.62}{8.63} & \fcell{90.8}{16.4} & \fcell{7.35}{0.92} & \fcell{98.1}{13.5} & \fcell{0.15}{0.17} & \fcell{10.1}{15.2} & \fcell{13.7}{17.9} & \fcell{97.8}{17.3} & \fcell{195}{161} & \unmeasured & \fcell{157}{61.2} & \fcell{181}{146} & \fcell{1.08}{0.29} & \fcell{0.77}{0.52} & \fcell{63.7}{65.5} \\
\bottomrule
\end{tabular}}
}

\end{appendixtable}

\begin{appendixtable}[90]
    \centering
    \caption{Energy per inference (mJ) on the MCU with 3\,s windows. Bold marks the lowest value per dataset. F: does not fit in flash, R: does not fit in RAM.}
    \label{tab:deploy-mcu-energy-w3s}
    \providecommand{\nofit}[1]{\textcolor{gray}{#1}}
\providecommand{\faileddeployment}{\textcolor{gray}{Failed}}
\providecommand{\todomissing}{\textcolor{red}{TODO}}
\providecommand{\unmeasured}{\textcolor{gray}{--}}
\providecommand{\fcell}[2]{\begin{tabular}[c]{@{}c@{}}#1\\[-2pt]{\scriptsize$\pm$#2}\end{tabular}}
\providecommand{\tableheaderangle}{90}
\providecommand{\tablemaxheight}{0.9\textheight}
\providecommand{\fittable}[1]{\sbox2{#1}%
  \sbox2{\resizebox{\linewidth}{!}{\usebox2}}%
  \dimen2=\ht2\advance\dimen2 by\dp2
  \ifdim\dimen2>\tablemaxheight\resizebox*{!}{\tablemaxheight}{\usebox2}\else\usebox2\fi}

{%
\setlength{\tabcolsep}{2pt}
\setlength{\aboverulesep}{0pt}\setlength{\belowrulesep}{0pt}
\renewcommand{\arraystretch}{1.15}
\footnotesize
\fittable{%
\begin{tabular}{@{}lccccccccccccccccccc|c@{}}
Dataset & \rotatebox{\tableheaderangle}{CNN-HAR\hspace{4pt}} & \rotatebox{\tableheaderangle}{Triple-Cross-Attn\hspace{4pt}} & \rotatebox{\tableheaderangle}{TinyHAR\hspace{4pt}} & \rotatebox{\tableheaderangle}{TinierHAR\hspace{4pt}} & \rotatebox{\tableheaderangle}{DANA\hspace{4pt}} & \rotatebox{\tableheaderangle}{LITEWAY-Light\hspace{4pt}} & \rotatebox{\tableheaderangle}{Attend-and-Discriminate\hspace{4pt}} & \rotatebox{\tableheaderangle}{DynamicWHAR\hspace{4pt}} & \rotatebox{\tableheaderangle}{DeepConvLSTM-Attn\hspace{4pt}} & \rotatebox{\tableheaderangle}{Random Forest\hspace{4pt}} & \rotatebox{\tableheaderangle}{MLP-HAR\hspace{4pt}} & \rotatebox{\tableheaderangle}{LITEWAY\hspace{4pt}} & \rotatebox{\tableheaderangle}{DeepConvLSTM\hspace{4pt}} & \rotatebox{\tableheaderangle}{MLP-Mixer\hspace{4pt}} & \rotatebox{\tableheaderangle}{Guan-LSTM\hspace{4pt}} & \rotatebox{\tableheaderangle}{SA-HAR\hspace{4pt}} & \rotatebox{\tableheaderangle}{DeepConvShallowLSTM\hspace{4pt}} & \rotatebox{\tableheaderangle}{SVM\hspace{4pt}} & \rotatebox{\tableheaderangle}{k-NN\hspace{4pt}} & \rotatebox{\tableheaderangle}{\textbf{Mean}\hspace{4pt}} \\
\midrule
SAD & \nofit{R} & \nofit{R} & \nofit{R} & 32.2 & \nofit{F} & 19.0 & \nofit{F} & \nofit{F} & \nofit{F} & \textbf{0.41} & \nofit{F} & 38.6 & \nofit{F} & \nofit{F} & \nofit{F} & \nofit{R} & \nofit{F} & \nofit{F} & \nofit{F} & 22.6\,{\scriptsize$\pm$16.9} \\
UCI-HAR & 99.9 & 127 & 33.0 & 8.87 & 240 & 3.34 & \nofit{F} & \nofit{F} & \nofit{F} & \textbf{0.08} & 12.5 & 6.83 & \nofit{F} & 205 & \nofit{F} & 317 & 453 & 2.42 & \nofit{F} & 116\,{\scriptsize$\pm$146} \\
MotionSense & \nofit{R} & \nofit{R} & 57.1 & 13.1 & \nofit{F} & 6.05 & \nofit{F} & \nofit{F} & \nofit{F} & \textbf{0.13} & 29.0 & 12.4 & \nofit{F} & 506 & \nofit{F} & \nofit{R} & 871 & \nofit{F} & \nofit{F} & 187\,{\scriptsize$\pm$325} \\
MHEALTH & \nofit{R} & \nofit{R} & 72.6 & 15.7 & \nofit{F} & 8.48 & \nofit{F} & \nofit{F} & \nofit{F} & \textbf{0.17} & 44.3 & 17.3 & \nofit{F} & 624 & \nofit{F} & \nofit{R} & \nofit{F} & \nofit{F} & \nofit{F} & 112\,{\scriptsize$\pm$227} \\
WEAR & 128 & 166 & 41.1 & 10.3 & 301 & \textbf{4.21} & \nofit{F} & \nofit{F} & \nofit{F} & \nofit{F} & 18.3 & 8.64 & \nofit{F} & 292 & \nofit{F} & 324 & 603 & \nofit{F} & \nofit{F} & 173\,{\scriptsize$\pm$191} \\
DSADS & \nofit{R} & \nofit{R} & 58.5 & 12.6 & \nofit{F} & 7.25 & \nofit{F} & \nofit{F} & \nofit{F} & \textbf{0.16} & \nofit{F} & 14.4 & \nofit{F} & 560 & \nofit{F} & \nofit{R} & \nofit{F} & \nofit{F} & \nofit{F} & 109\,{\scriptsize$\pm$222} \\
Opportunity (Loc.) & \nofit{F} & \nofit{R} & \nofit{R} & 39.5 & \nofit{F} & 27.7 & \nofit{F} & \nofit{F} & \nofit{F} & \textbf{0.52} & \nofit{F} & 55.7 & \nofit{F} & \nofit{F} & \nofit{F} & \nofit{R} & \nofit{F} & \nofit{F} & \nofit{F} & 30.9\,{\scriptsize$\pm$23.2} \\
WISDM & 14.0 & 16.5 & 5.32 & 2.50 & 42.1 & 0.60 & 85.2 & 6.66 & 98.8 & \textbf{0.03} & 1.23 & 1.27 & 92.9 & 45.3 & \nofit{F} & 95.1 & 38.8 & \nofit{F} & \nofit{F} & 34.1\,{\scriptsize$\pm$38.2} \\
REALDISP & \nofit{F} & \nofit{F} & \nofit{R} & 58.5 & \nofit{F} & \textbf{40.5} & \nofit{F} & \nofit{F} & \nofit{F} & \nofit{F} & \nofit{F} & 82.5 & \nofit{F} & \nofit{F} & \nofit{F} & \nofit{R} & \nofit{F} & \nofit{F} & \nofit{F} & 60.5\,{\scriptsize$\pm$21.1} \\
PAMAP2 & \nofit{R} & \nofit{F} & \nofit{R} & \nofit{R} & \nofit{F} & 36.3 & \nofit{F} & \nofit{F} & \nofit{F} & \textbf{0.72} & \nofit{F} & 75.3 & \nofit{F} & \nofit{F} & \nofit{F} & \nofit{R} & \nofit{F} & \nofit{F} & \nofit{F} & 37.4\,{\scriptsize$\pm$37.3} \\
HAPT & 72.1 & 86.8 & 25.1 & 7.48 & 160 & 2.38 & \nofit{F} & \nofit{F} & \nofit{F} & \textbf{0.06} & 8.25 & 4.89 & \nofit{F} & 200 & \nofit{F} & 298 & 307 & 1.71 & 1.41 & 83.9\,{\scriptsize$\pm$112} \\
HHAR & 72.4 & 96.5 & 25.2 & 7.51 & 162 & 2.64 & \nofit{F} & \nofit{F} & \nofit{F} & \textbf{0.06} & 9.14 & 5.48 & \nofit{F} & 204 & \nofit{F} & 331 & 338 & \nofit{F} & \nofit{F} & 104\,{\scriptsize$\pm$126} \\
Hang-Time & 44.3 & 52.8 & 17.1 & 6.11 & 86.8 & 1.61 & \nofit{F} & 9.32 & \nofit{F} & \textbf{0.04} & 4.37 & 3.40 & \nofit{F} & 108 & \nofit{F} & 317 & 191 & \nofit{F} & \nofit{F} & 64.7\,{\scriptsize$\pm$94.2} \\
w-HAR & \nofit{R} & \nofit{R} & \nofit{F} & \nofit{F} & \nofit{R} & 14.6 & \nofit{F} & \nofit{F} & \faileddeployment & \textbf{0.24} & \nofit{F} & 30.8 & \nofit{F} & \nofit{F} & \nofit{F} & \nofit{R} & \nofit{R} & 0.77 & 1.46 & 9.59\,{\scriptsize$\pm$13.3} \\
HuGaDB & \nofit{R} & \nofit{F} & 134 & 27.1 & \nofit{F} & 14.6 & \nofit{F} & \nofit{F} & \nofit{F} & \textbf{0.31} & \nofit{F} & 30.1 & \nofit{F} & \nofit{F} & \nofit{F} & \nofit{R} & \nofit{F} & \nofit{F} & \nofit{F} & 41.3\,{\scriptsize$\pm$53.3} \\
HARTH & 72.1 & 87.2 & 25.1 & 7.48 & 156 & 2.38 & \nofit{F} & \nofit{F} & \nofit{F} & \textbf{0.06} & 8.18 & 4.93 & \nofit{F} & 183 & \nofit{F} & 298 & 315 & \nofit{F} & \nofit{F} & 96.6\,{\scriptsize$\pm$116} \\
Daphnet & 131 & 173 & 43.4 & 11.7 & 306 & 4.66 & \nofit{F} & \nofit{F} & \nofit{F} & \textbf{0.09} & 20.2 & 9.74 & \nofit{F} & 261 & \nofit{F} & \nofit{R} & 682 & 0.61 & \nofit{F} & 137\,{\scriptsize$\pm$203} \\
USC-HAD & 152 & 171 & 53.0 & \nofit{R} & 306 & 4.68 & \nofit{F} & \nofit{F} & \nofit{F} & \textbf{0.11} & 23.4 & 9.85 & \nofit{F} & 419 & \nofit{F} & \nofit{R} & 699 & \nofit{F} & \nofit{F} & 184\,{\scriptsize$\pm$230} \\
HAR70+ & 72.1 & 86.9 & 25.1 & 7.47 & 158 & 2.37 & \nofit{F} & \nofit{F} & \nofit{F} & \textbf{0.06} & 8.19 & 4.89 & \nofit{F} & 188 & \nofit{F} & 298 & 323 & 2.10 & \nofit{F} & 90.5\,{\scriptsize$\pm$116} \\
GOTOV & \nofit{R} & \nofit{F} & \nofit{R} & 31.1 & \nofit{F} & 17.4 & \nofit{F} & \nofit{F} & \nofit{F} & \textbf{0.30} & \nofit{F} & 35.6 & \faileddeployment & \nofit{F} & \nofit{F} & \nofit{R} & \nofit{F} & \nofit{F} & \nofit{F} & 21.1\,{\scriptsize$\pm$15.9} \\
UTD-MHAD & 72.2 & 87.1 & 25.1 & 7.48 & 160 & 2.38 & \nofit{F} & \nofit{F} & \nofit{F} & \textbf{0.07} & 8.23 & 4.92 & \nofit{F} & 201 & \nofit{F} & 298 & 306 & 0.52 & 0.25 & 83.8\,{\scriptsize$\pm$112} \\
WISDM-19 (Watch) & 23.6 & 29.9 & 8.06 & 3.08 & 72.9 & \textbf{0.96} & 112 & \nofit{F} & 127 & \nofit{F} & 2.58 & 2.03 & 120 & 82.4 & \nofit{F} & 100 & 71.7 & \nofit{F} & \nofit{F} & 54.0\,{\scriptsize$\pm$48.8} \\
UCA-EHAR & 37.6 & 55.2 & 13.2 & 4.21 & 107 & 1.58 & \nofit{F} & \nofit{F} & \nofit{F} & \textbf{0.05} & 4.42 & 3.24 & \nofit{F} & 109 & \nofit{F} & 157 & 155 & \nofit{F} & \nofit{F} & 53.9\,{\scriptsize$\pm$61.5} \\
UP-Fall & \nofit{R} & \nofit{R} & 40.7 & 9.77 & \nofit{F} & 5.36 & \nofit{F} & \nofit{F} & \nofit{F} & \textbf{0.13} & \nofit{F} & 11.1 & \nofit{F} & 426 & \nofit{F} & 163 & \nofit{F} & \nofit{F} & \nofit{F} & 93.7\,{\scriptsize$\pm$157} \\
UMAFall & \nofit{R} & \nofit{F} & \nofit{F} & \nofit{F} & \nofit{F} & 49.6 & \nofit{F} & \nofit{F} & \nofit{F} & \textbf{1.01} & \nofit{F} & 103 & \nofit{F} & \nofit{F} & \nofit{F} & \nofit{R} & \nofit{F} & \nofit{F} & \nofit{F} & 51.2\,{\scriptsize$\pm$51.0} \\
RealLifeHAR & 52.1 & 70.5 & 16.2 & 4.81 & 165 & 2.07 & \nofit{F} & \nofit{F} & \nofit{F} & \textbf{0.06} & 7.80 & 4.27 & \nofit{F} & 157 & \nofit{F} & 116 & 168 & \nofit{F} & \nofit{F} & 63.6\,{\scriptsize$\pm$69.4} \\
UniMiB-SHAR & 44.3 & 47.5 & 17.1 & 6.09 & 86.0 & 1.44 & \nofit{F} & 8.40 & \nofit{F} & \textbf{0.05} & 3.95 & 3.04 & \nofit{F} & 108 & \nofit{F} & 286 & 174 & 1.72 & 0.69 & 52.5\,{\scriptsize$\pm$82.0} \\
AICOS-HAR & 108 & 156 & 36.7 & 9.51 & 267 & 4.02 & \nofit{F} & \nofit{F} & \nofit{F} & \textbf{0.08} & 15.7 & 8.30 & \nofit{F} & 329 & \nofit{F} & 357 & 554 & \nofit{F} & \nofit{F} & 154\,{\scriptsize$\pm$183} \\
WISDM-19 (Phone) & 23.6 & 29.9 & 8.06 & 3.08 & 73.0 & \textbf{0.97} & 112 & \nofit{F} & 127 & \nofit{F} & 2.59 & 2.02 & 120 & 82.6 & \nofit{F} & 100 & 71.6 & \nofit{F} & \nofit{F} & 54.1\,{\scriptsize$\pm$48.9} \\
\midrule
\textbf{Mean} & \fcell{71.7}{40.5} & \fcell{90.6}{51.5} & \fcell{35.5}{28.5} & \fcell{13.9}{13.6} & \fcell{168}{87.4} & \fcell{9.97}{13.0} & \fcell{103}{15.3} & \fcell{8.12}{1.35} & \fcell{117}{16.1} & \fcell{0.20}{0.24} & \fcell{12.2}{10.9} & \fcell{20.5}{26.7} & \fcell{111}{15.8} & \fcell{252}{166} & \unmeasured & \fcell{241}{98.2} & \fcell{351}{243} & \fcell{1.41}{0.77} & \fcell{0.95}{0.59} & \fcell{89.3}{103} \\
\bottomrule
\end{tabular}}
}

\end{appendixtable}

\clearpage

\section{Supporting Analyses for the Implications}
\label{sec:appendix-insights}

The figures in this section support the implications in \autoref{sec:discussion-implications}.
Accuracy figures use 3\,s windows unless they compare the two window lengths.

\subsection{Deployment Cost Details}
\label{sec:appendix-cost-details}

\paragraph{Memory on the MCU}
LITEWAY-Light has slightly fewer parameters than LITEWAY (\EffParamsLITEWAYLight{} vs.\ \EffParamsLITEWAY{} on average) and needs less memory on the phone and the watch, but on the MCU it requires more RAM in \LiteRamHigherShare\,\% of all combinations (\autoref{sec:appendix-deploy-mcu}).
The quantizer leaves the LeakyReLU activations after its first two convolution blocks and the exponential function of the ELU activation in its temporal module in 32-bit floating point, whereas all operators of LITEWAY are executed in 8-bit integers.
Around each of these operators, the 8-bit tensor is converted to floating point and back, so both representations, the floating-point one four times as large, must be held in the tensor arena at the same time.
Since the LeakyReLU activations operate on some of the largest intermediate tensors of the network, peak RAM grows accordingly.
Despite its higher RAM use, LITEWAY-Light remains about \LiteLatRatio{} times as fast as LITEWAY (\EffMcuLatLITEWAYLight{} vs.\ \EffMcuLatLITEWAY\,ms), since its strided convolutions shorten the sequence earlier.

\paragraph{Energy}
Energy per inference and latency correlate with $r = \HwEnergyLatR$ between the logarithms over a \HwLatencySpan-fold range of latencies, and the coefficient of variation of the average power is \HwPowerCv\,\%.
The energy per inference is thus, to within a few percent, the latency times about \McuPowerMedianMw\,mW, and \McuLowestEnergyModel{} requires the least energy with \McuLowestEnergyMj\,mJ per inference on average.

\paragraph{Cost proxies}
\autoref{fig:hw-proxies} compares how parameter and MAC counts agree with the rankings of architectures by latency on all three platforms, peak memory overhead on the phone and watch, and energy, peak RAM, and flash footprint on the MCU.
For each size measure and cost, we compute Spearman's rank correlation across all architectures with both measurements within each dataset and report the median over the datasets for which the correlation is defined.
Parameter and MAC counts show similar agreement with memory overhead rankings on the phone and watch.
On the MCU, parameter count agrees less closely with peak-RAM rankings ($\rho = \HwRhoParamsMcuRam$), and MAC count agrees more closely with peak RAM and less closely with flash than parameter count does.

\paragraph{Achieved throughput}
To describe the measured execution rate of each architecture, we divide its MAC count per inference by its inference latency and take the median over the datasets separately for each platform (\autoref{fig:hw-throughput}).
This achieved throughput ranges from \HwPhoneGmacsMin\,GMAC/s for \HwPhoneGmacsMinModel{} to \HwPhoneGmacsMax\,GMAC/s for \HwPhoneGmacsMaxModel{} on the phone.
Higher throughput means less inference time per counted MAC, but an architecture with more MACs can still take longer to classify a window.
The medians summarize each architecture's measured workloads, with MCU results limited to the datasets on which it runs, so their range does not establish latency differences between architectures with equal MAC counts.

\paragraph{Window length}
k-NN and SVM become faster with longer windows, since fewer training windows mean fewer stored neighbors and support vectors to compare against.
For most architectures, flash barely depends on the window length, for example \WinFlashKbLITEWAYLightTwoS{} to \WinFlashKbLITEWAYLightThreeS\,KB for LITEWAY-Light, whereas for \WinBigFlashModels, the median flash grows from \WinBigFlashMinMbTwoS{} to \WinBigFlashMaxMbTwoS\,MB for 2\,s windows to \WinBigFlashMinMbThreeS{} to \WinBigFlashMaxMbThreeS\,MB for 3\,s windows.

\begin{appendixfigure}
    \raggedright
    \subsection{Supporting Figures}
    \par\smallskip
    \centering
    \begin{subfigure}[t]{0.48\textwidth}
        \centering
        \includegraphics[width=\linewidth]{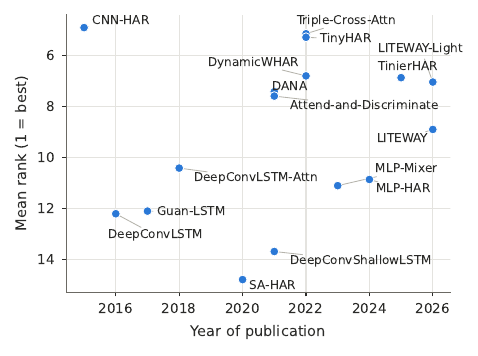}
        \caption{Year of publication}
        \label{fig:insight-progress-year}
    \end{subfigure}
    \hfill
    \begin{subfigure}[t]{0.48\textwidth}
        \centering
        \includegraphics[width=\linewidth]{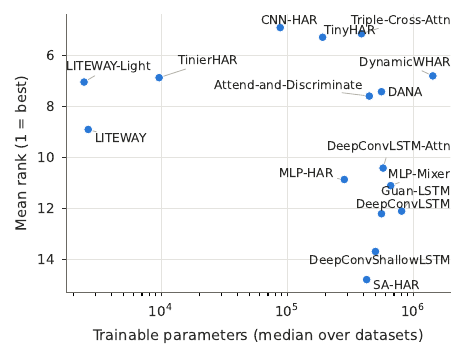}
        \caption{Trainable parameters}
        \label{fig:insight-progress-params}
    \end{subfigure}
    \caption{Mean rank of each deep architecture over the \RankNumDatasets{} datasets (1 = best) against its year of publication (a) and its number of trainable parameters, as the median over the datasets (b).
    Spearman's rank correlation over the architectures is $\rho = \InsYearRho$ ($p = \InsYearP$) for year of publication (a) and $\rho = \InsParamsRho$ ($p = \InsParamsP$) for trainable parameters (b).}
    \label{fig:insight-progress}
\end{appendixfigure}

\begin{appendixfigure}
    \centering
    \includegraphics[width=0.8\textwidth]{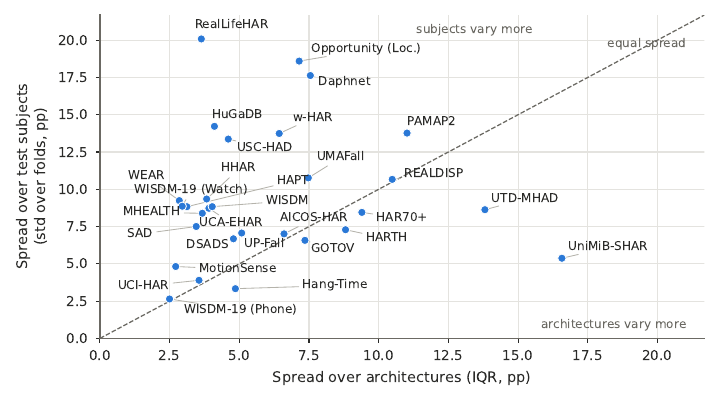}
    \caption{Variation between test subjects compared with variation between architectures.
    One point per dataset: the standard deviation of test macro-F1 over the folds, as the median over the deep architectures, against the interquartile range of test macro-F1 over the deep architectures, with points above the diagonal varying more between subjects than between architectures.}
    \label{fig:insight-subjects}
    \label{fig:insight-subjects-all}
\end{appendixfigure}

\begin{appendixfigure}
    \centering
    \begin{subfigure}[t]{0.48\textwidth}
        \centering
        \includegraphics[width=\linewidth]{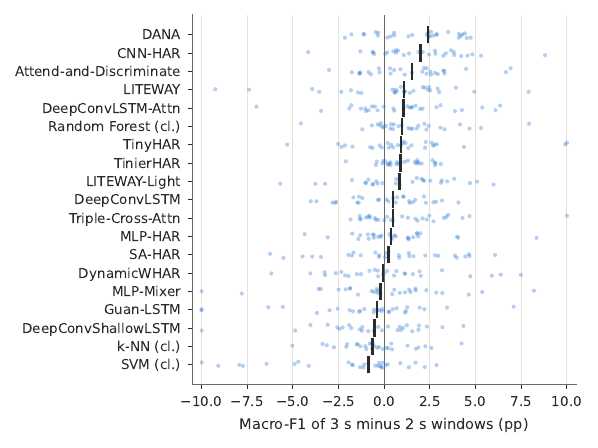}
        \caption{Per architecture}
        \label{fig:insight-window-models}
    \end{subfigure}
    \hfill
    \begin{subfigure}[t]{0.48\textwidth}
        \centering
        \includegraphics[width=\linewidth]{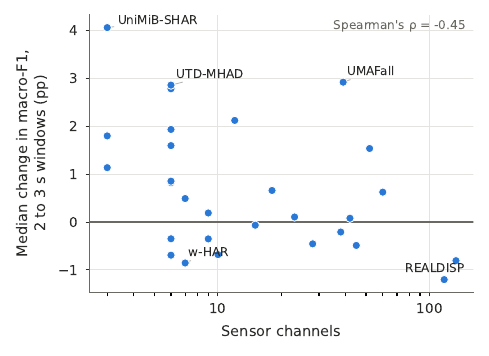}
        \caption{Per dataset}
        \label{fig:insight-window-channels}
    \end{subfigure}
    \caption{Change in test macro-F1 from 2 to 3\,s windows, in percentage points.
    (a) Each dot is one dataset (values beyond $\pm 10$ points are drawn at the border), the black tick marks the median per architecture, and architectures are sorted by their median.
    (b) Median change over the architectures per dataset against the number of sensor channels (Spearman's $\rho = \InsWinChanRho$).
    Datasets recorded below 25\,Hz are upsampled for 2\,s windows (\autoref{sec:method-protocol}), so for them the two window lengths also differ in sampling rate, and 3\,s windows also yield fewer training windows.}
    \label{fig:insight-window}
\end{appendixfigure}

\begin{appendixfigure}
    \centering
    \includegraphics[width=0.70\textwidth]{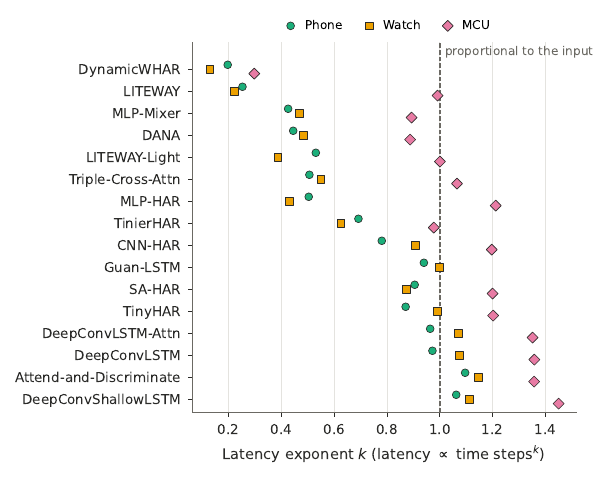}
    \caption{Growth of latency with the input length.
    Exponent $k$ of latency $\propto$ (time steps)$^k$ from 2 to 3\,s windows: per dataset $k = \log(\ell_3/\ell_2) / \log(T_3/T_2)$ for latencies $\ell$ and time steps $T$, median over the datasets whose input grows by at least 10\,\% and, on the MCU, on which the architecture fits at both window lengths.
    $k < 1$: latency grows less than the input, $k > 1$: faster.}
    \label{fig:insight-latency-scaling}
    \label{fig:insight-scaling}
\end{appendixfigure}

\noindent
\section{Declaration of Generative AI Use}
Generative AI tools, including Anthropic Opus 5.5 and OpenAI 6.1 Sol, were used to assist with writing all text passages in this manuscript, researching suitable analysis methods, and generating code for the analyses, figures, and tables.
The analyses were computed by executing this code on the benchmark results, with figures and tables produced from its outputs.
The authors take full responsibility for the choice and suitability of the methods, the correctness of the code, the validity of the reported results, and the accuracy, originality, and integrity of the manuscript, including any AI-assisted content.

\end{document}